\documentclass[journal]{IEEEtran}

\usepackage[utf8]{inputenc}
\usepackage{titling}

\usepackage{amsmath}
\usepackage{amssymb}
\usepackage{pifont}
\usepackage[ruled,vlined]{algorithm2e}
\usepackage{amsfonts}

\usepackage{graphicx}
\usepackage{subcaption}
\usepackage{rotating}
\usepackage{array, booktabs}
\usepackage{float}
\usepackage{caption}
\usepackage[pagebackref=false,breaklinks=true,colorlinks,bookmarks=false]{hyperref}
\usepackage{multirow}

\begin{document}

\title{MaskFaceGAN: High Resolution Face Editing with \\Masked GAN Latent Code Optimization\thanks{Supported by the Slovenian Research Agency ARRS through P2-0250(B), J2-2501(A) and the junior researcher program.}}
\author{Martin~Pernuš,~\IEEEmembership{Student Member,~IEEE,}
        Vitomir~Štruc,~\IEEEmembership{Senior Member,~IEEE,}
        and~Simon~Dobrišek,~\IEEEmembership{Member,~IEEE}\vspace{52mm}}
\markboth{Journal of \LaTeX\ Class Files}%
{Shell \MakeLowercase{\textit{et al.}}: Bare Demo of IEEEtran.cls for IEEE Journals}

\date{}
\maketitle

\noindent
\begin{minipage}[b]{\textwidth}
\vspace{-59mm}
    \begin{minipage}[b]{0.195\textwidth}
    \centering
        \includegraphics[width=\textwidth]{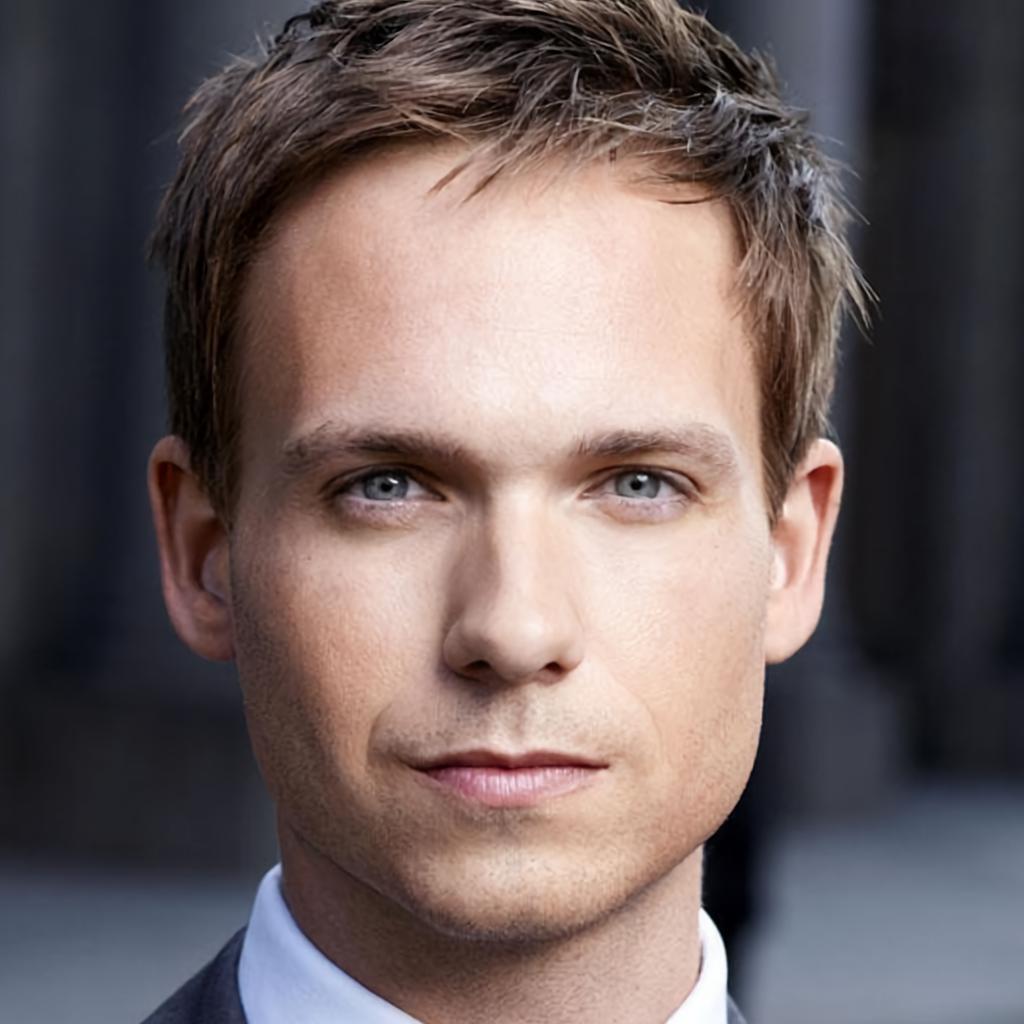}
        {\small Original}
    \end{minipage}
    \begin{minipage}[b]{0.195\textwidth}
    \centering
        \includegraphics[width=\textwidth]{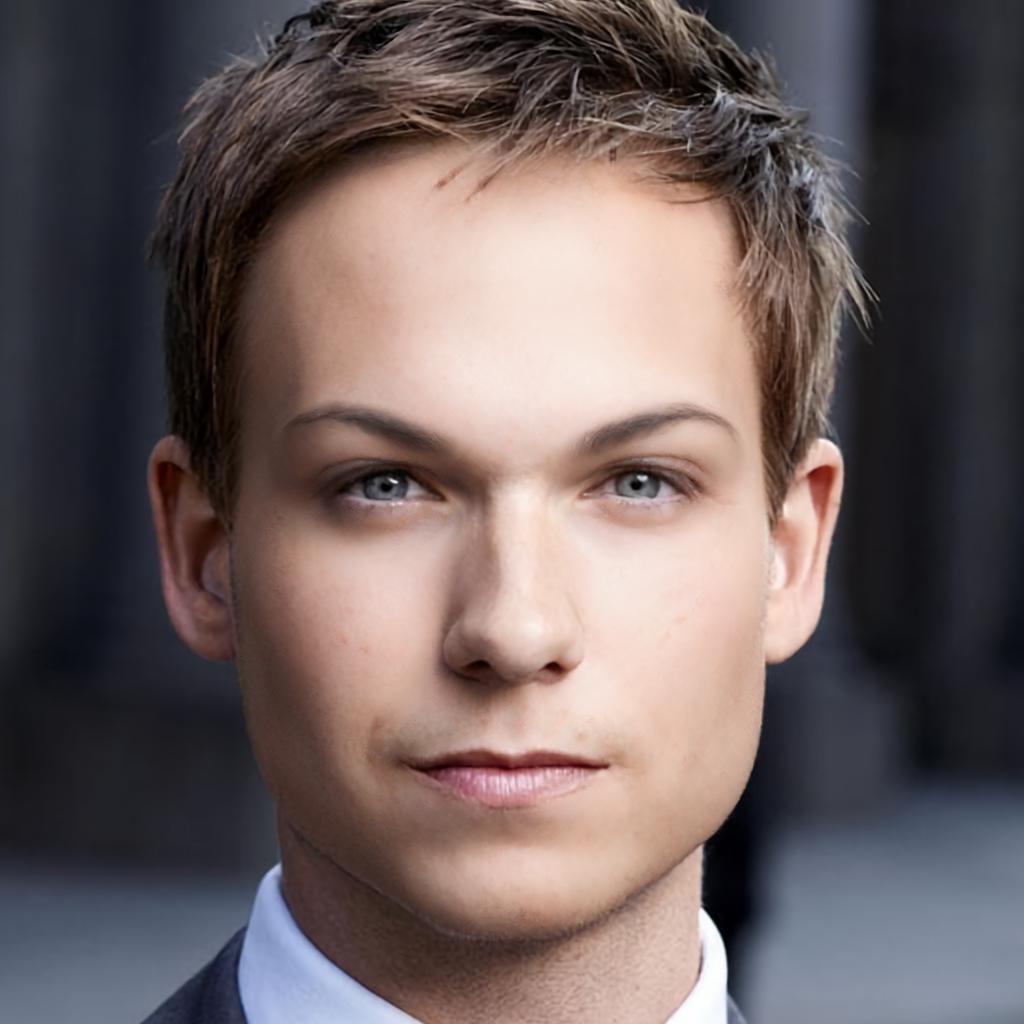}
         {\small Arched eyebrows}
    \end{minipage}
    \begin{minipage}[b]{0.195\textwidth}
    \centering
        \includegraphics[width=\textwidth]{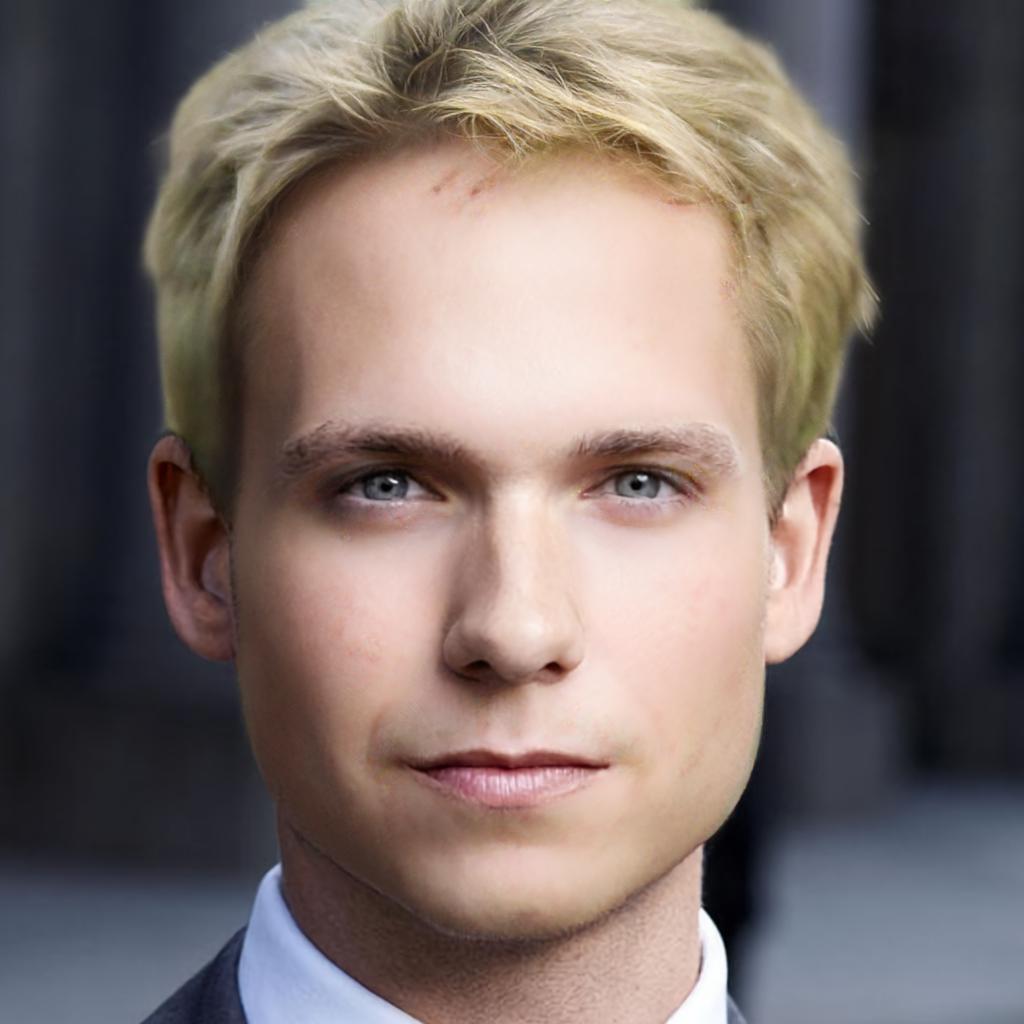}
         {\small Blond hair}
    \end{minipage}
    \begin{minipage}[b]{0.195\textwidth}
    \centering
        	\includegraphics[width=\textwidth]{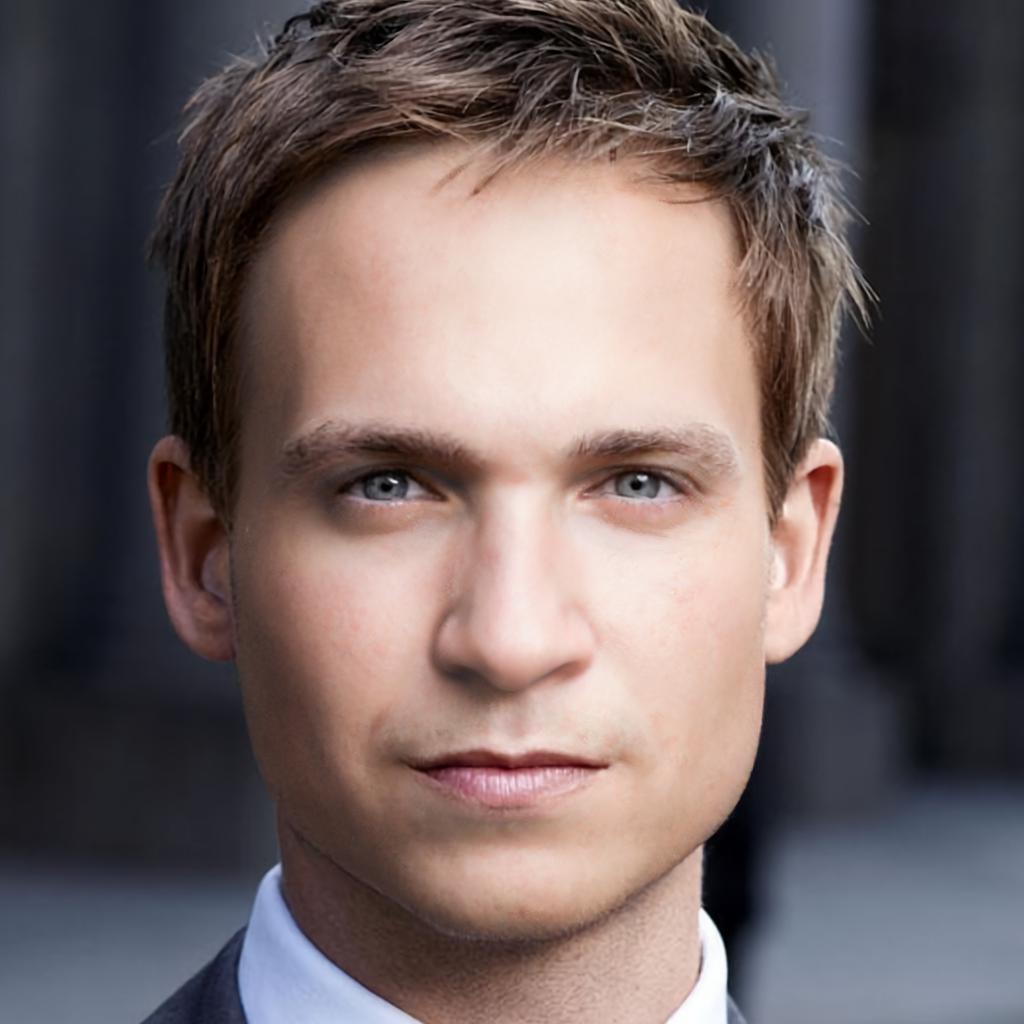}
        	 {\small Big nose}
    \end{minipage}
    \begin{minipage}[b]{0.195\textwidth}
    \centering
        \includegraphics[width=\textwidth]{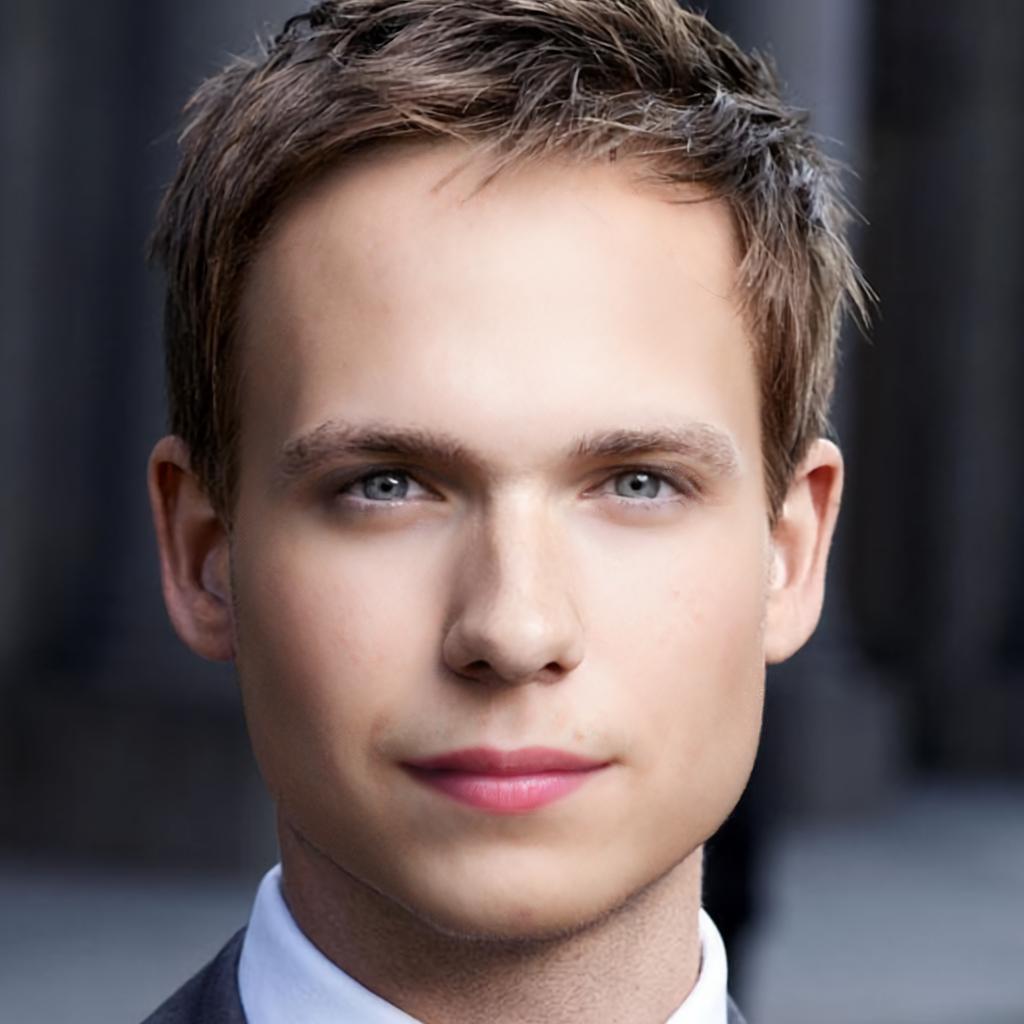}
         {\small Wearing lipstick}
    \end{minipage}
    \captionof{figure}{This paper introduces MaskFaceGAN, a novel approach to face attribute editing capable of generating high-resolution, artefact-free and photo-realistic editing results through a carefully designed optimization procedure applied over the StyleGAN2 latent space. The presented ($1024\times1024$) examples show editing results for four target attributes. Best viewed zoomed-in.\label{fig:cherry}}%
    \vspace{10mm}%
\end{minipage}

\vspace{-6mm}
\begin{abstract}
Face editing represents a popular research topic within the computer vision and image processing communities. While significant progress has been made recently in this area, existing solutions: $(i)$ are still largely focused on low-resolution images, $(ii)$ often generate editing results with visual artefacts, or $(iii)$ lack fine-grained control over the editing procedure and alter multiple (entangled) attributes simultaneously, when trying to generate the desired facial semantics. In this paper, we aim to address these issues through a novel editing approach, called MaskFaceGAN that focuses on local attribute editing. The proposed approach is based on an optimization procedure that directly optimizes the latent code of a pre-trained (state-of-the-art) Generative Adversarial Network (i.e., StyleGAN2) with respect to several constraints that ensure: $(i)$ preservation of relevant image content, $(ii)$ generation of the targeted facial attributes, and $(iii)$ spatially--selective treatment of local image regions. The  constraints are enforced with the help of an (differentiable) attribute classifier and face parser that provide the necessary reference information for the optimization procedure.    
MaskFaceGAN is evaluated in extensive experiments on the CelebA-HQ, Helen and SiblingsDB-HQf datasets and in comparison with several state-of-the-art techniques from the literature. 
Our experimental results show that the proposed approach is able to edit face images with respect to several local facial attributes with unprecedented image quality and at high-resolutions ($1024\times1024$), while exhibiting considerably less problems with attribute entanglement than competing solutions. The source code is  publicly available from: \url{https://github.com/MartinPernus/MaskFaceGAN}.
\end{abstract}%


\IEEEpeerreviewmaketitle

\newcommand{\figureimagewidth}{.3\columnwidth}
\newcommand{\figuretextwidth}{.02\columnwidth}
\newcommand\miniwidth{.162\textwidth}
\newcommand\miniwid{.162\textwidth}

\section{Introduction}
\renewcommand\miniwidth{.196\textwidth}

\IEEEPARstart{F}{ace attribute editing} refers to the task of manipulating facial images towards some predefined appearance. 
Techniques capable of automatic facial attributes editing (e.g., hair color, makeup, shape of facial components, age, identity) have important real-world applications not only in entertainment, graphics, arts, or the beauty industry, but also in problem domains related to visual privacy or security \cite{tolosana2020deepfakes,jiang2021geometrically,mirjalili2020privacynet,deng2021spatially}. As a result, considerable research effort has been directed towards face editing techniques over the years and resulted in powerful solutions capable of generating convincing photo-realistic editing results \cite{choi2018stargan,jo2019sc,he2019attgan,liu2019stgan,shen2020interpreting}.

Recent progress in face attribute editing has  largely been driven by the advances in convolutional neural networks (CNNs) and adversarial training objectives utilized in Generative Adversarial Networks (GAN) \cite{goodfellow2014generative}. 
Existing solutions can broadly be categorized into two main groups. The \textit{first} includes techniques that pose attribute editing as an \textit{image-to-image} translation task and utilize dedicated learning objective to generate the desired target semantics  \cite{shu2017neural, sengupta2018sfsnet,choi2018stargan, he2019attgan, liu2019stgan}. Such techniques typically rely on some sort of encoder-decoder architecture and are, therefore, computationally efficient, but primarily designed for low-resolution editing (e.g., $128\times 128$ or $256\times 256$). Moreover, due to the nature of the learning objective used, they often induce visual artefacts in the edited images. The \textit{second} (more recent) group of techniques is based the concept of \textit{GAN inversion}~\cite{xia2021gan} and exploits the generative capabilities of pre-trained GAN models for editing~\cite{shen2021interfacegan,abdal2019image2stylegan,abdal2020image2stylegan++}. Here, a target image is first converted (embedded) into a latent code and then edited through manipulations (optimization) in the latent space. The main advantage of this group of techniques is the high-resolution and impressive image quality of the editing results. However, because the latent code commonly represents a \textit{global image representation} with entangled attribute information, it is challenging to manipulate individual facial attributes without affecting others. Generating convincing local edits, therefore, represents a major challenge for this group of techniques.     

In this paper, we propose a novel GAN-inversion based approach to (local) facial attribute editing, called MaskFaceGAN, that is capable of generating high-resolution visually convincing editing results (illustrated in Fig.~\ref{fig:cherry}) but does not suffer from the entanglement problems discussed above. The approach is focused on editing facial attributes which are well--defined by specific image regions. At the core of MaskFaceGAN is a carefully designed optimization procedure operating directly over the latent space of the recent StyleGAN2 model~\cite{karras2020analyzing}. The procedure aims to determine a latent code that encodes the desired target semantics (i.e., presence/absence of a target attribute and original facial appearance) by considering multiple groups of optimization constraints during the process of GAN inversion. The \textit{first} group is enforced through a facial attribute classifier and ensures that the edited image contains the correct attribute information. The \textit{second} group of constraints is imposed through a face parser that defines image regions that belong to different facial components. Information on these components is then used as the basis for spatial constraints  that encourage the optimization procedure to either preserve or alter image regions corresponding to specific facial regions. A blending procedure is also incorporated into MaskFaceGAN to help preserve important aspects of the original input image.
MaskFaceGAN is evaluated on three high-resolution face datasets and in comparison to several state-of-the-art editing techniques from the literature. The results of rigorous (qualitative and quantitative) experiments show that the proposed approach generates highly competitive editing results, while exhibiting some unique characteristics not available with prior editing solutions. %
Overall, we make the following contributions in this paper:
\begin{itemize}
\item We present MaskFaceGAN, a novel approach to face image editing, capable of generating state-of-the-art, visually convincing, artefact-free, photo-realistic editing results at high resolutions, i.e., $1024\times 1024$ pixels.
\item We propose an efficient optimization procedure for estimating GAN latent  codes of facial images that encode the selected target semantics. The procedure enforces various optimization constraints through the use of differentiable models applied over the edited images. 
\item We show how MaskFaceGAN can be used for attribute-intensity control, multi-attribute editing and component size manipulation while requiring only binary attribute labels for optimization. 
\item  We demonstrate through comprehensive experiments that spatially constrained image editing significantly reduces entanglement problems compared to competing models.
\end{itemize}

\begin{figure*}[!t!]
    \centering
    \includegraphics[width=\textwidth]{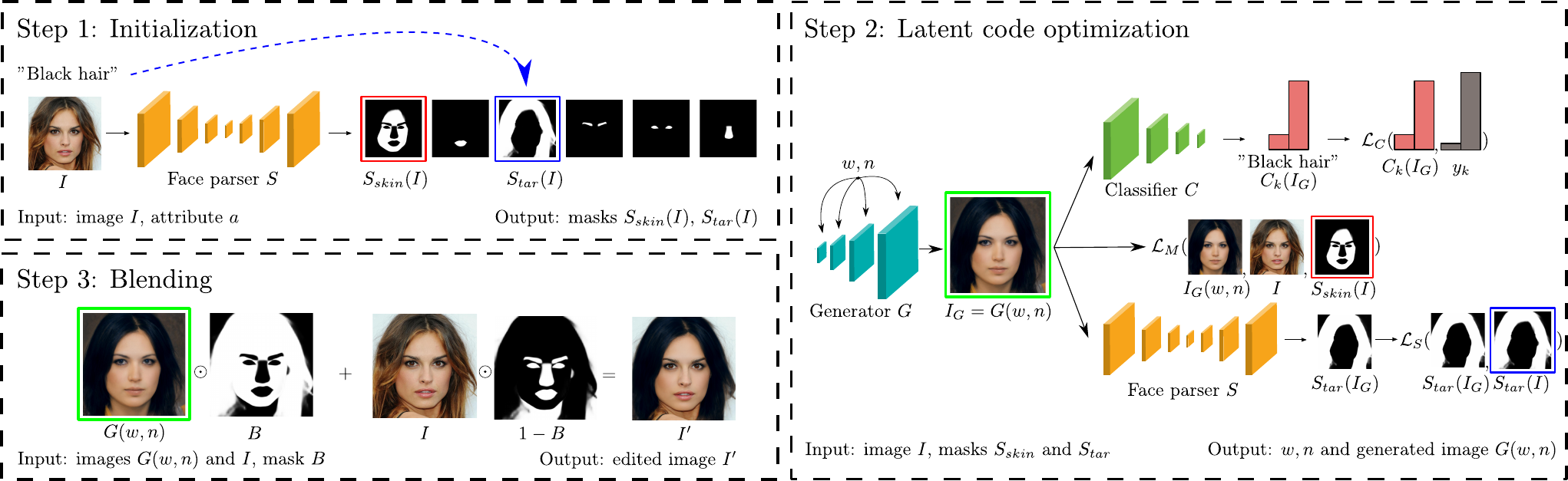}
    \caption{Overview of MaskFaceGAN, illustrated with the ``Black hair'' target attribute. To initialize the editing procedure, MaskFaceGAN uses a face parser to define masks that correspond to image regions that should be preserved ($S_{skin}$) and regions that should be altered ($S_{tar}$).  
    Next, latent code optimization is performed in accordance with semantic and spatial constraints to generate an intermediate image $I_G$ with the targeted characteristics. Finally, blending is used to combine the generated image  $I_G$ with the original $I$ and to produce the final editing result $I'$. The image is best viewed electronically.}
    \label{fig:skica}
\end{figure*}

\section{Related work}
\label{sec:RelatedWork}
In this section, we present prior work closely related to our paper. We discuss Generative Adversarial Networks (GANs), research on pre--trained GANs and face editing techniques.\vspace{-2mm}

\subsection{Generative Adversarial Networks}
Generative Adversarial Networks (GANs) are among the most popular generative models in the field of image processing and computer vision \cite{goodfellow2014generative}. 
Existing GANs  can be broadly split into two categories: $(i)$ unconditional and $(ii)$ conditional models. Unconditional GANs refer to models that rely only on random noise  to generate image data. No additional signal is used to steer the generation process. Conditional GANs, on the other hand, exploit additional inputs to control the semantic content of the generated images and typically utilize random noise to ensure diversity. Different forms of the conditional signals have been used in the literature, including 
class labels \cite{zhang2019self, brock2018large}, graph representations, \cite{johnson2018image, ashual2019specifying}, layouts of image objects \cite{park2019semantic} or text descriptions \cite{xu2018attngan, KT-GAN, multisentence-textgan} among others.

The progress in GAN-based image generation can for the most part be attributed to advances in model design and training.  
DCGAN \cite{DCGAN}, for example, proposed a convolutional GAN architecture and defined several useful design principles, such as the use of batch normalization in all model layers and specific activation functions for the generator and the discriminator. 
In \cite{karras2018progressive}, Karras \textit{et al.} introduced a progressive learning strategy for GANs that adds higher-resolution layers to the model once the lower-resolution layers have converged. The authors showed that using such a strategy results in GANs capable of generating convincing megapixel-sized images.
The progressively learned model was further improved with the introduction of StyleGAN \cite{karras2019style}, a GAN model inspired by the style--transfer literature. Different from traditional Gaussian-shaped latent spaces, StyleGAN proposed the use of a non-linear mapping from the Gaussian latent space to an intermediate latent space (with better interpolation and disentanglement properties) that was then fed to convolutional layers via an adaptive instance normalization operation. 
Additionally, the model also introduced noise inputs for generating stochastic image details. The next iteration of the model, StyleGAN2 \cite{karras2020analyzing}, modified the adaptive instance normalization operation to remove circular artefacts in the generated images, achieving state-of-the-art results on unconditional image generation. Considerable progress has also been made with GAN learning strategies, where different losses and regularizations were proposed to improve the generation quality \cite{mao2017least, arjovsky2017wasserstein, gulrajani2017improved, miyato2018spectral, mescheder2018training}. Additional, up-to-date information on GANs can be found in one of the recent surveys on this topic \cite{wang2019generative,saxena2021generative}.

\subsection{Studies on Pre-trained GANs}
Training GAN models can be highly resource intensive. For example, the computational effort required for developing and training the recent StyleGAN2 model was estimated to be around $51$ Volta GPU years \cite{karras2020analyzing}. This immense effort has motivated research into the capabilities of pre--trained GANs and resulted in powerful techniques that exploit existing models for various generative tasks. Bau \textit{et al.} \cite{baugan}, for example, analyzed pre--trained GANs to achieve localized deletion and insertion of objects. Jahanian \textit{et al.} \cite{jahanian2019steerability} investigated linear and non-linear walks in the GAN latent space that resulted in basic image manipulations, such as changes in brightness or the zoom factor. Goetschalckx \textit{et al.} \cite{goetschalckx2019ganalyze} navigated the latent code manifold to improve the image's memorability. Yang \textit{et al.} \cite{yang2019semantic} analyzed the relationship between image semantics and layer activations and manipulated image characteristics, such as layout, scene attributes or the employed color scheme. PULSE \cite{pulse} utilized StyleGAN to perform face superresolution. InsetGAN \cite{insetgan} utilized two separate pre--trained GANs to perform face--conditioned body synthesis and body--face image stitching, where the first GAN models the face region, whereas the other GAN models the entire body.

Similarly to the research discussed above, MaskFaceGAN also exploits a pre-trained GAN model to edit facial images. However, different from existing work, the GAN model used in our approach serves only as a proxy for the editing procedure that synthesizes semantically meaningful content in spatially constrained face regions. The synthesized content is combined later with the original image for the final result. 

\subsection{Face Editing}
Numerous approaches for face editing and manipulation have been presented in the literature, e.g.,~\cite{tolosana2020deepfakes,jiang2021geometrically,lin2021anycost,afifi2021histogan}. The work in \cite{portenier2018faceshop, jo2019sc}, for example, explored the use of user-supplied sketches to drive the editing procedure. The authors of \cite{shu2017neural, sengupta2018sfsnet} learned disentangled latent representations w.r.t. image formation to enable control of the image generation process. Lee \textit{et al.} \cite{CelebAMask-HQ} proposed MaskGAN model, a conditional GAN, capable of modifying specific facial components and demonstrated the benefit of using spatially local face editing.

Particularly convincing editing results have been reported with encoder-decoder models.
Choi \textit{et al.} \cite{choi2018stargan}, for instance, introduced StarGAN, an image-to-image encoder-decoder network with cycle consistency \cite{zhu2017unpaired}, capable of manipulating the appearance of several face attributes. He \textit{et al.} \cite{he2019attgan} proposed AttGAN, a notable encoder-decoder model that utilizes reconstruction constraints instead of cycle consistency. Liu \textit{et al.} \cite{liu2019stgan} improved on AttGAN with their STGAN design by modifying the input signal and improving the encoder-decoder architecture with selective transfer units. Encoder-decoder models were shown to generate visually convincing editing results, but are less suitable for editing high-resolution images, where visual artefacts are often observed. 

More recent solutions approach the problem of face editing with the use of pre-trained GANs. Abdal \textit{et al.} \cite{abdal2019image2stylegan} showed that it is possible to embed a large variety of images in the extended latent space of StyleGAN and to perform various forms of image manipulation in the latent space, including face morphing, style transfer and expression transfer. In their follow up work \cite{abdal2020image2stylegan++}, the authors improved the embedding algorithm by optimizing the StyleGAN noise component, and demonstrated additional capabilities, such as local editing or face inpainting. A convincing editing approach, called InterFaceGAN, was described by Shen \textit{et al.} in \cite{shen2020interpreting,shen2021interfacegan}. To edit an image, InterFaceGAN moves the latent code corresponding to the input image along a linear subspace. The direction of the displacement is determined through a support vector machine (SVM), trained on the StyleGAN latent space given labels of predefined facial attributes. While such approaches led to state-of-the-art editing results, they are based on linear operations applied over latent space representations and, therefore, often suffer from entanglement issues where changing one attribute also results in changes in other  (entangled) image attributes. 

With MaskFaceGAN we improve on previous methods by directly optimizing the latent code associated with an input image through multiple optimization constraints that not only control the manipulated semantic content but also the spatial area in which the editing occurs. This optimization procedure results in complex (non-linear) changes in the latent code of the input image and leads to editing results with significantly less entanglement problems than competing solutions, as we demonstrate in the experimental section. 

\section{Methodology}


\subsection{Background and Problem Formulation}

The goal of face attribute editing is to manipulate (in a photo--realistic manner) a given input image $I \in \mathbb{R}^{3 \times m \times n}$ in accordance with some specified target semantics $a$, i.e.,
\begin{equation}
    \psi_a: I \mapsto I' \in \mathbb{R}^{3 \times m \times n},   
\end{equation}
where $I'$ represents the edited image and the semantics are usually defined by predefined facial attributes (e.g., ``Blond hair'', ``Big nose'', etc.). As discussed in the previous section, the most recent methods implement the mapping $\psi_a$ through the process of GAN inversion~\cite{shen2021interfacegan,xia2021gan}. With these techniques the input image $I$ is first embedded in the  latent space of a pre-trained GAN model $G$, resulting in a latent representation (or latent code) $w$. Next, the latent code is modified in accordance with a target objective determined by $a$, i.e., $\psi_a: w\mapsto w^*$, and finally, the edited image $I'$ is generated by evaluating $w^*$ through $G$, that is, $I'=G(w^*)$.

MaskFaceGAN, presented in the following sections, follows this general GAN inversion framework, but different from competing solutions exhibits several unique characteristics, as illustrated in the experimental part of the paper. 

\subsection{Overview of MaskFaceGAN}
A high--level overview of MaskFaceGAN is presented in Fig.~\ref{fig:skica}. The key component of the proposed approach is a latent code optimization procedure that 
considers multiple constraints 
during optimization, including: $(i)$ an \textit{appearance--preservation constraint} that ensures that the edited image $I'$ is as close to the original $I$ in image areas that should not be altered by MaskFaceGAN, $(ii)$ a \textit{semantic constraint} that ensures meaningful target semantics within local image areas, and $(iii)$ a \textit{shape constraint} that determines the location and shape of the image regions to be manipulated during editing. The optimization procedure is used as part of a three--step editing operation in MaskFaceGAN, i.e.:  
\begin{itemize}
    \item \textbf{Step 1: Initialization.} Given an input image $I$ and a binary (target) attribute label $a$, MaskFaceGAN uses a face parser $S$ in the first step to identify facial regions $S_{skin}$ that should be preserved during editing and predict image areas $S_{tar}$ that need to be edited given $a$. See Fig.~\ref{fig:skica} for an example using the target attribute ``Black hair''. 
    \item \textbf{Step 2: Latent--code optimization.} The  second step of MaskFaceGAN involves the optimization of the GAN latent code. This step aims at estimating a latent code of the (intermediate) image $I_G$ that exhibits 
    the targeted semantics in image regions defined by $S_{tar}$ and preserved appearance in $S_{skin}$. 
    The semantics and spatial constraints are enforced through an attribute classifier $C$ 
    and 
    a face parser $S$, respectively. 
    A loss, defined 
    over these models, is backpropagated to the latent space to facilitate the optimization. 
    \item \textbf{Step 3: Blending.} Finally, a blending step is utilized to combine the (intermediate) image $I_G$, generated from the optimized latent code  (with locally manipulated target semantics), with the original image $I$. This step adds the background and other facial components not considered during optimization to the final editing result $I'$.   
\end{itemize}

\subsection{Models}

MaskFaceGAN relies on three distinct components to implement the optimization procedure, i.e., $(i)$ a GAN--based generator ($G$) capable of producing high--resolution facial images, $(ii)$ an attribute classifier ($C$) utilized for enforcing the targeted semantics, and $(iii)$ a face parser ($S$) 
used for imposing spatial constraints. 
\begin{itemize}
    \item \textbf{The Generator} ($G$) is based on StyleGAN2 \cite{karras2020analyzing}, a state-of-the-art GAN model specialized for generating photo--realistic facial images. Following established editing methodology \cite{abdal2019image2stylegan, abdal2020image2stylegan++} the extended latent space\footnote{The extended latent space $\mathcal{W}^+$ allows for the embedding of arbitrary facial images in StyleGAN2 and represents an extension of the model's original latent space to all layers of the model.} $\mathcal{W}^+$ of StyleGAN2 is used for encoding  image semantics. As a result, an image is represented through a concatenation of $n_c=18$ different $512$-dimensional latent vectors $w_i$, one for each layer of the model, i.e., $w = \{w_i\}_{i=1}^{n_c}$. To ensure photo-realism StyleGAN2 additionally uses $N=17$ stochastic (i.e., Gaussian noise) channels $n=\{n_i\}_{i=1}^{N}$ of different spatial resolutions (ranging from $4 \times 4$ to $1024 \times 1024$) that encode high-frequency image details.  
    The model, hence, generates output images $I_G$ based on the following mapping: $I_G = G(w,n)$. 
    \item \textbf{The Attribute Classifier} ($C$) is  designed around the multi–task tree neural network from \cite{vandenhende2019branched} and consists of several shared layers and $K$ classification heads (leaf branches), one for each of the $K$ attributes supported (for editing) by MaskFaceGAN. Given an image $I_G = G(w,n)$, each classification head $C_a$ predicts the probability of a facial attribute being present in $I_G$, i.e., $c_k = C_k(I')$ for the $k$-th attribute. 
    \item \textbf{The Face Parser} ($S$) is built around DeepLabV3 \cite{deeplabv3} and provides pixel-level probability predictions for various face components/regions, as illustrated in Fig.~\ref{fig:face_parser}. These facial regions are then associated with specific attributes that can be edited within the given region in accordance with Table~\ref{tab:attr2regions}. Formally, the model implements a mapping from an image $I$ to a tensor of probabilities along the channel dimension, i.e.: $S: \mathbb{R}^{3 \times n \times m} \to [0,1]^{L \times n \times m}$, where $L$ is the number of parsed categories (face components). For MaskFaceGAN, two principal channels are used. The first one is the skin region, $S_{skin} \in [0,1]^{n \times m}$, which facilitates preservation of facial characteristics unrelated to the change in the targeted semantics. The second one is determined (dynamically) based on the targeted facial attribute, $S_{{tar}}(I) \in [0,1]^{n \times m}$. 
\end{itemize}

\renewcommand{\miniwidth}{.175\columnwidth}
\renewcommand{\miniwid}{.03\columnwidth}
\newcommand{\im}[1]{\includegraphics[width=\miniwidth]{images/face_parser-GT_only/#1.jpg}}

\setlength{\tabcolsep}{0.5mm} 

\newcolumntype{C}{ >{\centering\arraybackslash} m{\miniwidth} }
\newcolumntype{D}{ >{\centering\arraybackslash} m{\miniwid} }
\newcommand{\turnedtextbegin}[1]{\begin{turn}{90} #1 \end{turn}}

\begin{figure}[t]
\begin{tabular}{DCCCCC}
    \turnedtextbegin{\hspace{-0.5cm} 
    \scriptsize Ground Truth \hspace{0.45cm} Input} & \im{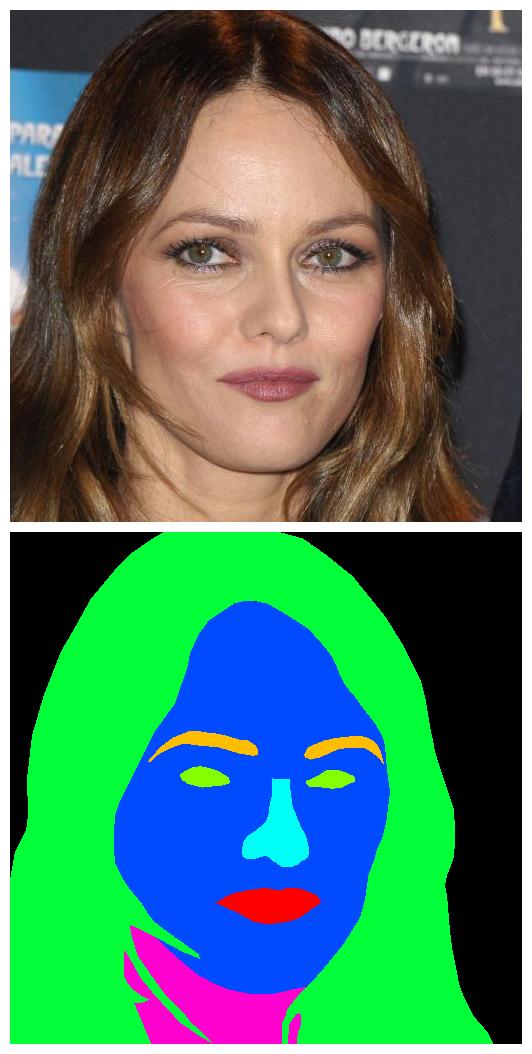} & \im{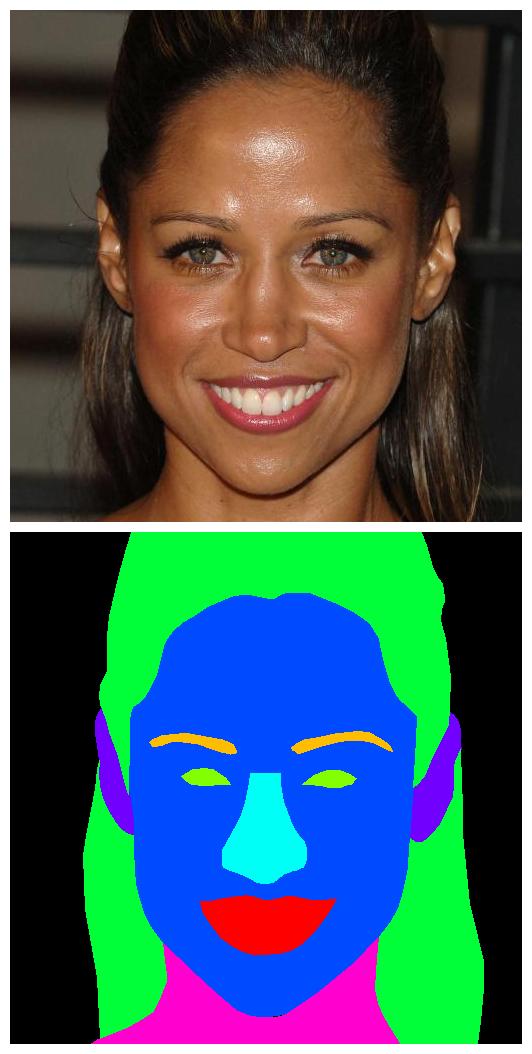} & \im{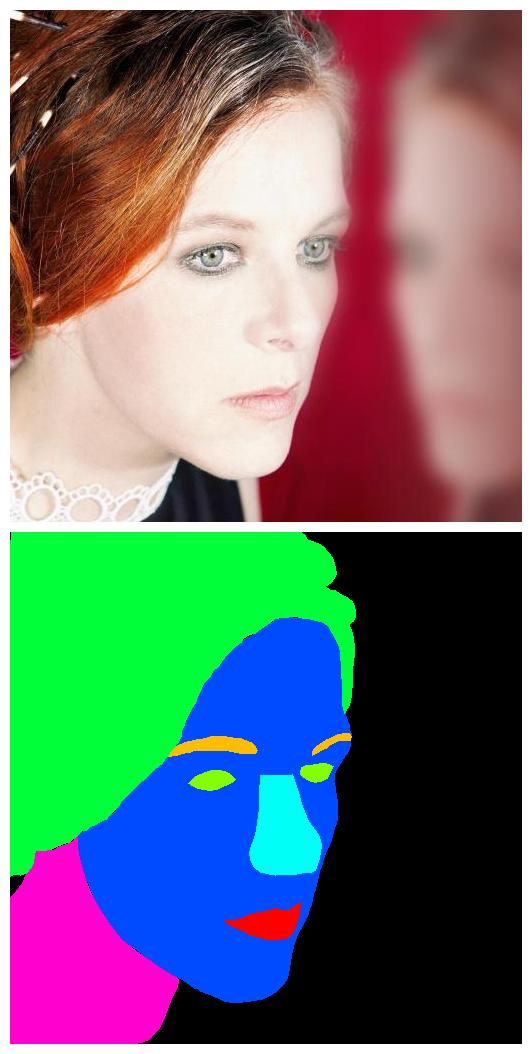} & \im{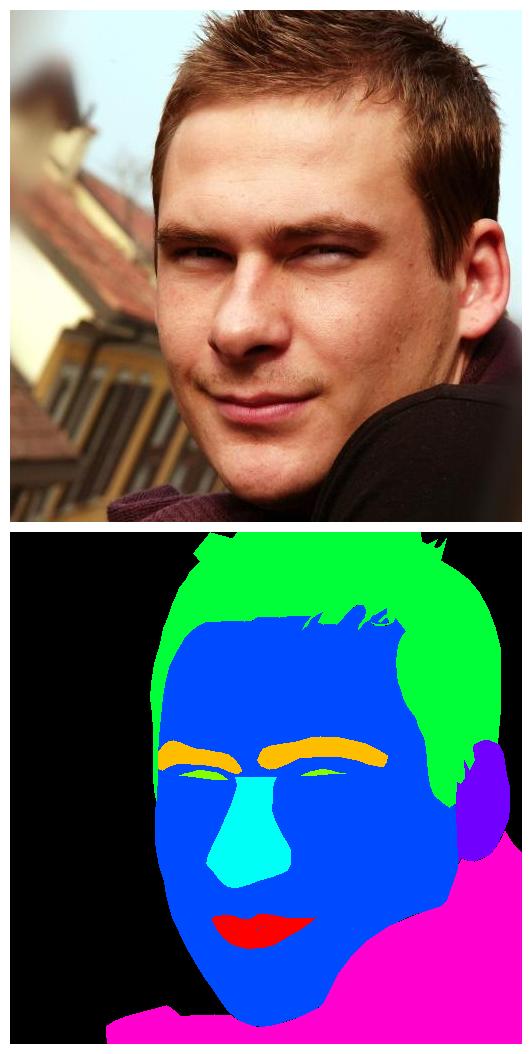} & \includegraphics[width=0.11\columnwidth]{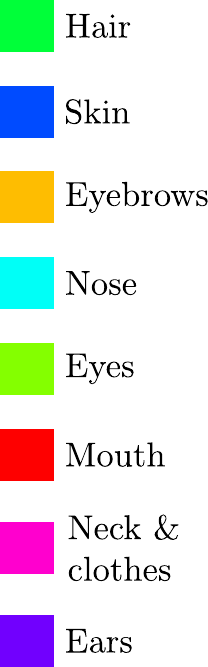}
\end{tabular}
\caption{Illustration of the ground truth used to train the face parser $S$. The presented regions are associated with specific (local) attributes that can be edited, as summarized in Table~\ref{tab:attr2regions}. 
}\vspace{-5mm}
\label{fig:face_parser}
\end{figure}

\subsection{Supported Attributes and Local Embedding}

MaskFaceGAN is designed around the assumption that changes in certain attributes are reflected only through changes in spatially local facial regions. As we demonstrate in the experimental section, this assumption is not only reasonable, but also helps to mitigate issues related to attribute entanglement often observed with competing techniques. 

Based on the attribute annotations and face components available with popular face datasets, such as CelebA~\cite{liu2015faceattributes} and CelebAMask-HQ~\cite{CelebAMask-HQ}, we identify $14$ facial attributes that can be associated with specific facial regions,
 as summarized in Table~\ref{tab:attr2regions}. MaskFaceGAN, hence, supports editing of $14$ different attributes,  comparable (in terms of numbers) to competing approaches from the literature, e.g., \cite{he2019attgan, liu2019stgan}. Additionally, the local nature of the editing procedure, allows MaskFaceGAN to embed only specific image regions into the StyleGAN latent space (similarly to~\cite{abdal2020image2stylegan++}) and enforce targeted semantics only within local spatial areas.    
For example, we only embed the nose region along with the adjacent face region when targeting the ``Pointy nose'' attribute and optimize the latent code with the goal of ensuring the desired semantics exclusively within the nose area -- irrespective of the visual changes to other facial parts\footnote{The modified visual information in other parts is corrected for through the blending procedure in the last step of the MaskFaceGAN editing procedure.}. This is achieved through a series of carefully designed optimization constraints presented in the next section. 
\begin{table}[!t!]
  \renewcommand{\arraystretch}{1.5}
  \caption{Attributes supported for editing by MaskFaceGAN and corresponding facial areas manipulated by the proposed approach to enforce desired semantics.}
  \label{tab:attr2regions}
  \centering
  \resizebox{\columnwidth}{!}{%
  \begin{tabular}{ll|ll}
  \hline\hline
    \textbf{Face attribute (for editing)} &&& \textbf{Face region (returned by $S$)} \\
    \hline
    Blond, Brown, Black, Grey, Straight, and Wavy hair &&& Hair \\
    Wearing lipstick, Smiling, Mouth slightly open &&& Lower and Upper lip, Mouth \\
    Bushy eyebrows, Arched eyebrows &&& Left and Right eyebrow  \\
    Pointy nose, Big nose &&& Nose \\
    Narrow eyes &&& Left eye, Right eye \\
    \hline\hline 
  \end{tabular}\vspace{-3mm}
  }
\end{table}

\subsection{Latent Code Optimization}

\textbf{Appearance Preservation.} The goal of facial attribute editing is to alter specific (targeted) image semantics, while preserving all (or most) other visual characteristics of the input images. To ensure that image regions not associated with the targeted attributes are preserved, a (local) appearance--preservation constraint is used during optimization. The constraint is defined as a masked mean squared error (MMSE): 
\begin{equation}
    \mathcal{L}_M = ||S_{skin}(I) \odot (I_G - I)||^2_2,
    \label{eq:mse}
\end{equation}
where $S_{skin}(I)$ is a probabilistic mask produced by the face parser $S$, $\odot$ is the Hadamard product, and $I_G=G(w,n)$. $\mathcal{L}_M$ encourages the generated image $I_G$ and the input image $I$ to be as similar as possible within the area defined by $S_{skin}$.

\textbf{Semantic Content.}
The constraint in Eq.~\eqref{eq:mse} forces certain image regions in $I_G$ to be preserved {w.r.t.} $I$, while the rest is allowed to change.  MaskFaceGAN, thus, synthesizes the remaining image pixels in accordance with the targeted semantics by considering a semantic--content constraint in the optimization procedure. The constraint ensures that the latent code $w$ produces an image $I_G$ with the desired facial attributes and is defined as the average Kullback--Leibler (KL) divergence $D_{KL}$ between the smoothed ground truth probability distribution and classifier predictions for the targeted attribute(s)~\cite{van2014renyi}, i.e.:

\begin{align}
    \mathcal{L}_C = \frac{1}{K} \sum_{k=1}^K D_{{KL}}(C_k(I_G), y_k),
    \label{eq:classf}
\end{align}
where $K$ denotes the number of targeted facial attributes, $C_k$ stands for the attribute classifier prediction corresponding to the $k$-th attribute and $y_k \in \{\epsilon,1-\epsilon\}$ is the smoothed ground truth that denotes the absence or the presence of the desired attribute, respectively. The value of $\epsilon$ can be used to set the intensity of the desired attribute, e.g. the intensity of lipstick presence when editing the ``Wearing lipstick'' attribute.

\textbf{Target Region Shape.} Because image content with the targeted semantics is first synthesized by MaskFaceGAN and later blended with the original image, it is critical that the shape of the targeted facial regions is preserved. To this end, the proposed approach constrains the shape of the targeted region with the help of the face parser $S$ using:   
\begin{equation}
    \mathcal{L}_S = ||S_{tar}(I) - S_{tar}(I_G)||^2_2,
    \label{eq:seg}
\end{equation}
where $S_{tar}$ is again a probabilistic mask of the spatial region associated with the targeted attribute -- see Table~\ref{tab:attr2regions}.  

Constraining the shape of the manipulated face components was found to be especially important for hair editing. If the synthesized hair in $I_G$ does not cover at least the original hair region in $I$, the blending steps generates visible artefacts that affect the perceived quality of the edited images.

\renewcommand{\miniwidth}{0.4\columnwidth}
\begin{figure}[t]
\centering
    \captionsetup[subfigure]{labelformat=empty}
        \begin{subfigure}{\miniwidth}
    		\centering
    		\includegraphics[width=\textwidth]{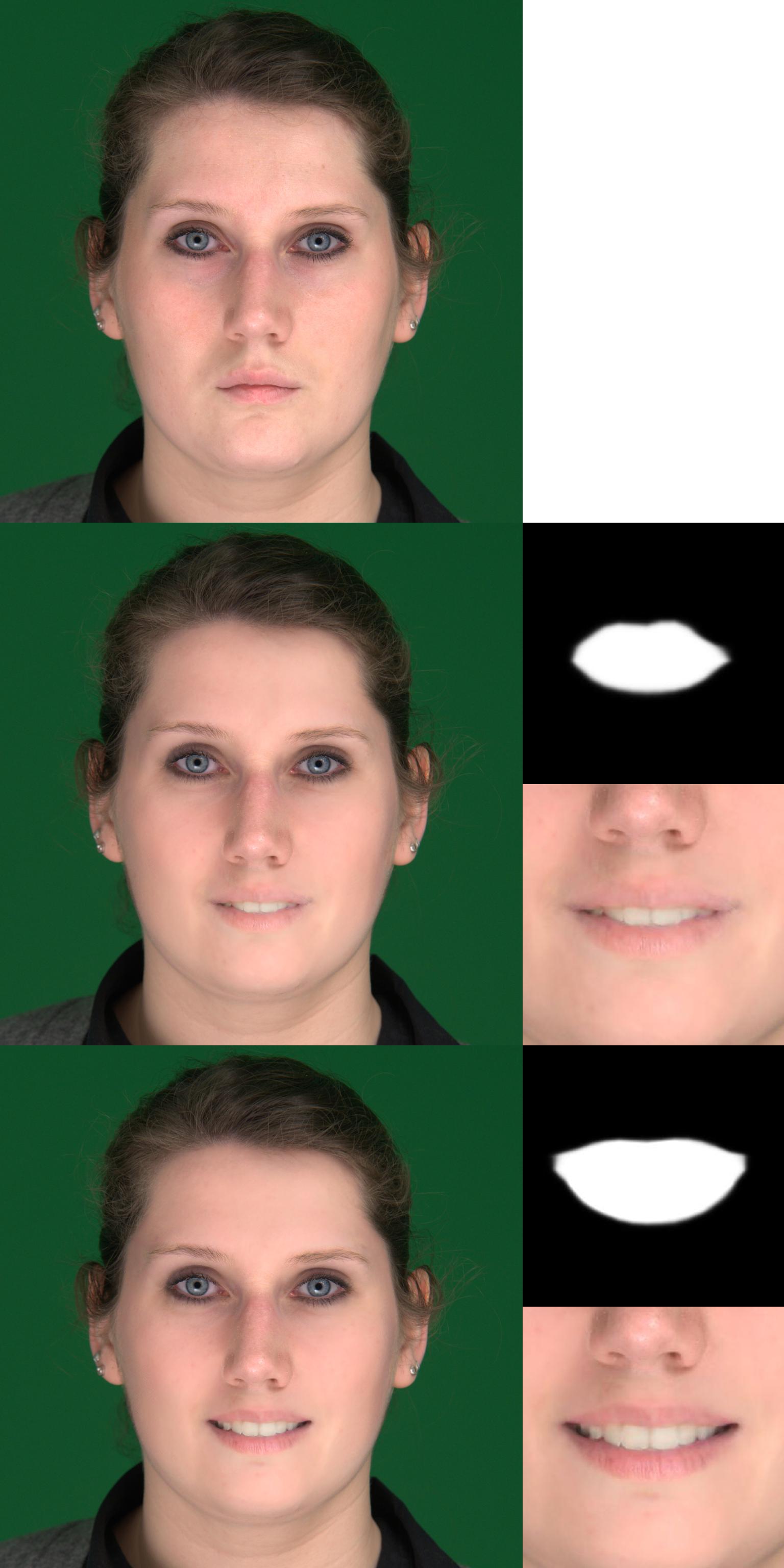}
    		\caption{\small Smiling}
    		\end{subfigure}
    \begin{subfigure}{\miniwidth}
    		\centering
    		\includegraphics[width=\textwidth]{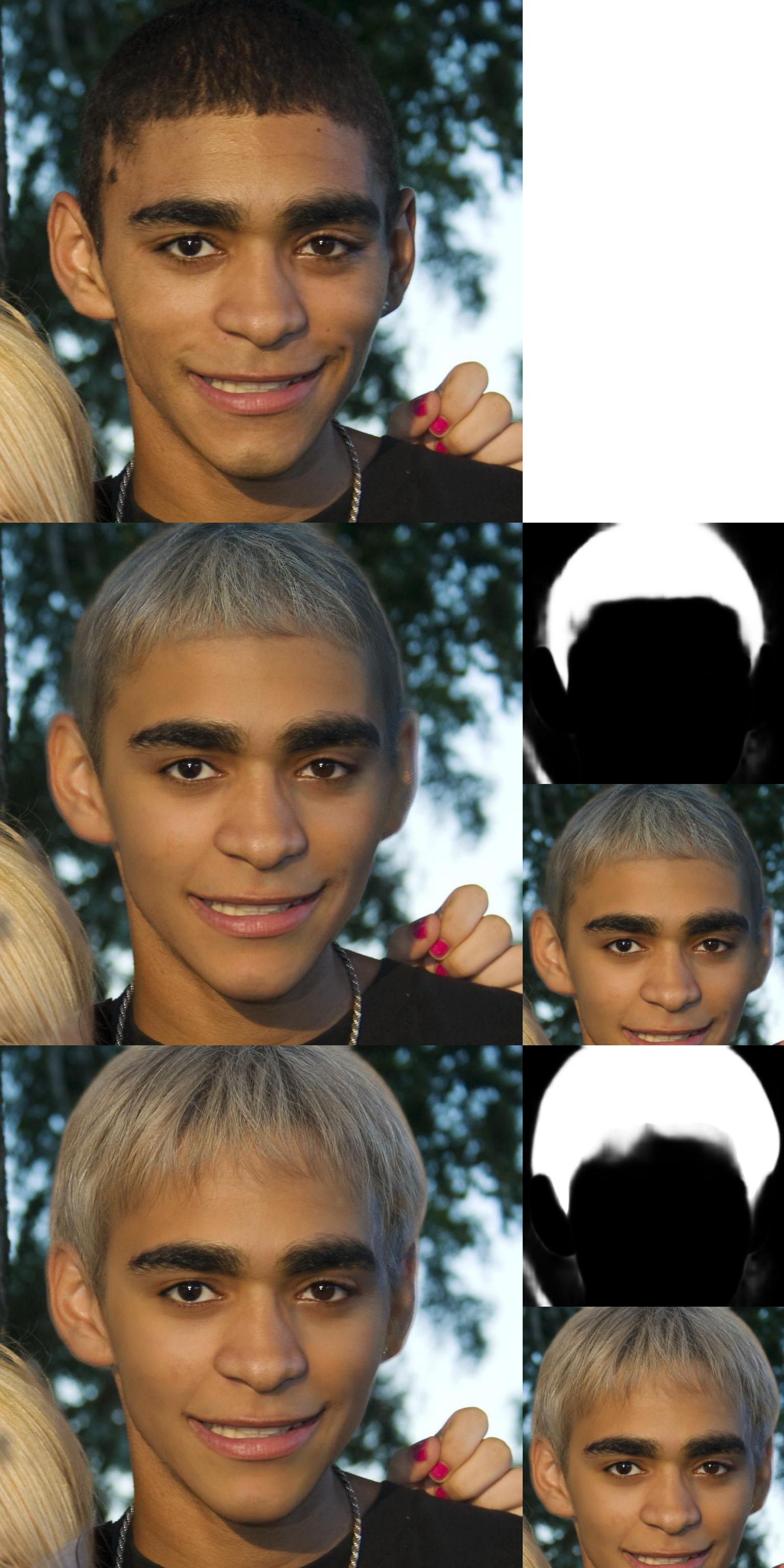}
    		\caption{\small Straight hair}
    	\end{subfigure}
\caption{Impact of flexible spatial constraints on the visual appearance of two sample images with two targeted attributes. The first row shows the original images, the middle row shows the editing results and the target region $S_{{tar}}(I)$ without flexible spatial constraints, and the final row shows the results with flexible constraints. Note how better semantics can be captured for the both the ``Smiling'' as well as the ``Straight hair'' target attributes by relaxing the spatial constraints. 
}\vspace{-2mm}
\label{fig:dynamic_masking}
\end{figure}

\textbf{Component Size.}
While the main goal of MaskFaceGAN is to produce convincing manipulations of existing image content, additional components can be incorporated into the framework to enable further editing capabilities. Specifically, MaskFaceGAN can manipulate the size of the target facial region by considering a suitable optimization objective. Let the portion of the image $s_{tar}\in[0,1]$ covered by a given target component be defined as:
\begin{equation}
s_{tar}(I) = \frac{\sum_{x,y} S_{tar}(I)}{|S_{tar}(I)|},
\end{equation}
where the operator $|S_{tar}(I)|$ denotes the number of pixels in $S_{tar}$.  
To be able to scale the size of the targeted facial region we introduce a scaling factor $\alpha$ and integrate it into an objective that considers the component size when optimizing for the latent code $w$. The objective is defined as the KL divergence between the initial component portion $s_{tar}(I)$ and the desired portion $s_{tar}(I_G)$ in the generated image $I_G$, i.e.: 
\begin{equation}
\mathcal{L}_P = D_{KL}\left(s_{tar}(I_G), \alpha s_{tar}(I)\right).
\label{eq:size_manipulation}
\end{equation}
We note that this term is optional and can be excluded from the optimization procedure by 
setting the corresponding weighting factor to $0$ - see final objective in Eq.~\eqref{eq:final} for details.

\textbf{Flexible Spatial Constraints.}
The appearance--preservation and target--shape optimization constraints, defined in Eqs. (\ref{eq:mse}) and (\ref{eq:seg}), impose significant limitations on the spatial regions associated with the targeted facial attributes. The appearance--preservation constraint does not allow to grow relevant facial components if they overlap with the skin region. Similarly, the target--shape objective forces the edited image to have exactly the same target component shape as the original image, which in some cases might not be desired. For example, when editing hair color, the target shape needs to be preserved, but when editing hair shape (e.g., ``Straight hair'' or ``Wavy hair'') modifications of the target image region must be allowed. 

To deal with such issues, we relax the optimization constraints from Eqs. (\ref{eq:mse}) and (\ref{eq:seg}) and incorporate mechanisms into MaskFaceGAN that allow for more flexible spatial editing. Specifically, in each iteration of the optimization procedure, we first compute the target region on the generated image $S_{tar}(I_G)$. Next, we subtract this region from $S_{{skin}}(I)$ for the appearance--preservation constraint to preserve less pixels. For the target--shape objective, the region is added to the target region of the original image $S_{{tar}}(I')$ to allow region growth. Here, we also require that the combined region covers at least the original component shape to avoid visual artefacts. The final (relaxed) appearance--preservation $\mathcal{L}_M$ and target--shape $\mathcal{L}_S$ constraint used by MaskFaceGAN are, hence, defined as:
\begin{equation}
\mathcal{L}_M = ||\min(S_{{skin}}(I) - S_{tar}(I_G), 0) \odot (I_G - I)||_2^2, \ \text{and}
\end{equation}
\begin{equation}
\mathcal{L}_S = ||\max(S_{tar}(I) + S_{tar}(I_G), 1) - S_{tar}(I_G)||_2^2,
\end{equation}
where $\min$ and $\max$ denote pixel--wise minimum and maximum operations. The impact of these constraints on the appearance of a few sample images is shown in Fig. \ref{fig:dynamic_masking}.

\textbf{Final Objective.} The overall optimization objective ($\mathcal{L}_{w}$) of MaskFaceGAN 
is defined as a linear combination of the objectives/constraints described above, i.e.:
\begin{align}
        \mathcal{L}_{w} = \lambda_M \mathcal{L}_M+ \lambda_C \mathcal{L}_C + \lambda_S \mathcal{L}_S + \lambda_P \mathcal{L}_P,
    \label{eq:final}
\end{align}
where $\lambda_M$, $\lambda_C$, $\lambda_S$ and $\lambda_P$ are weighting factors the control the contribution of the individual objectives. Minimizing $\mathcal{L}_{w}$ leads to an optimized latent code $w$ with respect to the targeted semantics $a$ that can be used to generate a (intermediate) synthetic attribute edited image $I_G$. 

\renewcommand\miniwidth{.2\columnwidth}
\begin{figure}[t]
\centering
    \captionsetup[subfigure]{labelformat=empty}
    \begin{subfigure}{\miniwidth}
    		\centering
    		\includegraphics[width=\textwidth]{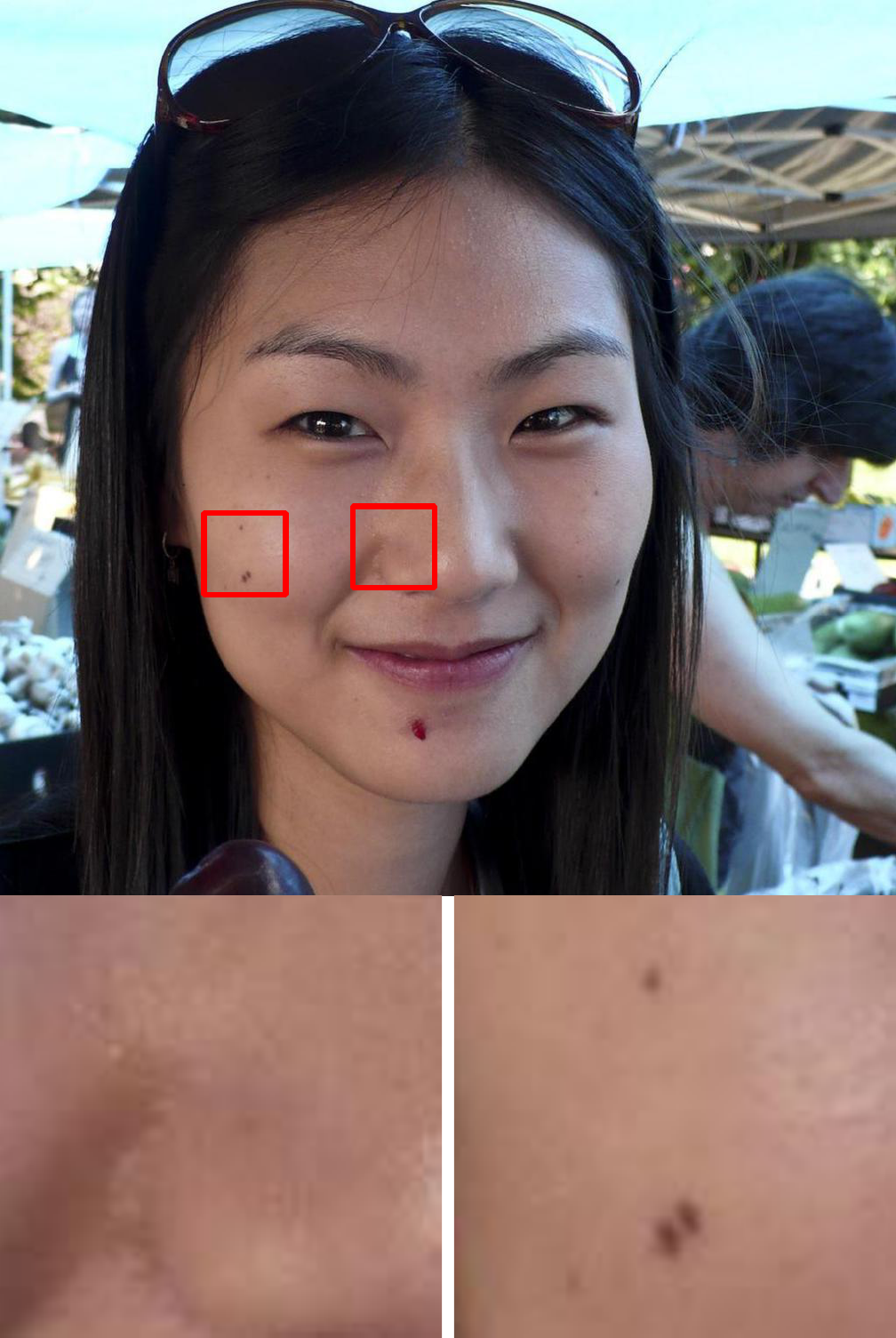}
    		\caption{Original}
    	\end{subfigure}
    \begin{subfigure}{\miniwidth}
    		\centering
    		\includegraphics[width=\textwidth]{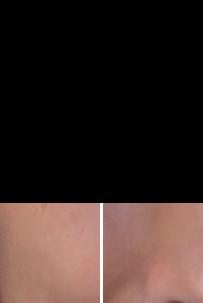}
    		\caption{$n=0$}
    	\end{subfigure}
    	    \begin{subfigure}{\miniwidth}
    		\centering
    		\includegraphics[width=\textwidth]{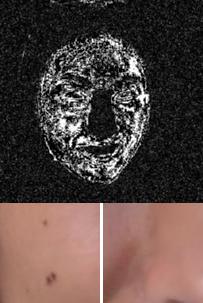}
    		\caption{$\lambda_R=0$}
    	\end{subfigure}
    	    \begin{subfigure}{\miniwidth}
   		\centering
  		\includegraphics[width=\textwidth]{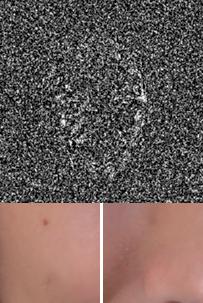}
		\caption{$\lambda_R \neq 0$}
		\end{subfigure}
	\caption{Embedding quality with respect to different noise optimization settings when editing a nose-region attribute. The first column shows the original image and a closeup of two face regions. Without noise optimization high-frequency details are not embedded, but smooth transitions are generated between the skin and the nose regions (second columns). Optimizing for the noise directly perfectly embeds fine skin details, but does not produce smooth transitions (third column), while the regularized optimization generates a reasonable trade--off between details and transitions (last column).}
	\label{fig:noise}
\end{figure}

\begin{table*}[!t!]
\centering
\caption{High--level summary of the dataset and experimental setup used with MaskFaceGAN. Note that datasets with different characteristics and diverse face images were selected for the experiments to demonstrate the merits of the proposed approach. The number of test images reported is used for the quantitative evaluations, e.g., the user study. 
}
    \label{tab:datasets}
\renewcommand{\arraystretch}{1.2}
\begin{tabular}{lccccr}
\hline\hline
\textbf{Dataset} & \textbf{Image Resolution} & \textbf{Purpose}  & \textbf{\#Training Images}$^\dagger$
& \textbf{\#Test Images} & \textbf{Variability Sources}\\ \hline
FFHQ~\cite{karras2020analyzing} & $1024\times 1024$ & Training of $G$  & $70,000$ & n/a & Age, ethnicity, background \\
CelebA~\cite{liu2015faceattributes} & $178\times 218$ & Training of $C$ & $182,636$ & n/a & Age, ethnicity, background,accessories\\
CelebA–HQ~\cite{CelebAMask-HQ}  & $1024\times 1024$ & Training of $S$, testing & $24 ,183$ & $100$ & Age, ethnicity, background, accessories\\
Helen \cite{le2012interactive} & $>500$ in width & Testing & n/a & 118 & Age, ethnicity, background clutter \\
SiblingsDB–HQf~\cite{Vieira2014} & $4256\times 2832$ & Testing & n/a & 163 & Age, gender \\
\hline\hline
\multicolumn{6}{l}{\footnotesize $^\dagger$ The number of training images reported includes both training and validation data.\vspace{-3mm}}
\end{tabular}
\end{table*}

\subsection{Noise Component Optimization}
While the semantic content of the edited images is controlled by the latent code $w$, the  high-frequency facial details that ensure photo realism are defined by the noise components $n$. After the latent code $w$ is optimized, $w$ is frozen and MaskFaceGAN proceeds to optimize $n$, similarly to \cite{abdal2020image2stylegan++}. Two key issues are considered during optimization, i.e.: 
\begin{itemize}
    \item \textit{Adversarial solutions:} Due to the high-dimensionality of $n$, a naive optimization procedure based on Eq.~\eqref{eq:noise_total} can lead to editing results akin to adversarial examples, i.e., generated images that satisfy all constraints but do not exhibit the desired semantics. To avoid such settings the objectives related to semantics and target--region shape are not considered when optimizing for $n$.
    \item \textit{Overfitting:} The optimization procedure can lead to solutions that perfectly reproduce all stochastic details of the original face (e.g., freckles, wrinkles) except for facial areas altered by the editing procedure. This mismatch between preserved and altered image regions results in unnatural appearances and a ``copy-paste'' look.  
    To avoid such overfitting and ensure a reasonable amount of details in the preserved as well as generated image regions, a noise regularization term is used, similarly to \cite{karras2020analyzing}. 
\end{itemize}
Based on the above considerations, MaskFaceGAN's noise-related optimization objective  takes the following form:
\begin{equation} \label{eq:noise_total}
    \mathcal{L}_n = \lambda_M \mathcal{L}_M + \lambda_R \sum_{i,j}L_{i,j},
\end{equation}
where $\lambda_M$ and $\lambda_R$ are again weighting factors and the noise regularization term $L_{i,j}$ is defined as:
\begin{align}
\begin{split}
L_{i,j} &= \left(\frac{1}{|n_{i,j}|} \sum_{x,y} n_{i,j}(x,y) \cdot n_{i,j}(x-1,y)\right)^2 \\
 &+ \left(\frac{1}{|n_{i,j}|} \sum_{x,y} n_{i,j}(x,y) \cdot n_{i,j}(x, y-1)\right)^2.
\end{split}
\end{align}
The goal of this regularization term is to ensure that the noise components follows a normal Gaussian probability distribution by preserving the mean and standard deviations of neighbouring values. At every step of the optimization, each noise component larger than $8 \times 8$ is downsampled in a pyramid-like fashion to a resolution of $8 \times 8$ by averaging $2 \times 2$ neighbouring values. In the above equation, $n_{i,j}$, thus, denotes the $i$-th noise component at the original resolution ($j=0$) or a given level of the downsampling pyramid ($j>0$). The number of elements of $n_{i,j}$ is denoted as $|n_{i,j}|$ and the corresponding regularization term as $L_{i,j}$. The impact of the term is illustrated in Fig. \ref{fig:noise}.


\subsection{Blending}
In the final step, MaskFaceGAN blends the image generated based on the optimized latent code $w$ and noise components $n$, $I_G = G(w,n)$, with image regions in the input image $I$ that were not considered during optimization. These regions correspond to the background and non-edited facial components. To facilitate this step, a blending mask is computed as $B = S_{skin}(I) + S_{tar}(I)$ for most attributes and $B = S_{skin}(I) + S_{tar}(I_G)$ when editing hair shape to account for potential modification of the hair region. The final image $I'$ is generated as
\begin{equation} \label{eq:blending}
    I' = B \odot I_G + (1-B) \odot I.
\end{equation}
The blending step is visualized in the bottom left of Fig.~\ref{fig:skica}. Note that the blending operation considers the skin region from the optimized image $I_G$. This is needed to allow MaskFaceGAN to also change the size and shape of the targeted attributes in a visually convincing manner, with smooth transitions and without visual artefacts.

\section{Experimental Setup}
\label{sec:ExperimentalSetup}


\subsection{Datasets and Experimental Splits}
\label{sec:datasets}
Five high-resolution images datasets are used in the experiment with MaskFaceGAN, i.e., Flickr-Faces-HQ (FFHQ) \cite{karras2020analyzing}, CelebA \cite{liu2015faceattributes}, CelebA--HQ \cite{CelebAMask-HQ}, Helen \cite{le2012interactive} and SiblingsDB--HQf~\cite{Vieira2014}. The datasets were selected 
based on different criteria, such as dataset size, image resolution, image quality, and available annotations. A brief summary of the datasets and experimental splits used is given below:    
\begin{itemize}
    \item \textbf{Flickr-Faces-HQ (FFHQ)} \cite{karras2020analyzing} contains $70,000$ high quality face images at a resolution of $1024 \times 1024$ pixels. Images from the dataset were crawled from Flickr and contain considerable variation in terms of age, ethnicity and image background. FFHQ is used to the train the generator model ($G$) of MaskFaceGAN.
    \item  \textbf{CelebA} \cite{liu2015faceattributes} is a large--scale face image dataset, consisting of more than $200,000$ celebrity images. Each image is annotated with identity information, $40$ binary attributes and $5$ landmark locations. CelebA is used to train the attribute classifier ($C$) of MaskFaceGAN in accordance with the  official training and validation splits~\cite{liu2015faceattributes}.
    \item \textbf{CelebA--HQ} \cite{karras2018progressive} is a recent dataset, derived from  CelebA. It consists of $30,000$ aligned, quality-improved images processed by JPEG artefact removal and  super-resolution. For the experiments, the CelebAMask--HQ\footnote{We use the \textit{CelebA--HQ} to refer to this dataset hereafter for brevity.} version from~\cite{CelebAMask-HQ} is utilized,  which comes with pixel-level annotations of semantic face classes. 
    CelebA--HQ is used to train the face parser ($S$) and to evaluate the performance of MaskFaceGAN. 
    Training and validation sets are defined by mapping the predefined experimental splits of CelebA to CelebA--HQ. For the quantitative evaluations (user study), a disjoint set of images is selected based on perceived face quality and data diversity.
    \item \textbf{Helen} \cite{le2012interactive} consists of $2330$ high-quality 
    face images annotated with facial landmark locations. In comparison to CelebA-HQ, there are considerably larger variations in age, race and lightning conditions present in this dataset, posing a greater challenge to face editing technology. A subset of test images is selected for the quantitative evaluations of MaskFaceGAN based on similar criteria as discussed above for CelebA-HQ. The selected test images are manually annotated with binary attributes.
    \item \textbf{SiblingsDB-HQf}~\cite{Vieira2014} contains frontal, expressionless images of $184$ subjects -- $92$ sibling pairs with a resolution of $4256 \times 2832$. Images in this dataset were captured in front of a  uniform background and under controlled lightning.  We process the images with the CelebA--HQ pipeline using cropping and alignment~\cite{karras2018progressive}, then manually annotate them with binary attributes. 
\end{itemize}

\setlength{\tabcolsep}{0.1mm} 
\renewcommand{\arraystretch}{0.3}  

\renewcommand{\miniwidth}{1.6cm}
\newcolumntype{C}{ >{\centering\arraybackslash} m{\miniwidth} }

\renewcommand{\miniwid}{0.2cm}
\newcolumntype{D}{ >{\centering\arraybackslash} m{\miniwid} }

\renewcommand{\figureimagewidth}{.08\textwidth}
\renewcommand{\im}[1]{ \includegraphics[width=\figureimagewidth]{#1}}

\begin{figure*}[!t]
\begin{subfigure}{\textwidth}
\begin{tabular}{DCCCCCCCCCCC}
 & \scriptsize{Original} & \scriptsize{Arched eyeb.} & \scriptsize{\textit{Big nose}} & \scriptsize{Black hair} & \scriptsize{Blond hair} & \scriptsize{Grey hair} & \scriptsize{Mth. sl. open} & \scriptsize{Narrow eyes} & \scriptsize{Smiling} & \scriptsize{Straight hair} & \scriptsize{Wearing lipst.}\\ [1.2mm] 
 
 \begin{turn}{90}
 \scriptsize{StarGAN}
 \end{turn}& \im{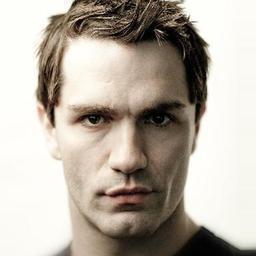} & \im{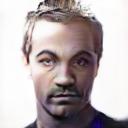} & \im{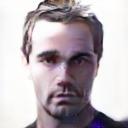} & \im{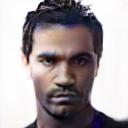} & \im{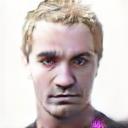} & \im{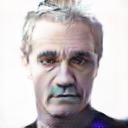} & \im{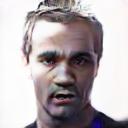} & \im{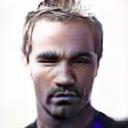} & \im{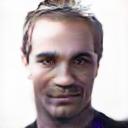} & \im{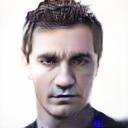} & \im{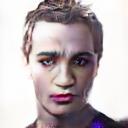}\\ 
 \begin{turn}{90}\scriptsize{AttGAN} \end{turn} & \im{images/all_edits/originals/1815.jpg} & \im{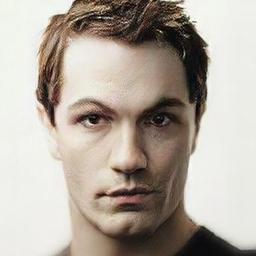} & \im{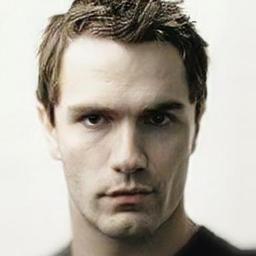} & \im{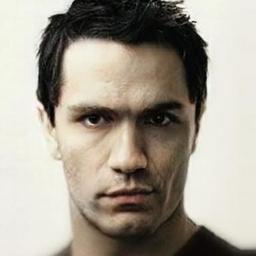} & \im{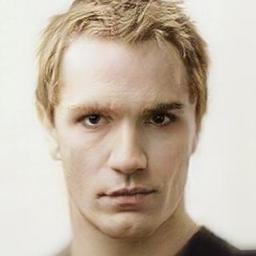} & \im{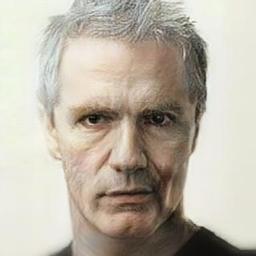} & \im{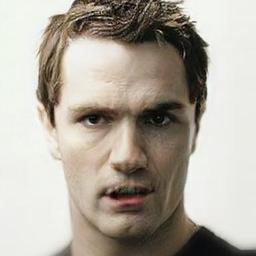} & \im{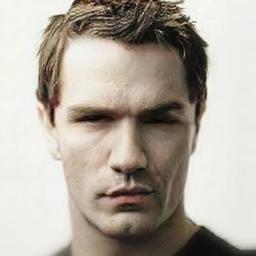} & \im{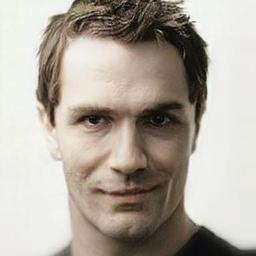} & \im{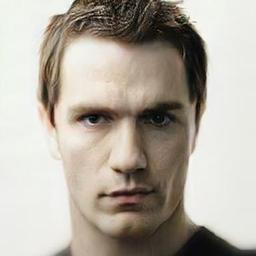} & \im{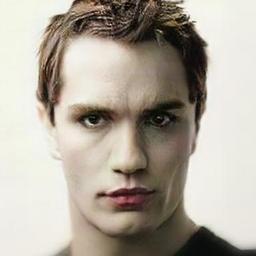}\\ 
 \begin{turn}{90}\scriptsize{STGAN} \end{turn} & \im{images/all_edits/originals/1815.jpg} & \im{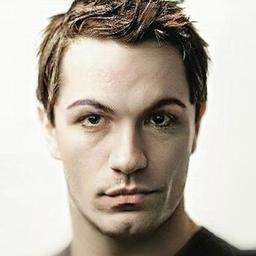} & \im{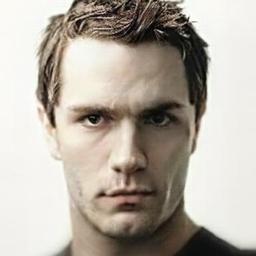} & \im{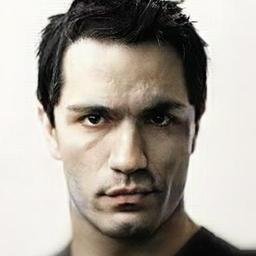} & \im{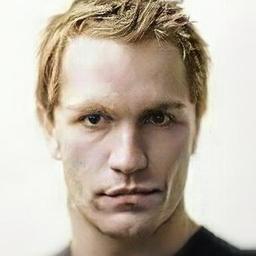} & \im{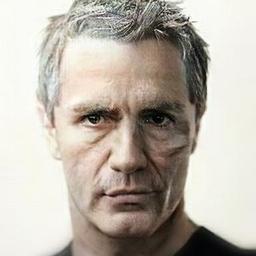} & \im{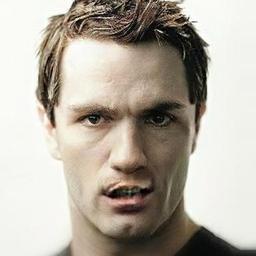} & \im{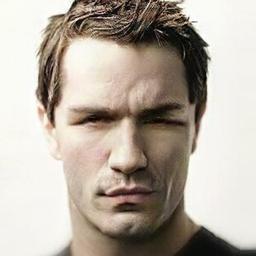} & \im{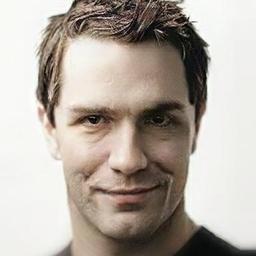} & \im{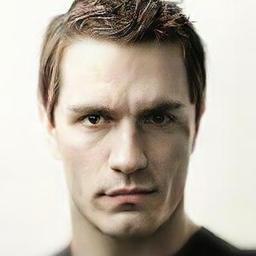} & \im{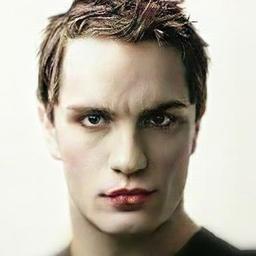}\\ 
 \begin{turn}{90}\scriptsize{InterFace} \end{turn} & \im{images/all_edits/originals/1815.jpg} & \im{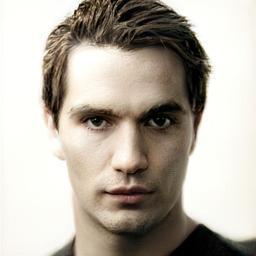} & \im{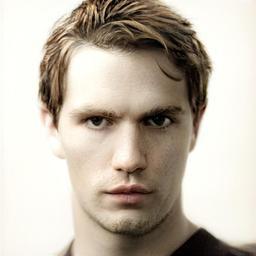} & \im{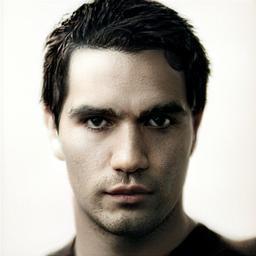} & \im{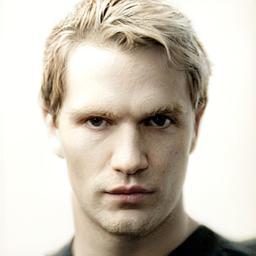} & \im{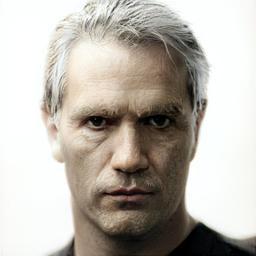} & \im{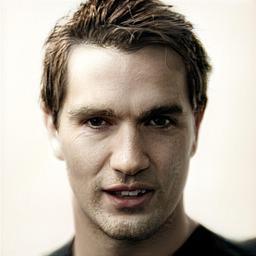} & \im{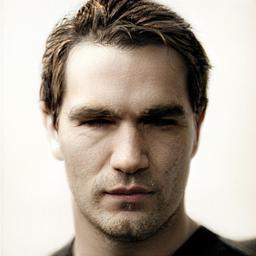} & \im{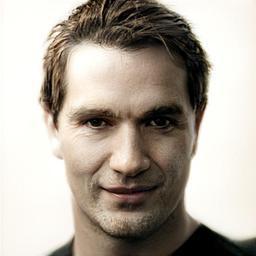} & \im{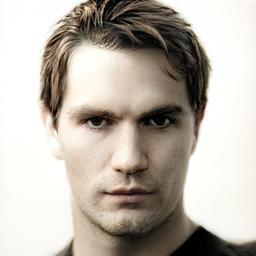} & \im{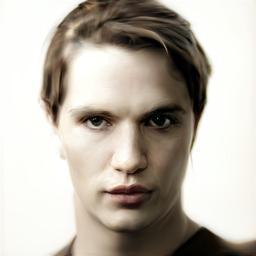}\\ 
 \begin{turn}{90}\scriptsize{InterFace-D} \end{turn} & \im{images/all_edits/originals/1815.jpg} & \im{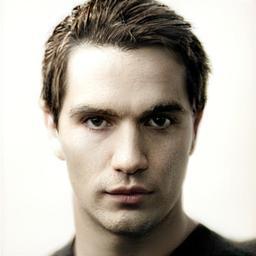} & \im{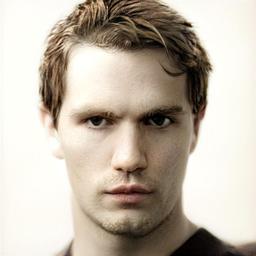} & \im{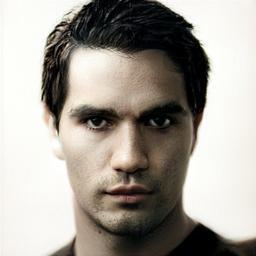} & \im{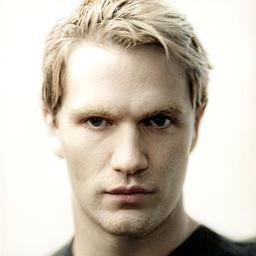} & \im{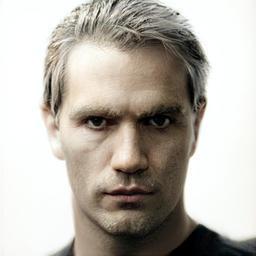} & \im{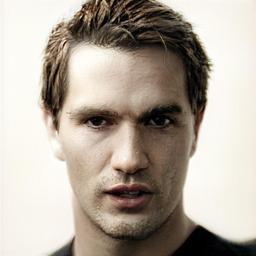} & \im{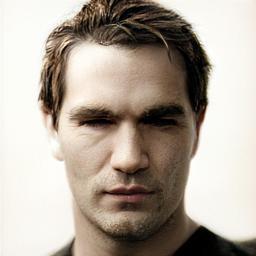} & \im{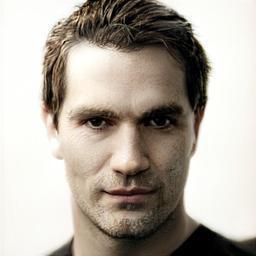} & \im{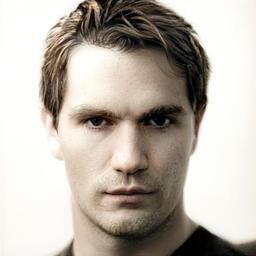} & \im{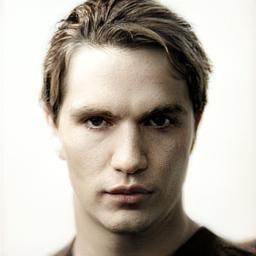}\\ 
 \begin{turn}{90}\scriptsize{MaskFaceGAN} \end{turn} & \im{images/all_edits/originals/1815.jpg} & \im{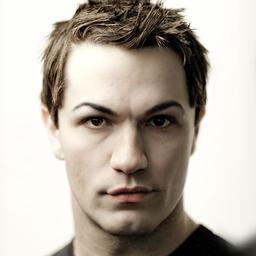} & \im{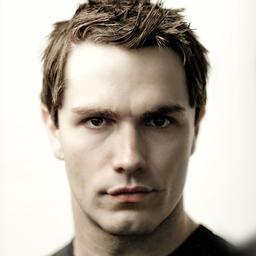} & \im{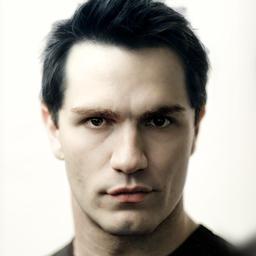} & \im{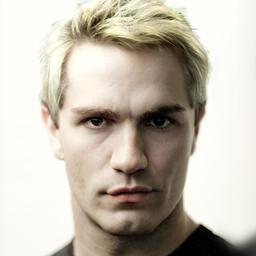} & \im{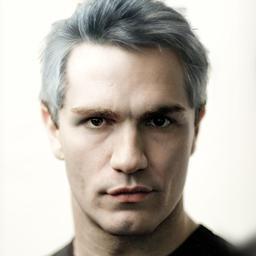} & \im{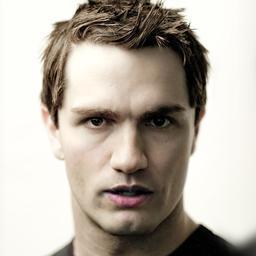} & \im{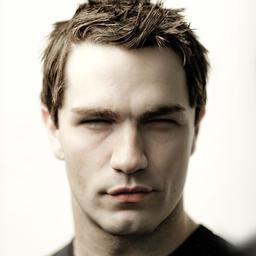} & \im{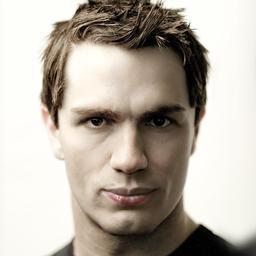} & \im{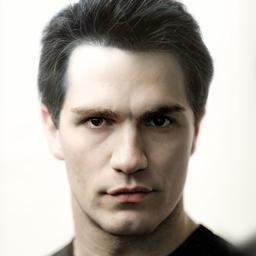} & \im{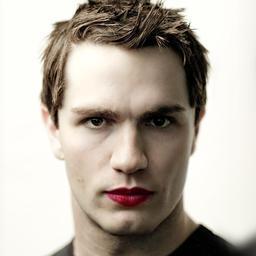}\\ 
\end{tabular}
\label{fig:all_edits_celebahq}
\end{subfigure}
\begin{subfigure}{\textwidth}
\begin{tabular}{DCCCCCCCCCCC}
 & \scriptsize{Original} & \scriptsize{\textit{Bushy eyeb.}} & \scriptsize{Blond hair} & \scriptsize{Brown hair} & \scriptsize{Grey hair} & \scriptsize{Mth. sl. open} & \scriptsize{Narrow eyes} & \scriptsize{Pointy nose} & \scriptsize{\textit{Smiling}} & \scriptsize{Wavy hair} & \scriptsize{Wearing lipst.}\\[1.2mm] 
 \begin{turn}{90}\scriptsize{StarGAN}\end{turn} & \im{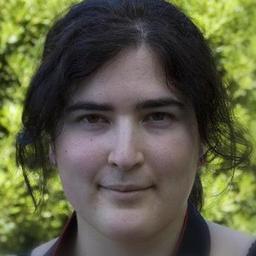} & \im{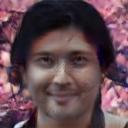} & \im{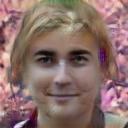} & \im{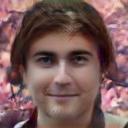} & \im{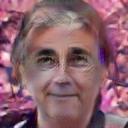} & \im{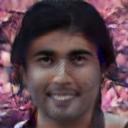} & \im{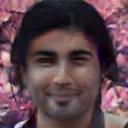} & \im{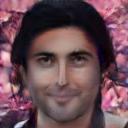} & \im{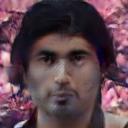} & \im{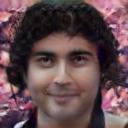} & \im{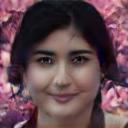}\\ 
 \begin{turn}{90}\scriptsize{AttGAN}\end{turn}  & \im{images/all_edits/originals/2642751678_1.jpg} & \im{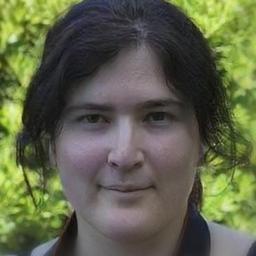} & \im{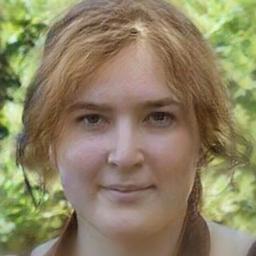} & \im{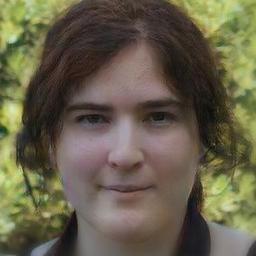} & \im{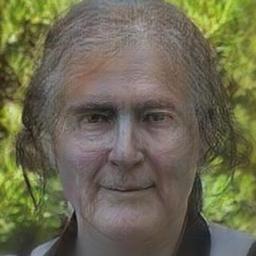} & \im{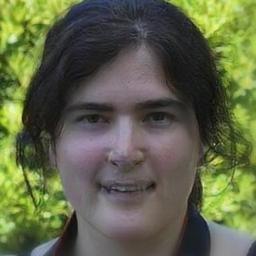} & \im{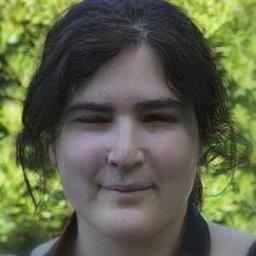} & \im{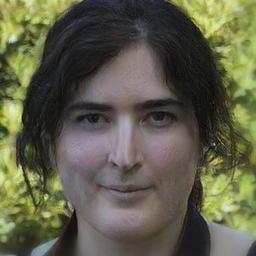} & \im{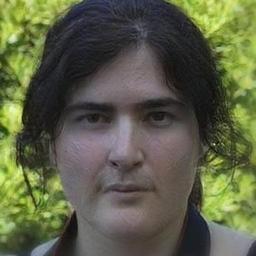} & \im{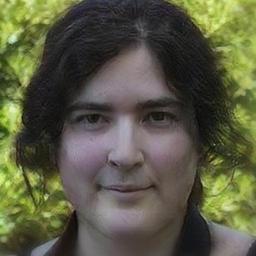} & \im{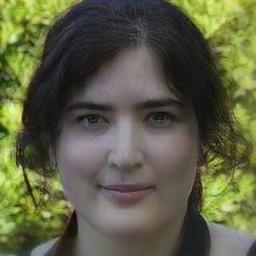}\\ 
 \begin{turn}{90}\scriptsize{STGAN}\end{turn}  & \im{images/all_edits/originals/2642751678_1.jpg} & \im{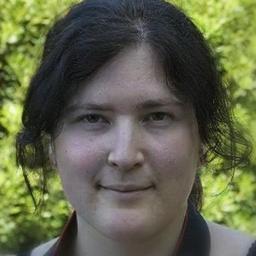} & \im{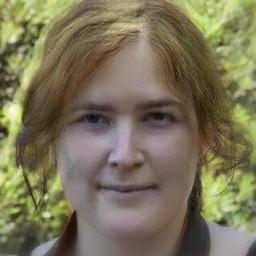} & \im{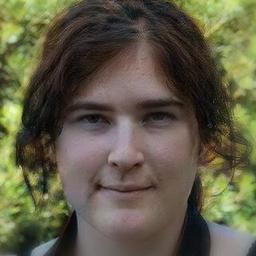} & \im{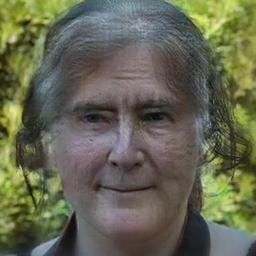} & \im{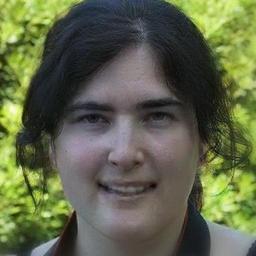} & \im{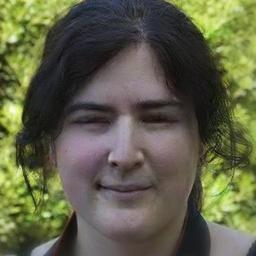} & \im{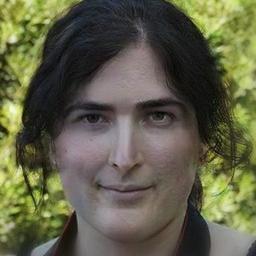} & \im{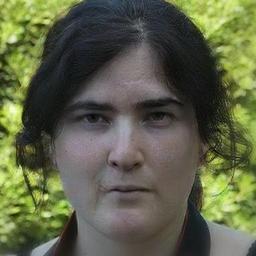} & \im{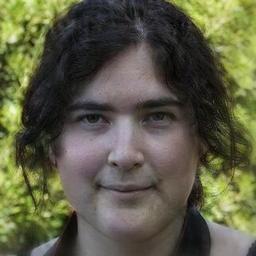} & \im{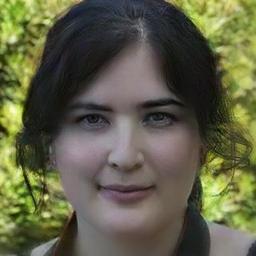}\\ 
 \begin{turn}{90}\scriptsize{InterFace}\end{turn}  & \im{images/all_edits/originals/2642751678_1.jpg} & \im{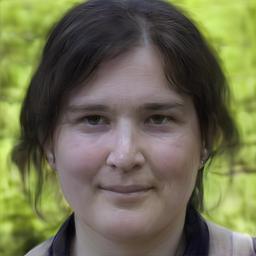} & \im{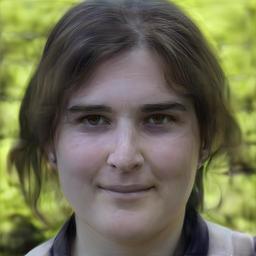} & \im{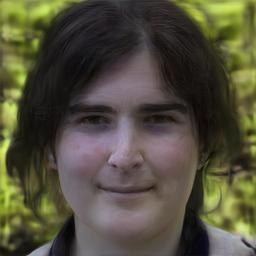} & \im{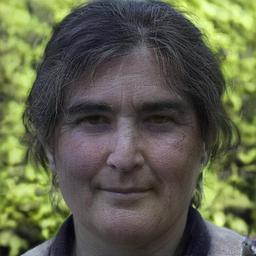} & \im{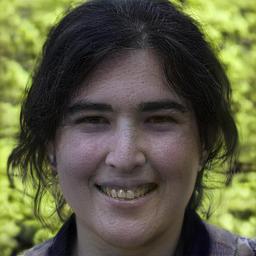} & \im{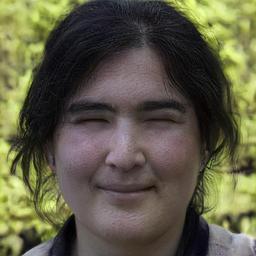} & \im{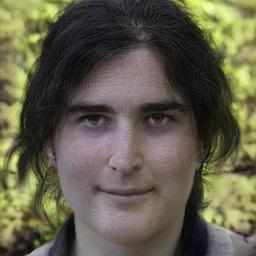} & \im{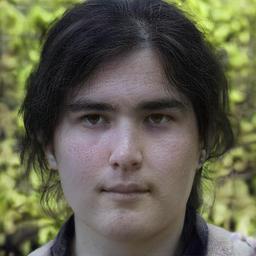} & \im{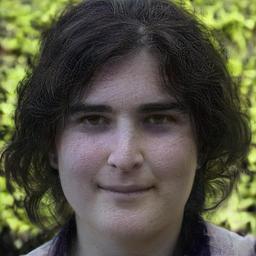} & \im{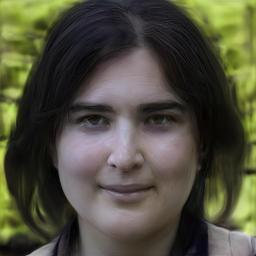}\\ 
 \begin{turn}{90}\scriptsize{InterFace-D}\end{turn}  & \im{images/all_edits/originals/2642751678_1.jpg} & \im{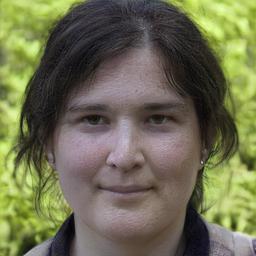} & \im{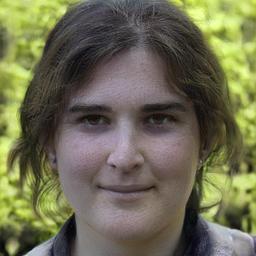} & \im{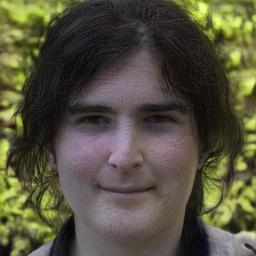} & \im{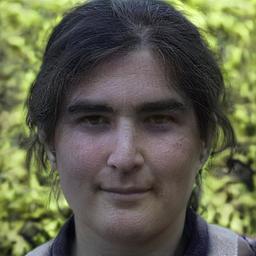} & \im{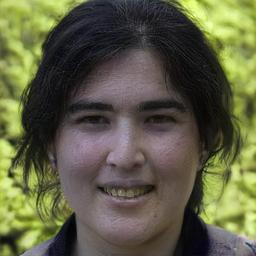} & \im{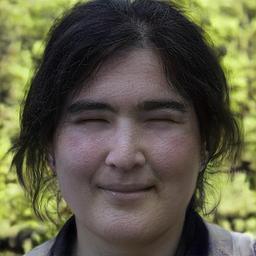} & \im{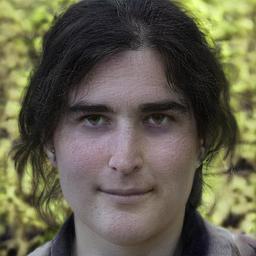} & \im{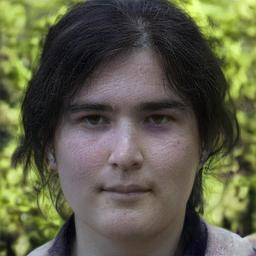} & \im{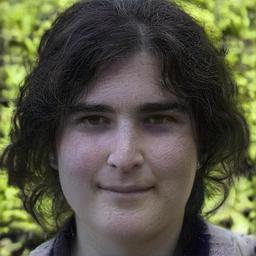} & \im{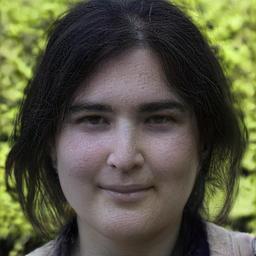}\\ 
 \begin{turn}{90}\scriptsize{MaskFaceGAN}\end{turn}  & \im{images/all_edits/originals/2642751678_1.jpg} & \im{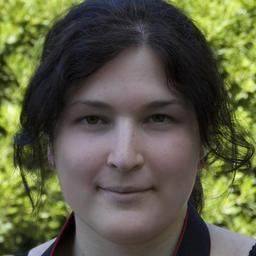} & \im{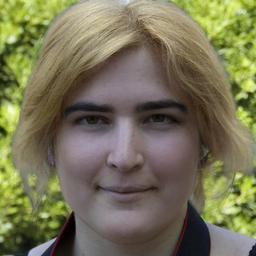} & \im{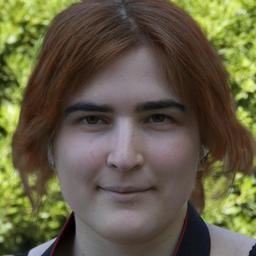} & \im{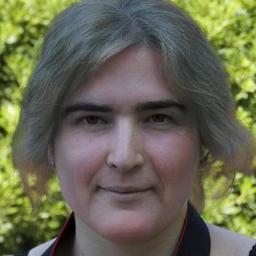} & \im{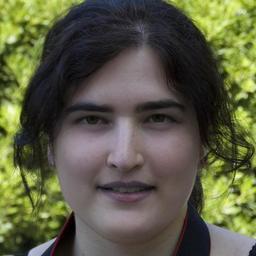} & \im{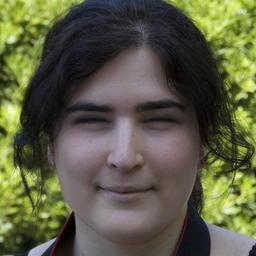} & \im{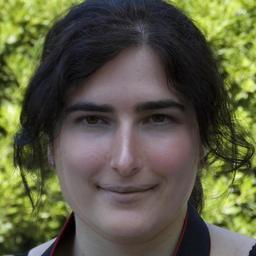} & \im{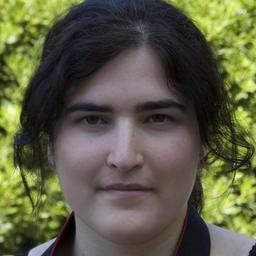} & \im{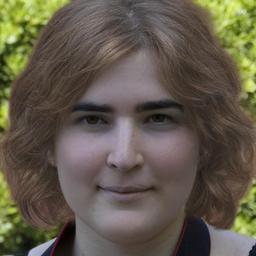} & \im{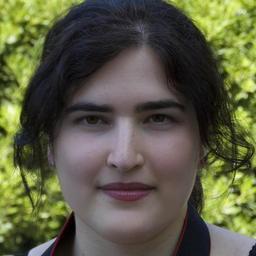}\\ 
\end{tabular}
\label{fig:all_edits_helen}
\end{subfigure}
\caption{Comparison of MaskFaceGAN and five state--of--the--art attribute editing models from the literature. Editing results are presented for $14$ distinct facial attributes with spatial correspondences. For attributes already present in the image, editing inverts the result (e.g., removes the lipstick for "wearing lipstick" if it is already there) - displayed in italic. Results on the top correspond to a sample image from CelebA--HQ and results at the bottom to an image from Helen. Best viewed zoom in.\vspace{-3mm}}
\label{fig:all_edits_celebahq_helen}
\end{figure*}

MaskFaceGAN is applied for attribute editing at a resolution of $1024\times 1024$ pixels with all experimental datasets. The facial images are, therefore, rescaled where necessary before applying the proposed face editing approach. A high--level summary of the datasets and experimental splits is presented in Table~\ref{tab:datasets}. The reported number of test image  corresponds to the amount of face imagery used for the quantitative evaluations.

\newcommand{\imrow}[1]{\im{#1}{orig-img} & \im{#1}{img-w-0000} & \im{#1}{img-w-0200} & \im{#1}{img-w-0400} & \im{#1}{img-w-0600} & \im{#1}{img-w-0800} & \im{#1}{img-n-0000} & \im{#1}{img-n-0200} & \im{#1}{img-n-0400} & \im{#1}{img-n-final} & \im{#1}{blend-mask} & \im{#1}{result}}

\subsection{Implementation details}
The models used by MaskFaceGAN are implemented with publicly available source code. Further details are given below. 
\begin{itemize}
    \item \textbf{The Generator} ($G$) of MaskFaceGAN is implemented using the official StyleGAN2 release~\cite{karras2020analyzing} to foster reproducibility and ensure a fair comparison with competing techniques from the literature designed around this model. 
    \item \textbf{The Attribute Classifier} ($C$) is implemented based on the model from \cite{vandenhende2019branched} and trained on CelebA \cite{liu2015faceattributes} for $16$ epochs with weighted binary cross entropy 
to account for the class imbalance in the training data. The learning rate is initially set to $0.05$ and decayed to $0.005$ on the $40,000$--th training step. The model is optimized with the Nesterov momentum algorithm \cite{sutskever2013importance} using a batch size of $32$. Augmentations with random horizontal flipping and affine transformations are used to avoid over-fitting. 
\item \textbf{The Face Parser} ($S$) is 
trained on the CelebA-HQ dataset to generate segmentation masks of the following seven classes: ``mouth'', ``eyebrows'', ``eyes'', ``earrings'', ``hair'', ``noise'' and ``skin''. Weighted cross entropy is again used as the learning objective. The model is learned for $5$ epochs using the Adam optimizer \cite{kingma2014adam}  with a  learning rate  of $3 \cdot 10^{-4}$ and a batch size of $24$. Data augmentation includes horizontal flipping and affine transformations.  
\end{itemize}

The latent code optimization procedure is conducted in several stages. First,  $w$ is initialized with the mean latent code $\bar{w}$, computed from $50,000$ sampled codes $w \in \mathcal{W}^+$. Next, it is optimized w.r.t. Eq.~\eqref{eq:mse}, so it approximately corresponds to the target face. This initial step was found to be important for the visual quality of the edited images. Finally, the remaining terms are added to enforce semantic and spatial constraints. Similarly to~\cite{abdal2020image2stylegan++}, the noise component $n$ is set to $0$ and kept constant while optimizing $w$. Once $w$ converges, it is frozen and the noise component $n$ is optimized independently of $w$. 

To identify parameter values that yield visually pleasing editing results, hyper-parameter optimization is used, resulting in weighting factors of $\lambda_M = 2, \lambda_C = 0.005, \lambda_S = 0.5, \lambda_R=1$. We set $\lambda_S=0$ for operations where no hair editing is done. The default value for Eq.~\eqref{eq:classf} is set to $\epsilon = 0.05$. The Adam algorithm \cite{kingma2014adam} is again for the optimization process. The learning rate  is set to $0.001$ for the latent code $w$ and to $0.1$ for the noise component $n$. Additional implementation details can be found in the publicly released code of MaskFaceGAN.

\renewcommand{\figuretextwidth}{.02\columnwidth}
\renewcommand{\figureimagewidth}{.23\columnwidth}
\newcolumntype{C}{ >{\centering\arraybackslash} m{\figureimagewidth} }

\renewcommand{\miniwid}{0.4cm}
\newcolumntype{D}{ >{\centering\arraybackslash} m{\miniwid} }

\setlength{\tabcolsep}{8pt} 

\begin{table}[!t]
\renewcommand{\arraystretch}{1.2}
\caption{FID scores produced by the evaluated models on three experimental datasets (lower is better).}
\label{tab:fid}
\centering
\resizebox{\columnwidth}{!}{%
    \begin{tabular}{l|rrr}
    \hline\hline
     \textbf{Method} & \textbf{CelebA--HQ} & \textbf{Helen} & \textbf{SiblingsDB--HQf}\\
     \hline
     {StarGAN} \cite{choi2018stargan} & $140.87$ & $152.64$ & $176.84$\\
     {AttGAN} \cite{he2019attgan} & $67.55$ & $79.90$ & $81.34$\\
     {STGAN} \cite{liu2019stgan}  & $49.53$ & $57.44$ & $48.88$\\
     {InterFaceGAN} \cite{shen2020interpreting,shen2021interfacegan} & $79.79$ & $90.93$ & $83.40$\\
     {InterFaceGAN--D} \cite{shen2020interpreting,shen2021interfacegan} & $76.87$ & $90.90$ & $81.57$\\
     \hline
     {MaskFaceGAN} (ours) & $\boldsymbol{32.56}$ & $\boldsymbol{34.00}$ & $\boldsymbol{34.53}$\\
     \hline\hline
    \end{tabular}
    }
\end{table}
\subsection{Methods}
MaskFaceGAN is evaluated in comparisons with multiple competing models, i.e., StarGAN \cite{choi2018stargan},  AttGAN \cite{he2019attgan}, STGAN \cite{liu2019stgan} and two versions of the InterFaceGAN approach from \cite{shen2020interpreting,shen2021interfacegan}. For a fair comparison, StarGAN, AttGAN and STGAN  are trained on the same attributes as MaskFaceGAN (see Table~\ref{tab:attr2regions}), using the models' official code repository. 
We implement InterFaceGAN \cite{shen2020interpreting} on the StyleGAN2 latent space, following the linear SVM framework. In addition to the vanilla 
InterFaceGAN, the authors of \cite{shen2021interfacegan} also introduced the concept of \textit{conditional manipulation} that tries to disentangle attributes when editing facial images. 
We also consider such type of model in the experiments and denote it as InterFaceGAN-D hereafter. Both InterFaceGAN models edit images by moving latent codes along attribute--dependent latent space directions.  The magnitude of this displacement/movement is set to $1$ based on preliminary experiments.

Note that the implemented models manipulate images at different resolutions, i.e., StarGAN produces edited images of $128 \times 128$ pixels, AttGAN and STGAN generate $384 \times 384$ images, while InterFaceGAN, InterFaceGAN--D and MaskFaceGAN edit images at a resolution of $1024 \times 1024$ pixels. We note again that MaskFaceGAN is specialized towards \textit{local face image editing}, and therefore excels at manipulating facial attributes that can be associated with specific image regions. 

\section{Results and Discussion}
\label{Sec: ResultsDiscussion}

This section reports results that:
$(i)$ compare MaskFaceGAN  to state-of-the-art attribute editing models, $(ii)$ highlight some unique characteristics of the proposed approach, $(iii)$ explore the contribution of various components through an ablation study, $(iv)$ study the impact of the attribute classifier and face parser, $(v)$ provide insight into the optimization procedure and blending, $(vi)$ illustrate MaskFaceGAN's global editing capabilities, and $(vii)$ analyze its limitations. 

\subsection{Comparison to the State-Of-The-Art}

\textbf{Visual Analysis.}
We first demonstrate the capabilities of MaskFaceGAN for the task of single attribute editing and include the $14$ binary attributes from  Table~\ref{tab:attr2regions} in the analysis. Images from different datasets are used for the experiments to explore the generalization capabilities of the evaluated approaches across various data distributions. If a face already exhibits a given attribute (e.g., a face wearing lipstick), we generate \textit{inverted} attributes, (i.e., a face without lipstick).
\setlength{\tabcolsep}{6pt} 
\renewcommand{\arraystretch}{1.3}  
\begin{table}[!t]
\renewcommand{\arraystretch}{1.3}
\caption{User study results, where human raters were shown editing results of all tested models and asked to select the best one. Reported is the fraction of times [in \%] a model was chosen as the overall best (higher is better). 
}

\label{tab:userstudy-best}
\centering
\resizebox{\columnwidth}{!}{%
    \begin{tabular}{l|rrr}
    \hline\hline
     \textbf{Method} & \textbf{CelebA--HQ} & \textbf{Helen} & \textbf{SiblingsDB--HQf}  \\
     \hline
    StarGAN \cite{choi2018stargan} & $2.96 \%$ & $2.76 \%$ & $4.26 \%$\\
AttGAN \cite{he2019attgan} & $4.95 \%$ & $6.53 \%$ & $6.38 \%$\\
STGAN \cite{liu2019stgan} & $13.93 \%$ & $8.70 \%$ & $12.60 \%$\\
InterFaceGAN \cite{shen2020interpreting,shen2021interfacegan} & $7.49 \%$ & $18.62 \%$ & $18.80 \%$\\
InterFaceGAN--D \cite{shen2020interpreting,shen2021interfacegan} & $9.99 \%$ & $14.71 \%$ & $10.82 \%$\\
\hline
MaskFaceGAN (ours) & $\boldsymbol{60.68 \%}$ & $\boldsymbol{48.68 \%}$ & $\boldsymbol{47.14 \%}$\\
\hline\hline
    \end{tabular}
    }

\end{table}

\begin{table}[!t]
\renewcommand{\arraystretch}{1.3}
\caption{User study results, where human raters were asked to rate the quality of the edited images on a $5$-point Lickert scale (higher is better). Reported is the average score and corresponding standard deviation.}
\label{tab:userstudy-scores}
\centering
\resizebox{\columnwidth}{!}{%
    \begin{tabular}{l|rrr}
    \hline\hline
     \textbf{Method} & \textbf{CelebA--HQ} & \textbf{Helen} & \textbf{SiblingsDB--HQf}  \\
     \hline
     StarGAN \cite{choi2018stargan}&     $1.46 \pm 0.92$ & $1.30 \pm 0.63$ & $1.52 \pm 0.89$ \\
     AttGAN \cite{he2019attgan} &      $2.85 \pm 1.01$ & $2.39 \pm 1.06$ & $2.48 \pm 0.90$ \\
     STGAN \cite{liu2019stgan} &       $3.07 \pm 1.13$ & $2.44 \pm 1.12$ & $2.66 \pm 0.98$ \\
     InterFaceGAN \cite{shen2020interpreting,shen2021interfacegan} &   $3.00 \pm 1.03$ & $3.03 \pm 1.18$ & $3.29 \pm 1.02$ \\
     InterFaceGAN--D \cite{shen2020interpreting,shen2021interfacegan} & $2.94 \pm 1.12$ & $2.78 \pm 1.22$ & $3.12 \pm 1.10$ \\
     \hline
     MaskFaceGAN (ours) & $\boldsymbol{4.07 \pm 1.21}$ & $\boldsymbol{3.80 \pm 1.23}$ & $\boldsymbol{3.85 \pm 1.15}$ \\
     \hline\hline
    \end{tabular}
    }
\end{table}

\begin{figure*}[!t!]
    \centering
    \includegraphics[width=0.8\textwidth]{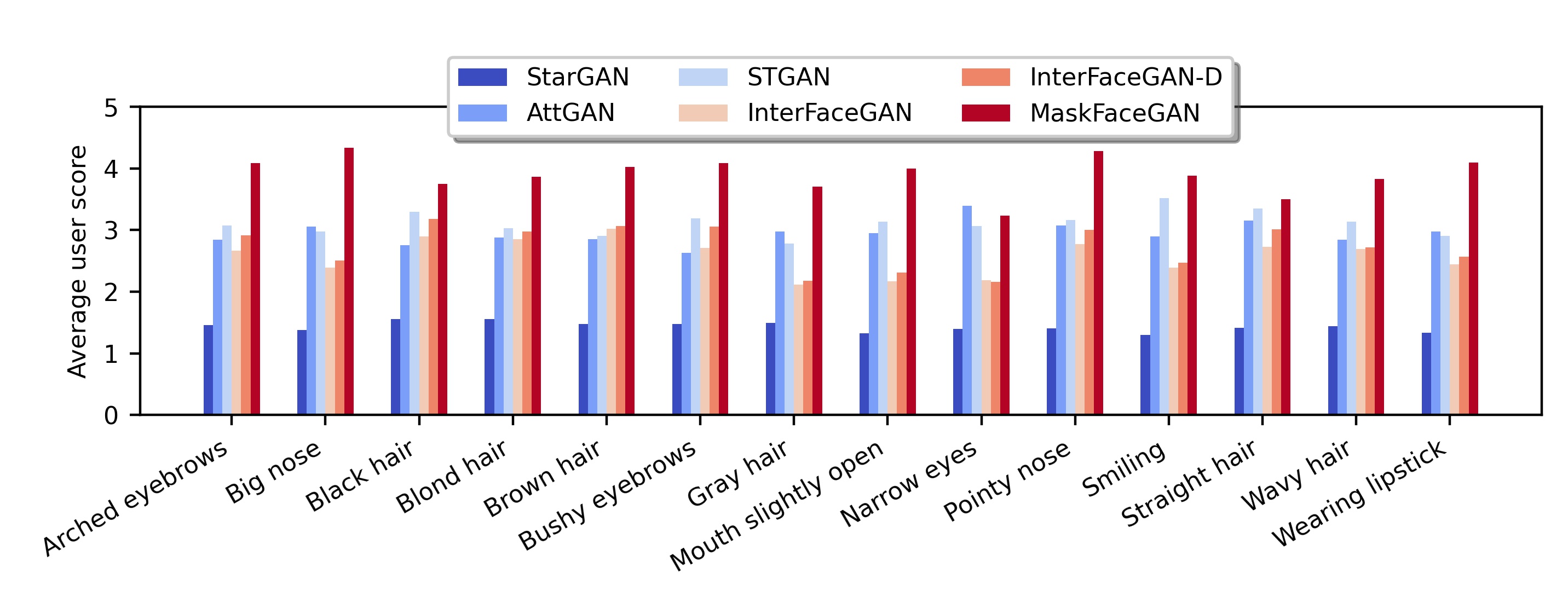}\vspace{-4mm}
    \caption{Comparison of user study scores, averaged across all three dataset for individual attributes. As can be seen, MaskFaceGAN achieves highly competitive results with all targeted attributes. The figure is best viewed in color. 
    \vspace{-4mm}}
    \label{fig:barplot}
\end{figure*}

\renewcommand{\figuretextwidth}{.02\columnwidth}
\setlength{\tabcolsep}{2pt} 
\renewcommand{\arraystretch}{1.3}
\newcommand{\turnedtext}[3]{\begin{turn}{270} \hspace{#1} #2 \hspace{#3} \end{turn}}
\renewcommand{\figureimagewidth}{.235\columnwidth}
\renewcommand{\im}[1]{ \includegraphics[width=\figureimagewidth]{#1}}

\begin{figure}[!t]
 \centering
 \begin{tabular}{CCCC}
 \im{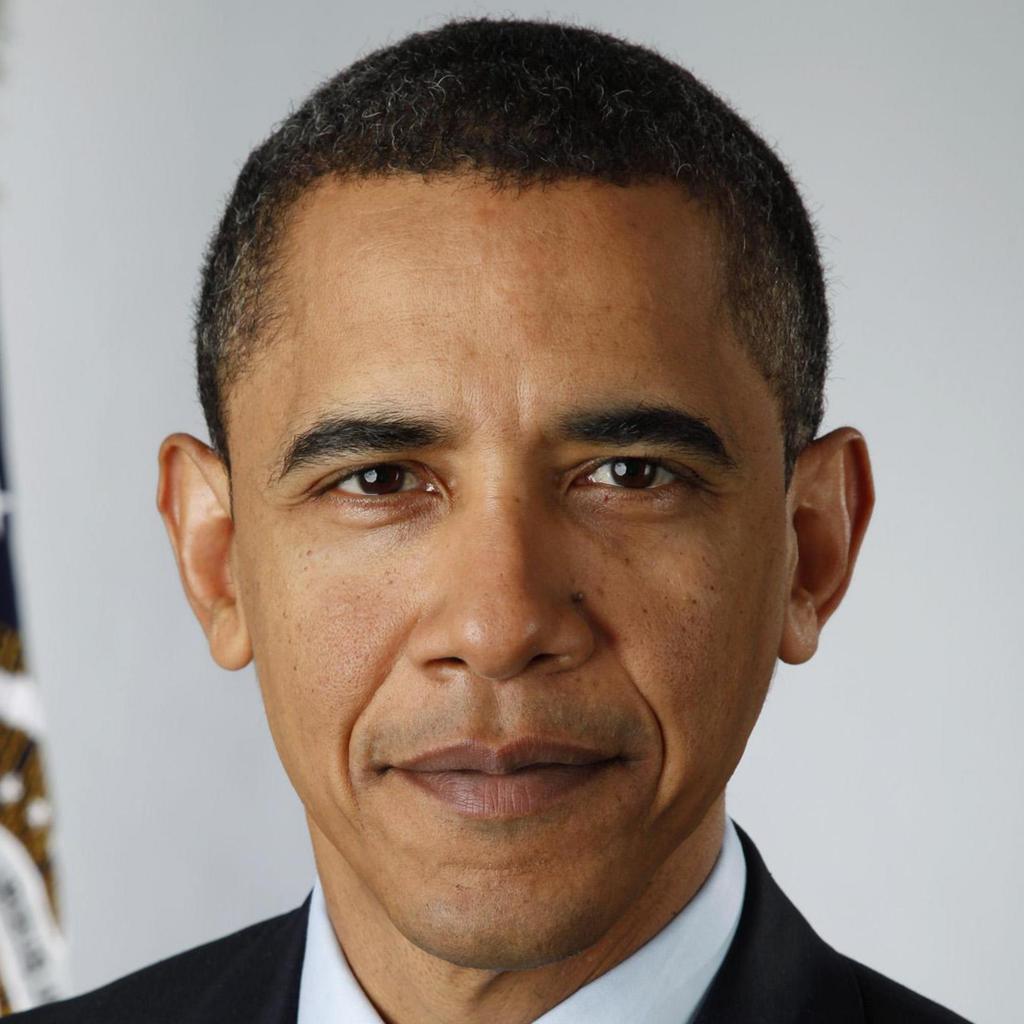} &  \im{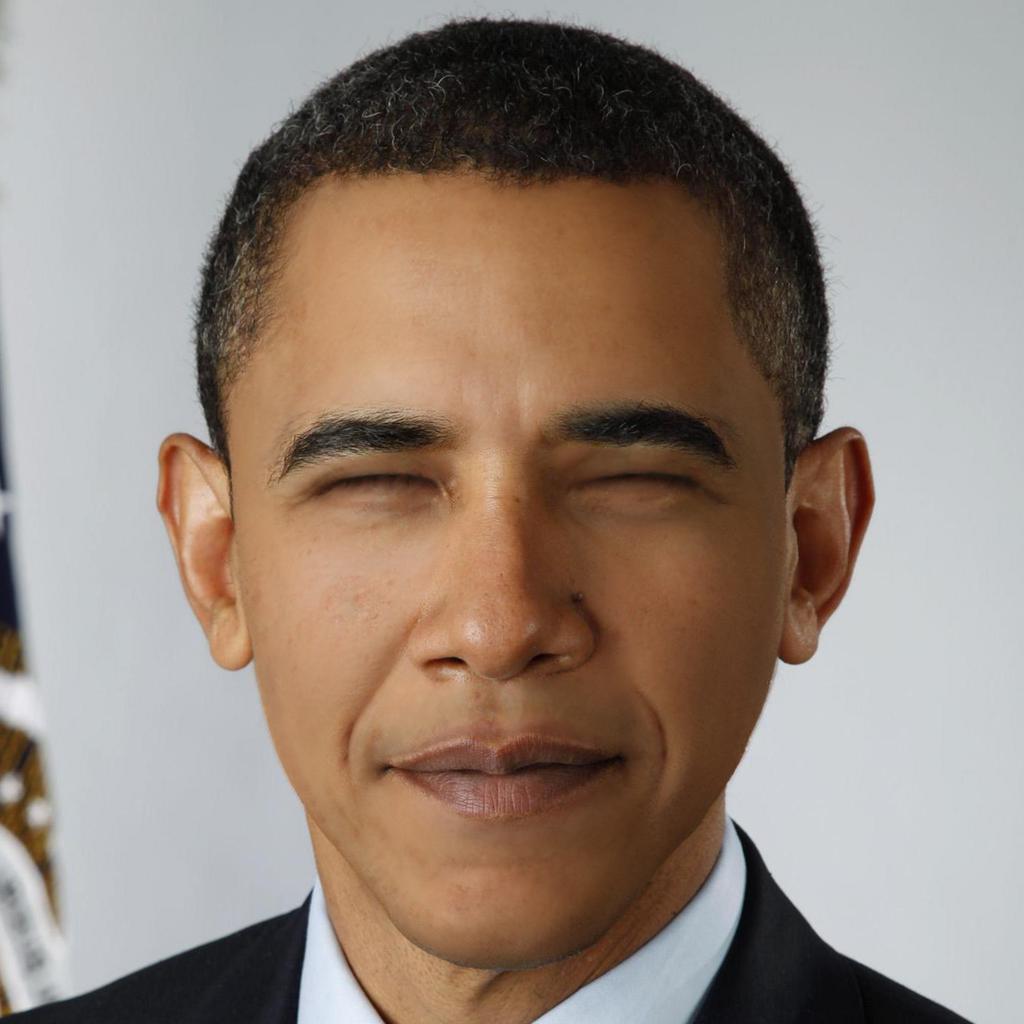} &  \im{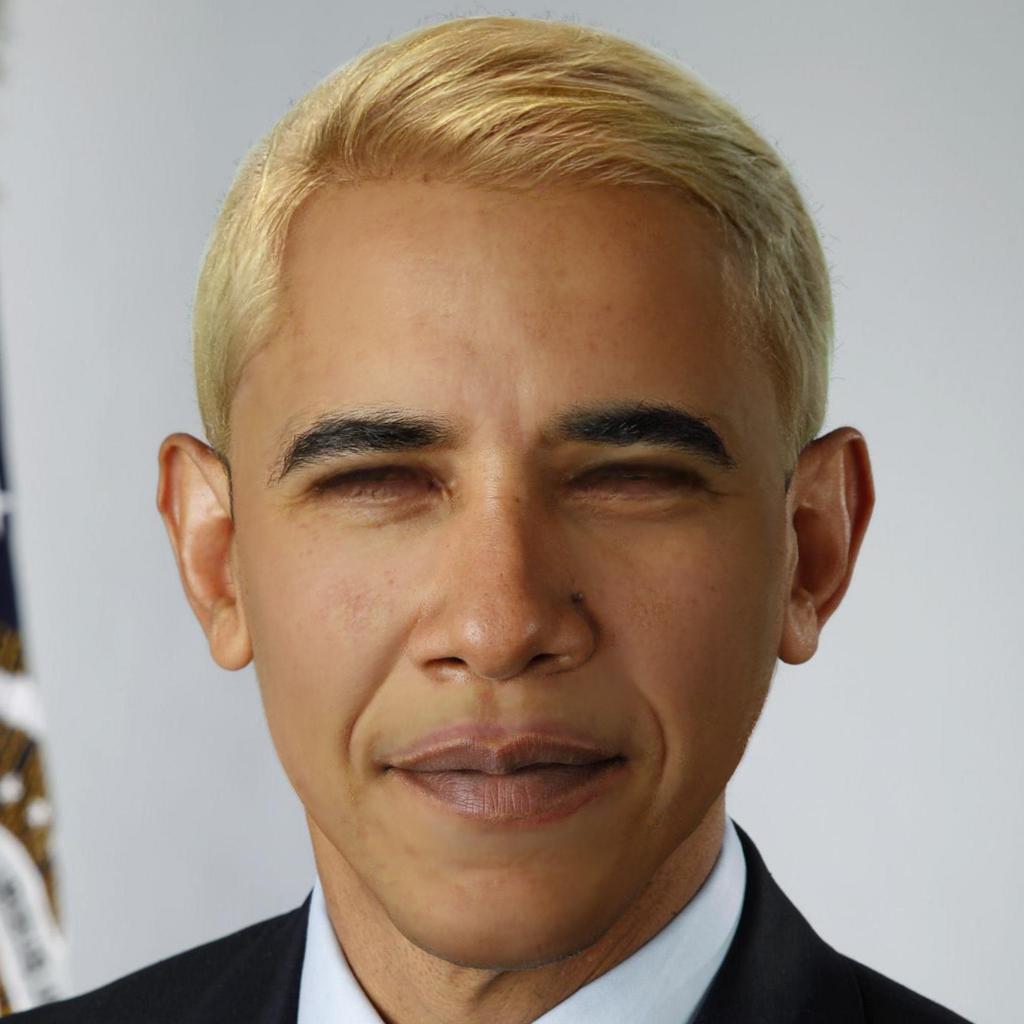} &  \im{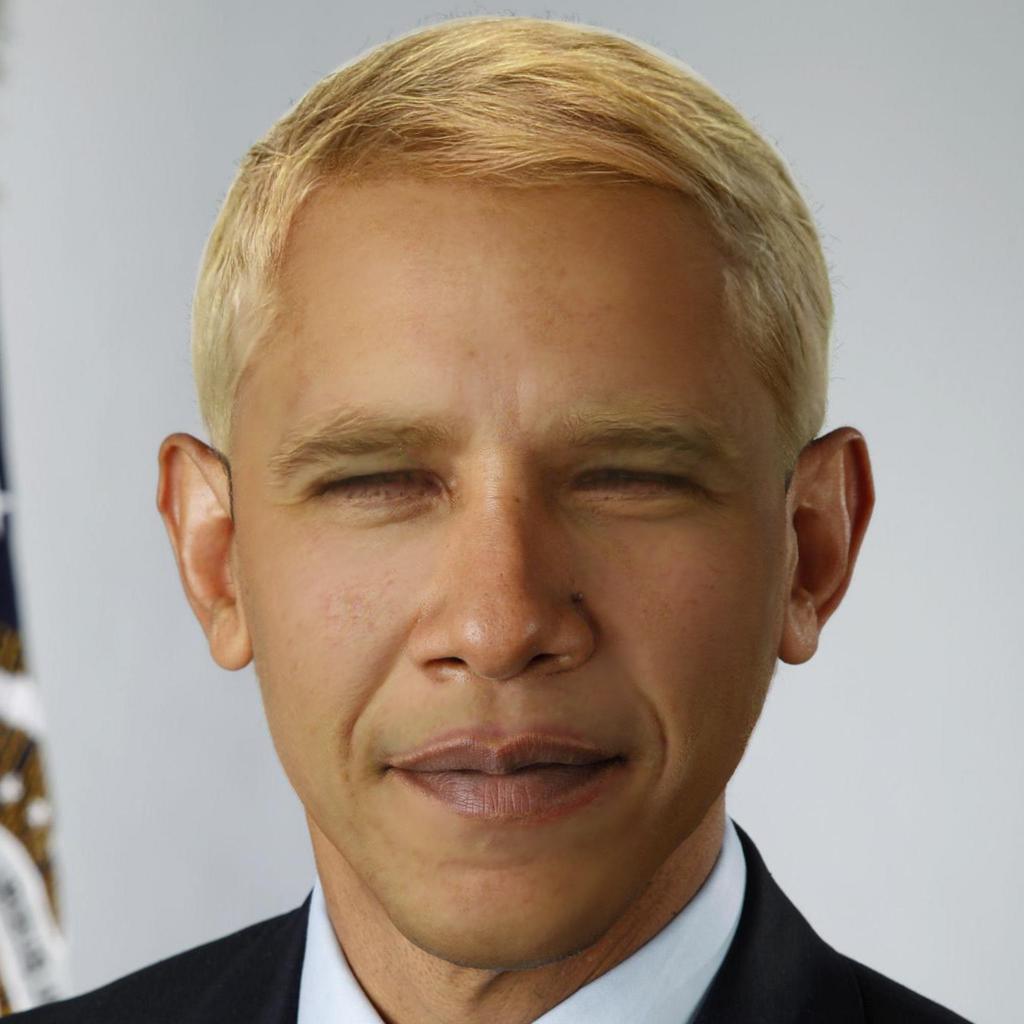} \\
  \im{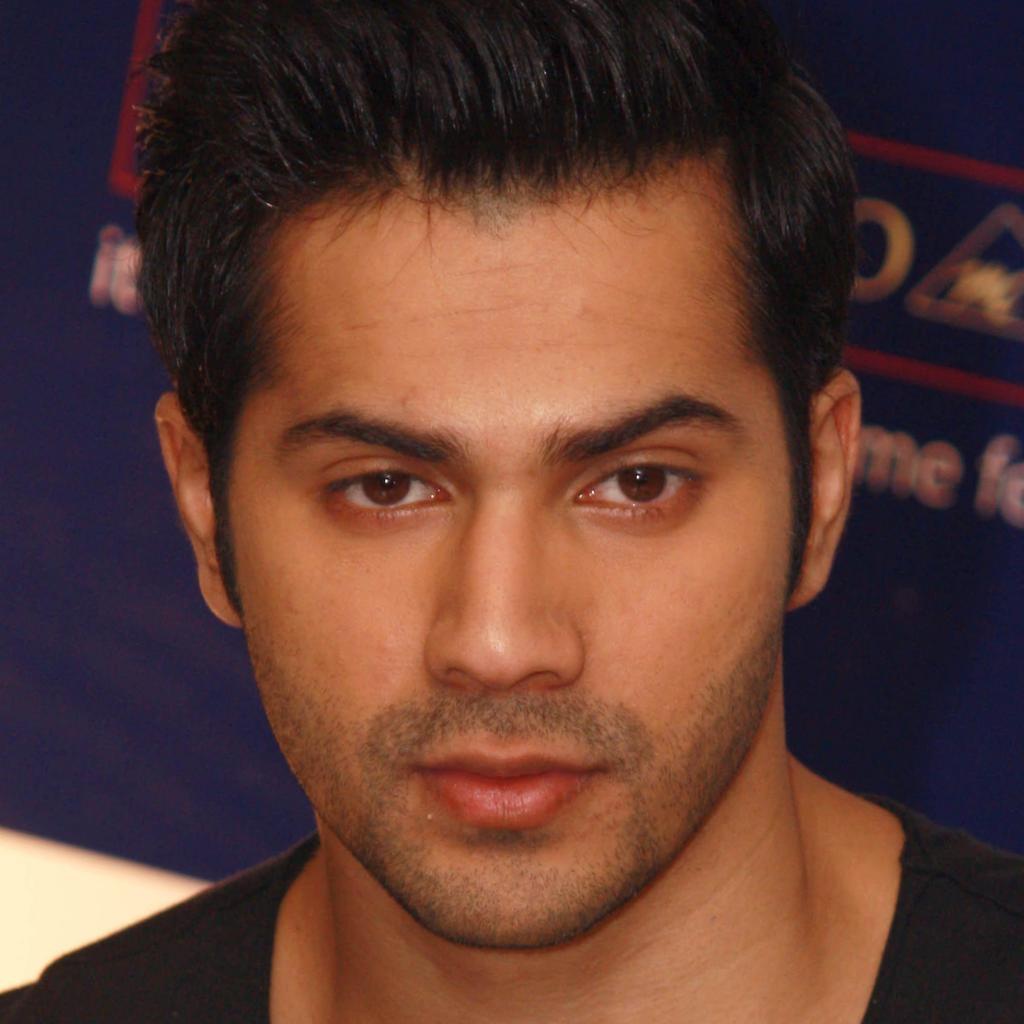} &  \im{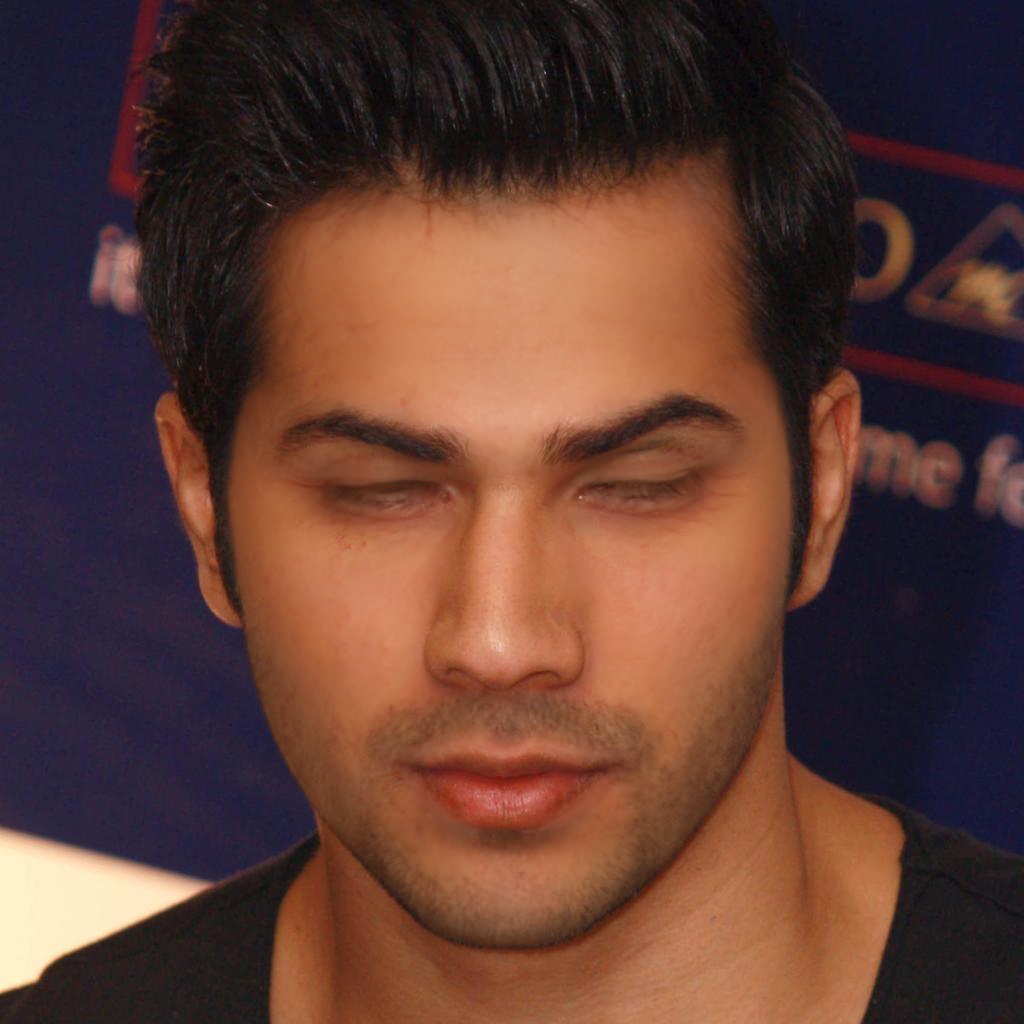} &  \im{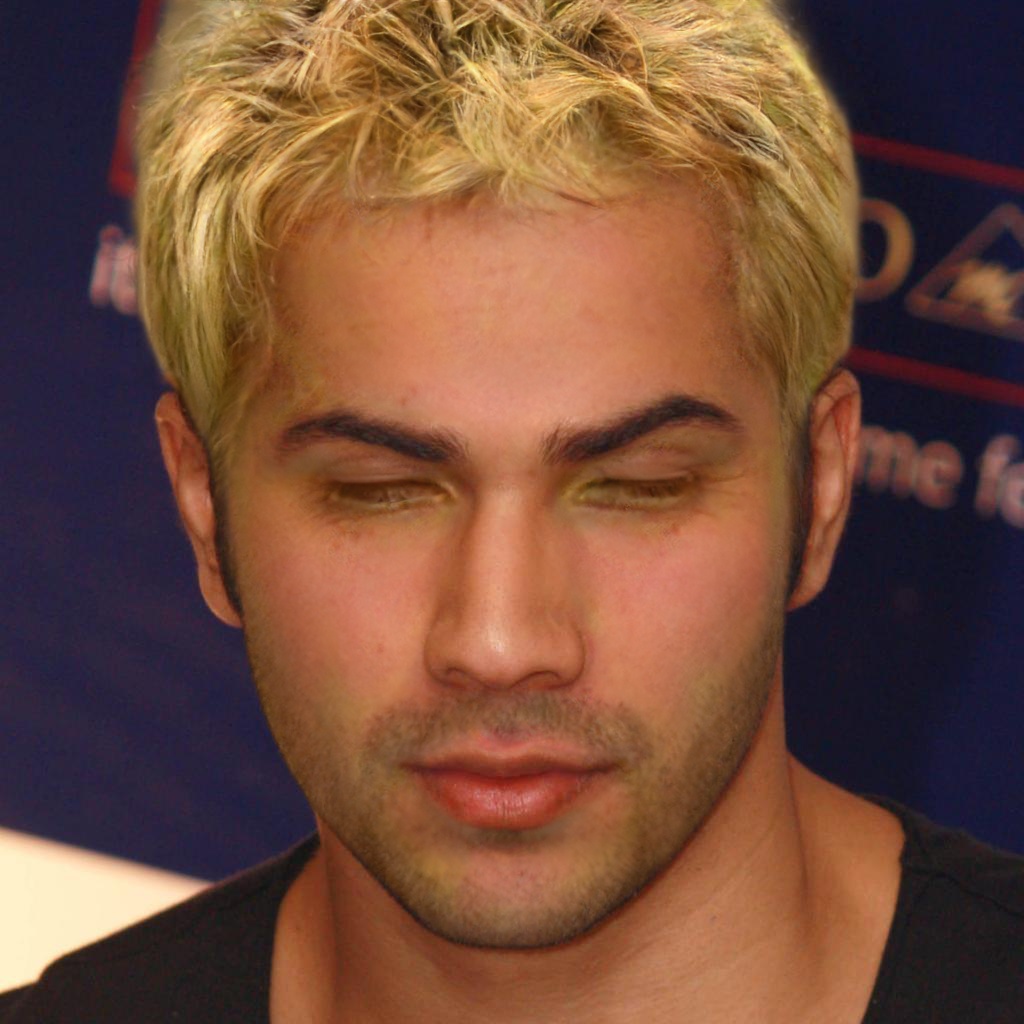} &  \im{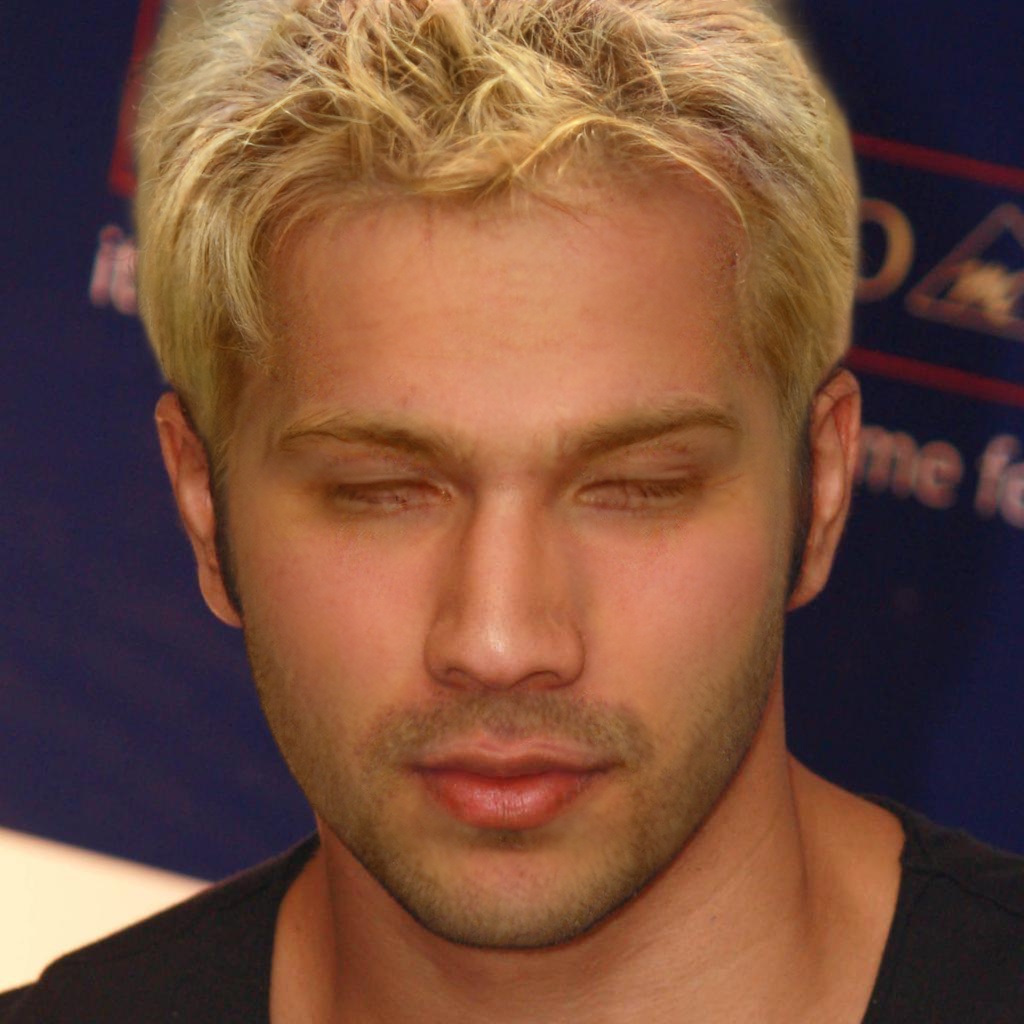} \\\hline
  \footnotesize{Original} & \multicolumn{1}{l}{\footnotesize{+ Narrow eyes}} & \multicolumn{1}{l}{\footnotesize{+ Narrow eyes}} & \multicolumn{1}{l}{\footnotesize{+ Narrow eyes}} \\[-2mm]
 \footnotesize{image} & \footnotesize{} & \multicolumn{1}{l}{\footnotesize{+ Blond hair  }} &\multicolumn{1}{l}{ \footnotesize{+ Blond hair}} \\[-2mm]
 \footnotesize{} & \footnotesize{} & \footnotesize{} & \multicolumn{1}{l}{\footnotesize{- Bushy eyebrows}} \\
 \hline
  \end{tabular}
\caption{Editing multiple attributes with MaskFaceGAN. Every image is the result of a separate optimization procedure and is generated independently from all others.  
\vspace{-2mm}}
\label{fig:multiple_attributes}
\end{figure}

Fig. \ref{fig:all_edits_celebahq_helen} compares editing results produced by MaskFaceGAN and the five competing models on a couple of sample images from CelebA-HQ and Helen. Note that $10$ attributes are considered per example image to ensure a reasonable image size for the presentation. Among the encoder--decoder models, StarGAN generates the highest amount of visual artefacts and also introduces a background change for some of the attributes, as illustrated with the sample image from Helen in Fig. \ref{fig:all_edits_celebahq_helen}. AttGAN and STGAN produce more convincing results, but still induce a certain amount of visual artefacts. These can, for example, be seen with the ``Mouth slightly open'' attribute in Fig. \ref{fig:all_edits_celebahq_helen}. 
The artefacts generated by the encoder-decoder methods stem from difficulties in balancing multiple loss terms commonly used when training such methods.   
\renewcommand{\im}[2]{\includegraphics[width=0.285\columnwidth]{images/multiple_attributes-hair_shapes/#1/#2.jpg}}
\renewcommand{\imrow}[1]{\im{#1}{orig_img} & \im{#1}{nocondition} & \im{#1}{attrforce}}
\renewcommand{\miniwid}{0.05cm}
\renewcommand{\figureimagewidth}{0.23\columnwidth}
\newcolumntype{C}{ >{\centering\arraybackslash} m{\figureimagewidth} }
\setlength{\tabcolsep}{6pt} 

\begin{figure}
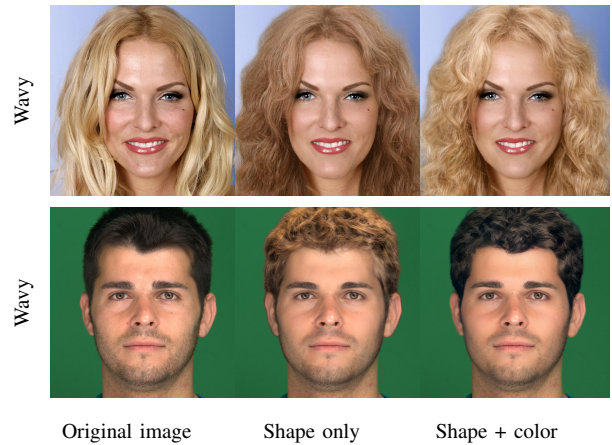

	\begin{tabular}{DCCC}
        \begin{turn}{90}{\footnotesize Wavy}\end{turn}		&	\imrow{235} \\
                \begin{turn}{90}{\footnotesize  Wavy}\end{turn}		&	\imrow{_DSC2164} \\
                    &			\footnotesize Original image & \footnotesize Shape only & \footnotesize Shape + color
	\end{tabular}
	\caption{Editing hair shape with MaskFaceGAN. Editing only the hair shape can lead to changes in hair color (second column). Adding hair color (from the input image) as an additional optimization constraint helps towards preserving the input hair color semantics.}\vspace{-2mm}
	\label{fig:multiple_attributes-hairshape}
\end{figure}

\setlength{\tabcolsep}{2pt} 
\renewcommand{\im}[1]{ \includegraphics[width=\figureimagewidth]{#1}}

InterFaceGAN and InterFaceGAN--D are most closely related to MaskFaceGAN and achieve higher--quality editing result than the encoder--decoder models due to the use of the StyleGAN2 generator. The vanilla version of InterFaceGAN yields  convincing target semantics, but due to the information entanglement in the latent codes,  often  changes correlated attributes in the process. This is best seen with the ``Grey hair'' attribute in Fig.~\ref{fig:all_edits_celebahq_helen}, 
where the edited faces appear much older than the originals. InterFaceGAN--D is able to remove some of this entanglement, but this requires a manual analysis of attribute correlations to exclude unwanted facial semantics from the editing procedure. 
We also observe an interesting behavior with the InterFaceGAN models, in that the same hyper--parameter setting (i.e., the magnitude of the latent code movement), results in attribute changes of different intensity for images of different characteristics -- see, for example, the  ``Blond hair'' results in Fig.~\ref{fig:all_edits_celebahq_helen}. 

Compared to the competing models, the proposed MaskFaceGAN approach: $(i)$ exhibits better disentanglement characteristics due to the latent space optimization procedure, which relies on semantic and spatial constraints, $(ii)$ ensures artefact--free high--resolution attribute editing with convincing image semantics, $(iii)$ preserves important image details (e.g., facial areas not related to the target attribute or background), and $(iv)$ does not require manual hyper--parameter tuning for each probe image separately.

\textbf{Quantitative evaluation.}
 To evaluate attribute editing performance in a quantitative manner, prior works \cite{he2019attgan, liu2019stgan} reported a measure quantifying attribute generation accuracy. Because MaskFaceGAN tries to maximize this exact measure during latent code optimization, we use an alternative approach 
 to ensure a fair comparison. Specifically, we first report  Fréchet Inception Distances (FID) to quantify performance and then present results of a user study, similarly to \cite{liu2019stgan}. 

\renewcommand{\figureimagewidth}{.27\columnwidth}
\begin{figure}[t]
\centering
\begin{tabular}{CCCD}
$y = 0.9$ & $y=0.95$ & $y=1.0$ & \\
\im{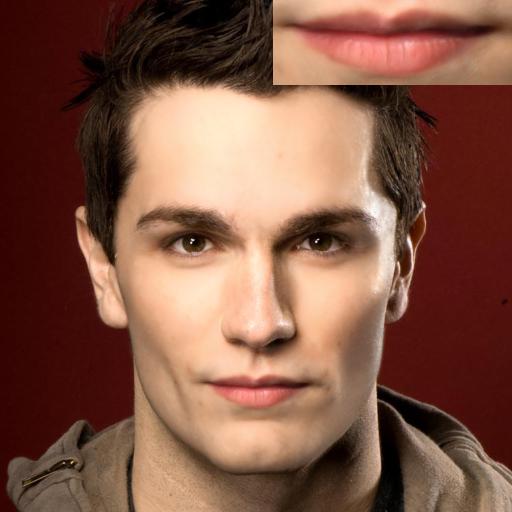} & \im{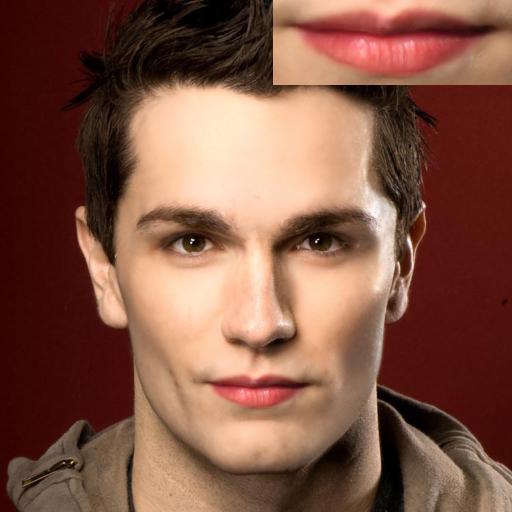} & \im{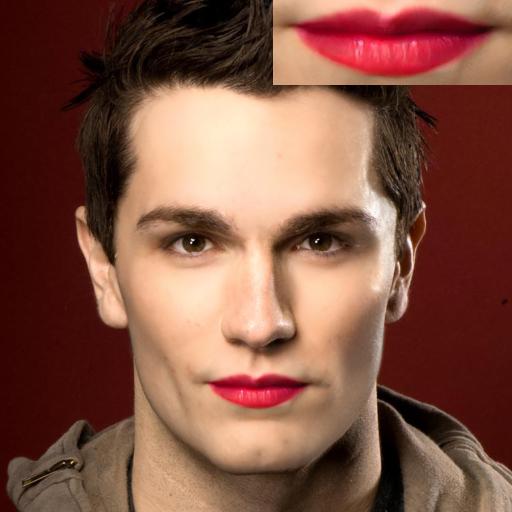} & \begin{turn}{270}  \hspace{-1.2cm} \footnotesize{Wearing lipstick} \end{turn} \\
\im{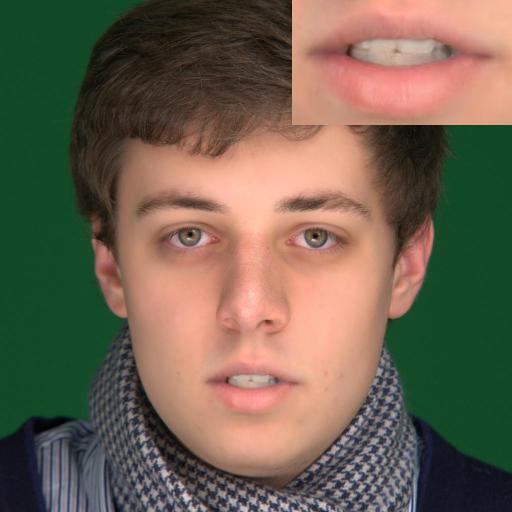} & \im{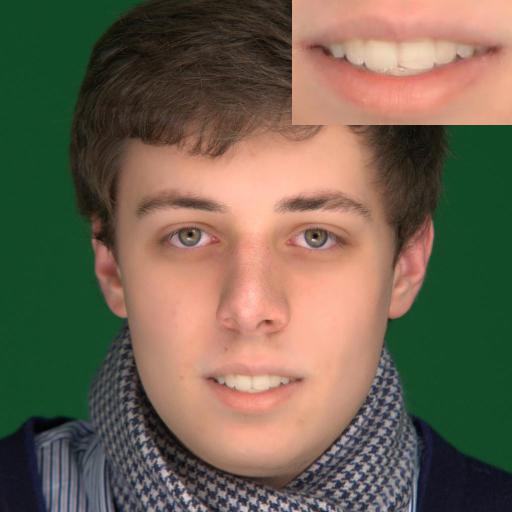}  & \im{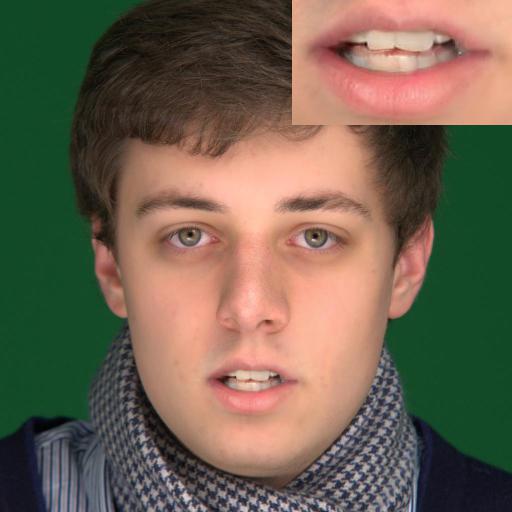} &  \begin{turn}{270} \hspace{-1.2cm} \footnotesize{Mouth sl. open} \end{turn} \\
\end{tabular}
	\caption{Visual examples of MaskFaceGAN's capability to control attribute intensity during editing. 
	\vspace{-3mm}}
	\label{fig:attribute_intensity}
\end{figure}

 \renewcommand{\figuretextwidth}{.02\columnwidth}
\setlength{\tabcolsep}{2pt} 
\renewcommand{\arraystretch}{1.3}

\begin{figure}[t]
\centering
\begin{tabular}{CCCD}
$\alpha=0.5$ & $\alpha = 1.$ & $\alpha = 1.5$ & \\
\im{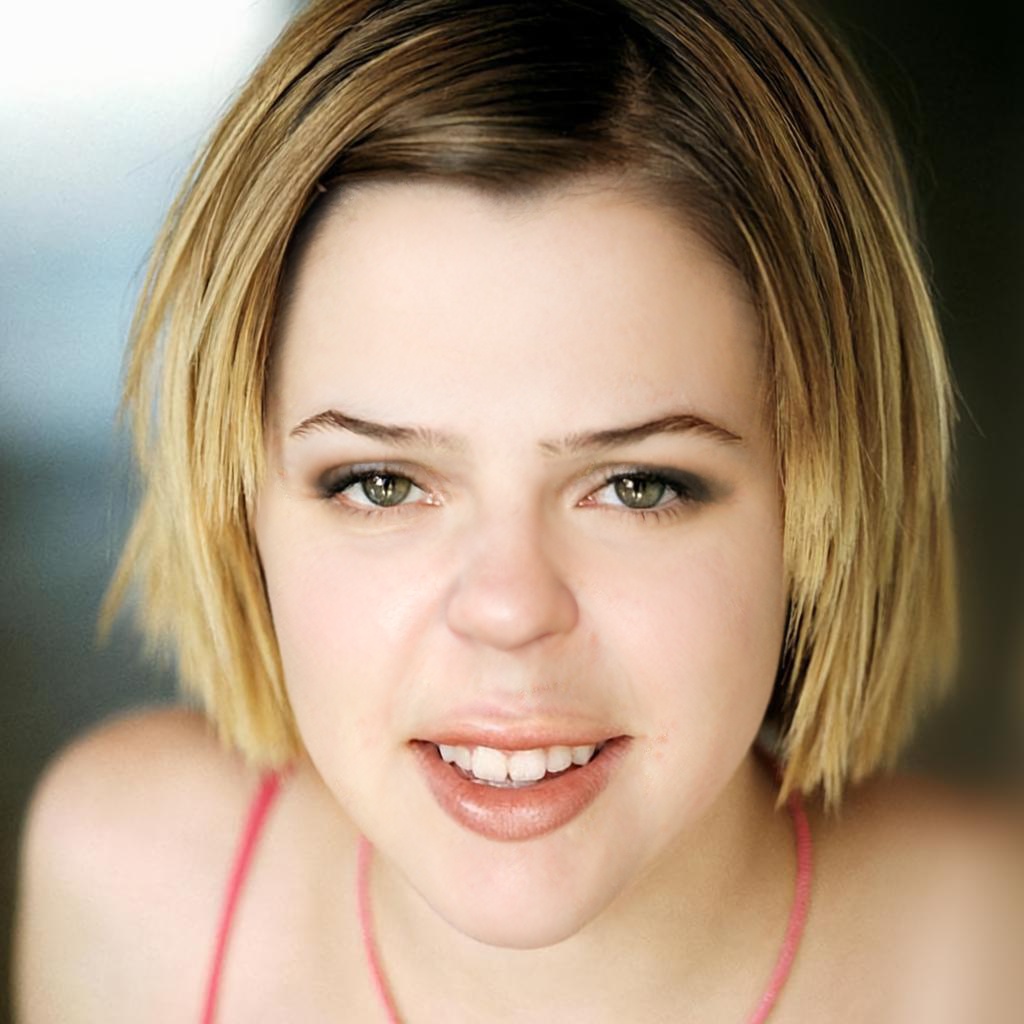} & \im{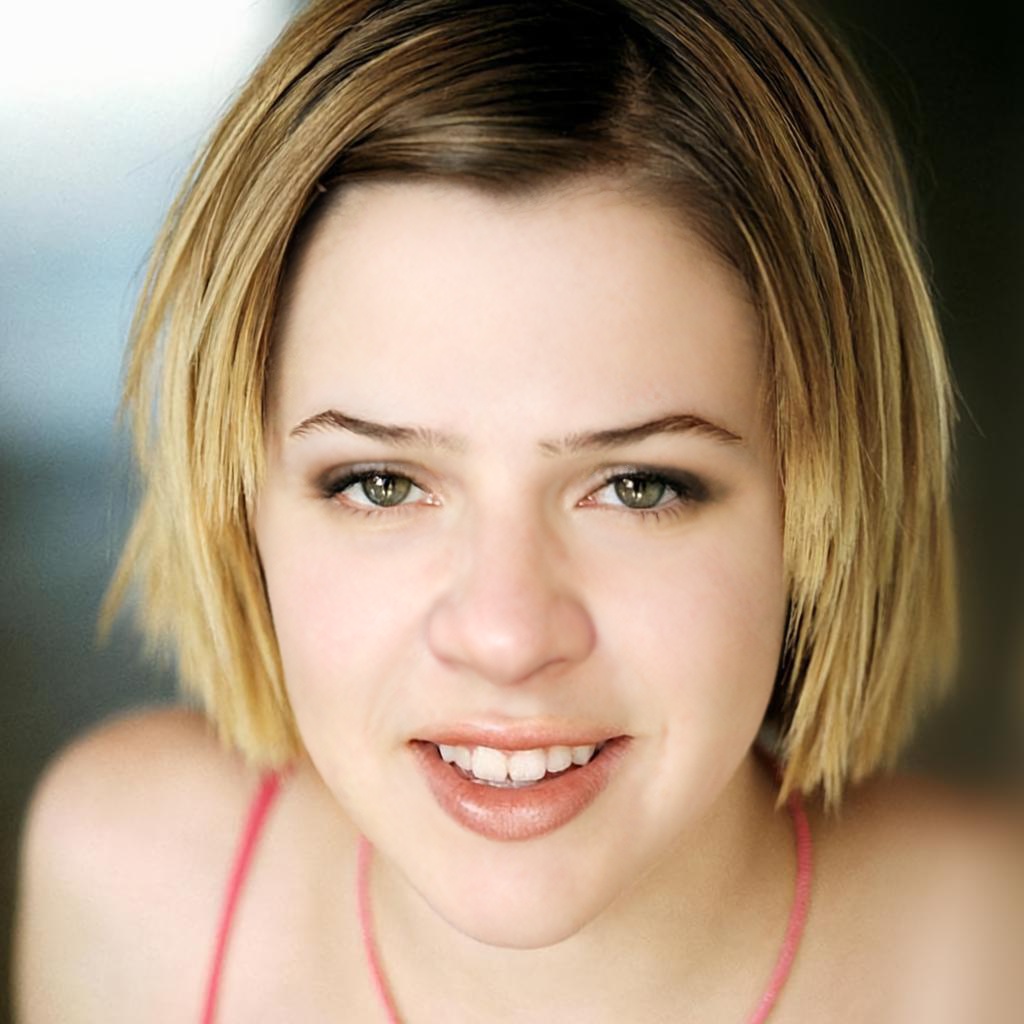} & \im{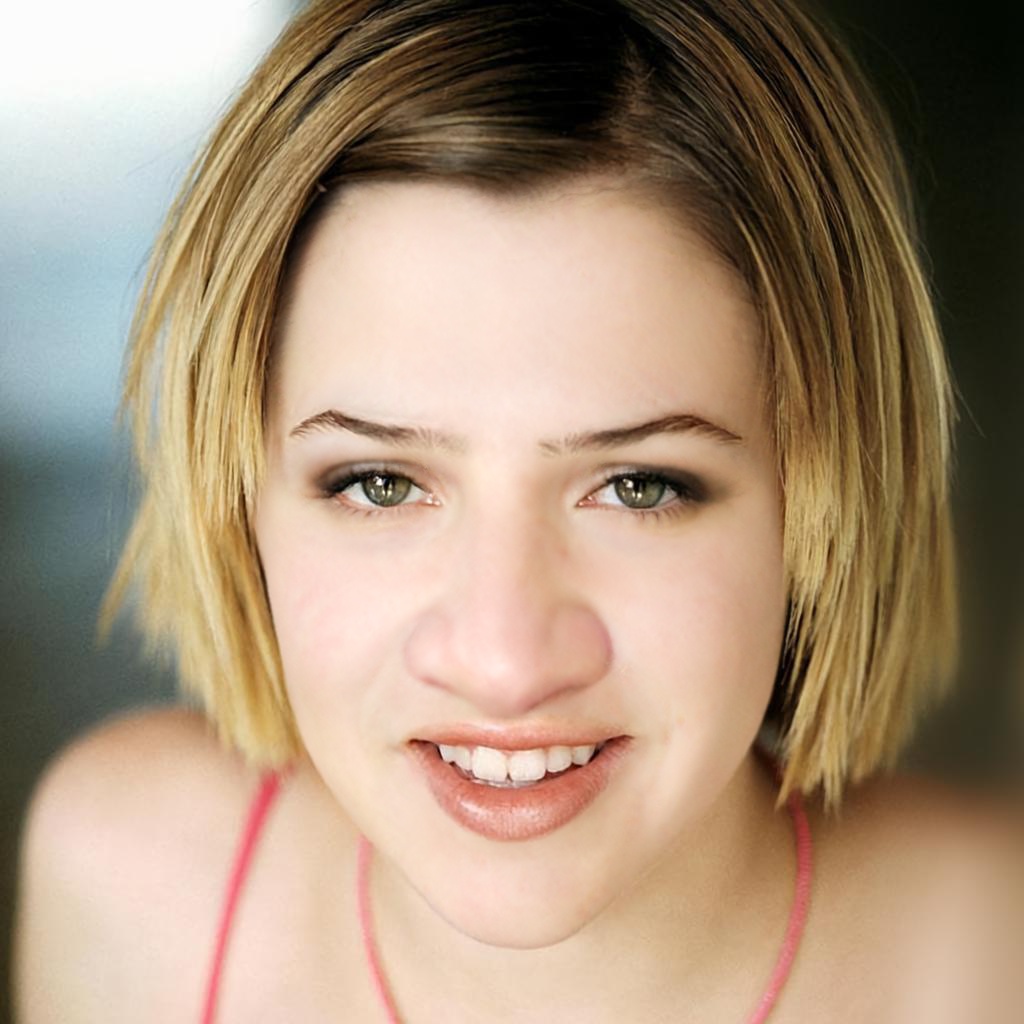} & \begin{turn}{270} \hspace{-0.8cm} \footnotesize{Big nose} \end{turn}  \\
\im{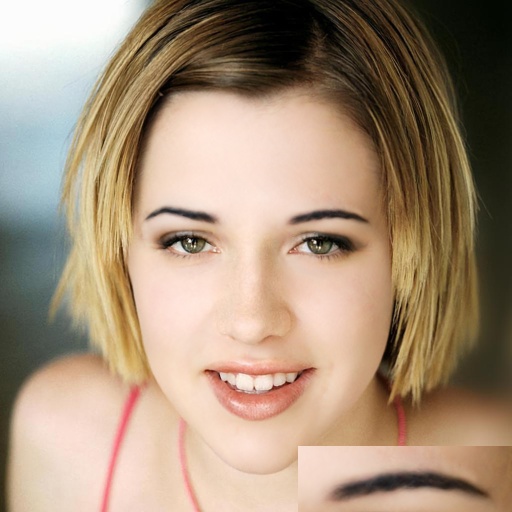} & \im{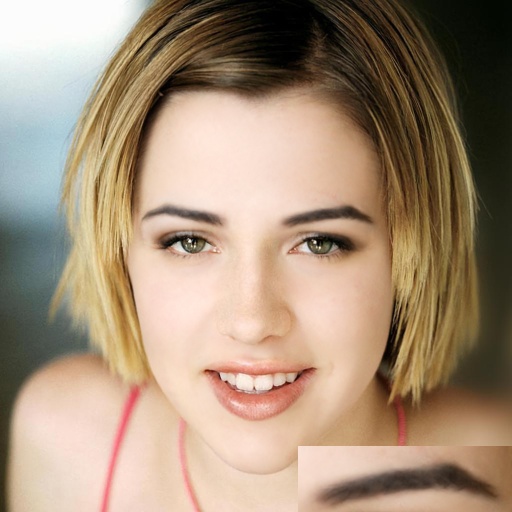} & \im{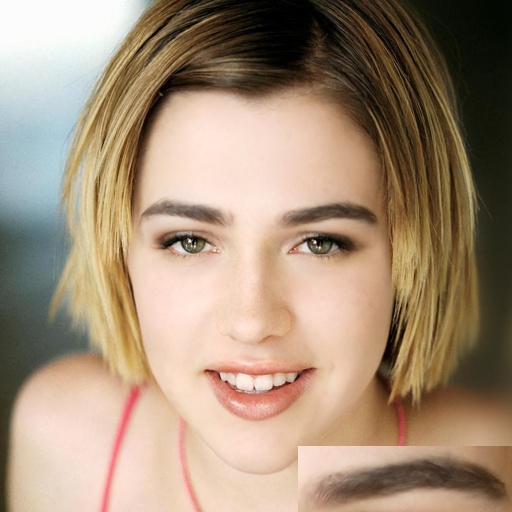} & \begin{turn}{270} \hspace{-1.2cm} \footnotesize{Bushy eyebrows} \end{turn} \\
\end{tabular}
\caption{Visual examples of MaskFaceGAN's capability to control the size of the edited facial components. 
\vspace{-3mm}}
\label{fig:size_manipulation}
\end{figure}

\begin{itemize}
    \item \textbf{FID Score Analysis.} The Fréchet Inception Distance (FID) \cite{heusel2017gans} represents a common measure of image quality, predominantly used in the evaluation of GANs. 
    We report FID scores for each dataset considered in our evaluation by first generating attribute specific FID scores and then averaging over all attributes. The facial images are rescaled to $299 \times 299$ pixels before extracting features. Table \ref{tab:fid} shows that MaskFaceGAN achieves the lowest FID scores on all three test datasets, significantly outperforming all five competing editing models. The lower scores can mostly be attributed to the high quality of the edited images and lack of artefacts, which are the result of spatially constrained image modifications that only alter a small portion of the image for a given target attribute, while keeping other parts of the images intact. 
    \item \textbf{User Study.} Following \cite{liu2019stgan}, we conduct a user study using a crowdsourcing platform to analyze the quality of the edited images. Here, the users (raters) were shown edited images of all considered models and asked to select the most convincing one based on the following instructions: \textit{``Choose the image that changes the attribute more successfully, is of higher image quality and better preserves the identity and fine details of the source image.''}. Additionally, they were also instructed to rate images 
    on a $5$--point Likert scale, where a higher number represents better image quality. A single user study covered all test images from a given dataset and was performed over all $14$ attributes. Images were shown in random order for a fair comparison. The results, reported in Tables \ref{tab:userstudy-best} and  \ref{tab:userstudy-scores},  show that MaskFaceGAN was most frequently selected  as the best among the evaluated techniques and also received the highest average scores (on the 5--point Likert scale) on all three dataset. These observations are further supported by the results in Fig.~\ref{fig:barplot}, where user scores are reported for each edited attribute separately. 
    The reported results speak of the excellent performance of MaskFaceGAN and competitiveness with respect to existing models. \vspace{-2mm}  
\end{itemize}
\renewcommand{\figureimagewidth}{.25\columnwidth}

\renewcommand{\miniwidth}{\textwidth}
\renewcommand{\im}[1]{\includegraphics[width=0.28\columnwidth]{images/cake/2/#1}}

\setlength{\tabcolsep}{2mm} 
\renewcommand{\miniwid}{0.05cm}
\newcolumntype{D}{ >{\centering\arraybackslash} m{\miniwid} }
\renewcommand{\turnedtext}[1]{\begin{turn}{90} #1 \end{turn}}

\begin{figure}[t]
    \captionsetup[subfigure]{labelformat=empty}    
    \begin{tabular}{DCCC}
 & $\epsilon=0.3$ & $\epsilon=0.0$ & \hspace{1mm} $\epsilon=0.0$ \\[-0.1mm]
 \turnedtext{$\alpha=0.75$}  & \im{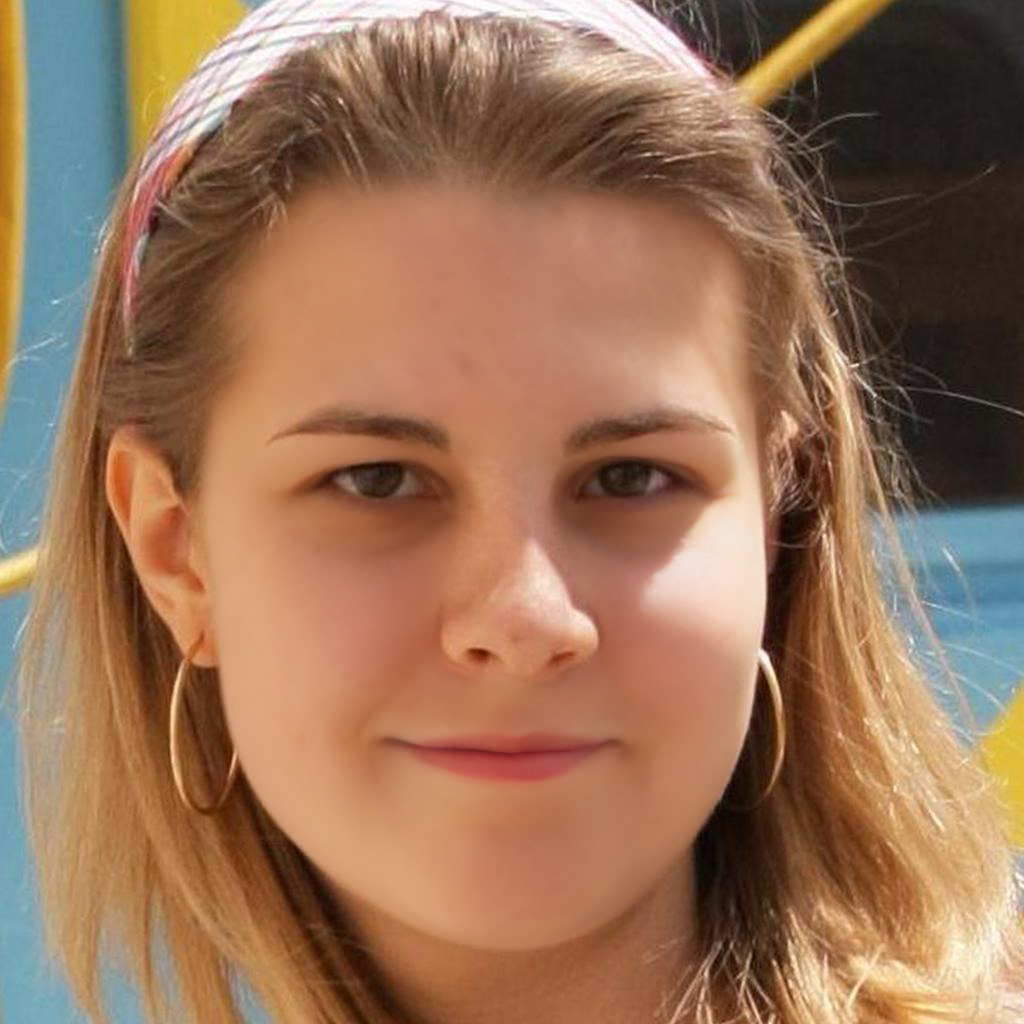} & \im{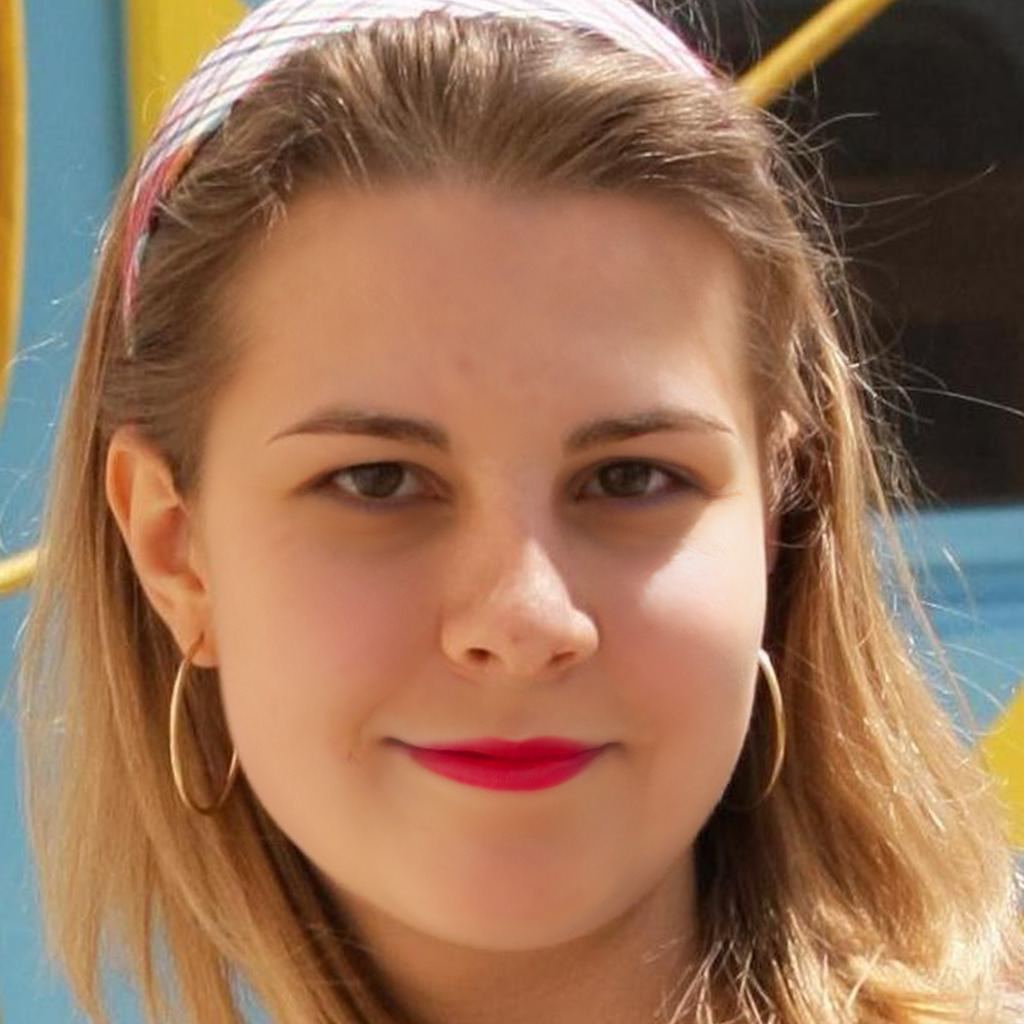}  & \im{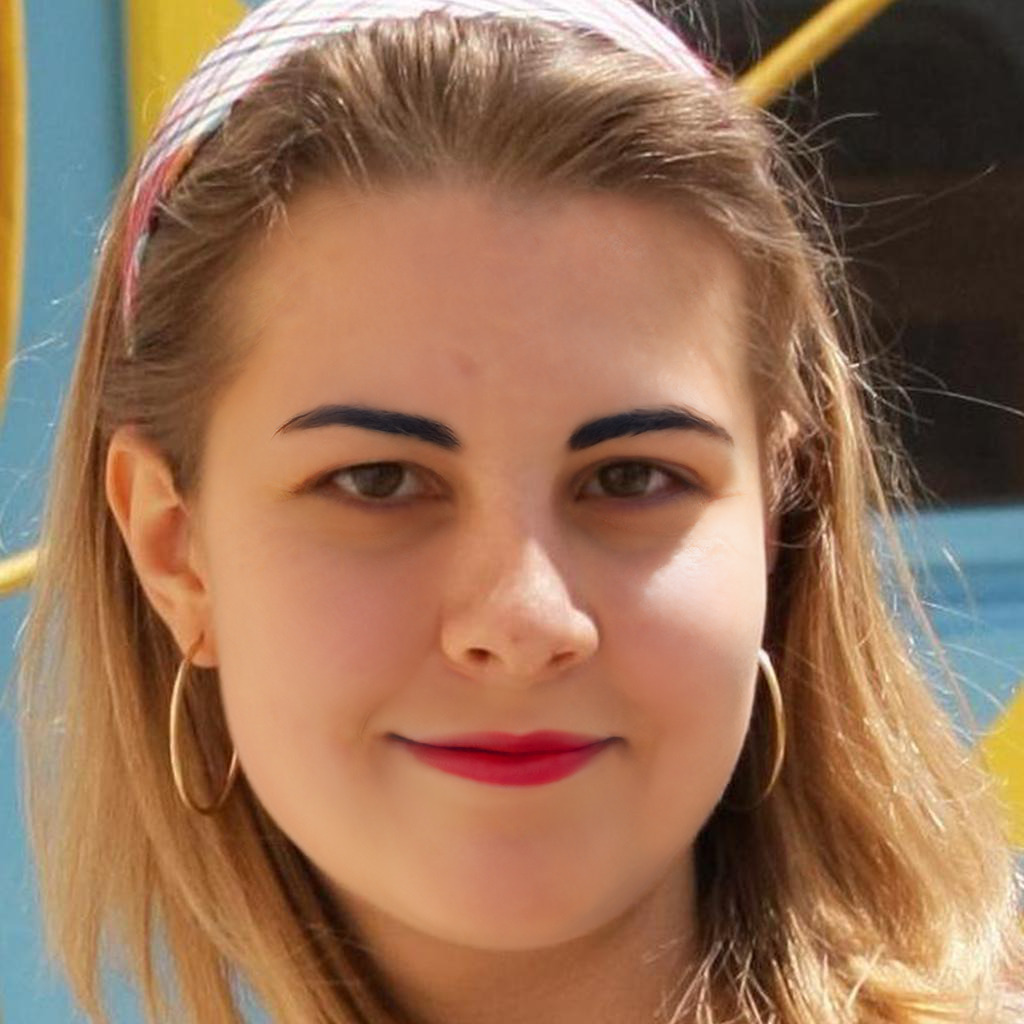} \\
 \turnedtext{$\alpha=1.25$} &  \im{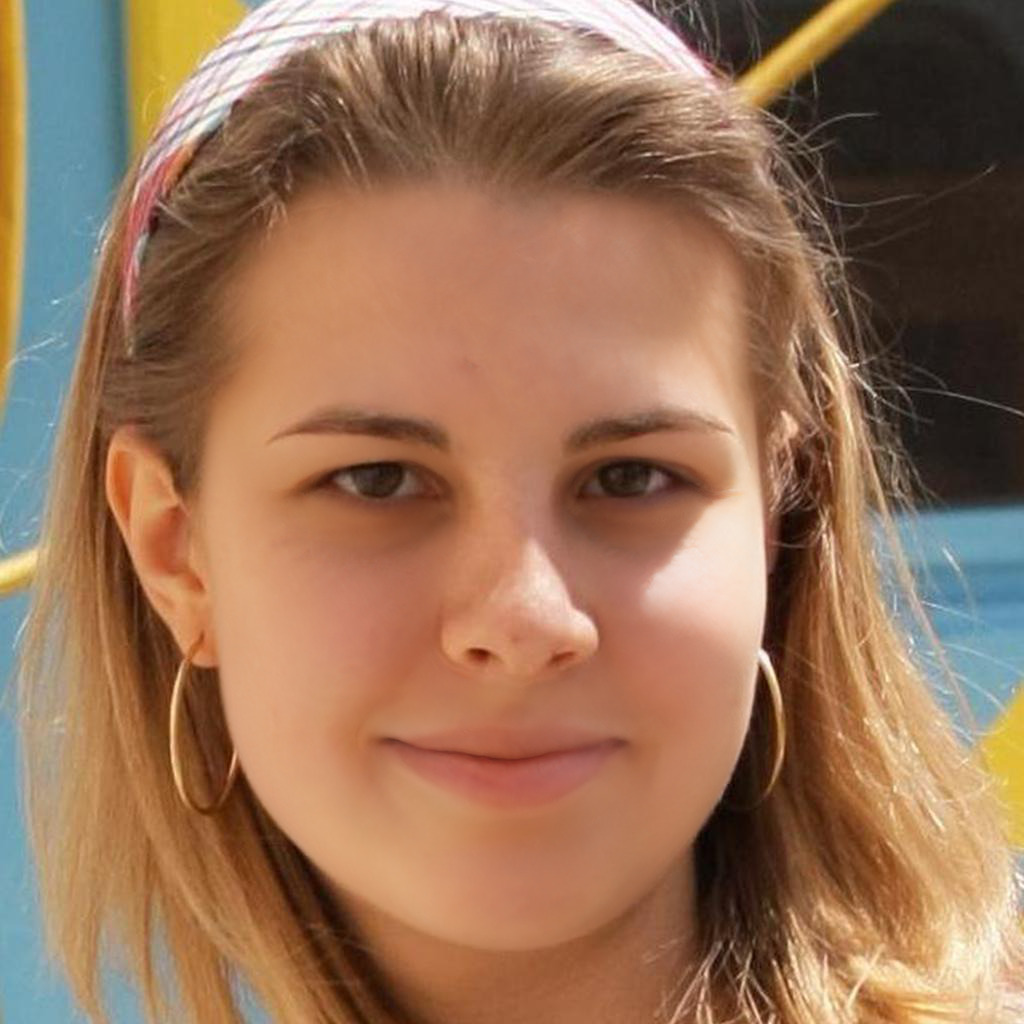} & \im{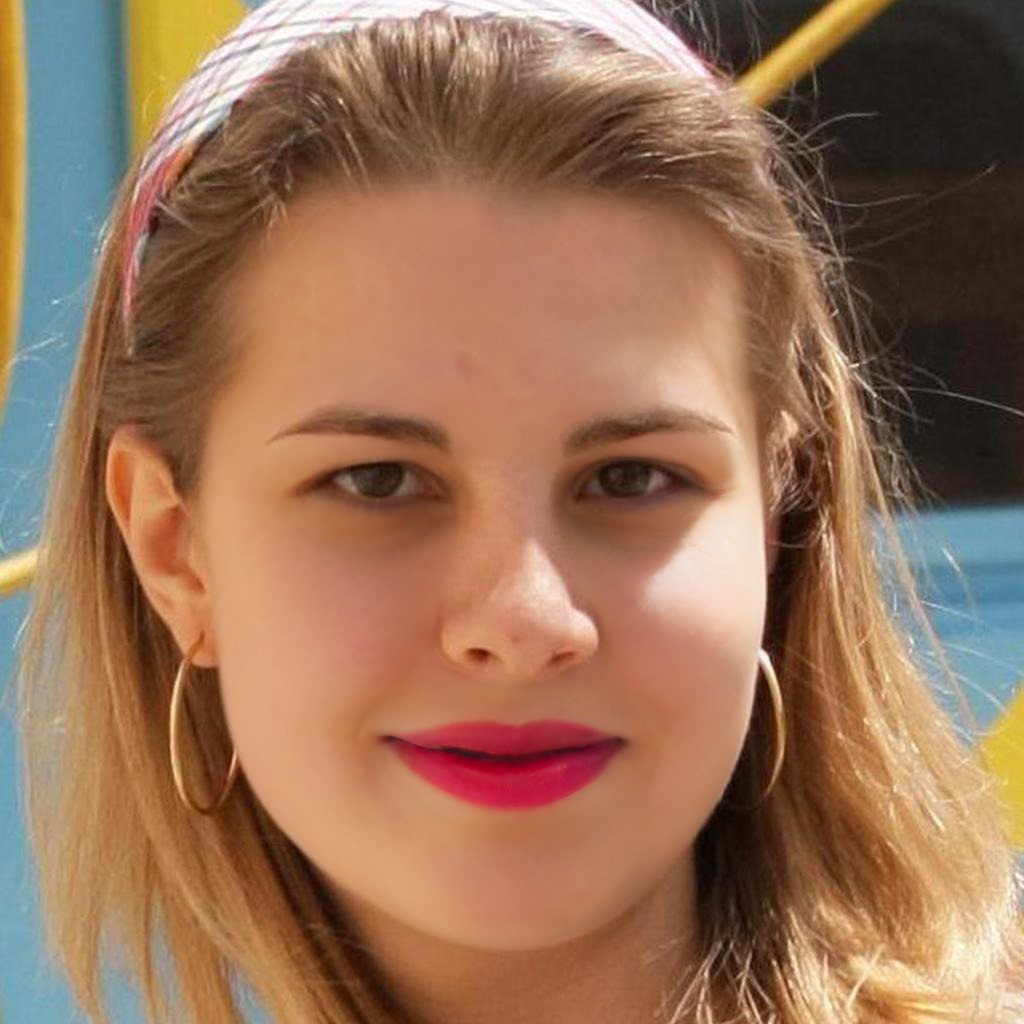} & \im{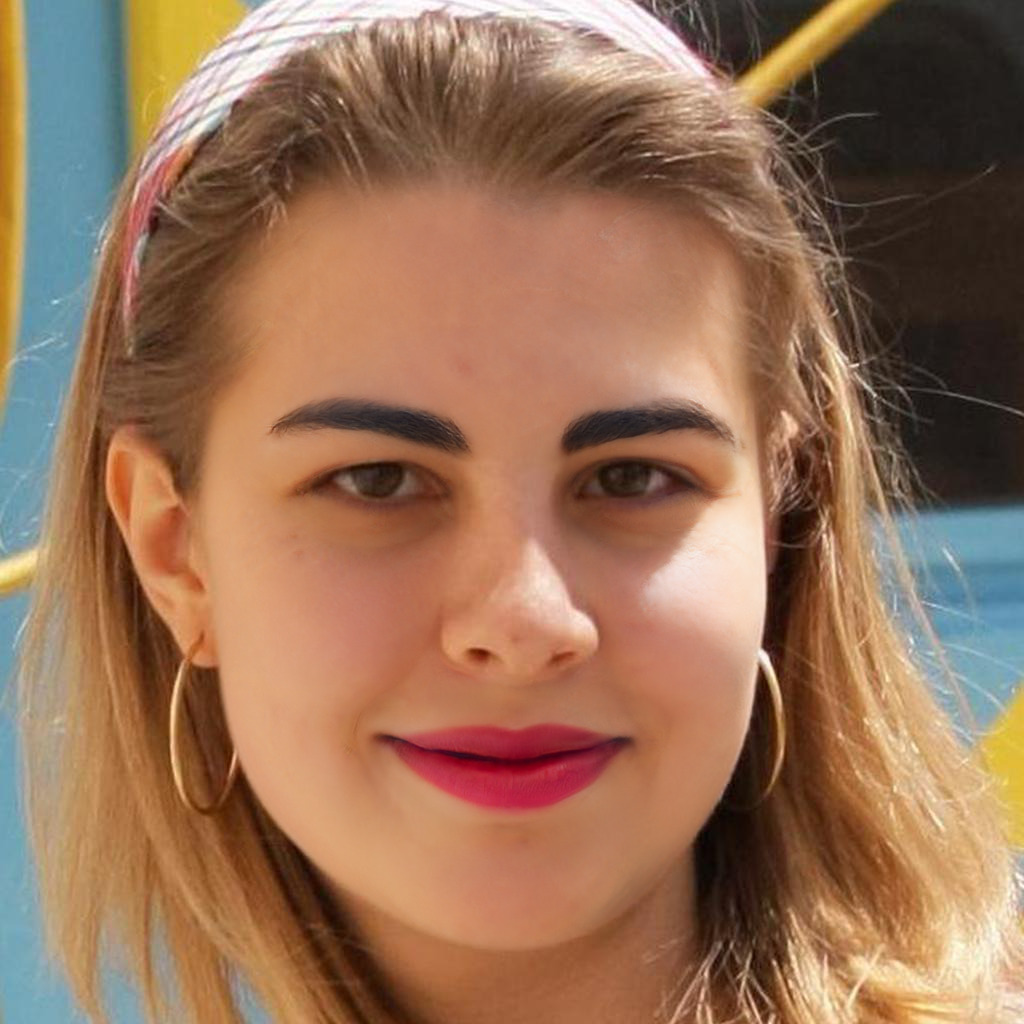}\\[-1mm]
  & \mbox{\footnotesize + Wearing lipstick} & \mbox{\footnotesize + Wearing lipstick} & \mbox{\footnotesize + Wearing lipstick} \\[-2mm]
   & \mbox{\footnotesize } & \mbox{\footnotesize } & \mbox{\footnotesize + Bushy eyebrows} \\[-1mm]
   \end{tabular}
\caption{An example of simultaneous component size manipulation and attribute intensity control. 
Results are shown for editing a single (i.e., ``Wearing lipstick'') or two attributes (i.e., ``Wearing lipstick'' and ``Bushy eyebrows'') at the same time.
\vspace{-3mm}}
\label{fig:simultaneous_all}
\end{figure}

\subsection{Characteristics of MaskFaceGAN}

MaskFaceGAN exhibits several desirable characteristics, such as the capability to $(i)$ edit multiple attributes with a single optimization procedure, $(ii)$ control the intensity of edited attribute, and $(iii)$ modify the size of the edited region. 
We illustrate these characteristics through several visual examples.  

\textbf{Multiple Attribute Editing.}
By averaging the KL divergence of the semantic constraint over multiple attributes, MaskFaceGAN can edit multiple binary attributes through a single optimization procedure. Examples of such editing results are presented in Fig. \ref{fig:multiple_attributes} for different numbers of attributes, i.e., $K=\{1,2,3\}$. 
Two interesting observations can be made here: $(i)$ even when multiple attributes are edited, the results are still visually convincing and artefact-free, and $(ii)$ the joint optimization of several attributes retains considerable correspondence with the original image. 

The multiple-attribute editing capabilities of MaskFaceGAN are especially useful when editing hair shape. Because the model is not explicitly aware of characteristics of the original facial region considered during the editing process, it can in a limited number of cases also alter some additional attributes in addition to the targeted attribute, e.g., change the hair color when hair shape edits are targeted. 
While this may not be desired, MaskFaceGAN can address such problems by defining multiple target attributes. For example, when editing hair shape, all other hair-related characteristics considered by the attribute classifier $C$ can be set to the same target value as in the input image. In Fig.~\ref{fig:multiple_attributes-hairshape}, we illustrate this characteristic on a couple of sample images. Observe how the inclusion of hair color in the optimization procedure helps to better preserve the initial image properties when editing hair shape. 

\textbf{Attribute Intensity Control.}
MaskFaceGAN's semantic constraint is defined by the KL divergence between the predictions of the attribute classifier ($C$)  and the corresponding ground truth. Because the ground truth is smoothed and for a given attribute consists of $y \in \{\epsilon, 1-\epsilon\}$, varying the smoothing parameter $\epsilon$ affects the strength (or intensity) of the targeted attribute in the edited images.
A few illustrative examples of the impact of  $\epsilon$  are presented in Fig. \ref{fig:attribute_intensity}. 
As can be seen, MaskFaceGAN allows for fine-grained control over the attribute intensity in the edited images, although the generated variations may not necessarily be smooth w.r.t. the visual appearance change. For example, the intensity change for the ``Mouth slightly open'' attribute in Fig. \ref{fig:attribute_intensity} does not simply open the mouth more, it actually turns into a half--smile. The results primarily depend on the trained classifier and what it considers an attribute presence with $1-\epsilon$ probability.

\setlength{\tabcolsep}{0.6mm} 
\renewcommand{\arraystretch}{0.8}
\renewcommand{\figureimagewidth}{.22\columnwidth}

\renewcommand{\im}[1]{\includegraphics[width=\figureimagewidth]{#1}}
\renewcommand{\miniwid}{0.3cm}

\renewcommand{\turnedtext}[3]{\begin{turn}{270}\hspace{#1} #2 \hspace{#3}\end{turn}}

\begin{figure}[t]
 \centering
 \begin{tabular}{CCCCD}
 \im{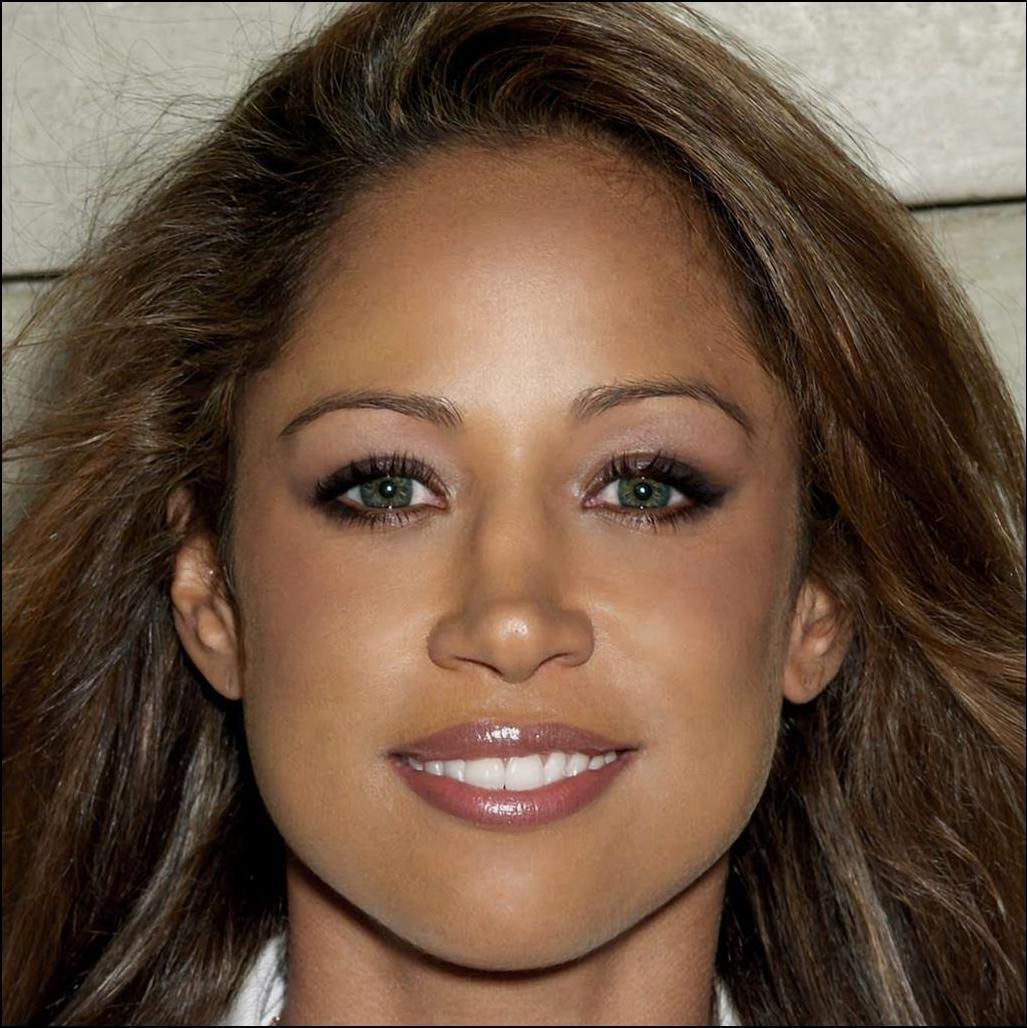} & \im{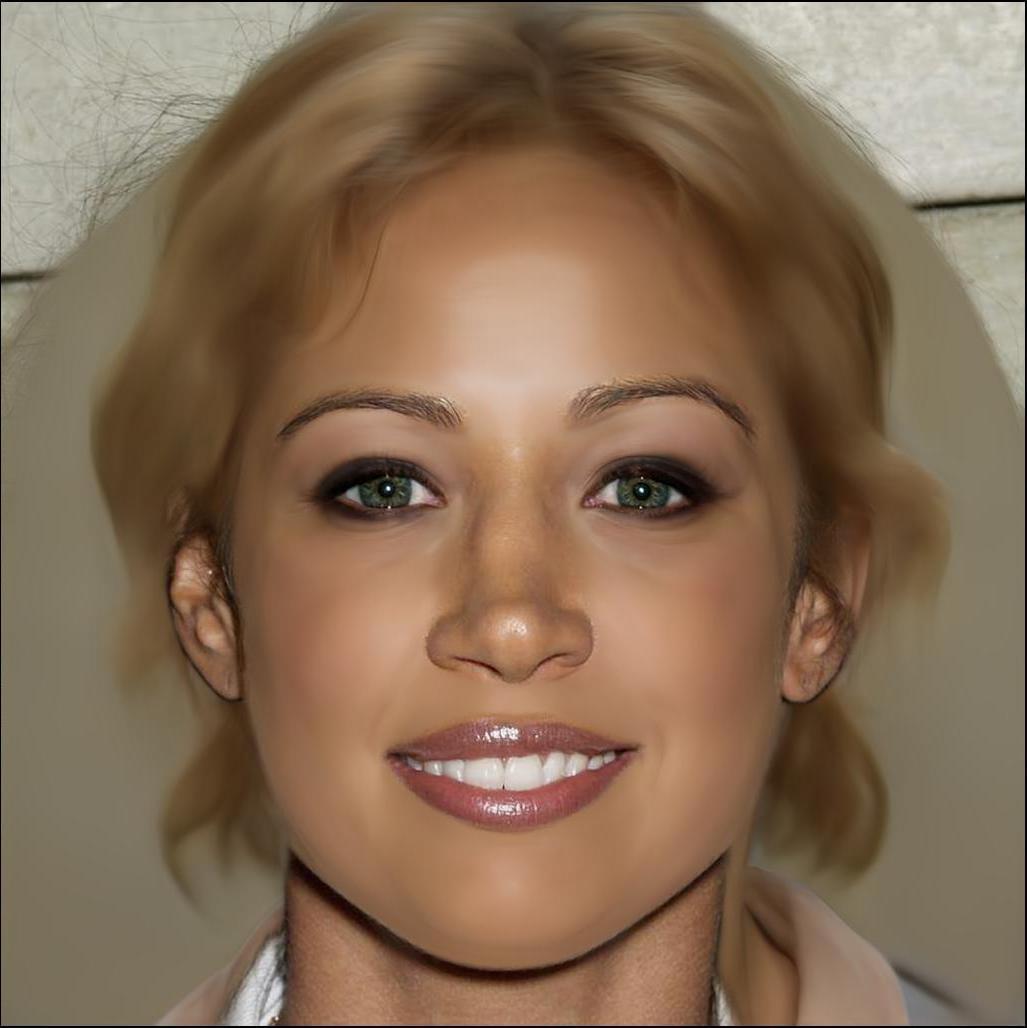} &  \im{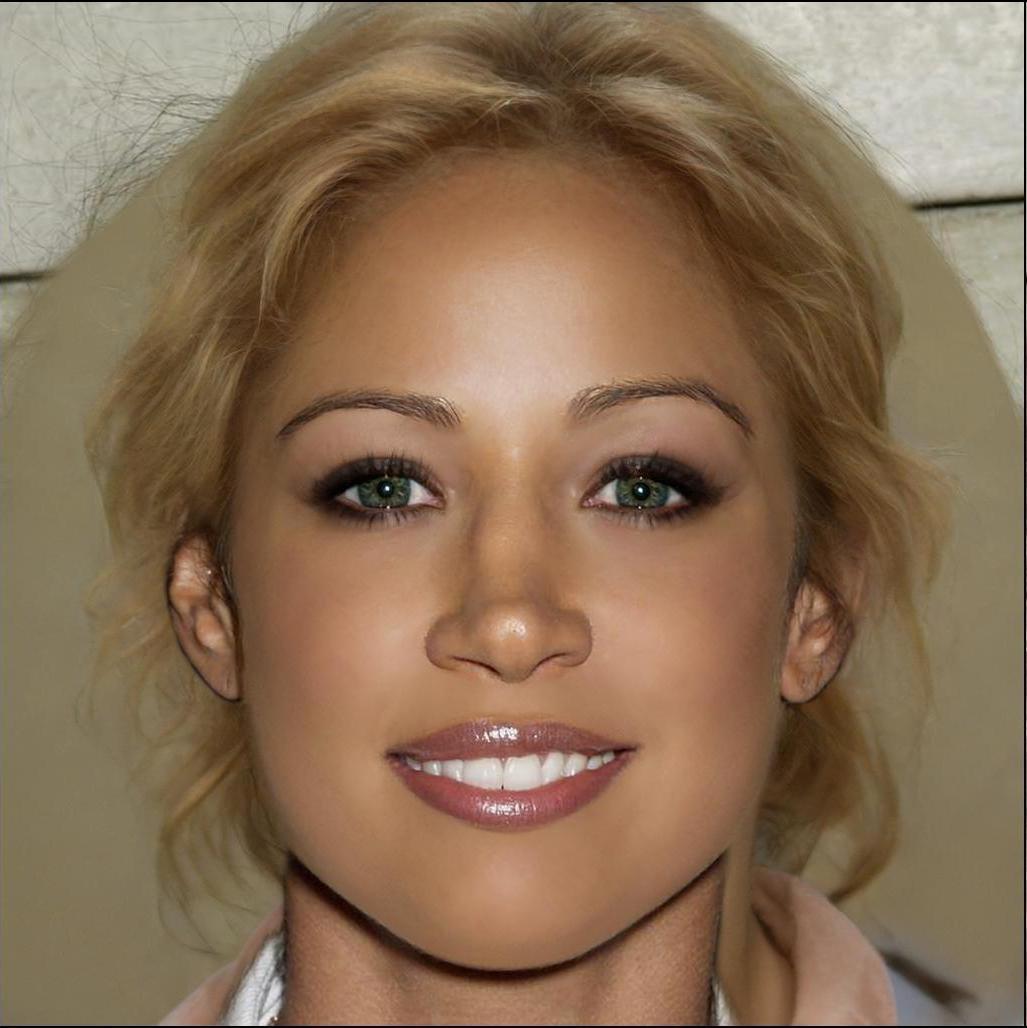} & \im{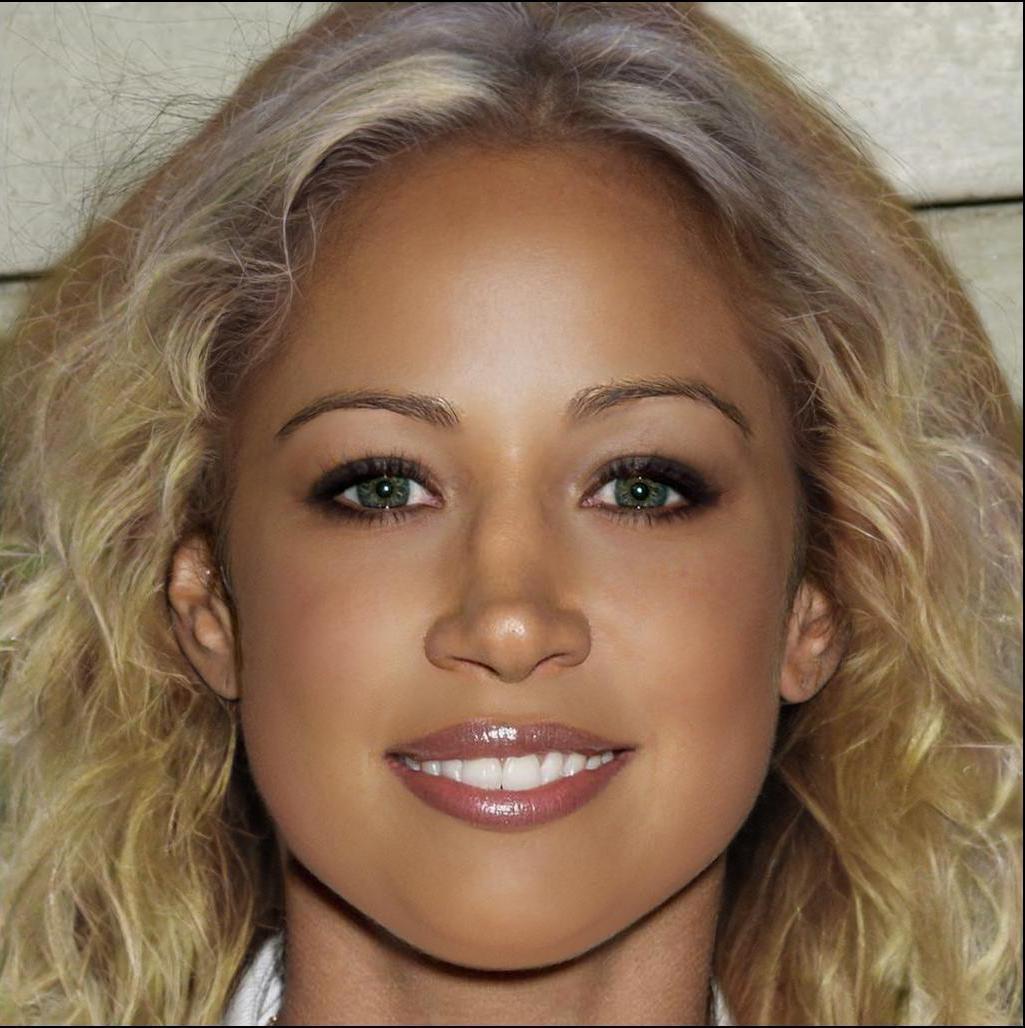} & \turnedtext{-8.8mm}{\footnotesize{Blond hair}}{0mm} \\
  \im{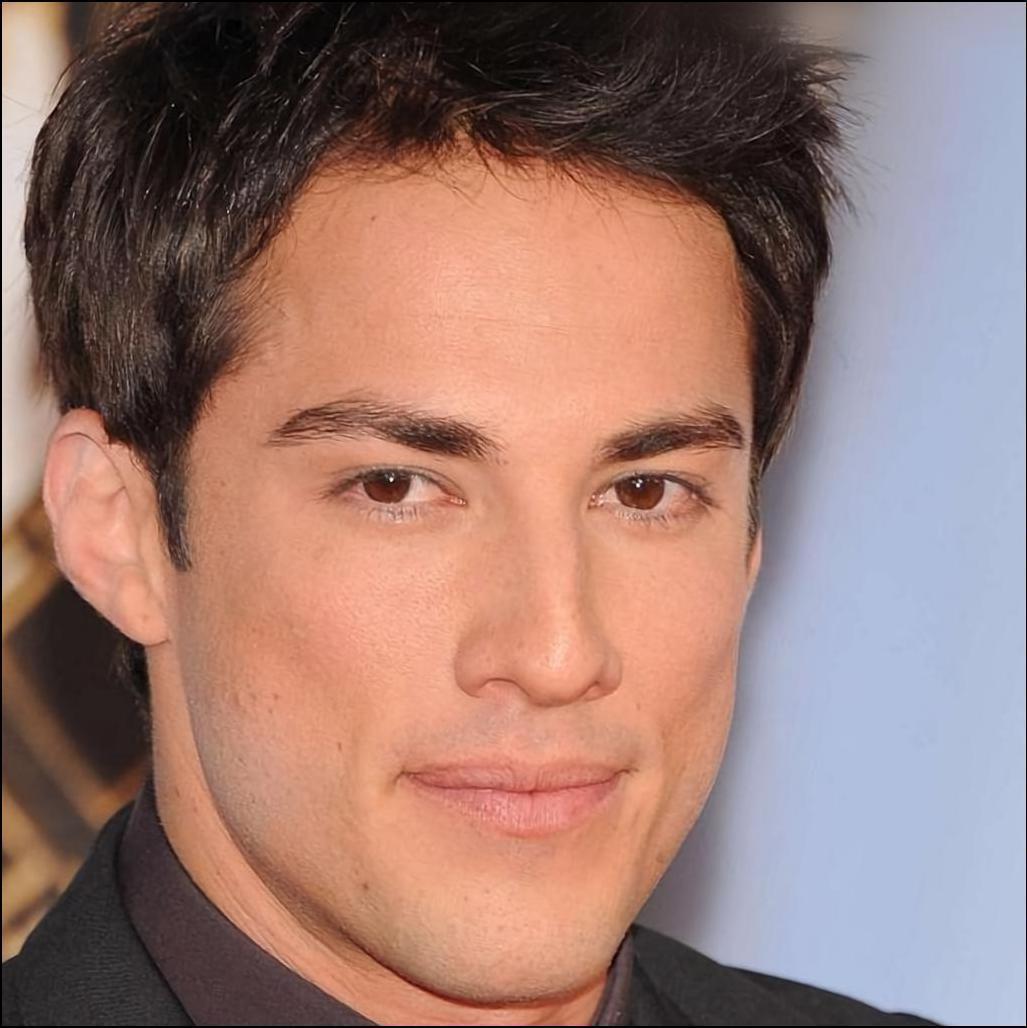} & \im{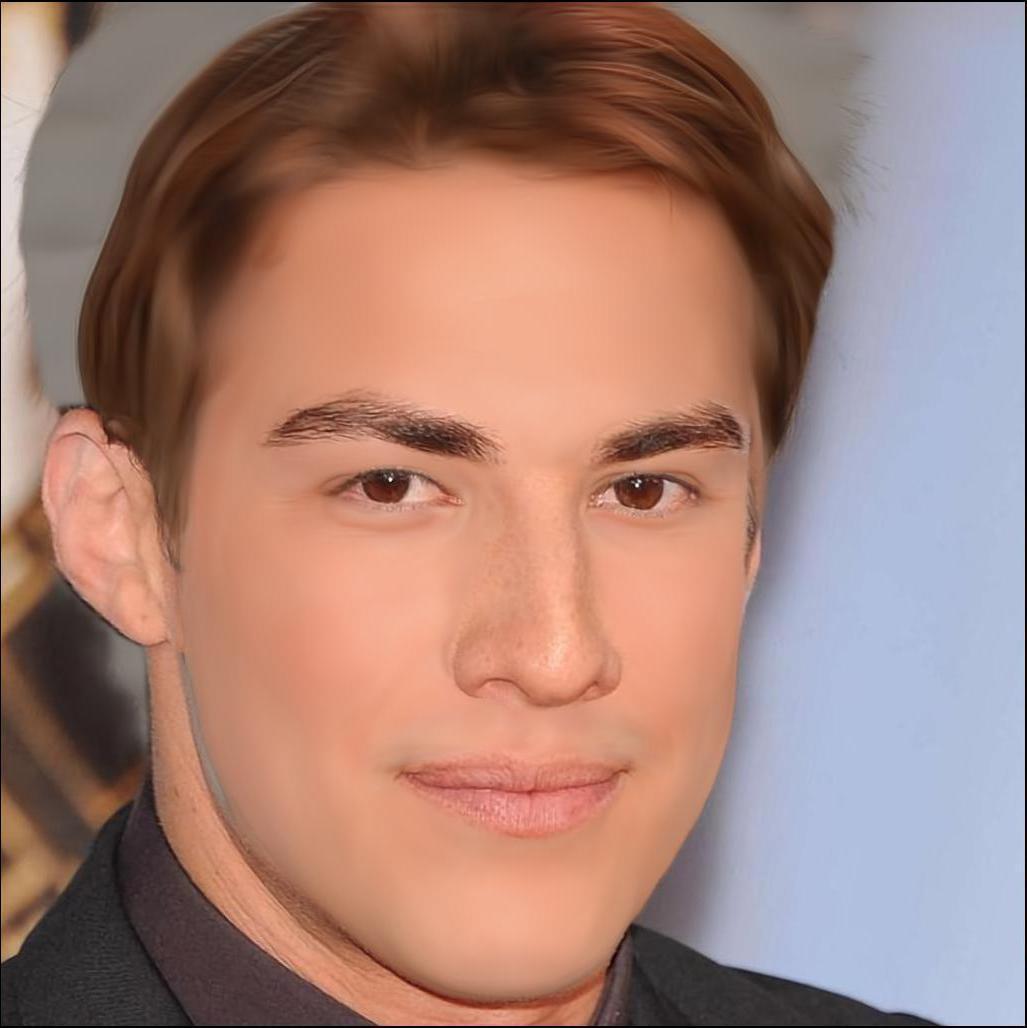} &  \im{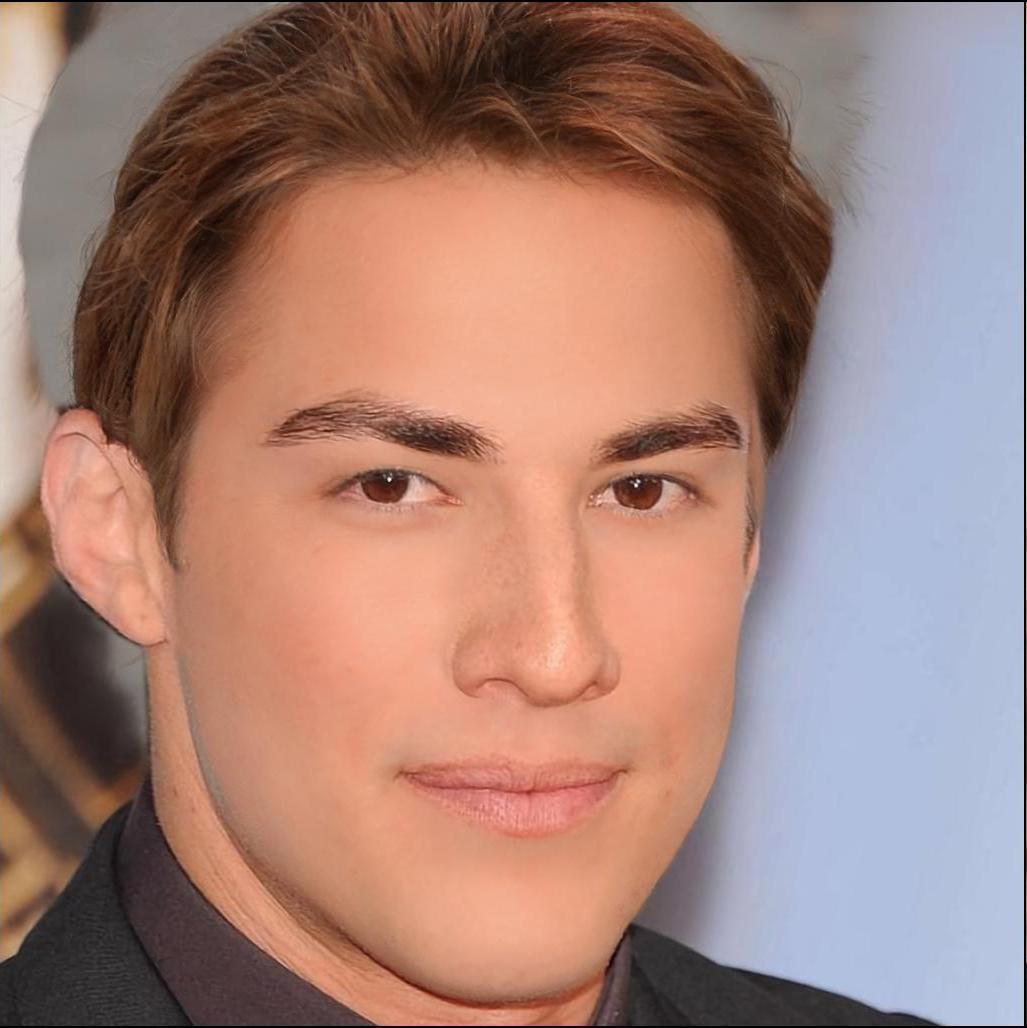} & \im{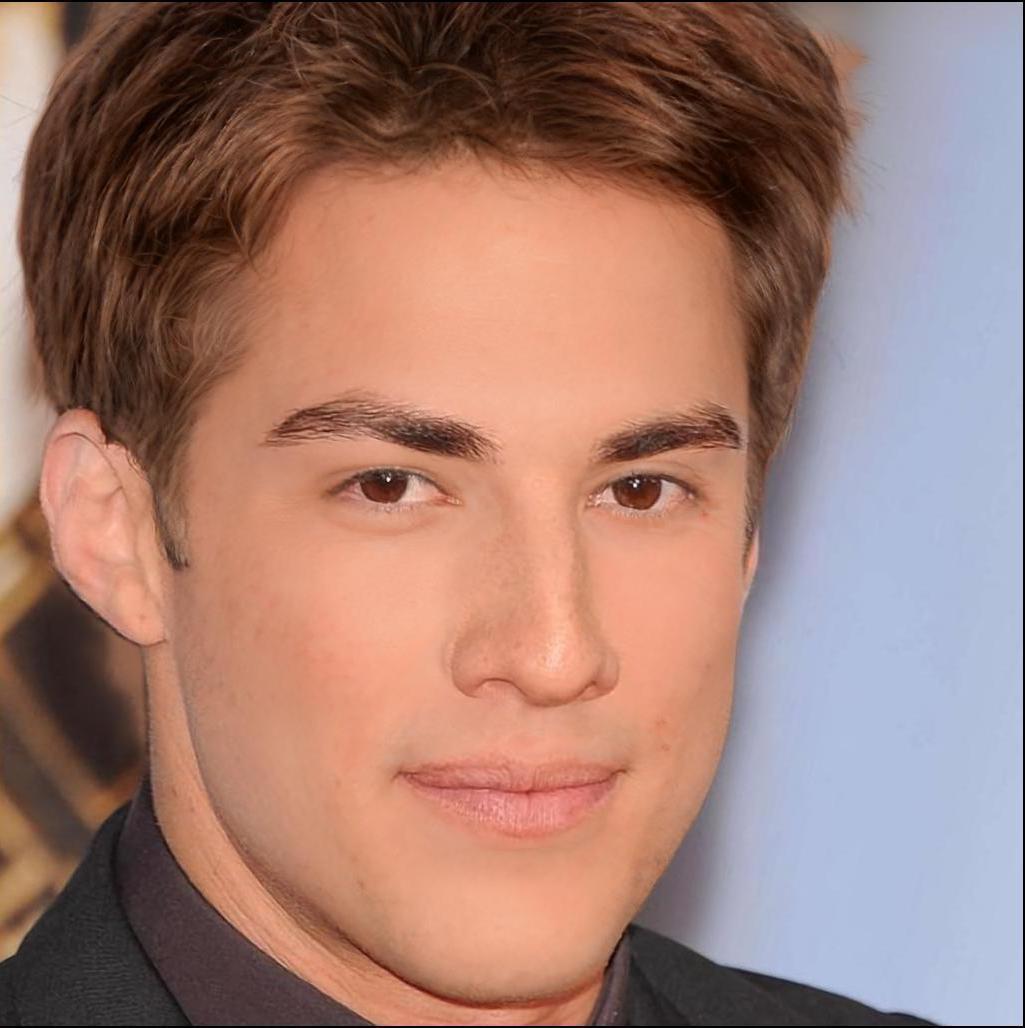}& \turnedtext{-0.8cm}{\footnotesize{Brown hair}}{0mm} \\
   \im{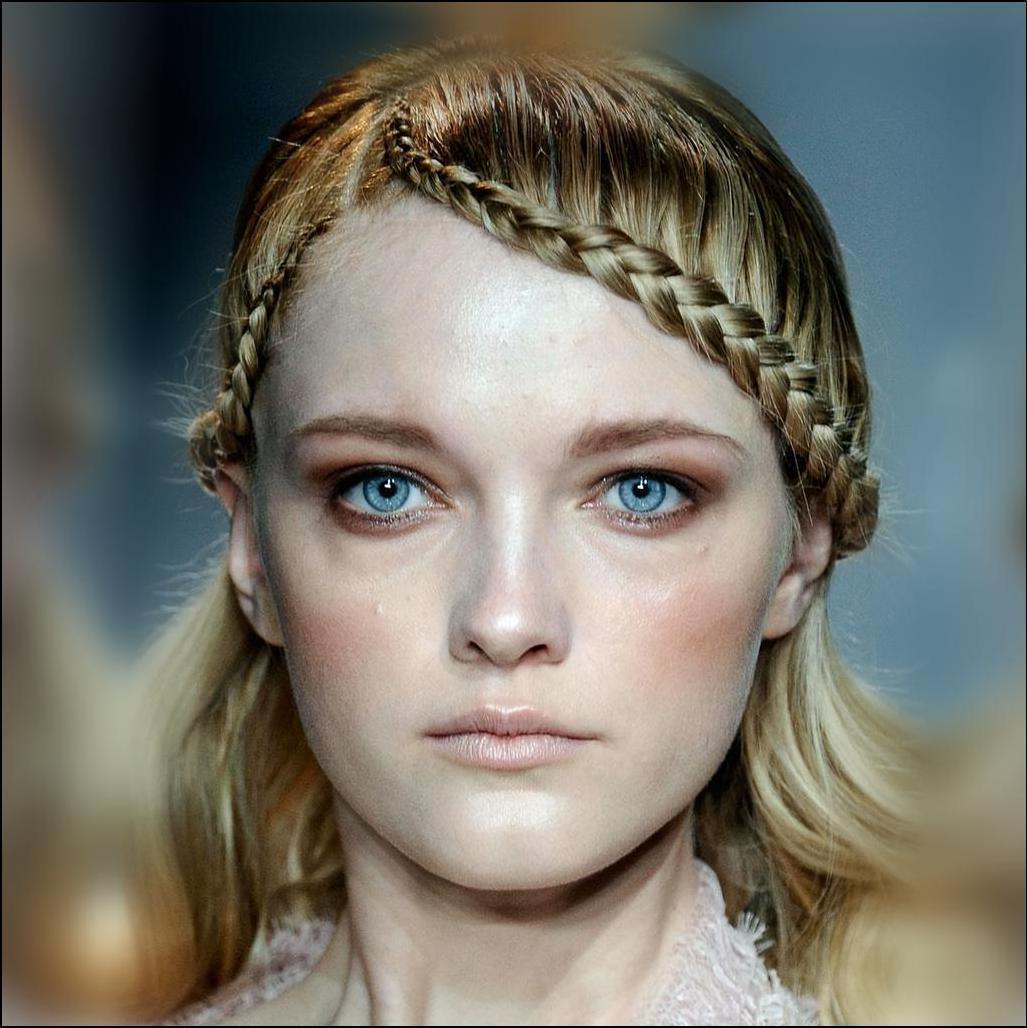} & \im{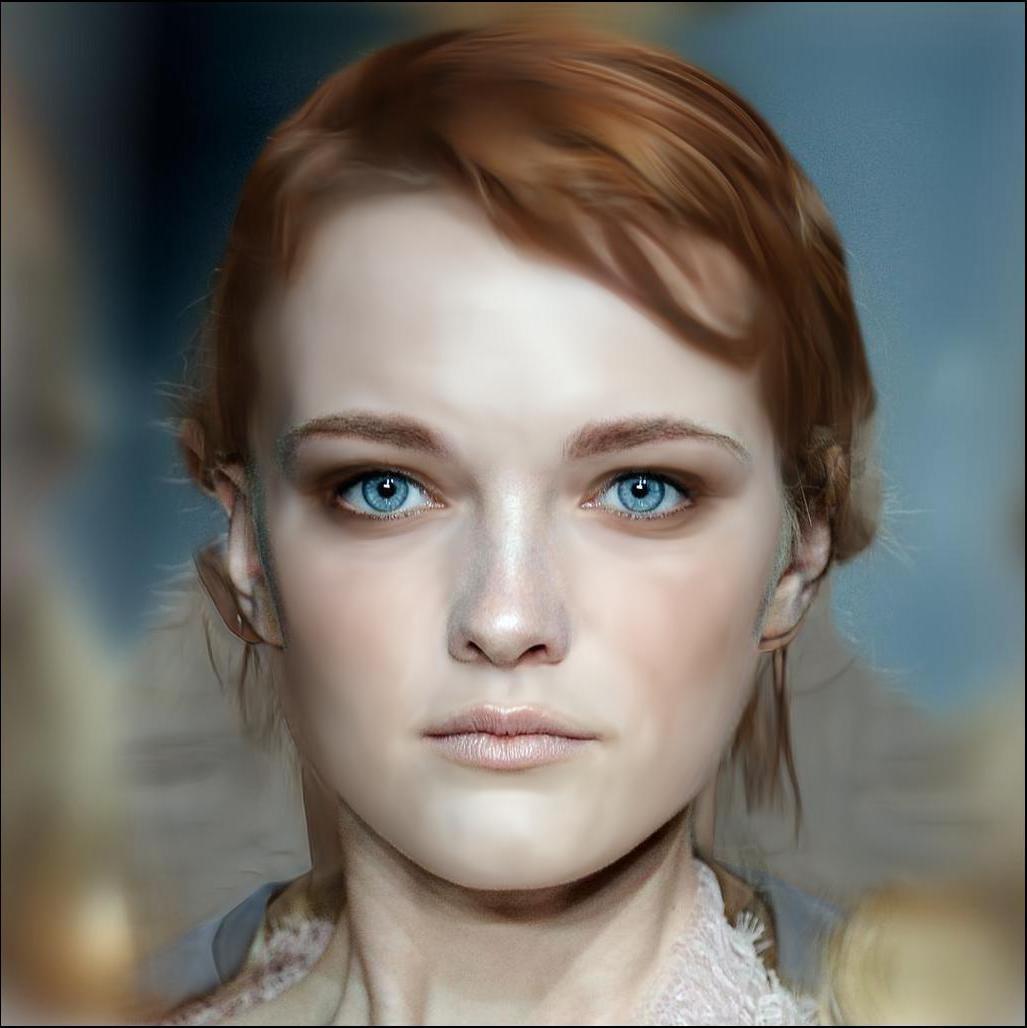} &  \im{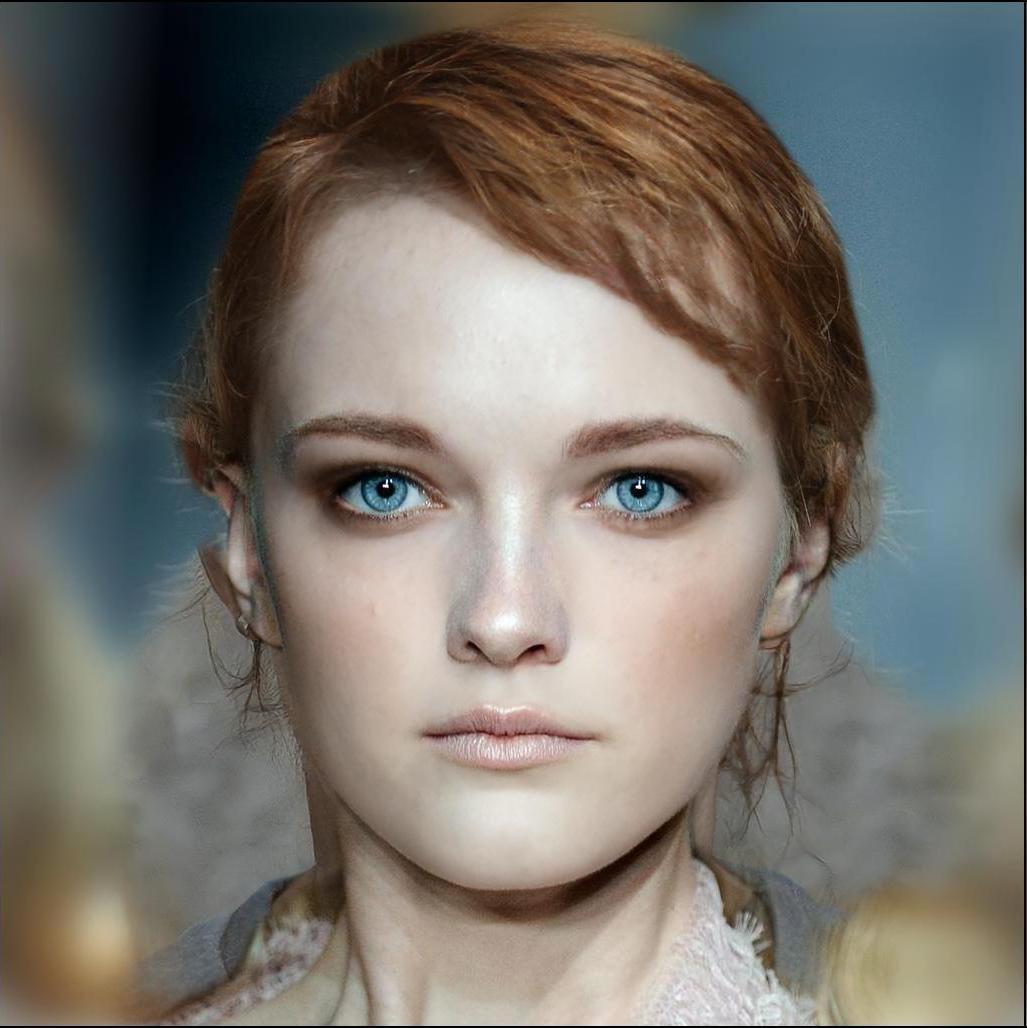} &  \im{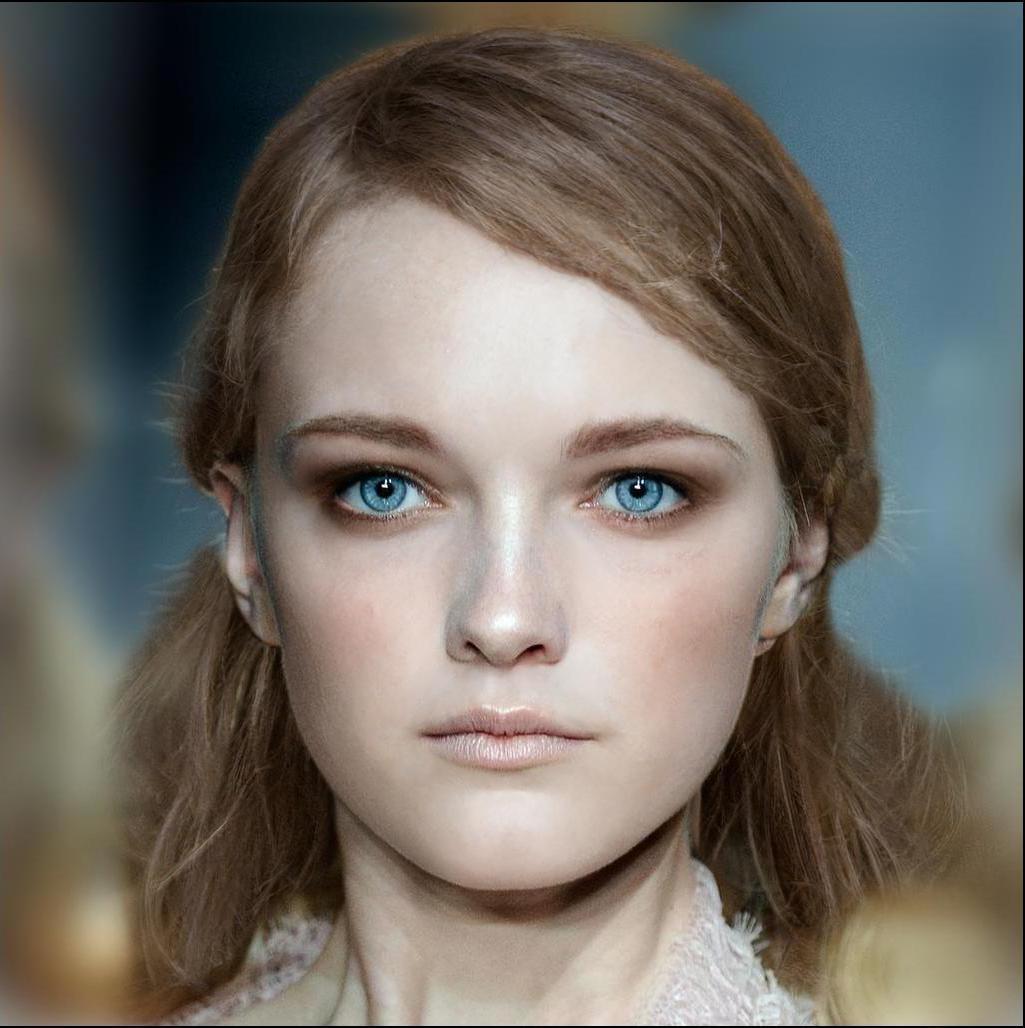}& \turnedtext{-0.8cm}{\footnotesize{Brown hair}}{0mm} \\
\footnotesize{Original} & \footnotesize{$n=0, \lambda_S=0$} & \footnotesize{$\lambda_S=0$} & \footnotesize{MaskFaceGAN} \\
  \end{tabular}
\caption{Qualitative results of the ablation study. The figure shows (left to right): $(i)$ original images, $(ii)$  results without noise optimization and without the shape constraint, $(iii)$ effect of the noise optimization, and $(iv)$ results with noise optimization and the shape term enabled.\vspace{-1mm}}
\label{fig:ablation}
\end{figure}

\textbf{Component Size Manipulation.}
MaskFaceGAN can be adapted to include additional constraints. An example constraint is 
the desired size of the facial component being manipulated. This is done by including the size manipulation objective from Eq.~(\ref{eq:size_manipulation}) in the overall optimization objective in Eq. (\ref{eq:final}). In Fig. \ref{fig:size_manipulation} we display results when specifying the portions of the original component size for $\alpha = \{0.5, 1.0, 1.5\}$. The presented examples show MaskFaceGAN's capability to  change the size of 
facial attributes in a photo realistic manner. 


\textbf{Combining Editing Constraints. }
Multiple attribute editing, intensity control and component size manipulation can also be used simultaneously to change several aspects of the input face image with a single application of MaskFaceGAN. Fig. \ref{fig:simultaneous_all} presents an example, where various aspects of the ``Wearing lipstick'' and ``Bushy eyebrows'' attributes are manipulated. Note that despite considerable changes to different facial attributes, the results still appear visually convincing.

\subsection{Ablation Study}

To evaluate the impact of different MaskFaceGAN components on the editing quality, we perform an ablation study on CelebA-HQ. 
Specifically, we focus on two major components: 
$(i)$ the shape constraint from Eq.~\eqref{eq:seg}, and $(ii)$ the noise optimization procedure. We note that the shape term only affects the hair region and does not impact other attributes. 

\textbf{Qualitative Analysis.}
Fig. \ref{fig:ablation} demonstrates the effect of different MaskFaceGAN settings. When the noise component is not optimized ($n=0$), the edited images contain low frequency image areas, which is most apparent in the hair region, as shown in the second column of Fig. \ref{fig:ablation}. Similarly, the skin region is also missing details, e.g., beauty marks. The noise optimization ensures that such facial details 
are present in the image, as can be seen in the third column of Fig. \ref{fig:ablation}. 

The absence of the shape term ($\lambda_S = 0$) results in suboptimal blending when dealing with hair modifications. In such settings, the  background synthesized by the generator ($G$) is blended with the original background, resulting in  unconvincing results with visible artefacts. The optimization of this term assures that the generator model considers information about the shape of the hair region during the synthesis step and produces photo realistic editing outputs.

\textbf{Quantitative Analysis.}
For a quantitative analysis of the ablation results, we report in Table \ref{tab:ablation} mean FID scores computed over the test images of CelebA--HQ dataset and averaged over all attributes.  Interestingly, the largest gain is obtained by the noise optimization procedure. Enabling the shape term to ensure blending consistency results in additional FID gains. We hypothesize that these gains are a consequence of visually more convincing images, as shown in  Fig. \ref{fig:ablation}

\setlength{\tabcolsep}{6pt} 
\renewcommand{\arraystretch}{1.3}  

\begin{table}[t]
\caption{Quantitative results of the ablation study. FID scores computed on CelebA-HQ  averaged over all attributes are reported (lower is better). 
}
\label{tab:ablation}
\centering
    \begin{tabular}{l|r}
    \hline\hline
     \textbf{MaskFaceGAN variant} & \textbf{FID} \\
     \hline
     Local latent code optimization, no shape term & $59.90$ \\
     + noise optimization & $34.86$ \\
     + shape term (complete MaskFaceGAN) & $\mathbf{32.56}$\\
      \hline\hline
    \end{tabular}\vspace{-2mm}
\end{table}

\subsection{Component Analysis}
The optimization procedure designed for MaskFaceGAN depends on the utilized attribute classifier ($C$) and face parser ($S$). In this section, we analyze the impact of these two components on the editing results.

\textbf{Attribute Classifier.}
To explore the impact of the attribute classifier $C$, we implement $5$ additional models in addition to the tree-like classifier from~\cite{vandenhende2019branched} that is used in the original MaskFaceGAN design:  two attribute classifiers based on the VGG architecture \cite{vgg}, i.e., VGG16 and VGG19, and three ResNet-based \cite{resnet} models, i.e.,  ResNet34, ResNet50 and ResNet101. These models come with different designs and complexity (in terms of parameters). We modify the models to predict multiple attributes by constructing a dedicated fully connected layer at the top with one output for each attribute. The models are trained on the CelebA dataset using the same optimization procedure and learning rate
schedule as utilized with the original tree-like classifier~\cite{vandenhende2019branched}. Table \ref{tab:ablation-models-classifiers} shows the classification accuracy (averaged across all attributes) on the CelebA test split achieved by the different models. We observe that despite differences in design and complexity most models weigh in at a classification accuracy around $85\%$, whereas the tree-like classifier performs slightly better. 

\renewcommand{\arraystretch}{1.3}  
\setlength{\tabcolsep}{2.3pt} 
\begin{table}
	\caption{Average prediction accuracy on  CelebA for the attribute classifiers selected for the component analysis.}
\label{tab:ablation-models-classifiers}
\resizebox{\columnwidth}{!}{%
\begin{tabular}{l|cccccc}
\hline
\hline
	\textbf{Model} & \textbf{VGG16} & \textbf{VGG19} & \textbf{ResNet34} & \textbf{ResNet50} & \textbf{ResNet101} & \textbf{Ours} \\
\hline
	\textbf{Accuracy} & $85.4 \%$ & $85.5 \% $ & $85.8 \%$ & $85.8 \%$ & $85.6 \% $ & $90.1 \%$ \\
\hline
\hline
\end{tabular}
}
\end{table}
\renewcommand{\im}[2]{\includegraphics[width=0.131\columnwidth]{images/component_analysis/attr_classifiers/#1/#2.jpg}}
\renewcommand{\imrow}[1]{\im{#1}{orig_img} & \im{#1}{vgg16} & \im{#1}{vgg19} & \im{#1}{resnet34} & \im{#1}{resnet50} & \im{#1}{resnet101} & \im{#1}{branchedtiny}}

\renewcommand{\arraystretch}{1.}  
\setlength{\tabcolsep}{1.5pt} 

\renewcommand{\miniwidth}{0.123\columnwidth}
\newcolumntype{C}{ >{\centering\arraybackslash} m{\miniwidth} }
\newcolumntype{D}{ >{\centering\arraybackslash} m{\miniwid} }

\begin{figure}[t]
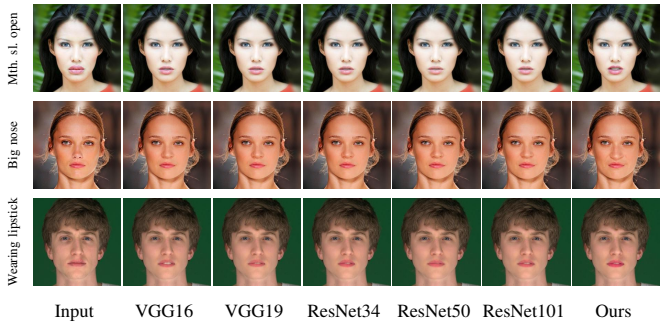

    \begin{tabular}{DCCCCCCC}
        \begin{turn}{90}{\tiny Mth. sl. open} \end{turn} & \imrow{030} \\
        \begin{turn}{90}{\tiny Big nose} \end{turn} & \imrow{044} \\
        \begin{turn}{90}{\tiny Wearing lipstick} \end{turn} & \imrow{337} \\ \vspace{-3mm}
                        &  \scriptsize Input &  \scriptsize VGG16 & \scriptsize VGG19 & \scriptsize ResNet34 & \scriptsize ResNet50 & \scriptsize ResNet101 & \scriptsize Ours \\
    \end{tabular}
    \caption{Impact of different attribute classifiers on the generated edits. For all results, the smoothing factor is set to $\epsilon=0.05$. The tree-based attribute classifier used in MaskFaceGAN \cite{vandenhende2019branched} leads to more expressive semantics in the final image compared to other attribute classification models.}\vspace{-2mm}
    \label{fig:component_analysis-classifiers}
\end{figure}

In Fig.~\ref{fig:component_analysis-classifiers} we investigate how the characteristics of the attribute classifiers impact results. As can be seen, all classifiers lead to semantically meaningful editing outputs and, even though their performance is close, they produce slight variations in the appearance of the targeted attributes - see, for example, the mouth region in the bottom row of Fig.~\ref{fig:component_analysis-classifiers}. This observation can be attributed to the different model topologies and differences in the gradients produced during the optimization procedure. Furthermore, it can be noted that the results generated with the tree-based attribute classifier (marked Ours) contain the most expressive semantic content with the most pronounced targeted attributes. We ascribe this fact to the superior classification performance of our attribute classifier, which consequently provides better gradient information compared to the competing models. 


\textbf{Face Parser.}
MaskFaceGAN uses DeepLabv3 \cite{deeplabv3} to implement the face parser ($S$). In this section, we analyze editing results produced with 3 other parsers. We select the following competing models for the comparison: UNet \cite{unet} and FCN \cite{fcn} with two different backbones, i.e., FCN--ResNet50 and FCN--ResNet101, and train them on  CelebA-HQ  using the same procedure/protocol and optimization method as with DeepLabv3.  
The performance in terms of the average Intersection over Union (IoU)~\cite{rot2020deep} over the test split of the data is reported in Table \ref{tab:ablation-models-parsers}. Note that the average IoU scores (in $\%$) are close and vary from around $75\%$ to up to $77\%$.  

A few qualitative editing results produced with the different parsers are shown in Fig.~\ref{fig:component_analysis-parsers}. We find the results to be less affected by changes in the parser than by changes in  the attribute classifier. The observed differences mostly consist of slight variations in the shape and look of the targeted facial attribute. However, all edits are still semantically reasonable and visually convincing.

\renewcommand{\arraystretch}{1.3}  
\begin{table}[t]
	\caption{Average Intersection over Union (IoU) [in \%] over the CelebA-HQ test data for the selected face parsers.}
\label{tab:ablation-models-parsers}
\resizebox{\columnwidth}{!}{%
\begin{tabular}{ll|cccccccc}
\hline
\hline
	\textbf{Model} && \textbf{UNet} && \textbf{FCN--ResNet50} && \textbf{FCN--ResNet101} && \textbf{DeepLabv3 (ours)} \\
\hline
	\textbf{IoU} && $75.34 \%$ && $77.18 \%$ && $76.70 \%$ && $77.24 \%$ \\
\hline
\hline
\end{tabular}
}
\end{table}

\renewcommand{\arraystretch}{0.5}  
\renewcommand{\im}[2]{\includegraphics[width=0.18\columnwidth]{images/component_analysis/parsers/#1/#2.jpg}}
\renewcommand{\imrow}[1]{\im{#1}{orig_img} & \im{#1}{unet} & \im{#1}{fcn_resnet50} & \im{#1}{fcn_resnet101} & \im{#1}{deeplabv3}}
\newcommand{\imrowmask}[1]{ & \im{#1}{unet-blendmask} & \im{fcn_resnet50-blendmask} & \im{#1}{fcn_resnet101-blendmask} & \im{#1}{deeplabv3-blendmask}}

\renewcommand{\miniwidth}{0.18\columnwidth}

\begin{figure}[t]
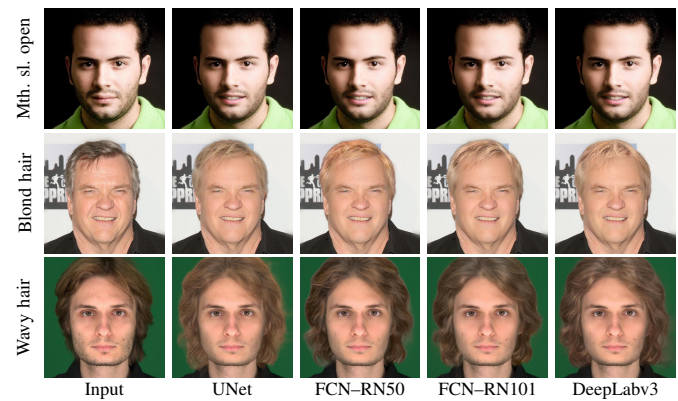

    \begin{tabular}{DCCCCC}
    \begin{turn}{90}{\scriptsize{Mth. sl. open}} \end{turn} & \imrow{025} \\
    \begin{turn}{90}{\scriptsize{Blond hair}} \end{turn} & \imrow{086} \\
    \begin{turn}{90}{\scriptsize{Wavy hair}} \end{turn} & \imrow{322} \\
    & \scriptsize{Input} & \scriptsize{UNet} & \scriptsize{FCN--RN50} & \scriptsize{FCN--RN101} & \scriptsize{DeepLabv3} \\
    \end{tabular}
    \caption{Impact of different face parsers on the generated edits. The impact on most attributes is negligible, as illustrated, for example, by the ``Mouth slightly open'' edit in the first row. When the target-region shape loss term is used (hair edits in the second and third row), minute differences can be observed in the generated images.}
    \label{fig:component_analysis-parsers}\vspace{-2mm}
\end{figure}

\subsection{Optimization Procedure and Blending}

\renewcommand{\figureimagewidth}{.0776\textwidth}
\renewcommand{\im}[2]{\includegraphics[width=\figureimagewidth]{images/intermediate_images/#1/#2.jpg}}
\renewcommand{\miniwidth}{\figureimagewidth}
\setlength{\tabcolsep}{0.5mm} 
\renewcommand{\arraystretch}{0.5}  

\renewcommand{\imrow}[1]{\im{#1}{orig-img} & \im{#1}{img-w-0000} & \im{#1}{img-w-0200} & \im{#1}{img-w-0400} & \im{#1}{img-w-0600} & \im{#1}{img-w-0800} & \im{#1}{img-n-0000} & \im{#1}{img-n-0200} & \im{#1}{img-n-0400} & \im{#1}{img-n-final} & \im{#1}{blend-mask} & \im{#1}{result}}

\begin{figure*}
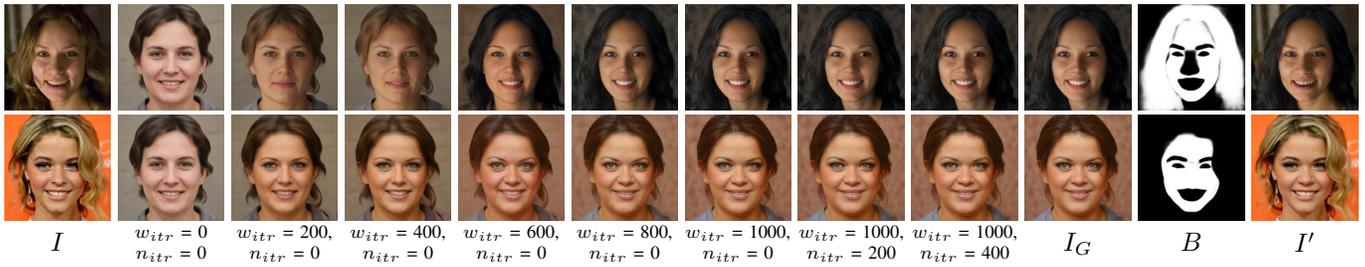

\begin{tabular}{CCCCCCCCCCCC}
	\imrow{0122} \\
	\imrow{0165} \\
    $I$ & \scriptsize $w_{itr}$ = 0 $n_{itr}$ = 0 & \scriptsize $w_{itr}$ = 200, $n_{itr}$ = 0 & \scriptsize $w_{itr}$ = 400, $n_{itr}$ = 0 & \scriptsize $w_{itr}$ = 600, $n_{itr}$ = 0 & \scriptsize $w_{itr}$ = 800, $n_{itr}$ = 0 & \scriptsize $w_{itr}$ = 1000, $n_{itr}$ = 0 & \scriptsize $w_{itr}$ = 1000, $n_{itr}$ = 200 & \scriptsize $w_{itr}$ = 1000, $n_{itr}$ = 400 & $I_G$ & $B$ & $I'$ \\
\end{tabular}
\caption{Evolution of the intermediate $G(w,n)$ results throughout the optimization procedure. The leftmost image shows the input $I$. The next examples show the intermediate  results for $G(w,n)$, where the numbers at the bottom correspond to the current optimization iteration of $w$ and $n$, that is, $w_{itr}$ and $n_{itr}$, respectively. The initial image ($w_{itr}=0,n_{itr}=0$) is always the same. During the $w$ latent code optimization procedure, the face shape and target-attribute appearance is adjusted in accordance with the considered constraints. After convergence, the noise component is optimized to introduce realistic fine image details. The final optimization result $I_G$ is then blended based on $B$ to generate the output image $I'$. 
The attributes edited in the presented examples are ``Black hair'' (top) and ``Big nose'' (bottom).}\vspace{-2mm}
\label{fig:intermediate_steps}
\end{figure*}

In this section, we study the optimization procedure and blending step used in MaskFaceGAN to provide better insight into their behavior and impact on the final results.

\textbf{Optimization Procedure.} A two-step process is used in MaskFaceGAN to find the optimal latent representation of the desired output image $G(w,n)$ given the constraints enforced through the attribute classifier and face parser. This process first optimizes the latent code $w$, and, after convergence, freezes the code and optimizes the high-frequency details encoded through the noise component $n$. In Fig.~\ref{fig:intermediate_steps}, we illustrate the evolution of the generated (intermediate) image $G(w,n)$ across different iterations of the optimization procedure for two attributes, i.e., ``Black hair'' and ``Big nose'' together with the computed blending mask and final output. Note how the first steps  gradually adjust the shape of the targeted facial region and introduce the desired semantics, while the latter steps  ensure that high-frequency details are present that contribute towards the realism of the generated images.

\textbf{Blending.} Next, we explore the impact of different blen\-ding-mask definitions, which lead to different trade-offs in the generated images. With MaskFaceGAN, the skin-region $S_{skin}(I)$ and targeted-attribute area $S_{tar}$ are used as the basis for the blending process. 
However, an important consideration here is whether the inclusion of $S_{skin}$ is truly required. Given that the GAN inversion cannot perfectly embed the skin region with the MMSE loss from Eq.~\eqref{eq:mse}, a blending mask without the skin region, such as $\hat{B}_1 = S_{tar}(I)$ or $\hat{B}_2 = S_{tar}(I_G)$ could also be used for the blending operation.

\setlength{\tabcolsep}{0.8mm} 
\renewcommand{\figureimagewidth}{0.14\columnwidth}
\renewcommand{\im}[2]{\includegraphics[width=\figureimagewidth]{images/blending_mask_choice/#1/#2.jpg}}
\renewcommand{\imrow}[1]{\im{#1}{imgorig} & \im{#1}{imggen} & \im{#1}{blendright} & \im{#1}{imgright} & \im{#1}{blendwrong} & \im{#1}{imgwrong}}
\renewcommand{\miniwidth}{\figureimagewidth}

\begin{figure}
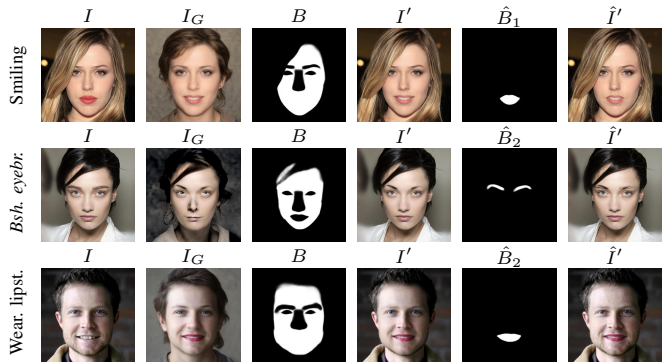

\begin{tabular}{DCCCCCC}
         &\scriptsize  $I$ &\scriptsize  $I_G$ &\scriptsize  $B$ &\scriptsize  $I'$ &\scriptsize  $\hat{B}_1$ &\scriptsize  $\hat{I}'$ \\
        \turnedtextbegin{\scriptsize Smiling} & \imrow{worse2/613--smiling} \\ \vspace{-2mm}
         &\scriptsize  $I$ &\scriptsize  $I_G$ &\scriptsize  $B$ &\scriptsize  $I'$ &\scriptsize  $\hat{B}_2$ &\scriptsize  $\hat{I}'$ \\ \vspace{-4mm}
        \turnedtextbegin{\scriptsize \hspace{-4mm} {\textit{Bsh. eyebr.}}} & \imrow{worse/bushy_eyebrows/5} \\ \vspace{-2mm}
         &\scriptsize  $I$ &\scriptsize  $I_G$ &\scriptsize  $B$ &\scriptsize  $I'$ &\scriptsize  $\hat{B}_2$ &\scriptsize  $\hat{I}'$ \\ \vspace{-4mm}
        \turnedtextbegin{\scriptsize \hspace{-4mm} Wear. lipst.} & \imrow{better/91} \\
\end{tabular}
    \caption{Analysis of the blending procedure with different blending masks. From left to right: The input image $I$, the optimized intermediate results $I_G$, the original blending mask $B$ and output image $I'$, the alternative blending mask ($\hat{B}_1$ or $\hat{B}_2$), and the alternative output $\hat{I}'$. Note how different definitions lead to different trade-offs in the results.} 
		\label{fig:blending_options}
\end{figure}

In Fig.~\ref{fig:blending_options}, we show a few illustrative examples, where such masks are utilized, as well as comparative results with the original process using $B$. In the top row, the blending procedure with $\hat{B}_1$ leads to poor semantics because the targeted region (the mouth) in $I_G$ has grown and changed shape during the optimization process. In the second row, the blending based on $\hat{B}_2$ produces visual artifacts in the blended output $\hat{I}'$, because of shrinkage of the targeted facial region (i.e., the eyebrows) in $I_G$. In the bottom row, the targeted facial area does not change in shape or size and the exclusion of the skin area in $\hat{B}_2$ leads to editing results with better high-frequency content compared to what is generated with MaskFaceGAN. Nonetheless, both results, $I$ and $I'$, are visually convincing. 

While different formulations of blending masks could be defined for specific tasks in MaskFaceGAN, this would introduce an additional layer of complexity to the model. MaskFaceGAN, therefore, opts for an approach that supports attribute growth and shrinking without being overly complex.

\subsection{Global Editing}
\begin{figure}[t]
\centering
 \includegraphics[width=0.8\columnwidth]{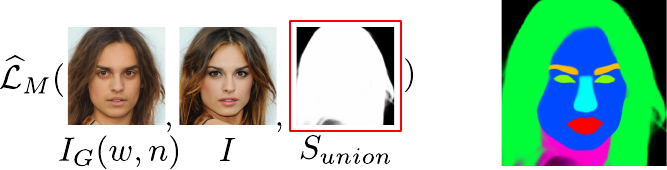}
\caption{MaskFaceGAN uses a modified appearance-preser\-vation constraint to enable editing of global image characteristics. The constraint (left) is based on the LPIPS perceptual similarity \cite{zhang2018unreasonable} and is applied over the entire face region, shown in parsed form on the the right.}
\label{fig:new constrain}
\end{figure}

While the primary purpose of MaskFaceGAN is editing of local attributes that correspond to specific facial regions, the approach can also be extended towards editing of global facial characteristics. To facilitate global editing, we relax the original appearance-preservation constraint from Eq.~\eqref{eq:mse}, so it allows for global image changes. Specifically, instead of using the MMSE-based constraint over the skin area to preserve the initial image appearance, we introduce a perceptual loss over the whole face region, as illustrated in Fig.~\ref{fig:new constrain}. The perceptual loss allows us to preserve high-level semantics of the original image, while providing room for global appearance modifications. Formally, the modified appearance-preservation constraint is defined as: 
\begin{equation}
\begin{split}
     \mathcal{\widehat{L}}_M &=  \sum_{l=1}^L \Big| \Big| S_{union}^l(I) \odot \bigl( \phi^l (I_G) - \phi^l (I) \bigr) \Big| \Big|^2_2
\end{split}
\end{equation}
where $\phi^l(\cdot)$ are LPIPS \cite{zhang2018unreasonable} activations from the $l-th$ layer of a pretrained VGG network, $S_{union}^l$ is a mask constructed through the union of all facial regions returned by the face parser (see right part of Fig.~\ref{fig:new constrain}), downscaled to match the spatial resolution of $l$-th activations. Similarly to the LPIPS reference implementation, we use $L=5$ layers to compute the Learned Perceptual Image Patch Similarity (LPIPS).  The generated image is then blended with the original based on $S_{union}$, and not $S_{skin}$. All other part of the optimization framework are kept unchanged.  

To demonstrate the global editing capabilities of MaskFaceGAN, we select two challenging attributes that affect the whole facial area, i.e., ``Young'' and ``Male'', and show some sample results in Fig. \ref{fig:global_editing}. We again compare the results to the competitors, StarGAN, AttGAN, STGAN and InterFaceGAN. We observe that StarGAN, AttGAN, and STGAN are able to enforce the desired semantics to some extent, but especially with images with cluttered background also often introduce visual artefacts. InterfaceGAN and MaskFaceGAN generate higher-resolution results, again with the desired semantic content. However, while the targeted global attributes are clearly expressed with InterfaceGAN, much of the correspondence with the input image is lost. MaskFaceGAN, on the other hand, produces the desired semantic content but also better preserves correspondences with the initial input image both in terms of facial appearance as well as background.

\renewcommand{\figureimagewidth}{.14\columnwidth}
\renewcommand{\im}[1]{\includegraphics[width=\figureimagewidth]{images/global_editing/#1.jpg}}
\renewcommand{\imrow}[2]{\im{#1--orig} & \im{#1--#2--stargan} & \im{#1--#2--attgan} & \im{#1--#2--stgan} & \im{#1--#2--interface} & \im{#1--#2--maskfacegan}}

\begin{figure}[t]
\begin{tabular}{DCCCCCC}
\turnedtextbegin{\scriptsize Male} & \imrow{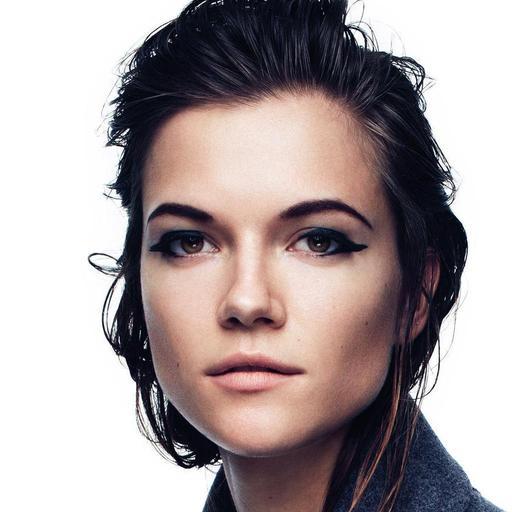}{male} \\
\turnedtextbegin{\scriptsize Male} & \imrow{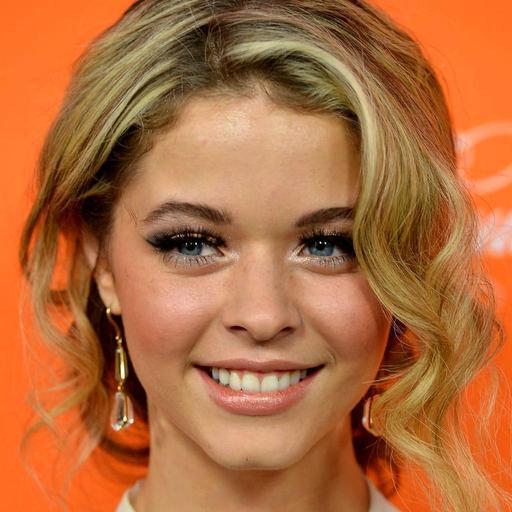}{male} \\
\turnedtextbegin{\scriptsize Young} & \imrow{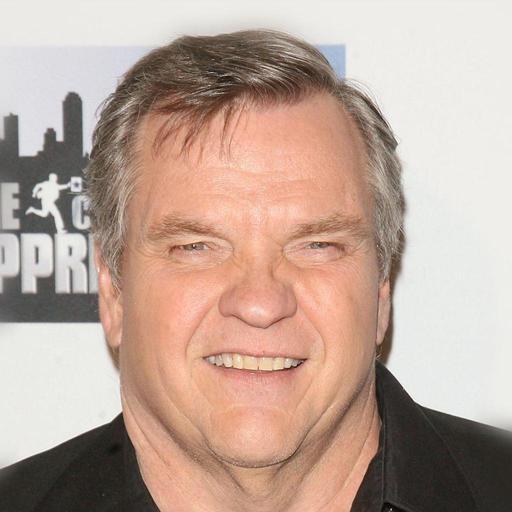}{young} \\
\turnedtextbegin{\scriptsize \textit{Young}} & \imrow{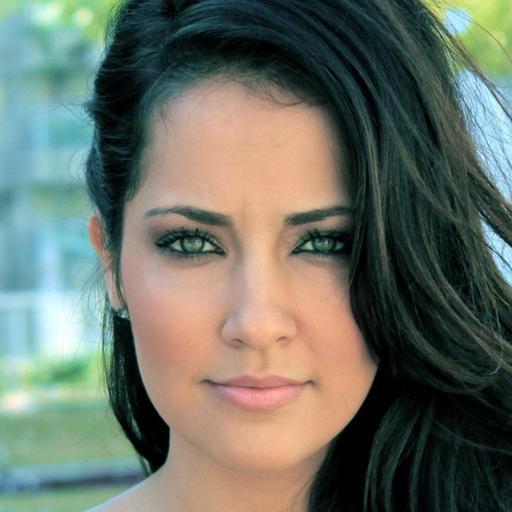}{young} \\
& \scriptsize Input & \scriptsize StarGAN & \scriptsize AttGAN & \scriptsize STGAN & \scriptsize \hspace{-2mm} InterFaceGAN & \scriptsize Ours \\
\end{tabular}
\caption{Illustration of the global editing capabilities of MaskFaceGAN for the ``Young'' and ``Male'' attributes. 
Compared to the competitors, MaskFaceGAN generates higher-resolution edits without visual artefacts and better preserves correspondences with the input image (appearance characteristics, background), while still producing the desired global semantics.}
\label{fig:global_editing}
\end{figure}

\subsection{Limitations}
The results presented so far show that MaskFaceGAN generates competitive (high--quality) editing results when compared to state--of--the--art models from the literature. Nevertheless, the approach still exhibits a number of limitations.

MaskFaceGAN is based on gradient optimization that takes between $2$ and $5$ minutes per image on a GeForce GTX $1080$. In comparison with encoder--decoder methods that are capable of editing images in milliseconds, the proposed approach is slower by orders of magnitude. However, when compared to related methods, e.g., InterFaceGAN \cite{shen2021interfacegan}, the local embedding procedure requires considerably fewer steps to converge. 

\begin{figure}[t]
\captionsetup[subfigure]{labelformat=empty}    
\centering
    \begin{minipage}[b]{0.3\columnwidth}
    		\centering
    		\includegraphics[width=0.99\textwidth,height=0.99\textwidth]{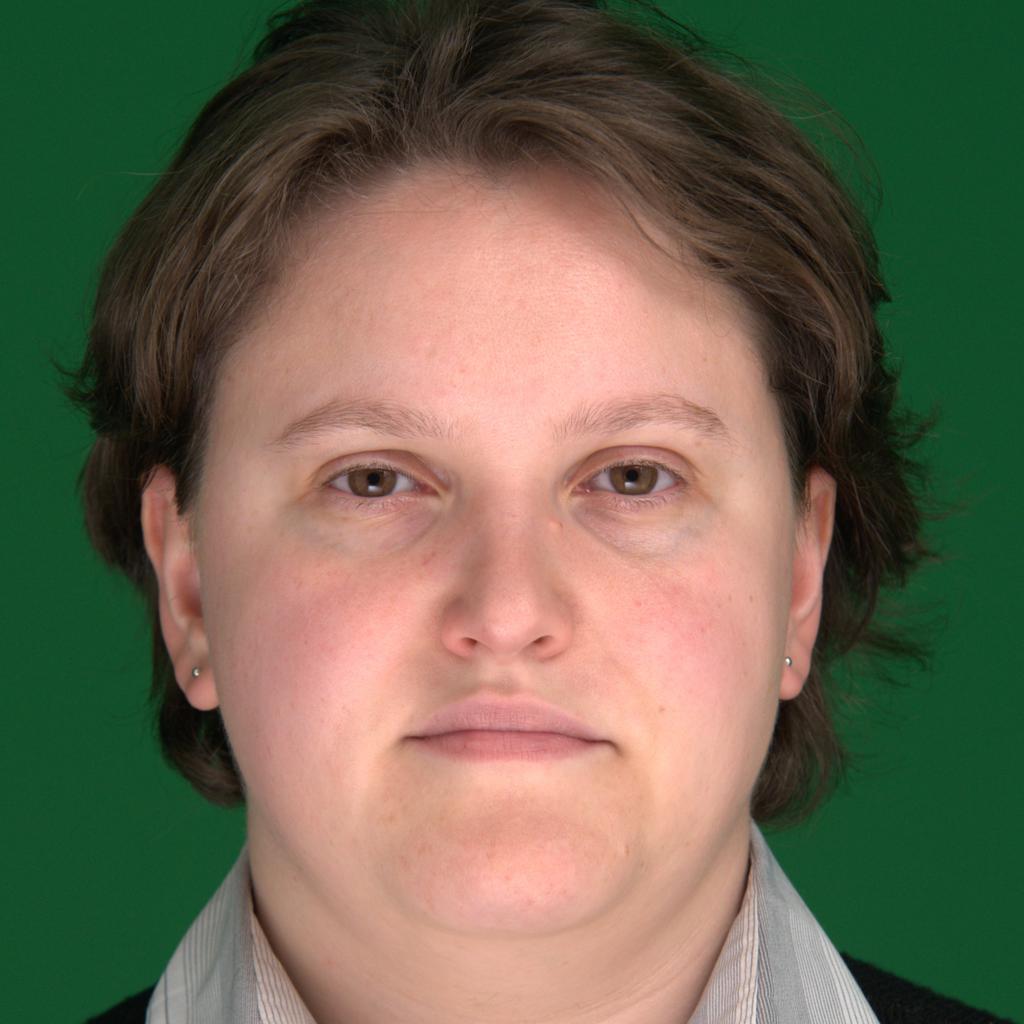}\\[1mm] \includegraphics[width=0.99\textwidth,height=0.99\textwidth]{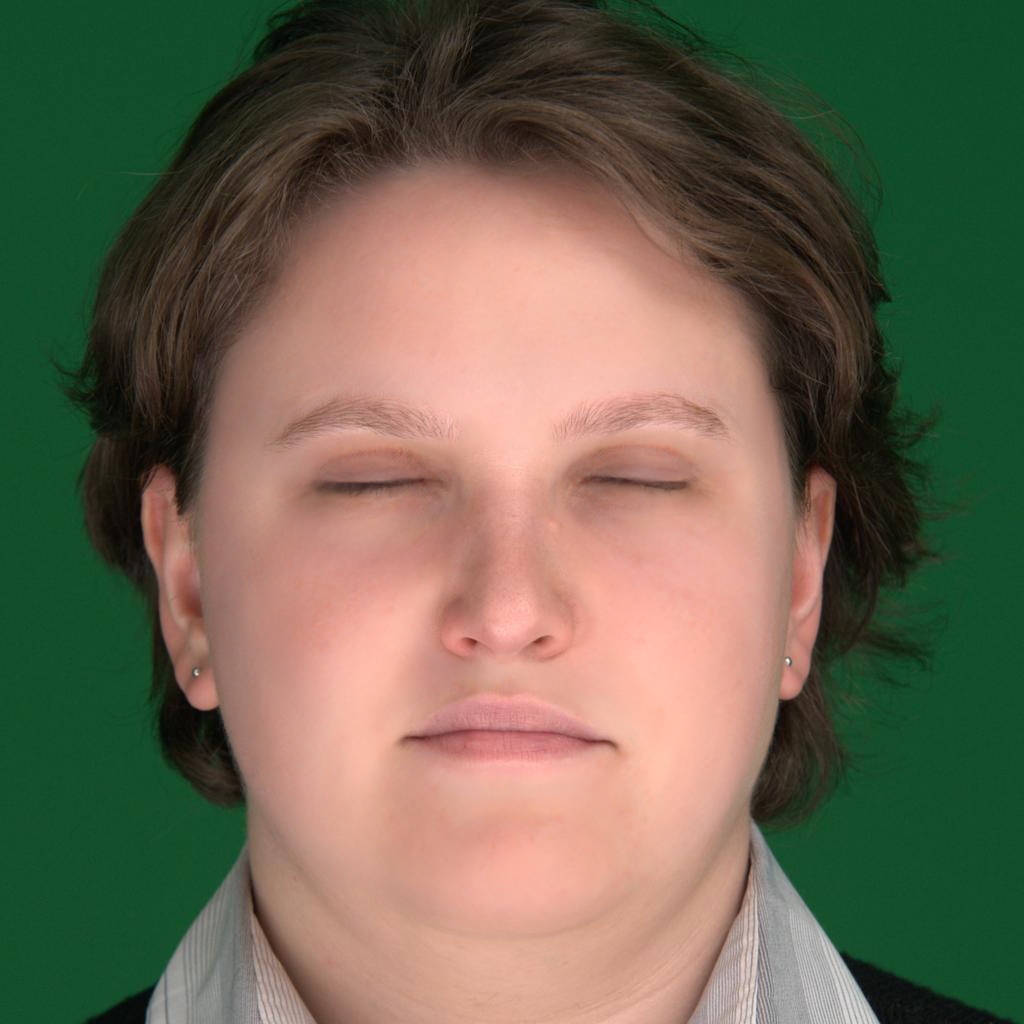}\\
    		{\footnotesize Narrow eyes}
    	\end{minipage} 
    	\begin{minipage}[b]{0.3\columnwidth}
    		\centering
\includegraphics[width=0.99\textwidth,height=0.99\textwidth]{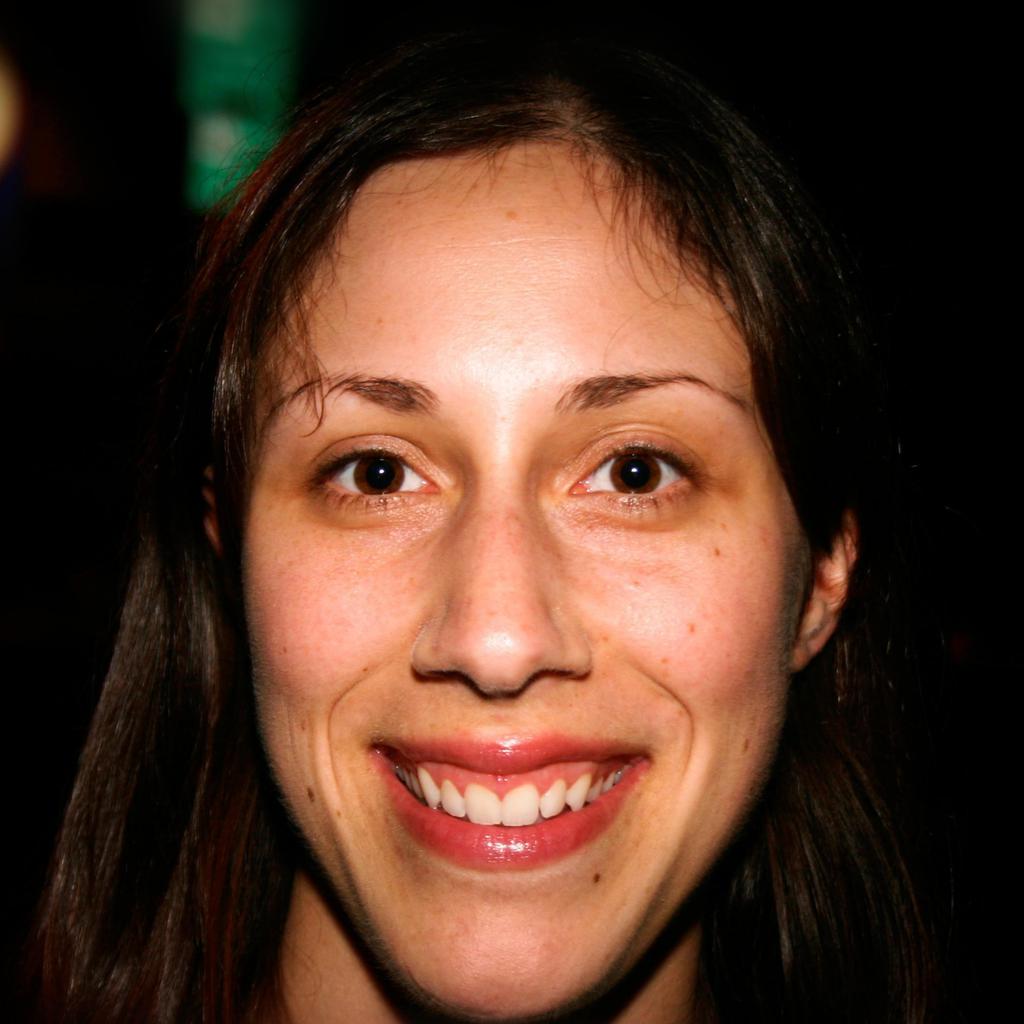}\\[1mm] \includegraphics[width=0.99\textwidth,height=0.99\textwidth]{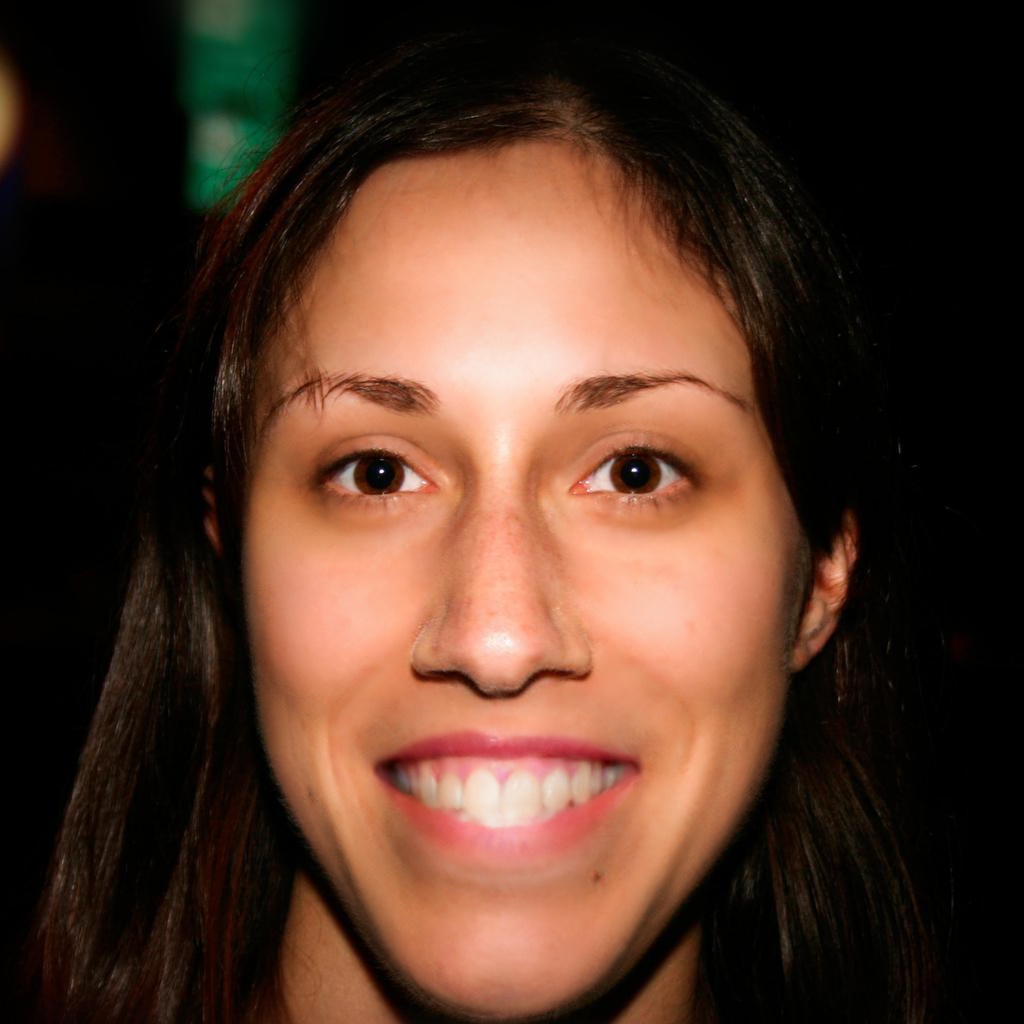}\\ 
    {\footnotesize \textit{Smiling}}
    	\end{minipage} 
    	 \begin{minipage}[b]{0.3\columnwidth}
    		\centering
\includegraphics[width=0.99\textwidth,height=0.99\textwidth]{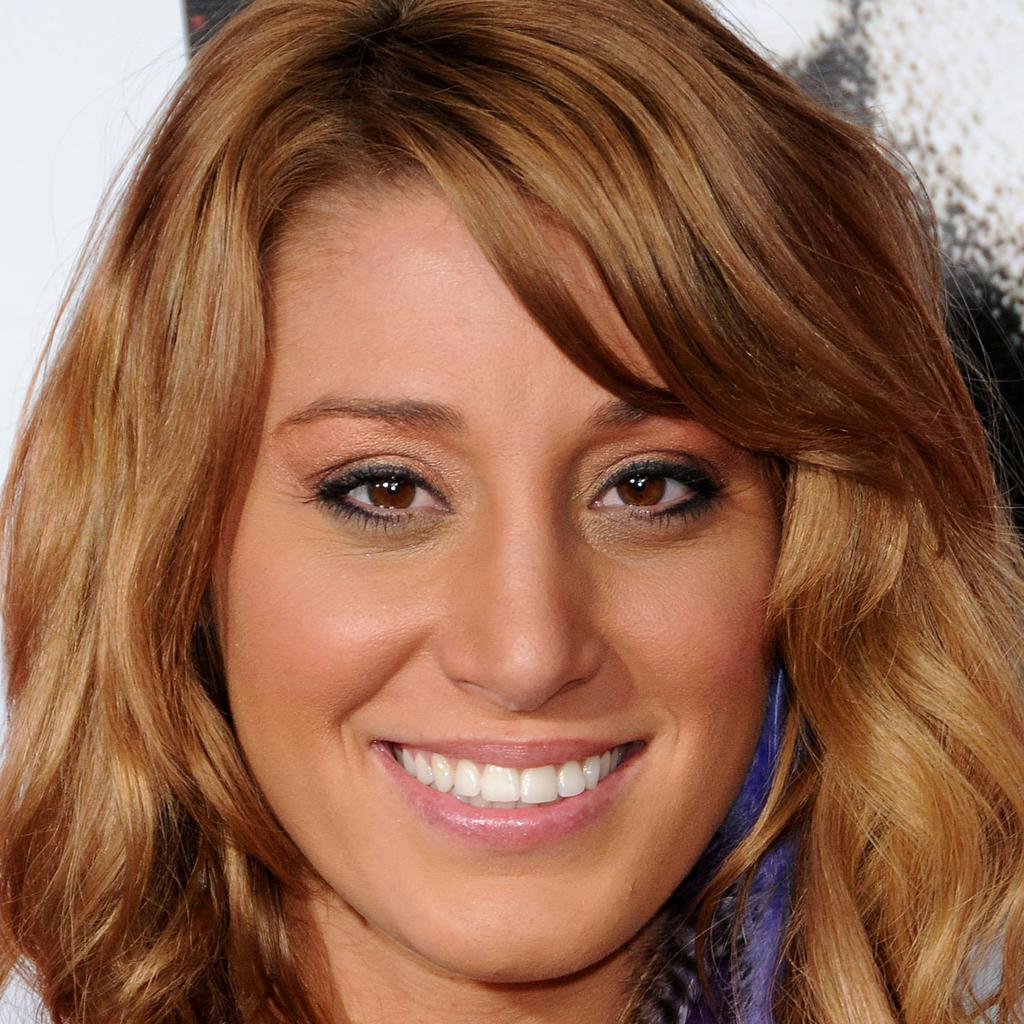}\\[1mm] \includegraphics[width=0.99\textwidth,height=0.99\textwidth]{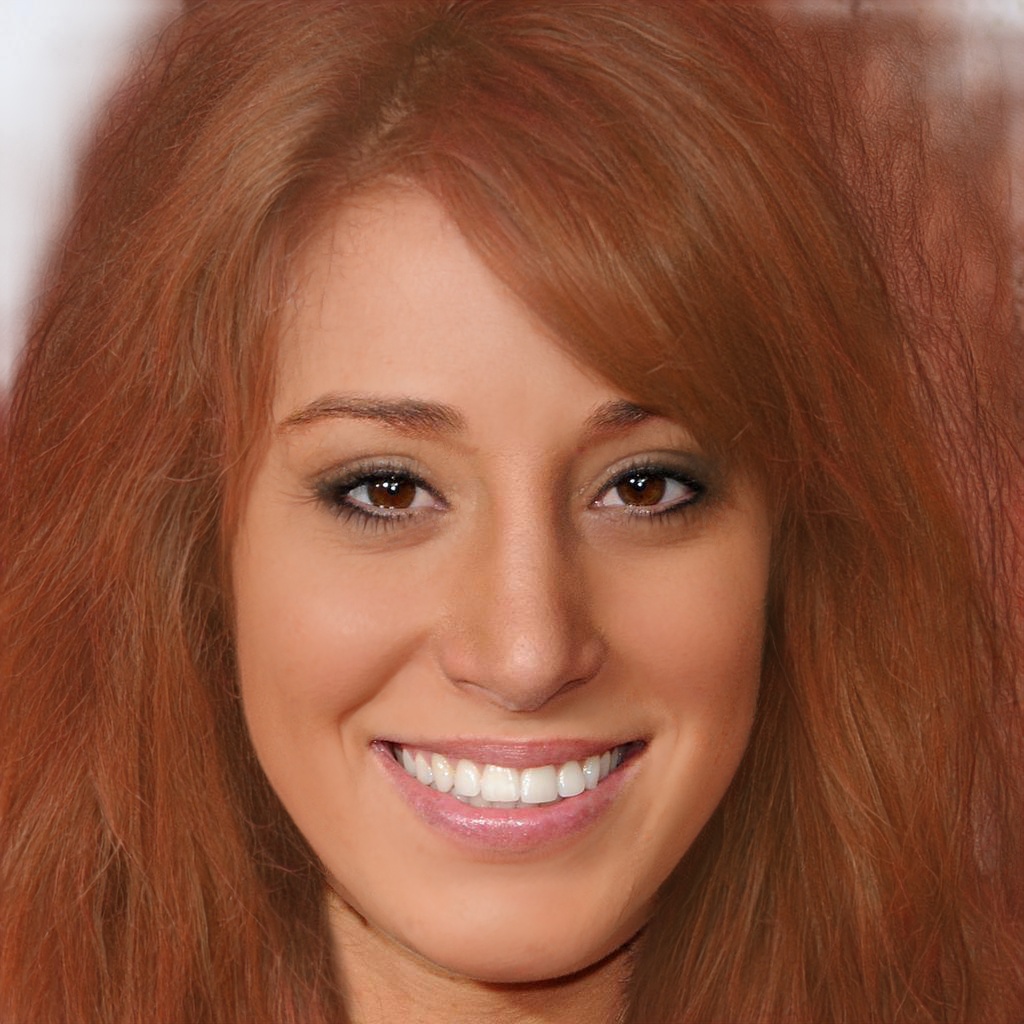} 
{\footnotesize Straight hair}
    	\end{minipage} 
    	\caption{Examples of MaskFaceGAN limitations. The original input images on the top are edited according to the listed target attributes. MaskFaceGAN is affected by the performance of the attribute classifier ($C$) and the face parser ($S$). Difficulties with these components are reflected in the editing results. 
    	\vspace{-3mm}}
    	\label{fig:limitations}
\end{figure}

MaskFaceGAN  relies on an attribute classifier ($C$) to steer the editing process. While this is an effective way of controlling the semantic content 
in the edited image, it may produce inconsistent results for certain attributes. Our user study showed (see Fig.~\ref{fig:barplot}) that especially for the ``Narrow eyes'' attribute MaskFaceGAN does not outperform STGAN and AttGAN in terms of user scores. An analysis of this observation showed that MaskFaceGAN exhibits a tendency to close the eyes instead of trying to narrow them, 
as illustrated in the first column of Fig.~\ref{fig:limitations}. While the edited image still looks convincing, such inconsistencies represent one of the limitations of MaskFaceGAN.

The second source of errors 
is the face parser ($S$). For images, where $S$ produces incorrect parsing results, the editing procedure operates with inappropriate spatial constraints and results in image changes in (partially) incorrect  regions. A couple of examples of such editing results are presented in the second and third column of Fig.~\ref{fig:limitations}, where the ``Smiling'' and ``Straight hair'' attributes were considered. 

\renewcommand{\miniwidth}{0.3\columnwidth}
\renewcommand{\arraystretch}{1.3}  

\section{Conclusion}
In this paper, we introduced MaskFaceGAN, a novel approach to high--resolution face image editing. At the core of the approach is a GAN latent code optimization procedure that generates targeted image regions in accordance with spatial and semantic constraints, enforced 
by pre--trained face parsing and classification networks. Through rigorous experiments on three face datasets, MaskFaceGAN was shown to convincingly alter a wide variety of facial attributes and ensure 
competitive performance when compared to the state-of-the-art. 
Additionally, the approach was demonstrated to enable unique editing characteristics, including attribute intensity control and component size manipulation.

\bibliographystyle{IEEEtran}
\bibliography{bibliography}  

\newpage
\title{MaskFaceGAN: High Resolution Face Editing with Masked GAN Latent Code Optimization --
Supplementary Material}
\author{Martin~Pernuš,~\IEEEmembership{Student Member,~IEEE,}
        Vitomir~Štruc,~\IEEEmembership{Senior Member,~IEEE,}
        and~Simon~Dobrišek,~\IEEEmembership{Member,~IEEE}}
\date{}
\maketitle

\begin{abstract}
In the main part of the paper, we introduced MaskFaceGAN for face editing and presented several experiments to demonstrate its capabilities. This \textit{Supplementary material} provides further details on the proposed approach and reports additional results to highlight its merits. Specifically, the supplementary material: $(i)$ elaborates on the importance and effect of the local GAN inversion process used by MaskFaceGAN, $(ii)$ further motivates the use of spatial constraints for face attribute editing, $(iii)$ presents additional editing results for single attribute editing, attribute intensity control, component size manipulation, multiple attribute editing and compares editing quality at higher resolution, $(iv)$ reports per--attribute results for the ablation experiments, FID analysis and user study, $(v)$ discusses details with respect to the implementation of the user study, $(vi)$ elaborates on the implementation details of the competing models included in the experimental evaluations, and $(vii)$ provides information on the reproducibility of the experiments.  
\end{abstract}%


\renewcommand{\figureimagewidth}{.3\columnwidth}
\renewcommand{\figuretextwidth}{.02\columnwidth}
\renewcommand\miniwidth{.162\textwidth}

\renewcommand{\figureimagewidth}{.095\textwidth}
\renewcommand{\figuretextwidth}{.08\textwidth}
\setlength{\tabcolsep}{0.5mm} 
\renewcommand{\arraystretch}{0.3}  

\newcolumntype{C}{ >{\centering\arraybackslash} m{\figureimagewidth} }
\newcolumntype{D}{ >{\centering\arraybackslash} m{\figuretextwidth} }

\section{Embedding quality and local GAN inversion}

Unlike most competing approaches to face attribute editing that rely on latent code optimization, e.g., \cite{shen2020interpreting, abdal2019image2stylegan}, MaskFaceGAN does not embed the entire face image into the GAN latent space. Instead, it only embeds facial region(s) needed for editing, which results in higher quality embeddings. In the main part of paper, we showed a visual example to illustrate this characteristic. Here, we report results of a simple experiment aimed at evaluating image quality when embedding smaller image regions and blending the result with the rest of the original input image. 

For this experiment, the target face region is embedded with a masked MSE loss (MMSE), as defined by Eq. (2) in the main part of the paper. We optimize the latent code $w$ for $2000$ iterations, reconstruct the output image, blend it with the original and then report average PSNR, SSIM, MSE and perceptual loss (PL) (computed over conv1-conv5 features of the VGG ImageNet model as in \cite{zhang2018unreasonable}) scores computed over $10$ randomly selected face images from CelebA--HQ. The  results reported Table \ref{tab:embedding} reflect the similarity between the blended and the original input image and (among others) serve as indicators of the visual quality of the outputs of MaskFaceGAN. We note at this point that image manipulation techniques based on GAN inversion need to ensure that the edited images are as close to the originals as possible and that important characteristics, such as the identity of the subject shown, the background and other semantically critical image aspects are captured by the embedding process. 
\begin{table}[t]
\renewcommand{\arraystretch}{1.3}
\setlength{\tabcolsep}{1.5mm}
\caption{Comparison of StyleGAN latent code embedding quality, where the quality is measured by the similarity of the image generated from the optimized latent code and the original input image. The arrows next to the performance scores indicate whether a higher ($\uparrow$) or lower ($\downarrow$) score corresponds to better performance.}
\label{tab:embedding}
\centering
    \begin{tabular}{lrrrr} 
    \hline\hline
     \textbf{Method} & \textbf{MSE $\mathbf{\cdot 10^4}$} $\downarrow$ & \textbf{PSNR} $\uparrow$ & \textbf{SSIM} $\uparrow$ & \textbf{PL} $\downarrow$ \\ 
     \hline 
     Image2StyleGAN++ \cite{abdal2020image2stylegan++} & $89.9$ & $20.87$ & $0.81$ & $0.23$  \\
     MMSE -- face & $24.0$ & $26.73$ & $0.90$ & $0.20$ \\
     MMSE -- no hair & $8.0$ & $31.01$ & $0.97$ & $0.06$ \\
     MMSE -- skin only & $3.9$ & $34.11$ & $0.99$ & $0.05$ \\
     \hline\hline
    \end{tabular}
\end{table}
\renewcommand{\miniwidth}{.22\textwidth}
\newcommand{\razmik}{\hspace{.5cm}}
\begin{figure}[t]
\centering
    \captionsetup[subfigure]{labelformat=empty}    
    \begin{subfigure}{\miniwidth}
    		\centering
    		\includegraphics[width=\textwidth]{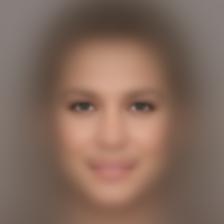}\vspace{-1mm}
    		\caption{Average face\vspace{3mm}}
    	\end{subfigure}
    	\razmik
    \begin{subfigure}{\miniwidth}
    		\centering
    		\includegraphics[width=\textwidth]{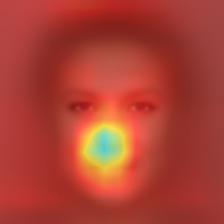}\vspace{-1mm}
    		\caption{Big nose\vspace{3mm}}
    	\end{subfigure}
    	\vspace{5mm}
    	    \begin{subfigure}{\miniwidth}
    		\centering
    		\includegraphics[width=\textwidth]{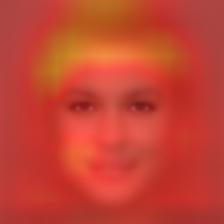}
    		\caption{Brown hair}\vspace{-2mm}
    	\end{subfigure}
    	\razmik
    	    \begin{subfigure}{\miniwidth}
    		\centering
    		\includegraphics[width=\textwidth]{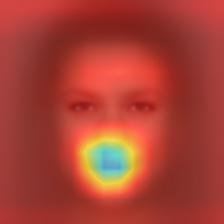}
    		\caption{Smiling}\vspace{-2mm}
    	\end{subfigure}\vspace{-1mm}
\caption{Visualization of image regions with the greatest impact on the attribute classifier $C$ for a given attribute. The heatmaps were generated with GradCAM. The left image in the top row shows the average CelebA--HQ face. The remaining images contain the average face overlayed with average GradCAM heatmaps for a few sample attributes. Note that the attributes are reflected in local image regions.}
\label{fig:gradcam}
\end{figure}

Table \ref{tab:embedding} shows the performance scores achieved by MaskFaceGAN, when embedding different (local) facial regions, and a comparison with the  state--of--the--art Image2StyleGAN++ embedding algorithm from \cite{abdal2020image2stylegan++}. For the latter, the whole face is embedded in the latent space using a combination of pixel--wise MSE and perceptual losses \cite{johnson2016perceptual, dosovitskiy2016generating}.   We observe that choosing to embed a smaller portion of the image results in a considerable gain in embedding quality. It is important to note that the results should not be regarded as an improvement over existing embedding algorithms -- one could achieve a perfect score by simply choosing not to embed any part of the image. Nevertheless, the results motivate the idea of focusing the editing procedure on local image areas. This local embedding step allows us to perform a lower number of iterations during the optimization procedure and still generate visually convincing and photo--realistic images. We therefore report the editing results on larger datasets than, for example, \cite{abdal2020image2stylegan++}. 

\section{Spatial constraints and facial attribute correspondence}

MaskFaceGAN relies on spatial constraints when optimizing the StyleGAN2 latent--codes for face attribute editing. The targeted attributes, therefore, need to have spatial correspondences in the image and, similarly, visually convincing changes in the attributes need to be reflected in local image regions only. To demonstrate that utilizing spatial constraints for image editing is reasonable, we visualize image areas contributing most to the decision of MaskFaceGAN's attribute classifier $C$ for a number of target attributes in Fig. \ref{fig:gradcam}. Here, GradCAM \cite{selvaraju2017grad} is utilized to generate spatial heatmaps for the visualization. 

For this experiment, we extract heatmaps from a subset of images from  CelebA--HQ. Due to the tree--like architecture of the classifier used in our implementation of MaskFaceGAN, only gradients and activations from the leaf convolutional layer of a given target facial attribute are considered. All other leaf convolutional layers do not receive any gradient and, therefore, have zero contribution to the heatmaps. In Fig. \ref{fig:gradcam} the average CelebA--HQ face as well average GradCam heatmaps (computed over a subset of CelebA--HQ images) for a few example attributes are presented. 
Note that the classifier focuses predominantly on local image regions, suggesting that changing the image within a local image region is a viable approach for editing such attributes.

\section{Additional editing results}

In this section, we present additional visual results for all targeted facial attributes and across images from all three experimental datasets.  

\subsection{Single Attribute Editing}

Figs. \ref{fig:celebahq}, \ref{fig:helen} and \ref{fig:siblings} show editing examples for images from CelebA--HQ, Helen and SiblingsDB--HQf, respectively. The StarGAN model shows a somewhat weaker performance than other editing techniques considered and exhibits limited generalization capabilities across different datasets, as shown in Figs. \ref{fig:helen} and \ref{fig:siblings}. AttGAN and STGAN produce higher--quality editing results, but often struggle with the entanglement of different attributes, which is especially apparent with the ``Grey hair'' attribute. The disentangled version of InterFaceGAN, InterFaceGAN--D, is more successful when editing attributes that exhibit a high degree of entanglement. For example, it relatively convincingly removes age--dependent image characteristics, such as wrinkles, when trying to generate ``Grey hair'', as shown in Fig. \ref{fig:celebahq}. However, disentangling ``Pale skin'' from hair color does not have any apparent effect compared to the vanilla InterFaceGAN version, as shown in Fig. \ref{fig:celebahq} under ``Blond hair'' and in Fig. \ref{fig:helen} under ``Black hair'' despite the fact that these attributes are highly correlated. MaskFaceGAN, on the other hand, is less affected by entanglement problems due to the local nature of the editing procedure and produces convincing results for the majority of edited attributes.
\renewcommand{\figureimagewidth}{.3\columnwidth}
\renewcommand{\figuretextwidth}{.02\columnwidth}
\renewcommand\miniwidth{.162\textwidth}
\renewcommand{\figureimagewidth}{.095\textwidth}
\renewcommand{\figuretextwidth}{.1\textwidth}
\setlength{\tabcolsep}{0.5mm} 
\renewcommand{\arraystretch}{0.3}  
\newcolumntype{C}{ >{\centering\arraybackslash} m{\figureimagewidth} }
\newcolumntype{D}{ >{\centering\arraybackslash} m{\figuretextwidth} }
\begin{figure*}[!t!]
\centering
\begin{subfigure}{\textwidth}
\centering
\begin{tabular}{DCCCCCCCC}
 & \scriptsize{Original} & \scriptsize{\textit{Arched eyeb.}} & \scriptsize{Big nose} & \scriptsize{Black hair} & \scriptsize{Blond hair} & \scriptsize{Brown hair} & \scriptsize{Bushy eyeb.} & \scriptsize{Grey hair}\\[1mm] \scriptsize{StarGAN} & \includegraphics[width=.095\textwidth]{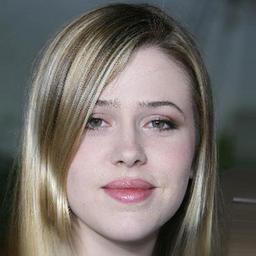} & \includegraphics[width=.095\textwidth]{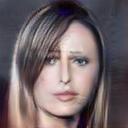} & \includegraphics[width=.095\textwidth]{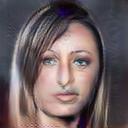} & \includegraphics[width=.095\textwidth]{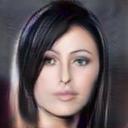} & \includegraphics[width=.095\textwidth]{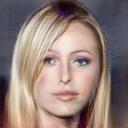} & \includegraphics[width=.095\textwidth]{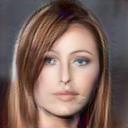} & \includegraphics[width=.095\textwidth]{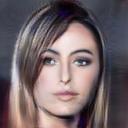} & \includegraphics[width=.095\textwidth]{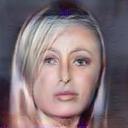}\\ 
 \scriptsize{AttGAN} & \includegraphics[width=.095\textwidth]{images/all_edits/originals/524.jpg} & \includegraphics[width=.095\textwidth]{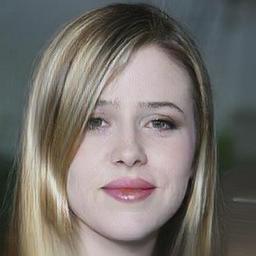} & \includegraphics[width=.095\textwidth]{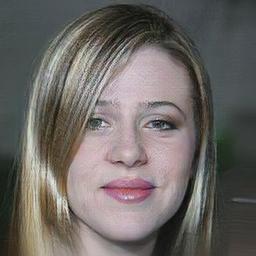} & \includegraphics[width=.095\textwidth]{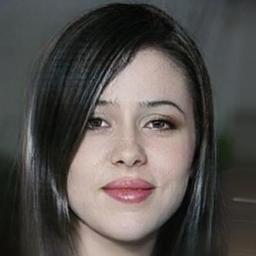} & \includegraphics[width=.095\textwidth]{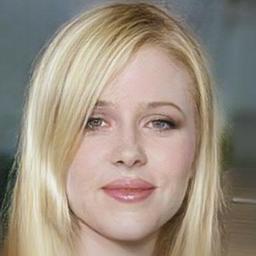} & \includegraphics[width=.095\textwidth]{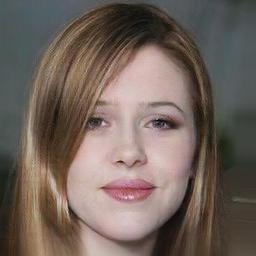} & \includegraphics[width=.095\textwidth]{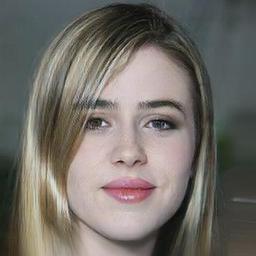} & \includegraphics[width=.095\textwidth]{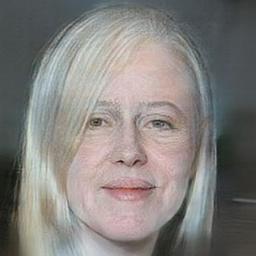}\\ 
\scriptsize{STGAN} & \includegraphics[width=.095\textwidth]{images/all_edits/originals/524.jpg} & \includegraphics[width=.095\textwidth]{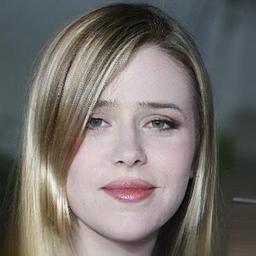} & \includegraphics[width=.095\textwidth]{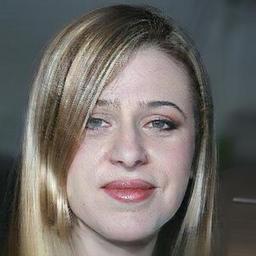} & \includegraphics[width=.095\textwidth]{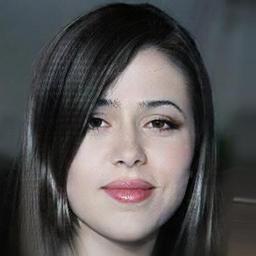} & \includegraphics[width=.095\textwidth]{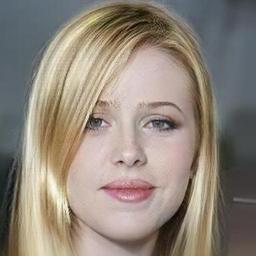} & \includegraphics[width=.095\textwidth]{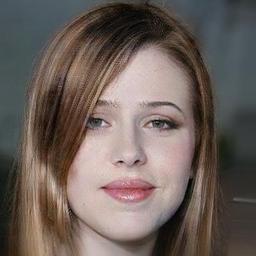} & \includegraphics[width=.095\textwidth]{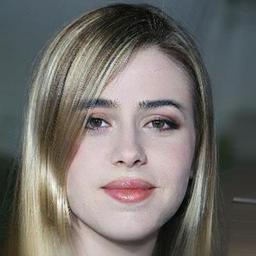} & \includegraphics[width=.095\textwidth]{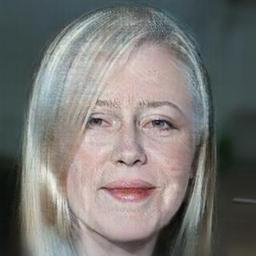}\\ 
\scriptsize{InterFaceGAN} & \includegraphics[width=.095\textwidth]{images/all_edits/originals/524.jpg} & \includegraphics[width=.095\textwidth]{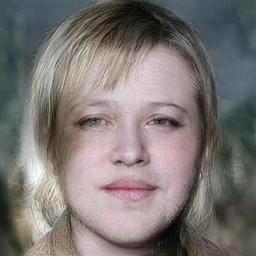} & \includegraphics[width=.095\textwidth]{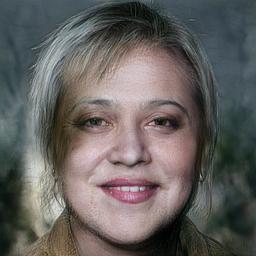} & \includegraphics[width=.095\textwidth]{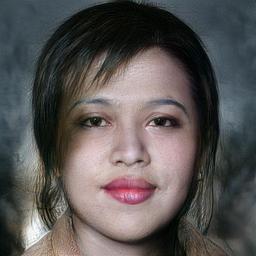} & \includegraphics[width=.095\textwidth]{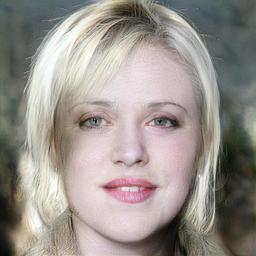} & \includegraphics[width=.095\textwidth]{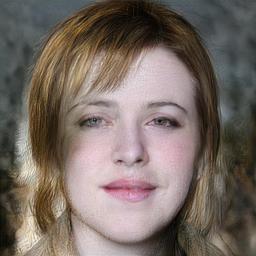} & \includegraphics[width=.095\textwidth]{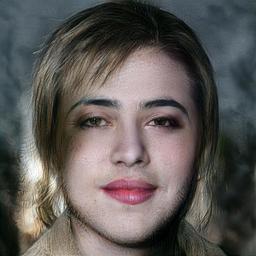} & \includegraphics[width=.095\textwidth]{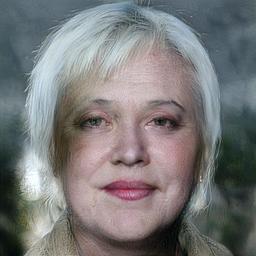}\\ 
\scriptsize{InterFaceGAN--D} & \includegraphics[width=.095\textwidth]{images/all_edits/originals/524.jpg} & \includegraphics[width=.095\textwidth]{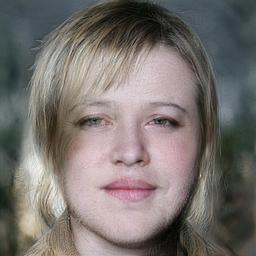} & \includegraphics[width=.095\textwidth]{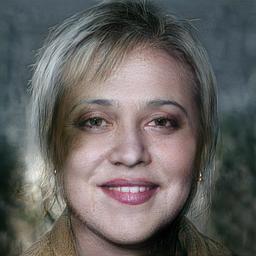} & \includegraphics[width=.095\textwidth]{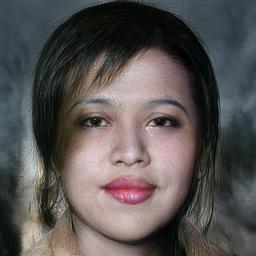} & \includegraphics[width=.095\textwidth]{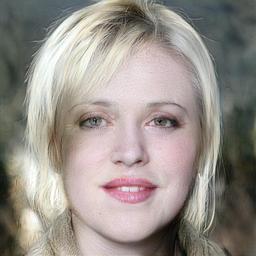} & \includegraphics[width=.095\textwidth]{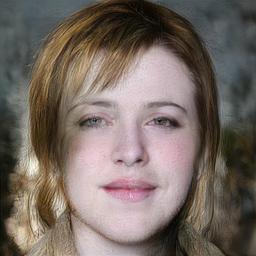} & \includegraphics[width=.095\textwidth]{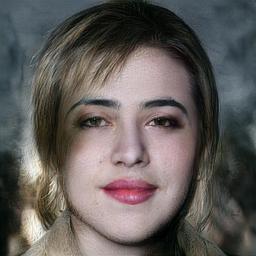} & \includegraphics[width=.095\textwidth]{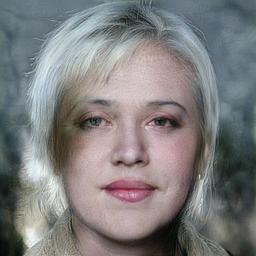}\\ 
\scriptsize{MaskFaceGAN} & \includegraphics[width=.095\textwidth]{images/all_edits/originals/524.jpg} & \includegraphics[width=.095\textwidth]{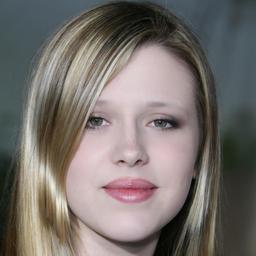} & \includegraphics[width=.095\textwidth]{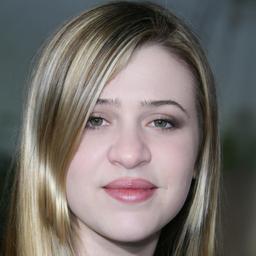} & \includegraphics[width=.095\textwidth]{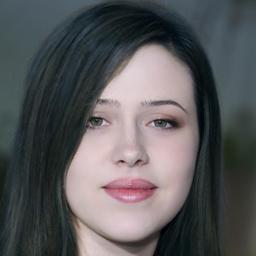} & \includegraphics[width=.095\textwidth]{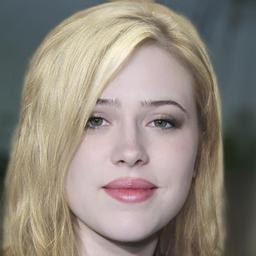} & \includegraphics[width=.095\textwidth]{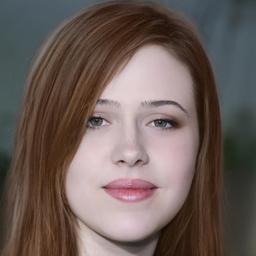} & \includegraphics[width=.095\textwidth]{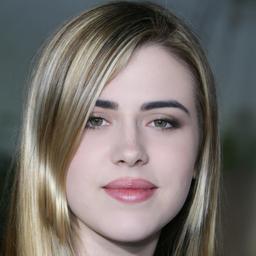} & \includegraphics[width=.095\textwidth]{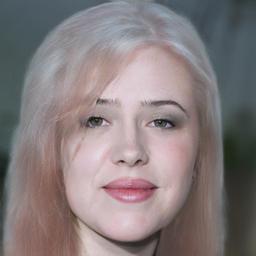}\\[3mm]  
&\\[2mm] 
 & \scriptsize{Original} & \scriptsize{Mth. sl. open} & \scriptsize{Narrow eyes} & \scriptsize{\textit{Pointy nose}} & \scriptsize{Smiling} & \scriptsize{\textit{Straight hair}} & \scriptsize{Wavy hair} & \scriptsize{\textit{Wearing lipst.}}\\[1mm]  \scriptsize{StarGAN} & \includegraphics[width=.095\textwidth]{images/all_edits/originals/524.jpg} & \includegraphics[width=.095\textwidth]{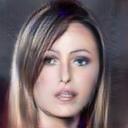} & \includegraphics[width=.095\textwidth]{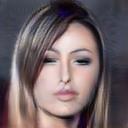} & \includegraphics[width=.095\textwidth]{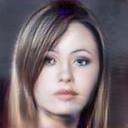} & \includegraphics[width=.095\textwidth]{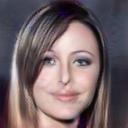} & \includegraphics[width=.095\textwidth]{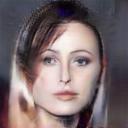} & \includegraphics[width=.095\textwidth]{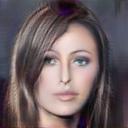} & \includegraphics[width=.095\textwidth]{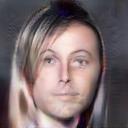}\\ 
 \scriptsize{AttGAN} & \includegraphics[width=.095\textwidth]{images/all_edits/originals/524.jpg} & \includegraphics[width=.095\textwidth]{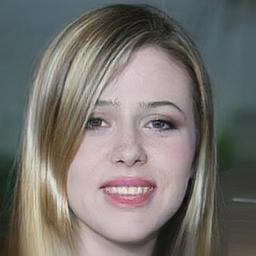} & \includegraphics[width=.095\textwidth]{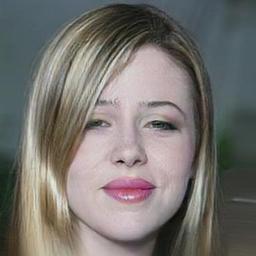} & \includegraphics[width=.095\textwidth]{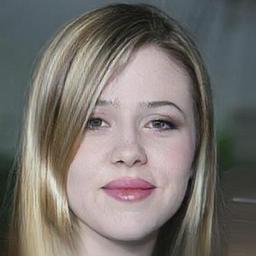} & \includegraphics[width=.095\textwidth]{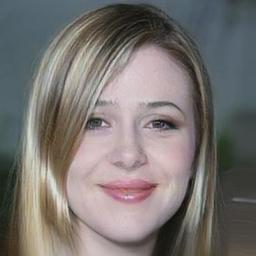} & \includegraphics[width=.095\textwidth]{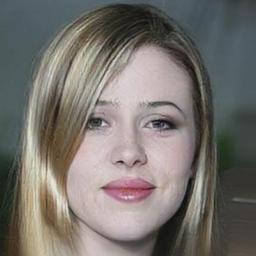} & \includegraphics[width=.095\textwidth]{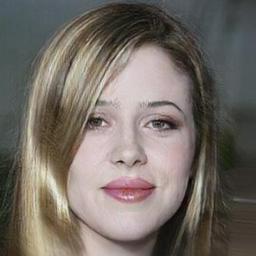} & \includegraphics[width=.095\textwidth]{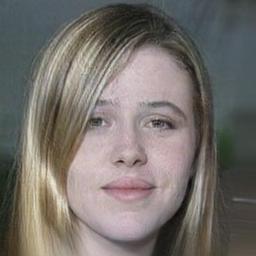}\\ 
\scriptsize{STGAN} & \includegraphics[width=.095\textwidth]{images/all_edits/originals/524.jpg} & \includegraphics[width=.095\textwidth]{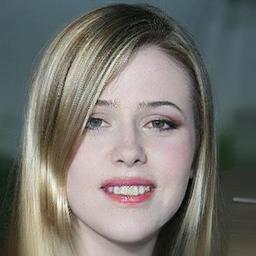} & \includegraphics[width=.095\textwidth]{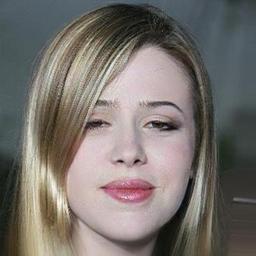} & \includegraphics[width=.095\textwidth]{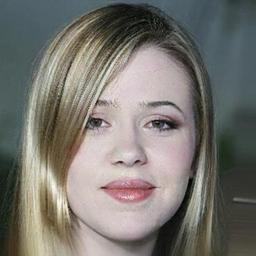} & \includegraphics[width=.095\textwidth]{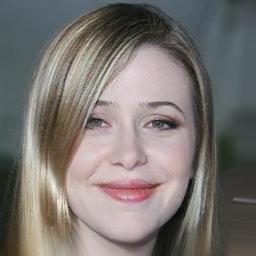} & \includegraphics[width=.095\textwidth]{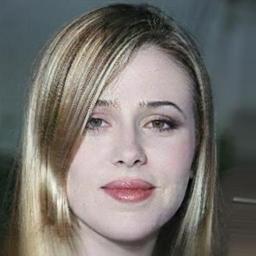} & \includegraphics[width=.095\textwidth]{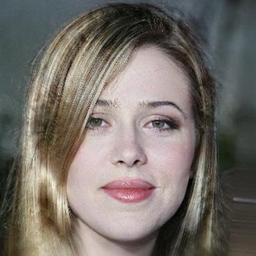} & \includegraphics[width=.095\textwidth]{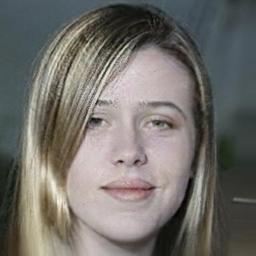}\\ 
\scriptsize{InterFaceGAN} & \includegraphics[width=.095\textwidth]{images/all_edits/originals/524.jpg} & \includegraphics[width=.095\textwidth]{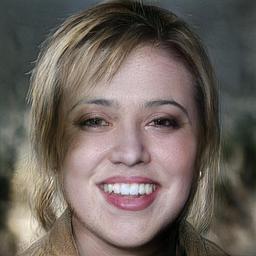} & \includegraphics[width=.095\textwidth]{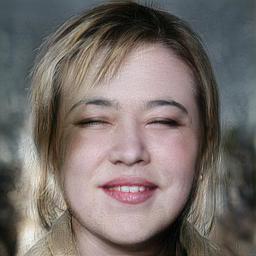} & \includegraphics[width=.095\textwidth]{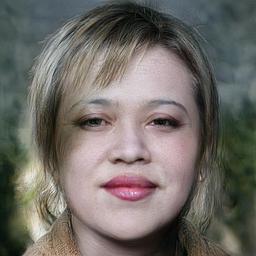} & \includegraphics[width=.095\textwidth]{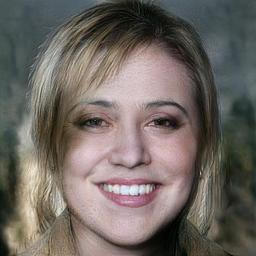} & \includegraphics[width=.095\textwidth]{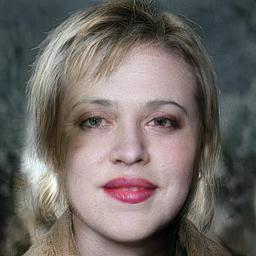} & \includegraphics[width=.095\textwidth]{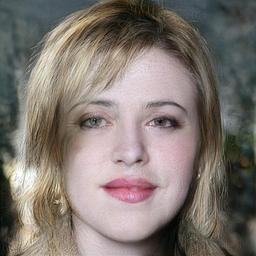} & \includegraphics[width=.095\textwidth]{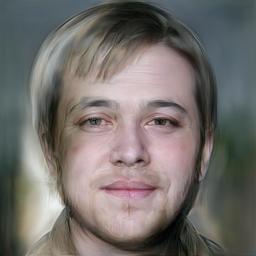}\\ 
\scriptsize{InterFaceGAN--D} & \includegraphics[width=.095\textwidth]{images/all_edits/originals/524.jpg} & \includegraphics[width=.095\textwidth]{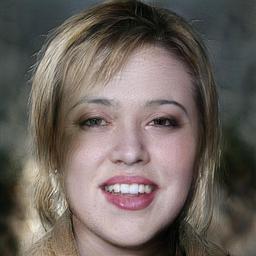} & \includegraphics[width=.095\textwidth]{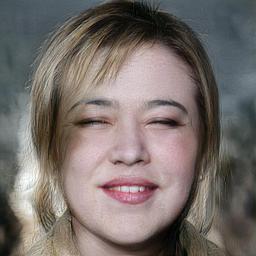} & \includegraphics[width=.095\textwidth]{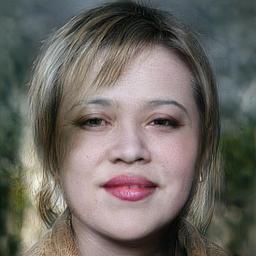} & \includegraphics[width=.095\textwidth]{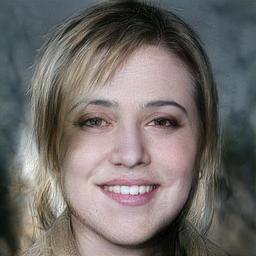} & \includegraphics[width=.095\textwidth]{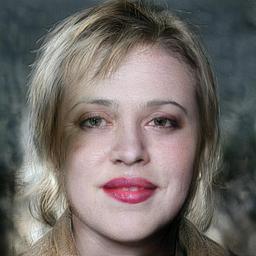} & \includegraphics[width=.095\textwidth]{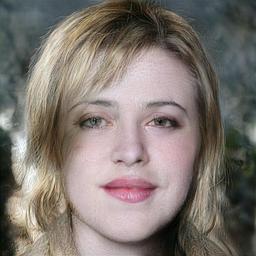} & \includegraphics[width=.095\textwidth]{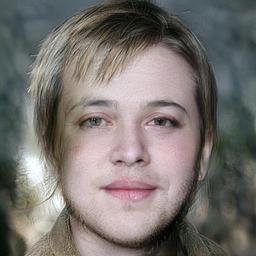}\\ 
\scriptsize{MaskFaceGAN} & \includegraphics[width=.095\textwidth]{images/all_edits/originals/524.jpg} & \includegraphics[width=.095\textwidth]{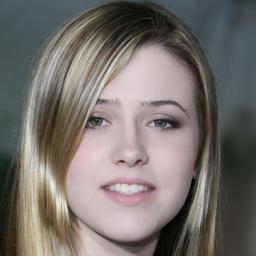} & \includegraphics[width=.095\textwidth]{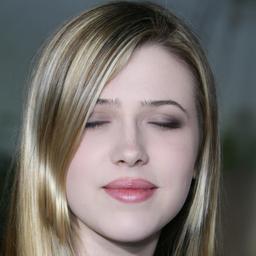} & \includegraphics[width=.095\textwidth]{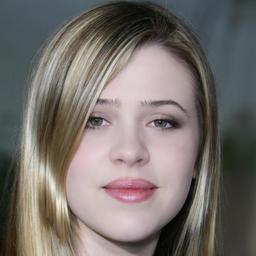} & \includegraphics[width=.095\textwidth]{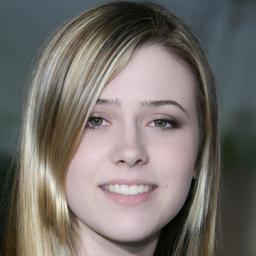} & \includegraphics[width=.095\textwidth]{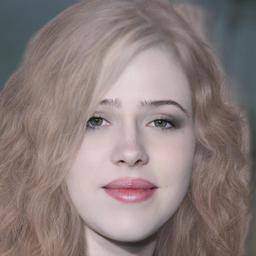} & \includegraphics[width=.095\textwidth]{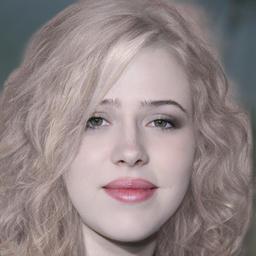} & \includegraphics[width=.095\textwidth]{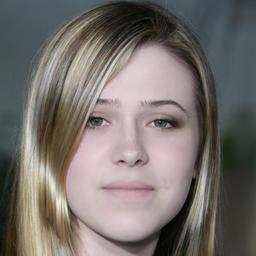}\\ 
\end{tabular}
\end{subfigure}
\vspace{2mm}
\caption{Attributes editing examples for a sample image from the CelebA--HQ dataset. The results show manipulation of a single attribute and a comparison to the results generated with competing models. Inverted attributes are marked italic. The figure is best viewed electronically and zoomed in for details.}
\label{fig:celebahq}
\end{figure*}
\begin{figure*}[!th!]
\centering
\begin{subfigure}{\textwidth}
\centering
\begin{tabular}{DCCCCCCCC}
 & \scriptsize{Original} & \scriptsize{Arched eyeb.} & \scriptsize{\textit{Big nose}} & \scriptsize{Black hair} & \scriptsize{Blond hair} & \scriptsize{Brown hair} & \scriptsize{\textit{Bushy eyeb.}} & \scriptsize{Grey hair}\\[1mm] \scriptsize{StarGAN} & \includegraphics[width=.095\textwidth]{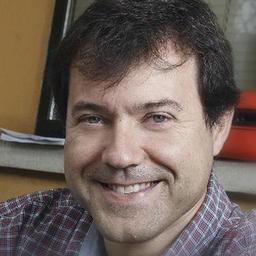} & \includegraphics[width=.095\textwidth]{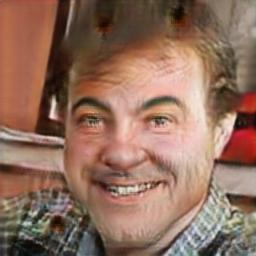} & \includegraphics[width=.095\textwidth]{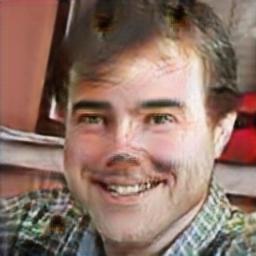} & \includegraphics[width=.095\textwidth]{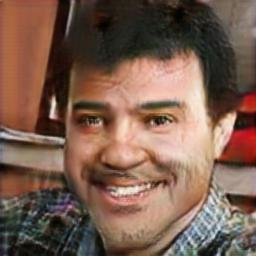} & \includegraphics[width=.095\textwidth]{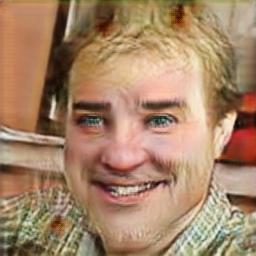} & \includegraphics[width=.095\textwidth]{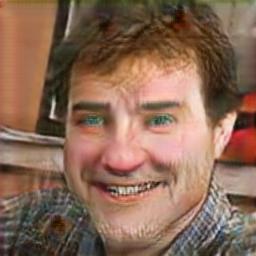} & \includegraphics[width=.095\textwidth]{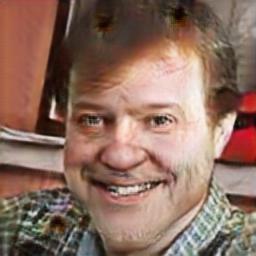} & \includegraphics[width=.095\textwidth]{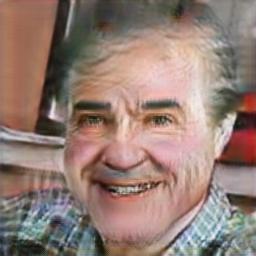}\\ 
 \scriptsize{AttGAN} & \includegraphics[width=.095\textwidth]{images/all_edits/originals/2882152659_1.jpg} & \includegraphics[width=.095\textwidth]{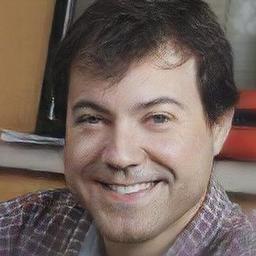} & \includegraphics[width=.095\textwidth]{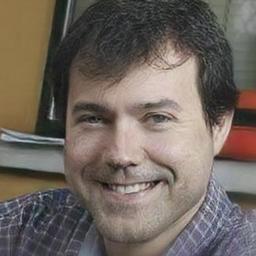} & \includegraphics[width=.095\textwidth]{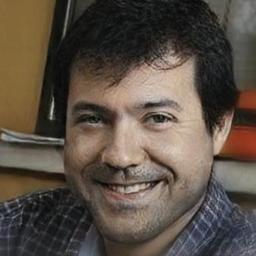} & \includegraphics[width=.095\textwidth]{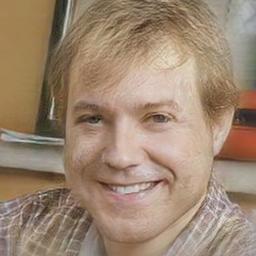} & \includegraphics[width=.095\textwidth]{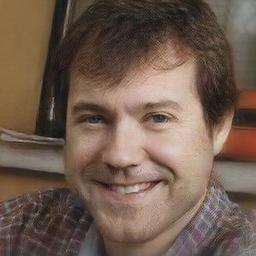} & \includegraphics[width=.095\textwidth]{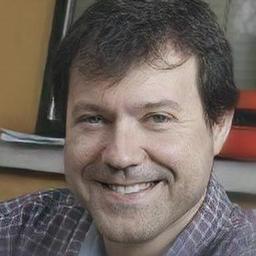} & \includegraphics[width=.095\textwidth]{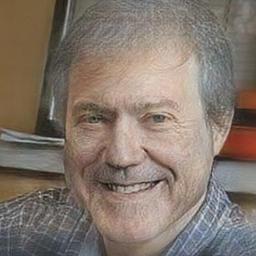}\\ 
\scriptsize{STGAN} & \includegraphics[width=.095\textwidth]{images/all_edits/originals/2882152659_1.jpg} & \includegraphics[width=.095\textwidth]{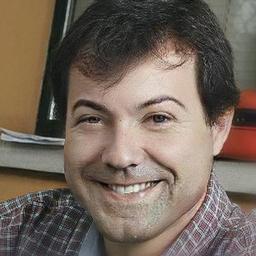} & \includegraphics[width=.095\textwidth]{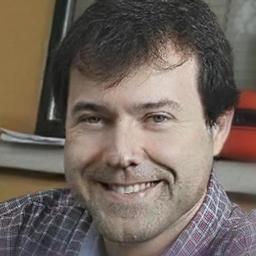} & \includegraphics[width=.095\textwidth]{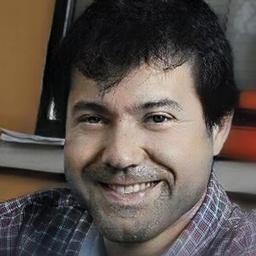} & \includegraphics[width=.095\textwidth]{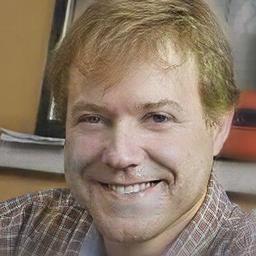} & \includegraphics[width=.095\textwidth]{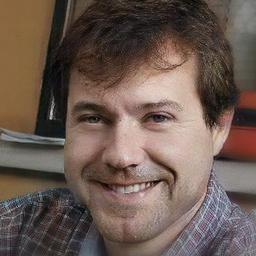} & \includegraphics[width=.095\textwidth]{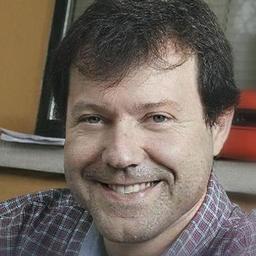} & \includegraphics[width=.095\textwidth]{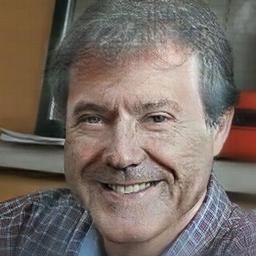}\\ 
\scriptsize{InterFaceGAN} & \includegraphics[width=.095\textwidth]{images/all_edits/originals/2882152659_1.jpg} & \includegraphics[width=.095\textwidth]{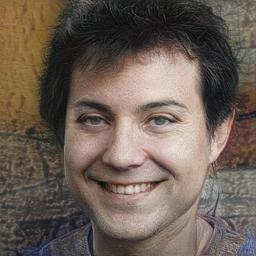} & \includegraphics[width=.095\textwidth]{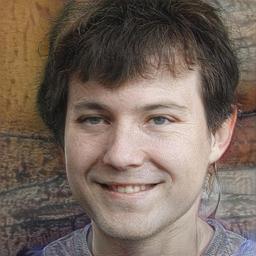} & \includegraphics[width=.095\textwidth]{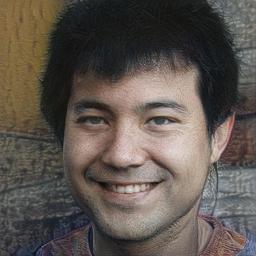} & \includegraphics[width=.095\textwidth]{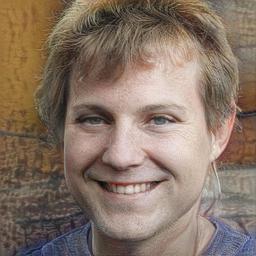} & \includegraphics[width=.095\textwidth]{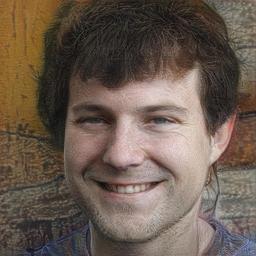} & \includegraphics[width=.095\textwidth]{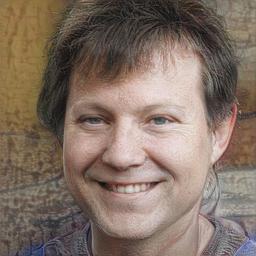} & \includegraphics[width=.095\textwidth]{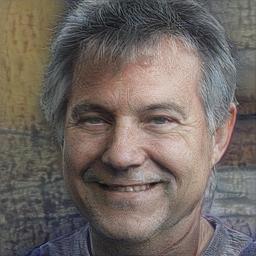}\\ 
\scriptsize{InterFaceGAN--D} & \includegraphics[width=.095\textwidth]{images/all_edits/originals/2882152659_1.jpg} & \includegraphics[width=.095\textwidth]{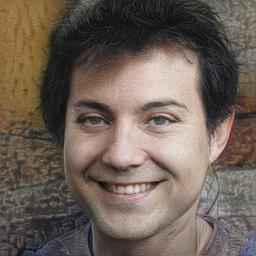} & \includegraphics[width=.095\textwidth]{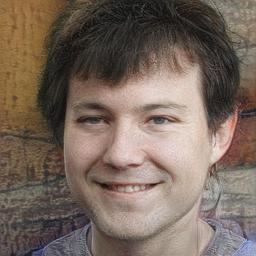} & \includegraphics[width=.095\textwidth]{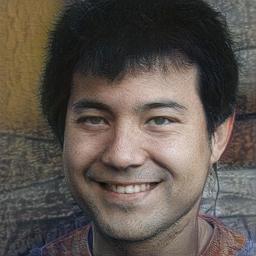} & \includegraphics[width=.095\textwidth]{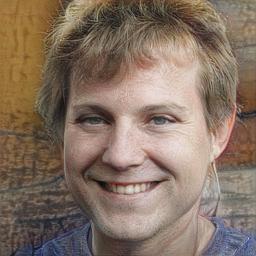} & \includegraphics[width=.095\textwidth]{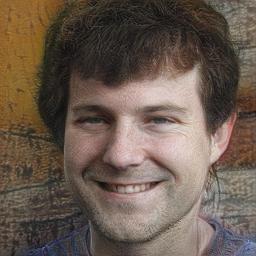} & \includegraphics[width=.095\textwidth]{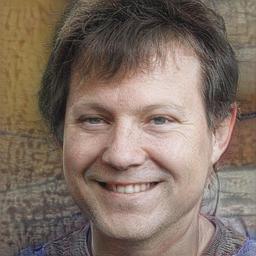} & \includegraphics[width=.095\textwidth]{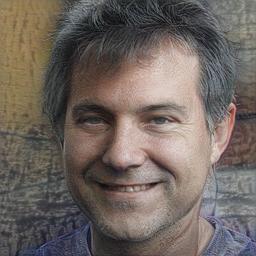}\\ 
\scriptsize{MaskFaceGAN} & \includegraphics[width=.095\textwidth]{images/all_edits/originals/2882152659_1.jpg} & \includegraphics[width=.095\textwidth]{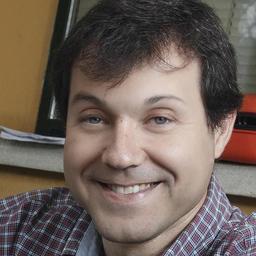} & \includegraphics[width=.095\textwidth]{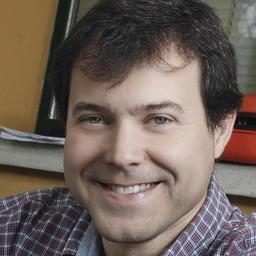} & \includegraphics[width=.095\textwidth]{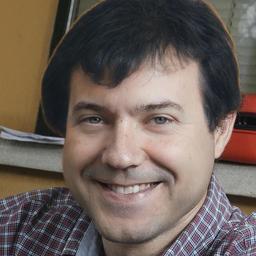} & \includegraphics[width=.095\textwidth]{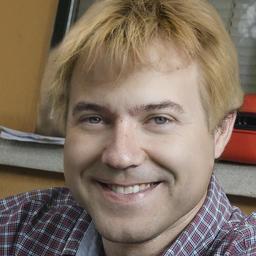} & \includegraphics[width=.095\textwidth]{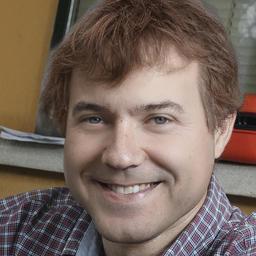} & \includegraphics[width=.095\textwidth]{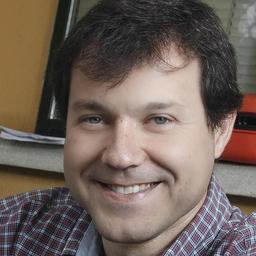} & \includegraphics[width=.095\textwidth]{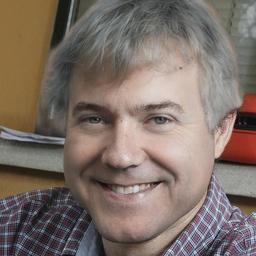}\\[3mm]  
&\\[2mm] 
 & \scriptsize{Original} & \scriptsize{\textit{Mth. sl. open}} & \scriptsize{Narrow eyes} & \scriptsize{Pointy nose} & \scriptsize{\textit{Smiling}} & \scriptsize{Straight hair} & \scriptsize{Wavy hair} & \scriptsize{Wearing lipst.}\\[1mm] \scriptsize{StarGAN} & \includegraphics[width=.095\textwidth]{images/all_edits/originals/2882152659_1.jpg} & \includegraphics[width=.095\textwidth]{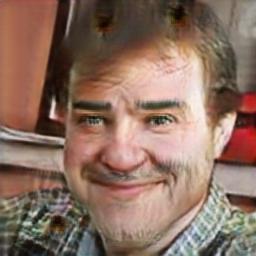} & \includegraphics[width=.095\textwidth]{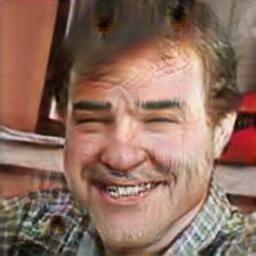} & \includegraphics[width=.095\textwidth]{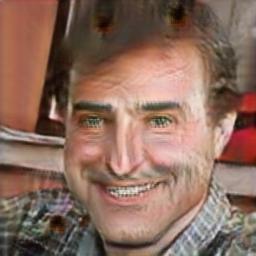} & \includegraphics[width=.095\textwidth]{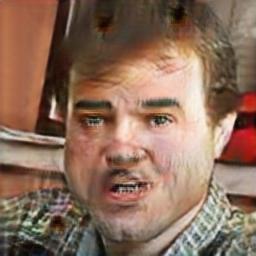} & \includegraphics[width=.095\textwidth]{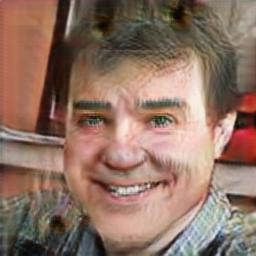} & \includegraphics[width=.095\textwidth]{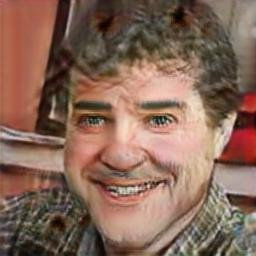} & \includegraphics[width=.095\textwidth]{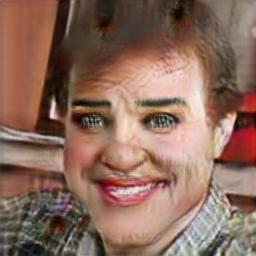}\\ 
 \scriptsize{AttGAN} & \includegraphics[width=.095\textwidth]{images/all_edits/originals/2882152659_1.jpg} & \includegraphics[width=.095\textwidth]{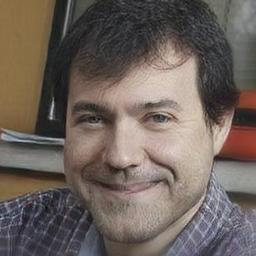} & \includegraphics[width=.095\textwidth]{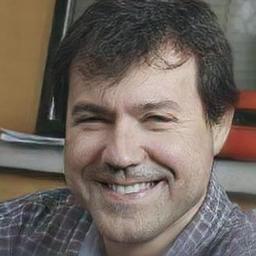} & \includegraphics[width=.095\textwidth]{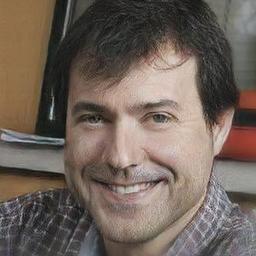} & \includegraphics[width=.095\textwidth]{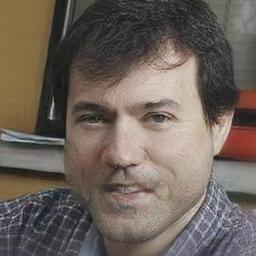} & \includegraphics[width=.095\textwidth]{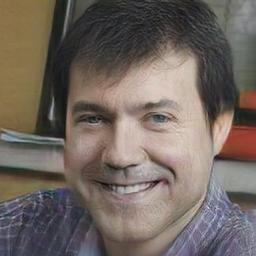} & \includegraphics[width=.095\textwidth]{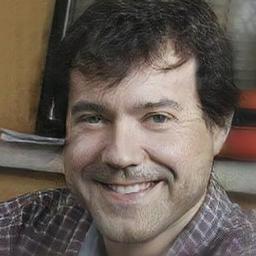} & \includegraphics[width=.095\textwidth]{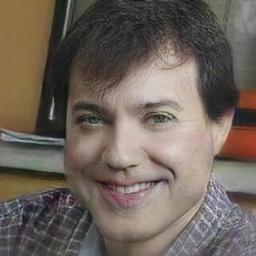}\\ 
\scriptsize{STGAN} & \includegraphics[width=.095\textwidth]{images/all_edits/originals/2882152659_1.jpg} & \includegraphics[width=.095\textwidth]{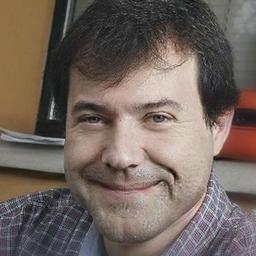} & \includegraphics[width=.095\textwidth]{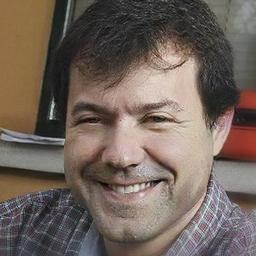} & \includegraphics[width=.095\textwidth]{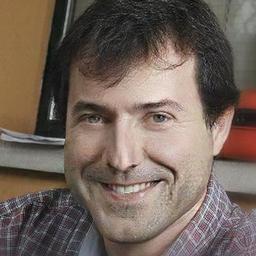} & \includegraphics[width=.095\textwidth]{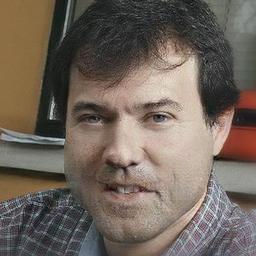} & \includegraphics[width=.095\textwidth]{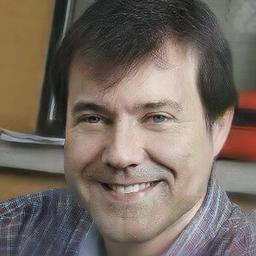} & \includegraphics[width=.095\textwidth]{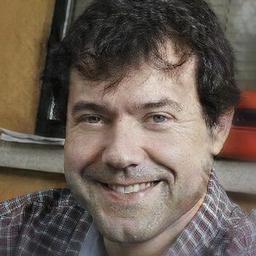} & \includegraphics[width=.095\textwidth]{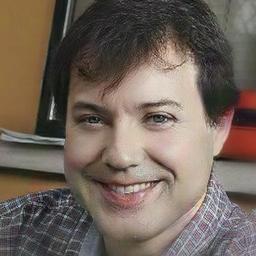}\\ 
\scriptsize{InterFaceGAN} & \includegraphics[width=.095\textwidth]{images/all_edits/originals/2882152659_1.jpg} & \includegraphics[width=.095\textwidth]{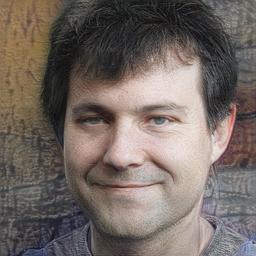} & \includegraphics[width=.095\textwidth]{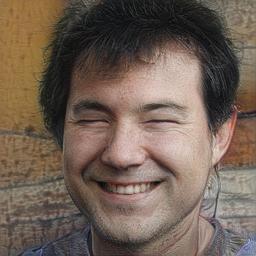} & \includegraphics[width=.095\textwidth]{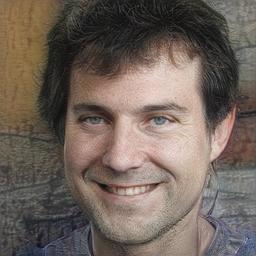} & \includegraphics[width=.095\textwidth]{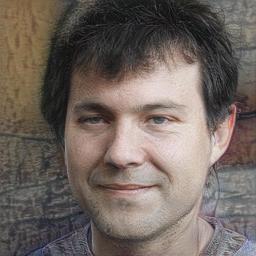} & \includegraphics[width=.095\textwidth]{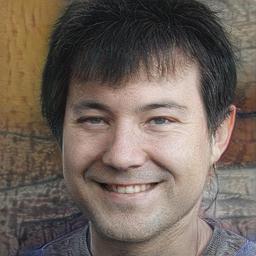} & \includegraphics[width=.095\textwidth]{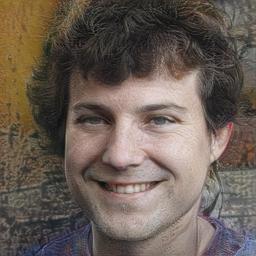} & \includegraphics[width=.095\textwidth]{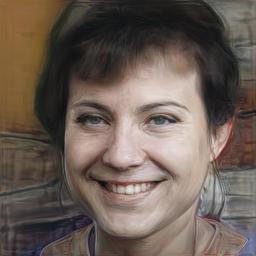}\\ 
\scriptsize{InterFaceGAN--D} & \includegraphics[width=.095\textwidth]{images/all_edits/originals/2882152659_1.jpg} & \includegraphics[width=.095\textwidth]{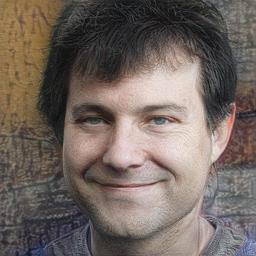} & \includegraphics[width=.095\textwidth]{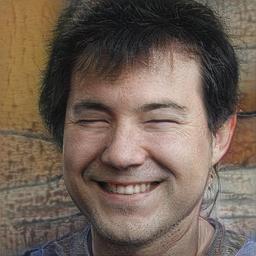} & \includegraphics[width=.095\textwidth]{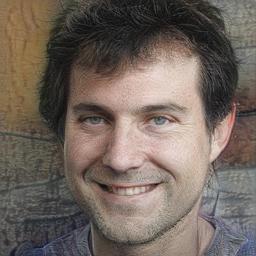} & \includegraphics[width=.095\textwidth]{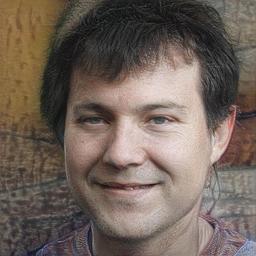} & \includegraphics[width=.095\textwidth]{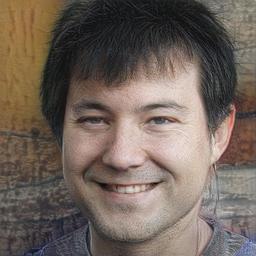} & \includegraphics[width=.095\textwidth]{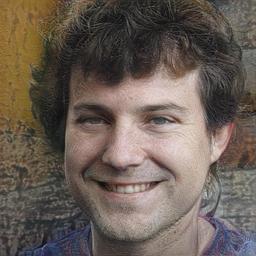} & \includegraphics[width=.095\textwidth]{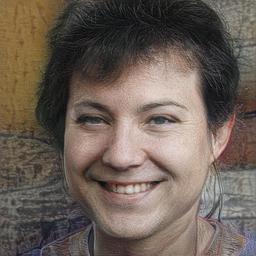}\\ 
\scriptsize{MaskFaceGAN} & \includegraphics[width=.095\textwidth]{images/all_edits/originals/2882152659_1.jpg} & \includegraphics[width=.095\textwidth]{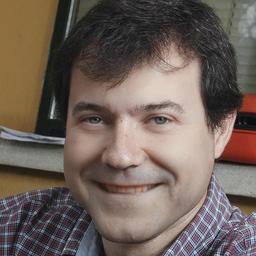} & \includegraphics[width=.095\textwidth]{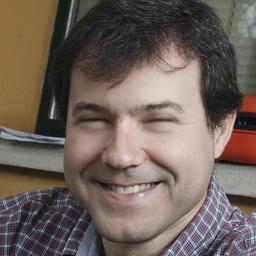} & \includegraphics[width=.095\textwidth]{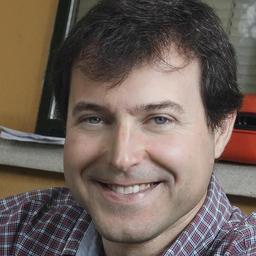} & \includegraphics[width=.095\textwidth]{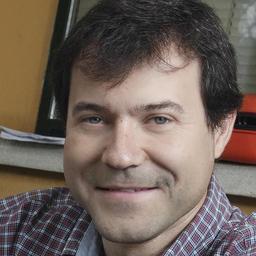} & \includegraphics[width=.095\textwidth]{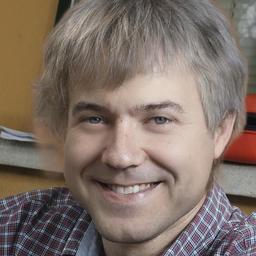} & \includegraphics[width=.095\textwidth]{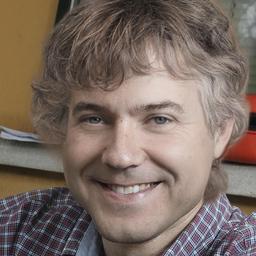} & \includegraphics[width=.095\textwidth]{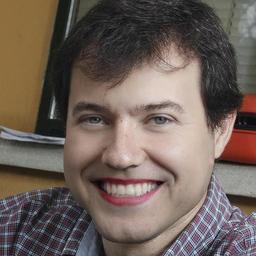}\\ 
\end{tabular}
\end{subfigure}
\caption{Attributes editing examples for a sample image from the Helen dataset. The results show manipulation of a single attribute and a comparison to the results generated with competing models. Inverted attributes are marked italic. The figure is best viewed electronically and zoomed in for details.}
\label{fig:helen}
\end{figure*}
\begin{figure*}[!th!]
\centering
\begin{subfigure}{\textwidth}
\centering
\begin{tabular}{DCCCCCCCC}
 & \scriptsize{Original} & \scriptsize{Arched eyeb.} & \scriptsize{Big nose} & \scriptsize{Black hair} & \scriptsize{Blond hair} & \scriptsize{Brown hair} & \scriptsize{\textit{Bushy eyeb.}} & \scriptsize{\textit{Grey hair}}\\[1mm] \scriptsize{StarGAN} & \includegraphics[width=.095\textwidth]{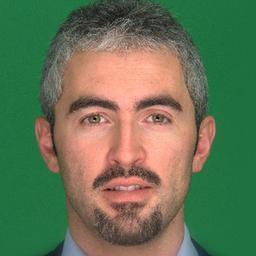} & \includegraphics[width=.095\textwidth]{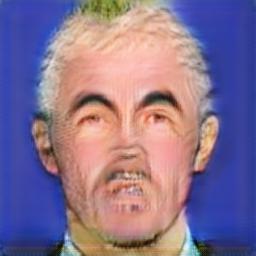} & \includegraphics[width=.095\textwidth]{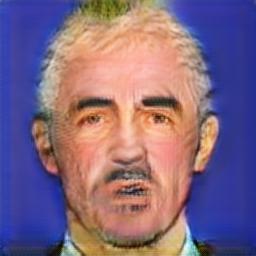} & \includegraphics[width=.095\textwidth]{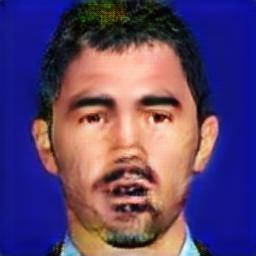} & \includegraphics[width=.095\textwidth]{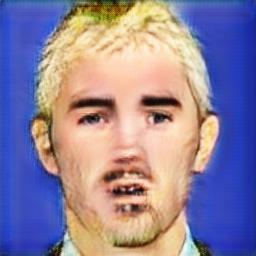} & \includegraphics[width=.095\textwidth]{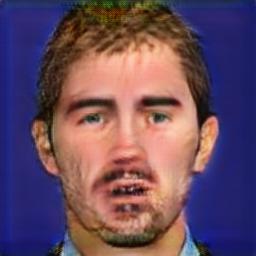} & \includegraphics[width=.095\textwidth]{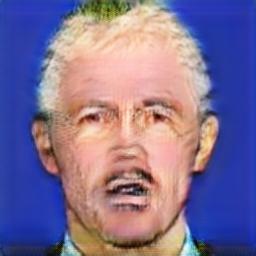} & \includegraphics[width=.095\textwidth]{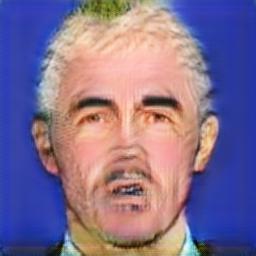}\\ 
 \scriptsize{AttGAN} & \includegraphics[width=.095\textwidth]{images/all_edits/originals/_DSC0195.jpg} & \includegraphics[width=.095\textwidth]{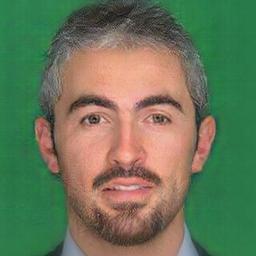} & \includegraphics[width=.095\textwidth]{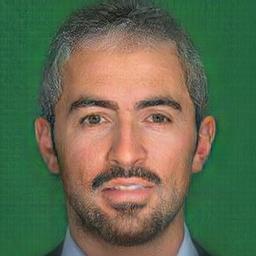} & \includegraphics[width=.095\textwidth]{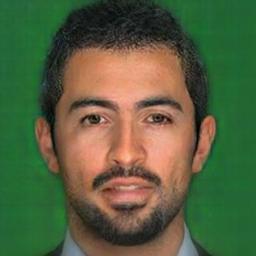} & \includegraphics[width=.095\textwidth]{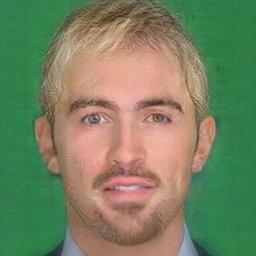} & \includegraphics[width=.095\textwidth]{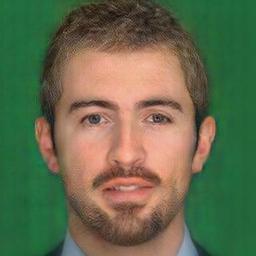} & \includegraphics[width=.095\textwidth]{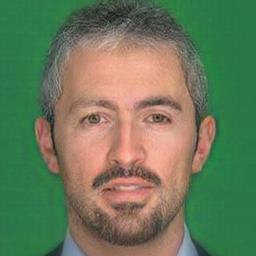} & \includegraphics[width=.095\textwidth]{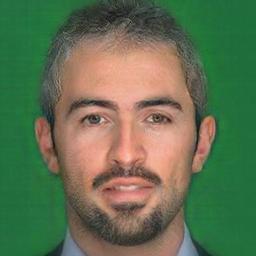}\\ 
\scriptsize{STGAN} & \includegraphics[width=.095\textwidth]{images/all_edits/originals/_DSC0195.jpg} & \includegraphics[width=.095\textwidth]{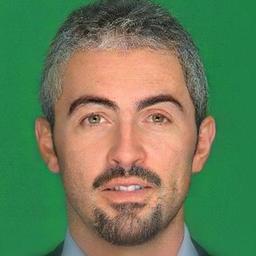} & \includegraphics[width=.095\textwidth]{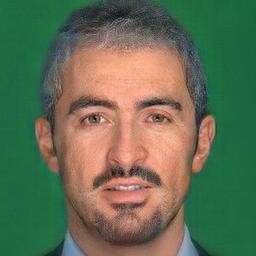} & \includegraphics[width=.095\textwidth]{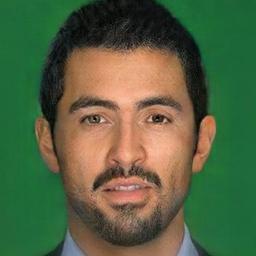} & \includegraphics[width=.095\textwidth]{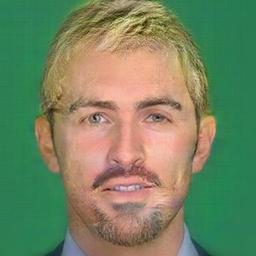} & \includegraphics[width=.095\textwidth]{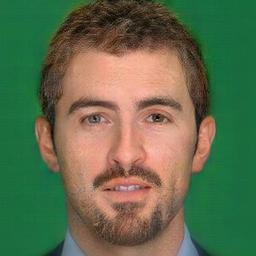} & \includegraphics[width=.095\textwidth]{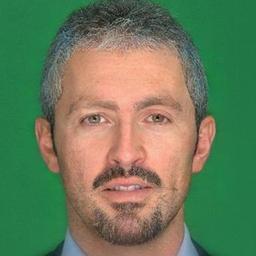} & \includegraphics[width=.095\textwidth]{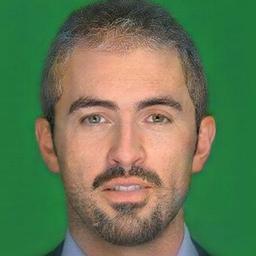}\\ 
\scriptsize{InterFaceGAN} & \includegraphics[width=.095\textwidth]{images/all_edits/originals/_DSC0195.jpg} & \includegraphics[width=.095\textwidth]{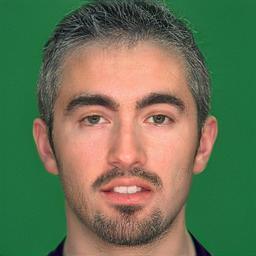} & \includegraphics[width=.095\textwidth]{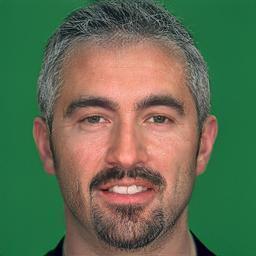} & \includegraphics[width=.095\textwidth]{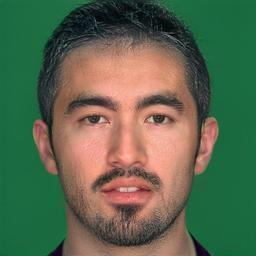} & \includegraphics[width=.095\textwidth]{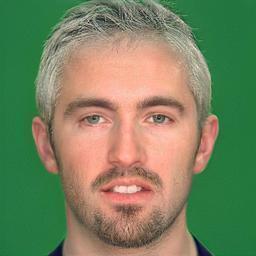} & \includegraphics[width=.095\textwidth]{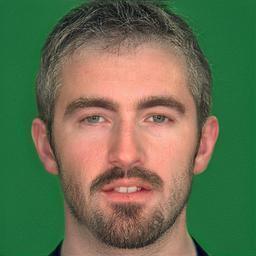} & \includegraphics[width=.095\textwidth]{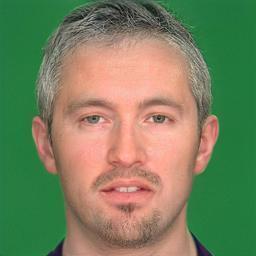} & \includegraphics[width=.095\textwidth]{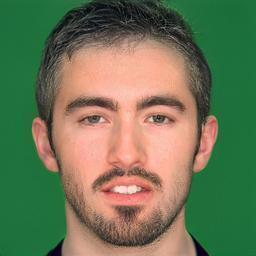}\\ 
\scriptsize{InterFaceGAN--D} & \includegraphics[width=.095\textwidth]{images/all_edits/originals/_DSC0195.jpg} & \includegraphics[width=.095\textwidth]{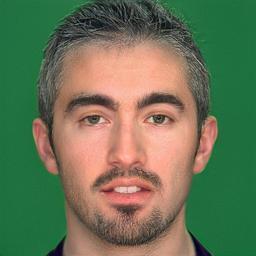} & \includegraphics[width=.095\textwidth]{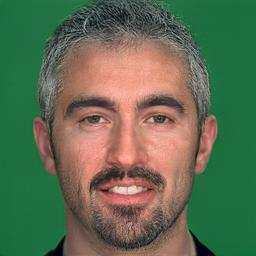} & \includegraphics[width=.095\textwidth]{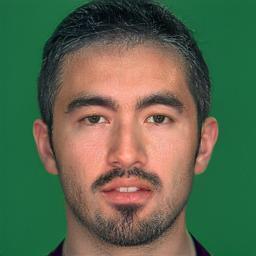} & \includegraphics[width=.095\textwidth]{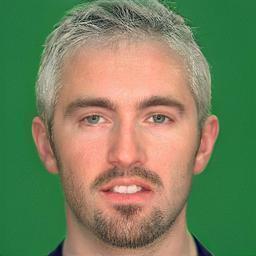} & \includegraphics[width=.095\textwidth]{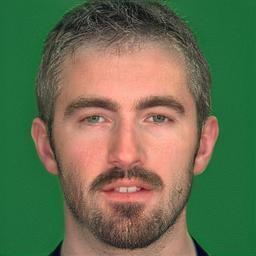} & \includegraphics[width=.095\textwidth]{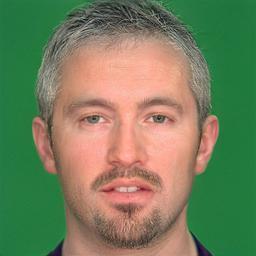} & \includegraphics[width=.095\textwidth]{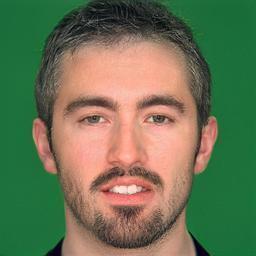}\\ 
\scriptsize{MaskFaceGAN} & \includegraphics[width=.095\textwidth]{images/all_edits/originals/_DSC0195.jpg} & \includegraphics[width=.095\textwidth]{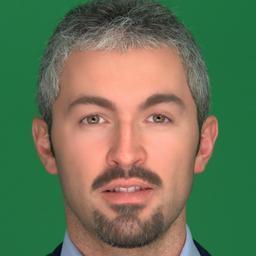} & \includegraphics[width=.095\textwidth]{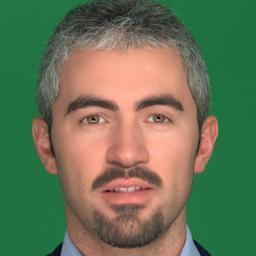} & \includegraphics[width=.095\textwidth]{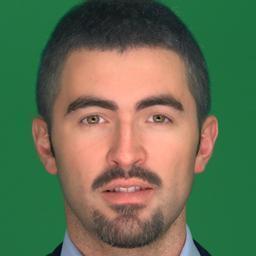} & \includegraphics[width=.095\textwidth]{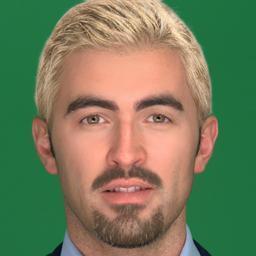} & \includegraphics[width=.095\textwidth]{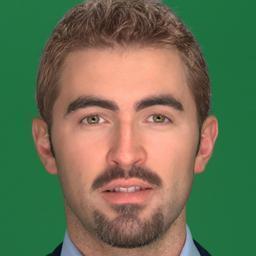} & \includegraphics[width=.095\textwidth]{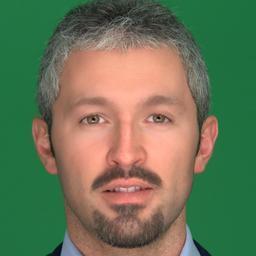} & \includegraphics[width=.095\textwidth]{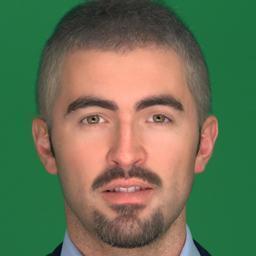}\\[3mm] 
%
&\\[2mm]
 & \scriptsize{Original} & \scriptsize{\textit{Mth. sl. open}} & \scriptsize{Narrow eyes} & \scriptsize{Pointy nose} & \scriptsize{Smiling} & \scriptsize{Straight hair} & \scriptsize{Wavy hair} & \scriptsize{Wearing lipst.}\\[1mm] \scriptsize{StarGAN} & \includegraphics[width=.095\textwidth]{images/all_edits/originals/_DSC0195.jpg} & \includegraphics[width=.095\textwidth]{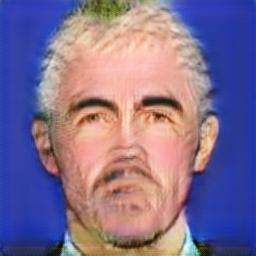} & \includegraphics[width=.095\textwidth]{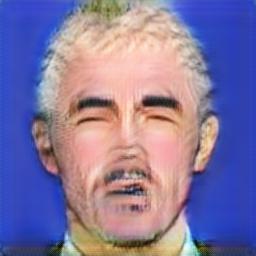} & \includegraphics[width=.095\textwidth]{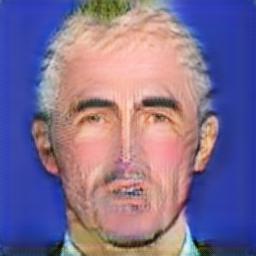} & \includegraphics[width=.095\textwidth]{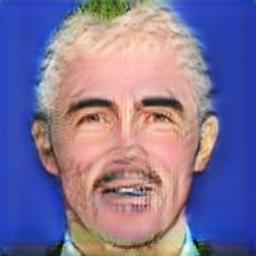} & \includegraphics[width=.095\textwidth]{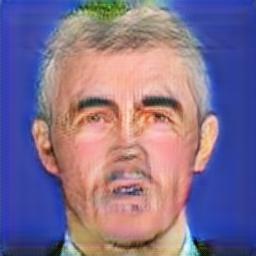} & \includegraphics[width=.095\textwidth]{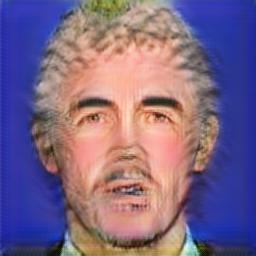} & \includegraphics[width=.095\textwidth]{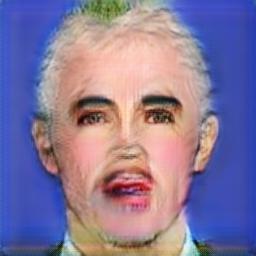}\\ 
\scriptsize{AttGAN} & \includegraphics[width=.095\textwidth]{images/all_edits/originals/_DSC0195.jpg} & \includegraphics[width=.095\textwidth]{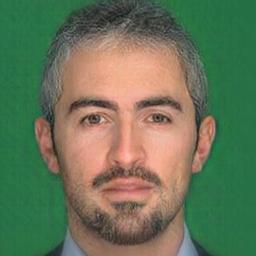} & \includegraphics[width=.095\textwidth]{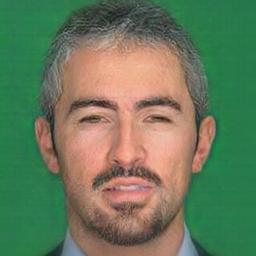} & \includegraphics[width=.095\textwidth]{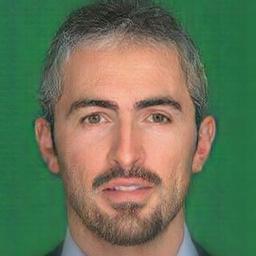} & \includegraphics[width=.095\textwidth]{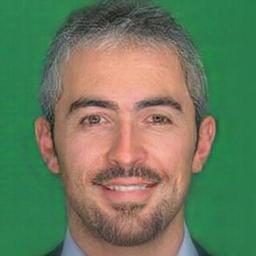} & \includegraphics[width=.095\textwidth]{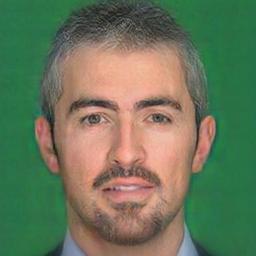} & \includegraphics[width=.095\textwidth]{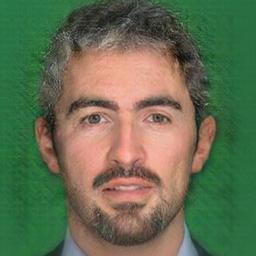} & \includegraphics[width=.095\textwidth]{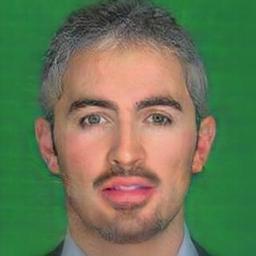}\\ 
\scriptsize{STGAN} & \includegraphics[width=.095\textwidth]{images/all_edits/originals/_DSC0195.jpg} & \includegraphics[width=.095\textwidth]{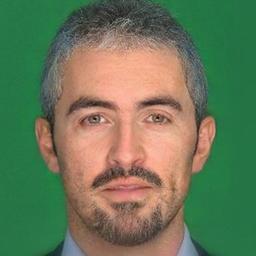} & \includegraphics[width=.095\textwidth]{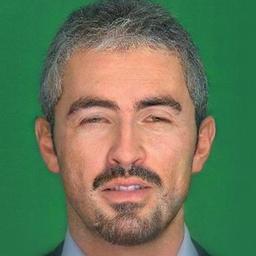} & \includegraphics[width=.095\textwidth]{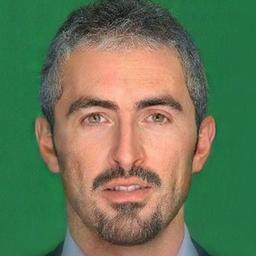} & \includegraphics[width=.095\textwidth]{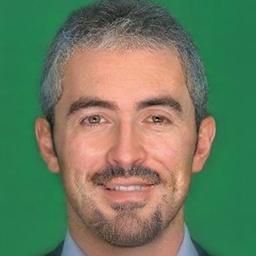} & \includegraphics[width=.095\textwidth]{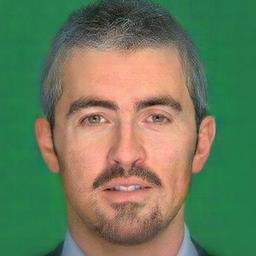} & \includegraphics[width=.095\textwidth]{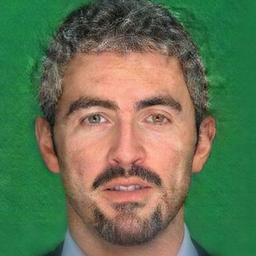} & \includegraphics[width=.095\textwidth]{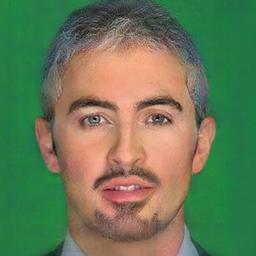}\\ 
\scriptsize{InterFaceGAN} & \includegraphics[width=.095\textwidth]{images/all_edits/originals/_DSC0195.jpg} & \includegraphics[width=.095\textwidth]{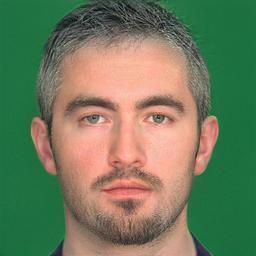} & \includegraphics[width=.095\textwidth]{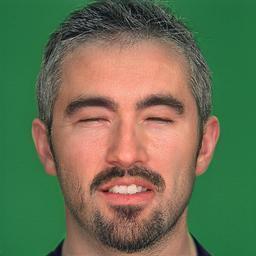} & \includegraphics[width=.095\textwidth]{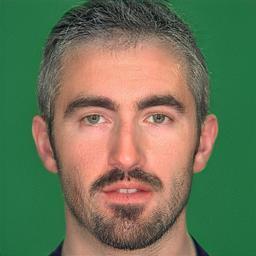} & \includegraphics[width=.095\textwidth]{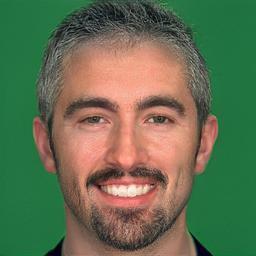} & \includegraphics[width=.095\textwidth]{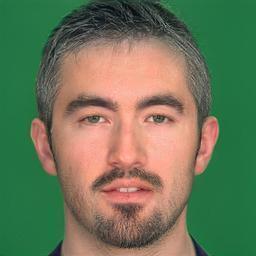} & \includegraphics[width=.095\textwidth]{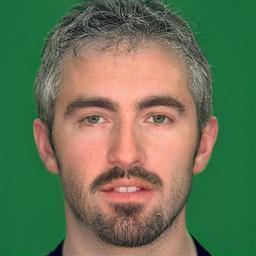} & \includegraphics[width=.095\textwidth]{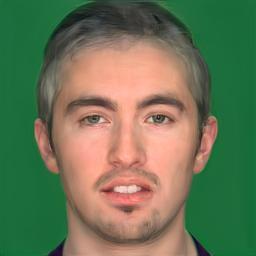}\\ 
\scriptsize{InterFaceGAN--D} & \includegraphics[width=.095\textwidth]{images/all_edits/originals/_DSC0195.jpg} & \includegraphics[width=.095\textwidth]{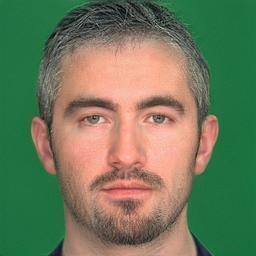} & \includegraphics[width=.095\textwidth]{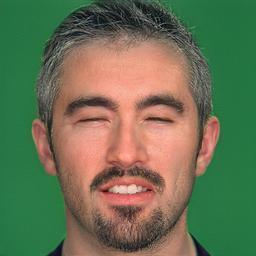} & \includegraphics[width=.095\textwidth]{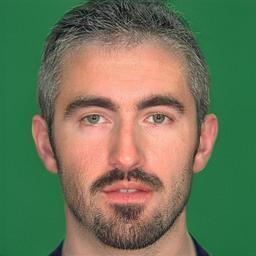} & \includegraphics[width=.095\textwidth]{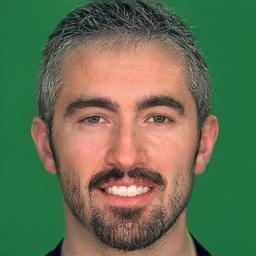} & \includegraphics[width=.095\textwidth]{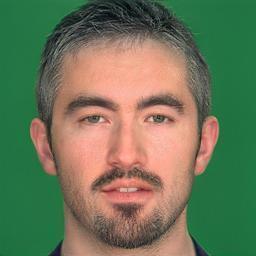} & \includegraphics[width=.095\textwidth]{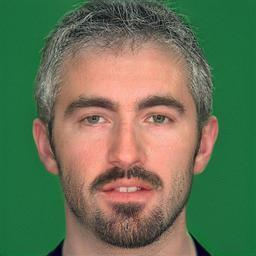} & \includegraphics[width=.095\textwidth]{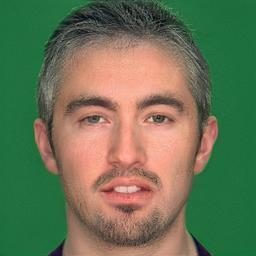}\\ 
\scriptsize{MaskFaceGAN} & \includegraphics[width=.095\textwidth]{images/all_edits/originals/_DSC0195.jpg} & \includegraphics[width=.095\textwidth]{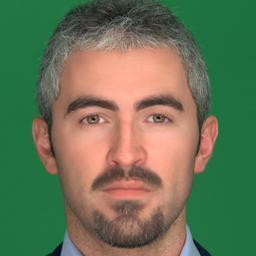} & \includegraphics[width=.095\textwidth]{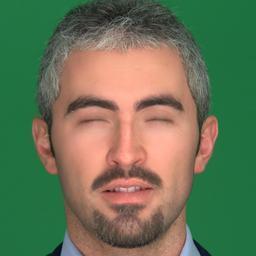} & \includegraphics[width=.095\textwidth]{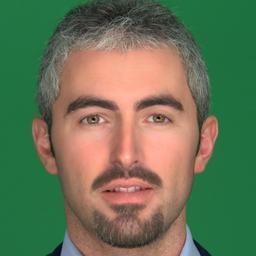} & \includegraphics[width=.095\textwidth]{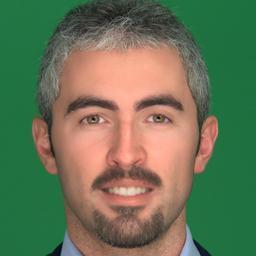} & \includegraphics[width=.095\textwidth]{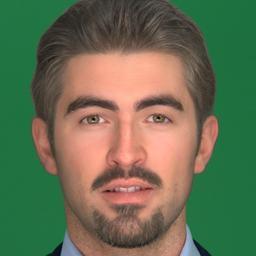} & \includegraphics[width=.095\textwidth]{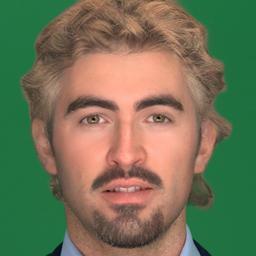} & \includegraphics[width=.095\textwidth]{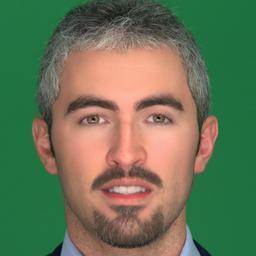}\\ 
\end{tabular}
\end{subfigure}
\caption{Attributes editing examples for a sample image from the SiblingsDB--HQf dataset. The results show manipulation of a single attribute and a comparison to the results generated with competing models. Inverted attributes are marked italic. The figure is best viewed electronically and zoomed in for details.}
\label{fig:siblings}
\end{figure*}
\subsection{High Resolution Results}
Fig. \ref{fig:HR} presents high--resolution editing results for the ``Blond hair''  attribute. In this example, the StarGAN model produces the most blurry result due to the low image resolution of the model. AttGAN and STGAN achieve better image quality. However, the edited images still exhibit visual artefacts. For the presented results, InterFaceGAN--D is disentangled with respect to the ``Pale skin'' attribute. Nonetheless, both InterFaceGAN as well as its disentangled version, InterFaceGAN--D, edit the image with very similar results that slightly change the skin tone as well as the color and shape of clothes, the background, the ears and the eye region. The proposed model does not exhibit such problems and preserves both appearance as well as identity of the original face well. Additionally, even at higher resolutions, no apparent artefacts are present in the edited image. 
\renewcommand{\figureimagewidth}{.3\textwidth}
\renewcommand{\im}[1]{
\begin{minipage}{\figureimagewidth}
\centering
\includegraphics[width=\textwidth]{images/HR/#1.jpg}
{\small #1}
\end{minipage}
}
\renewcommand{\arraystretch}{13.3}
\setlength{\tabcolsep}{6pt}
\begin{figure*}[!th!]
\begin{center}
\begin{tabular}{CCC}
 &  \im{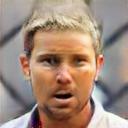} & \im{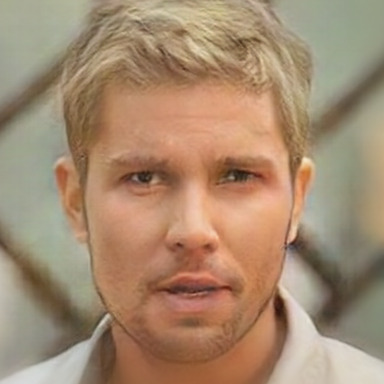} \\[-7mm]
\im{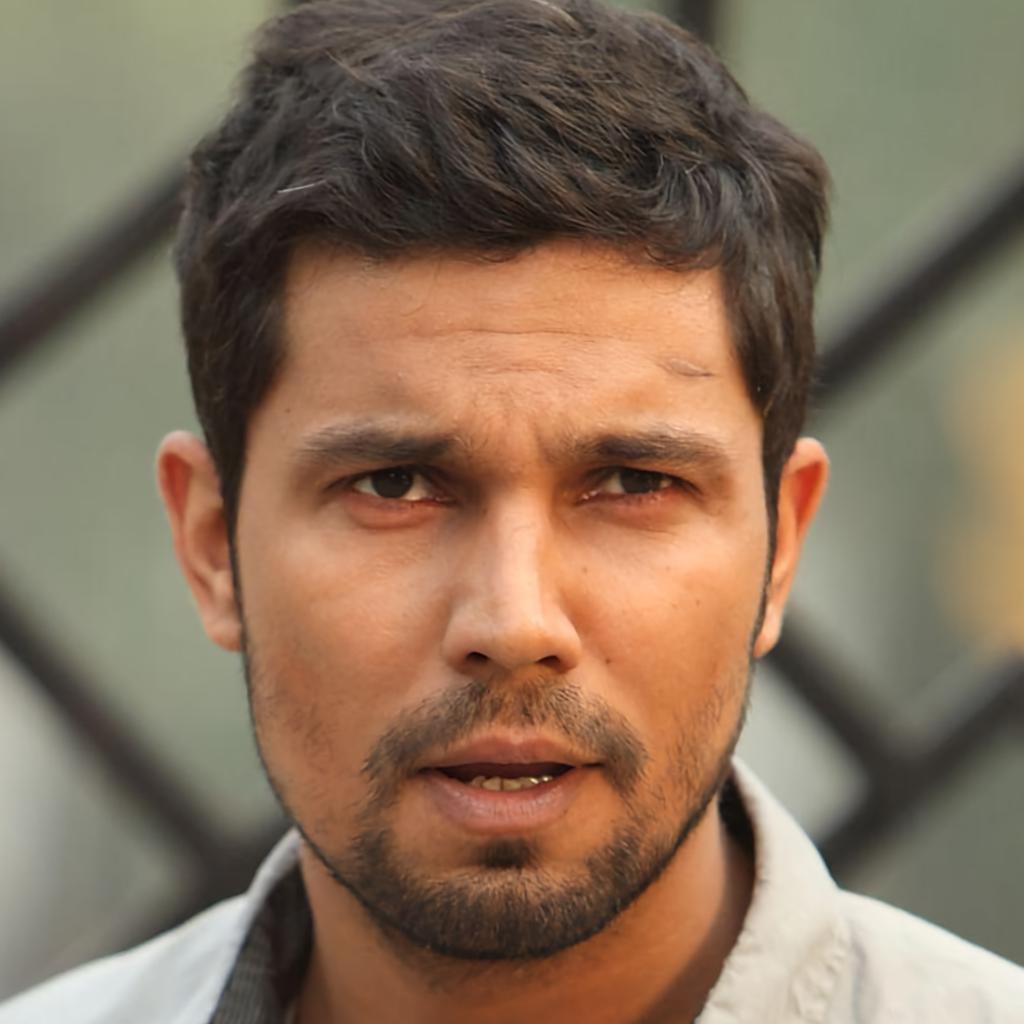} &  \im{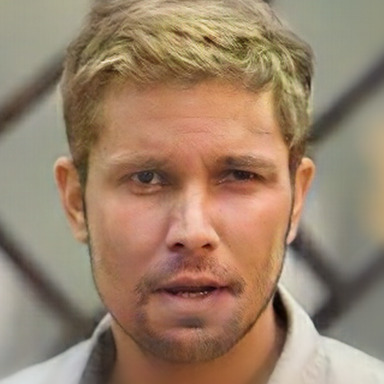} & \im{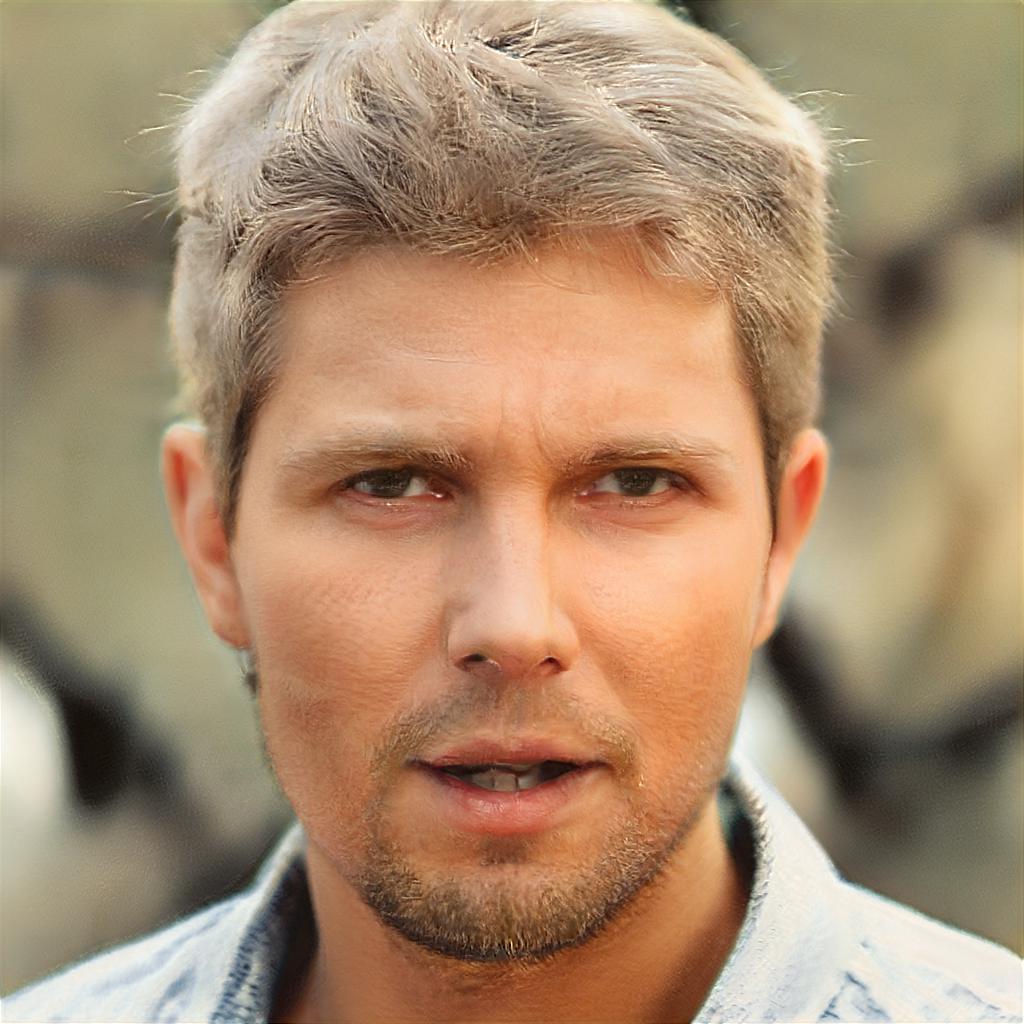} \\[-7mm]
 &  \im{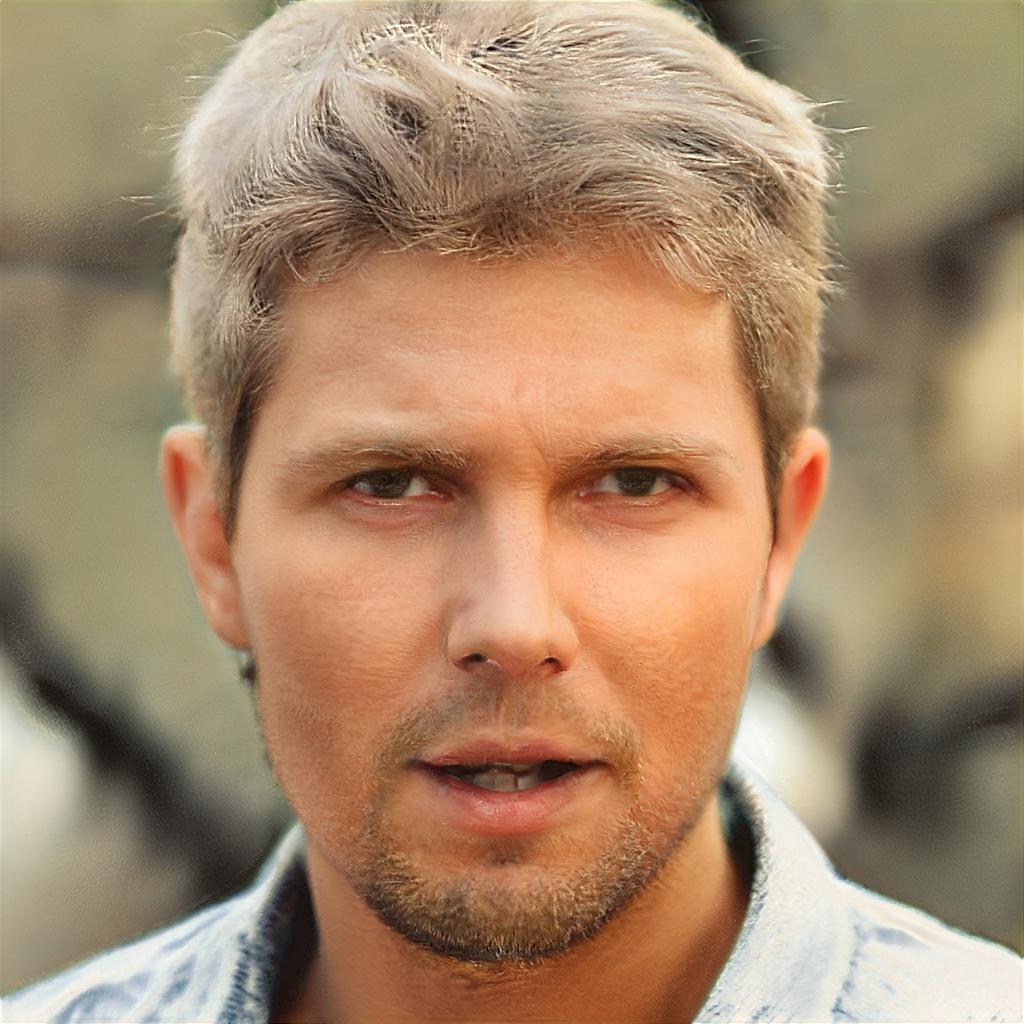} & \im{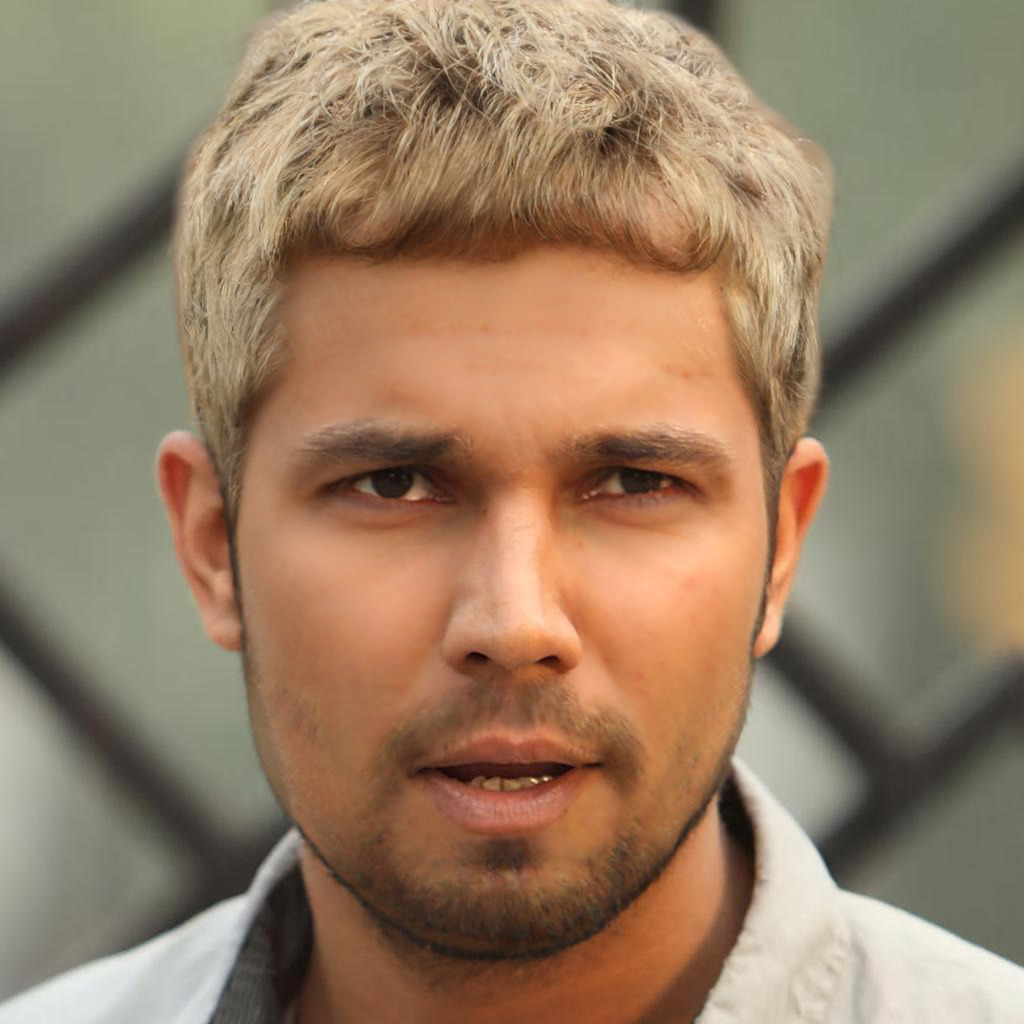} 
\end{tabular}
\caption{High--resolution comparison of  MaskFaceGAN and several state-of-the-art editing techniques from the literature for the target attribute ``Blond hair''. Observe the quality of the generated images (the right two columns) and the details retained from the original input image presented on the left. While some of the competing models ensure solid results, MaskFaceGAN produces the most convincing image manipulations with the least amount of visible artefacts.} 
\label{fig:HR}
\end{center}
\end{figure*}

\subsection{Attribute Intensity Control}
Fig. \ref{fig:intensity} shows editing results generated by varying the $\epsilon$ parameter of MaskFaceGAN to achieve different intensities of the attribute presence/absence in the edited image. Lowering the value of the $\epsilon$ parameter towards $0$ results in stronger semantic content (i.e., stronger presence) of the targeted attribute. As illustrated in Fig. \ref{fig:intensity}, MaskFaceGAN enables a considerable level of control over the intensity of different facial attributes, such as hair color, eyebrow shape, smiling intensity and nose shape. Note how all edited images appear photo-realistic despite varying intensities of the targeted attributes. 

\renewcommand{\figuretextwidth}{.12\textwidth}
\renewcommand{\figureimagewidth}{.20\textwidth}

\renewcommand{\im}[1]{ \includegraphics[width=\figureimagewidth]{images/attribute_intensity/appendix/#1}}
\setlength{\tabcolsep}{3pt}
\renewcommand{\arraystretch}{1.3}  

\begin{figure*}[!th!]
\begin{tabular}{DCCCC}
Blond hair & \im{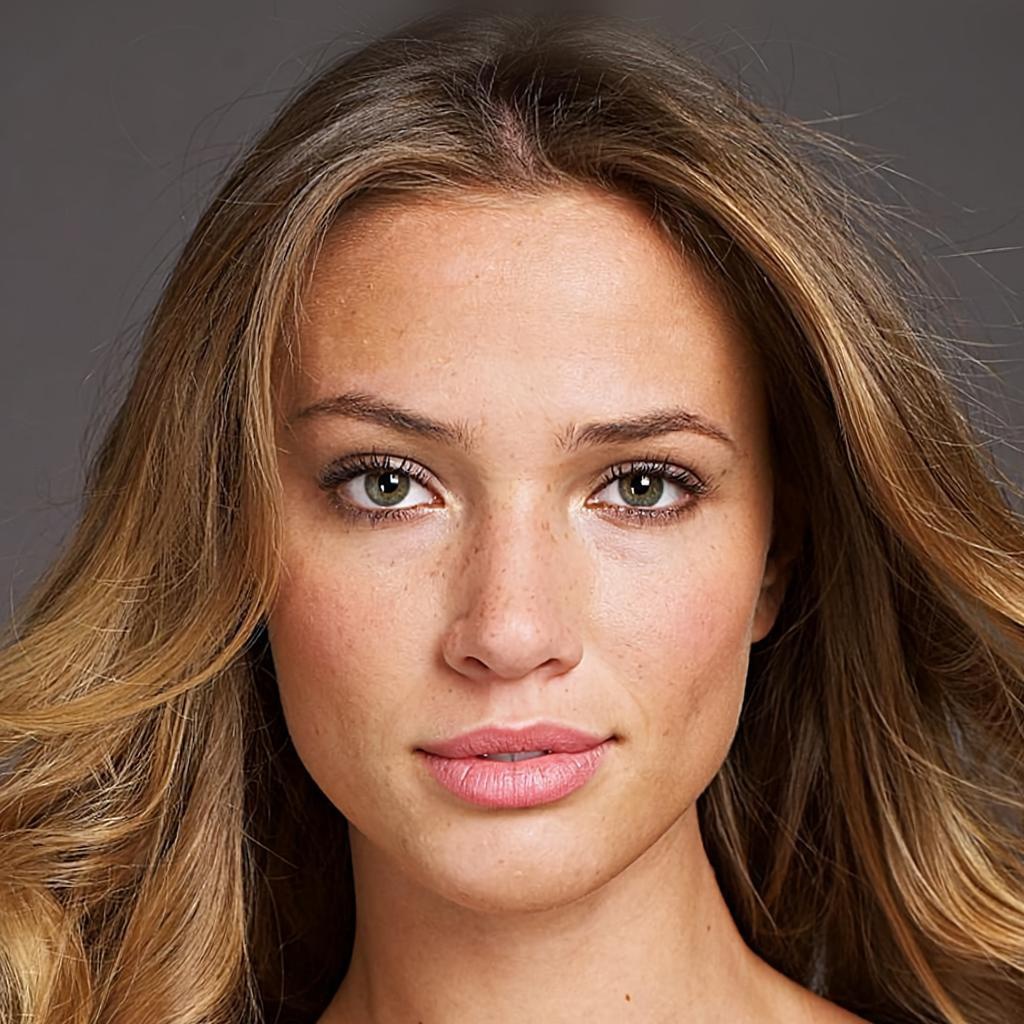} & \im{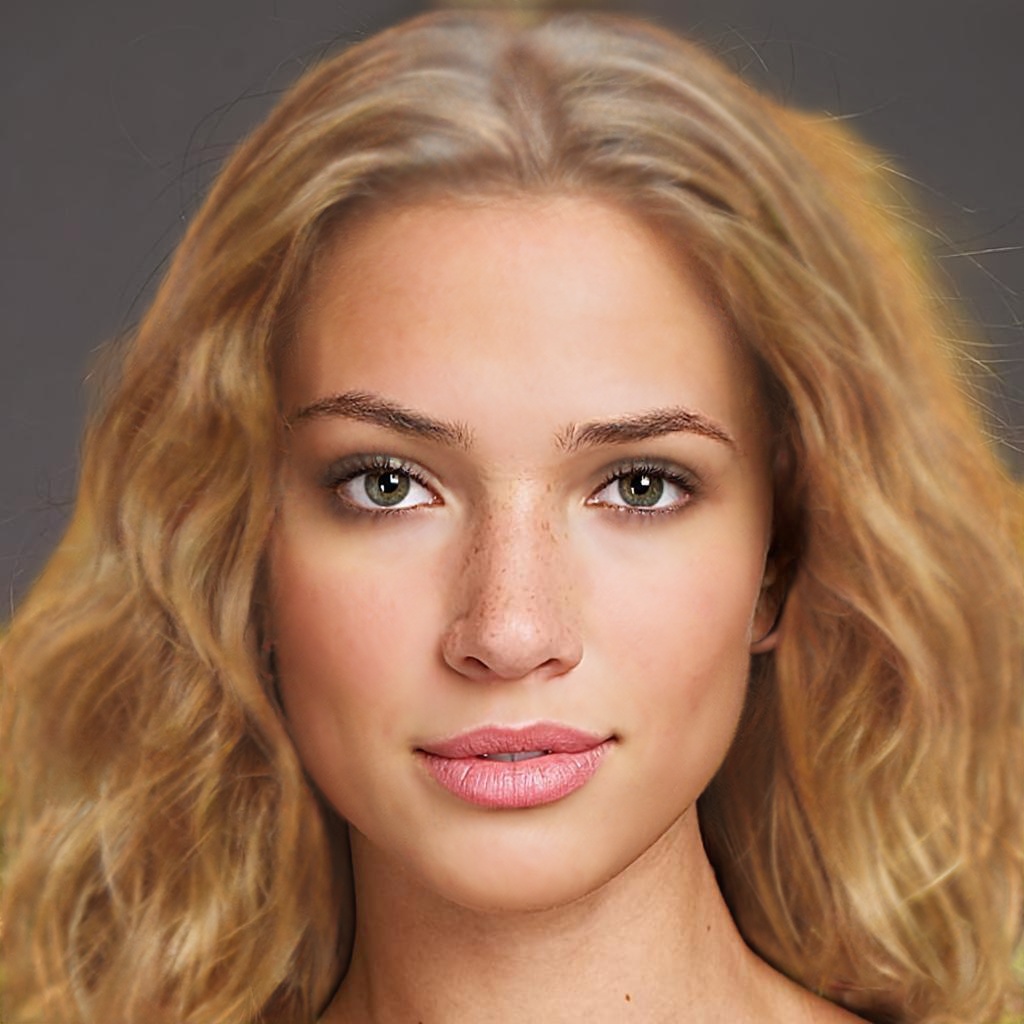} & \im{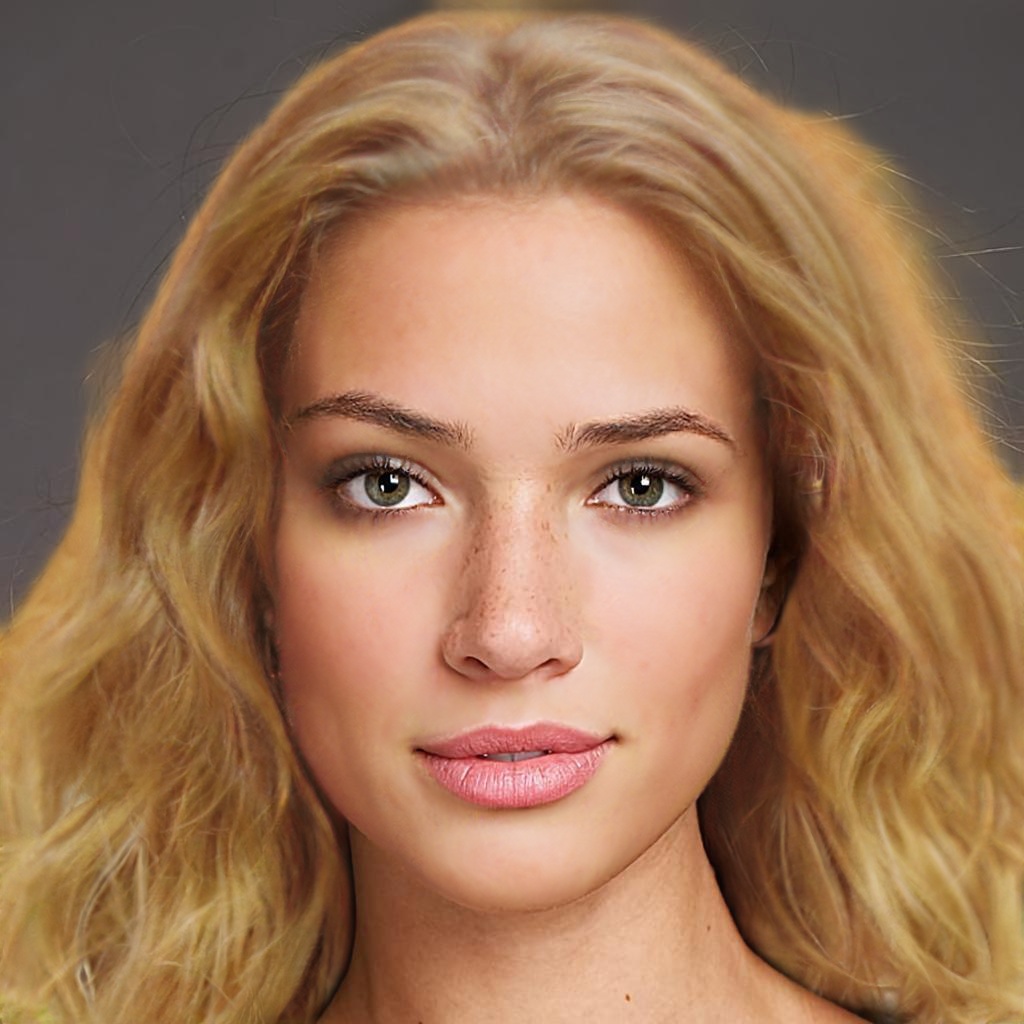} & \im{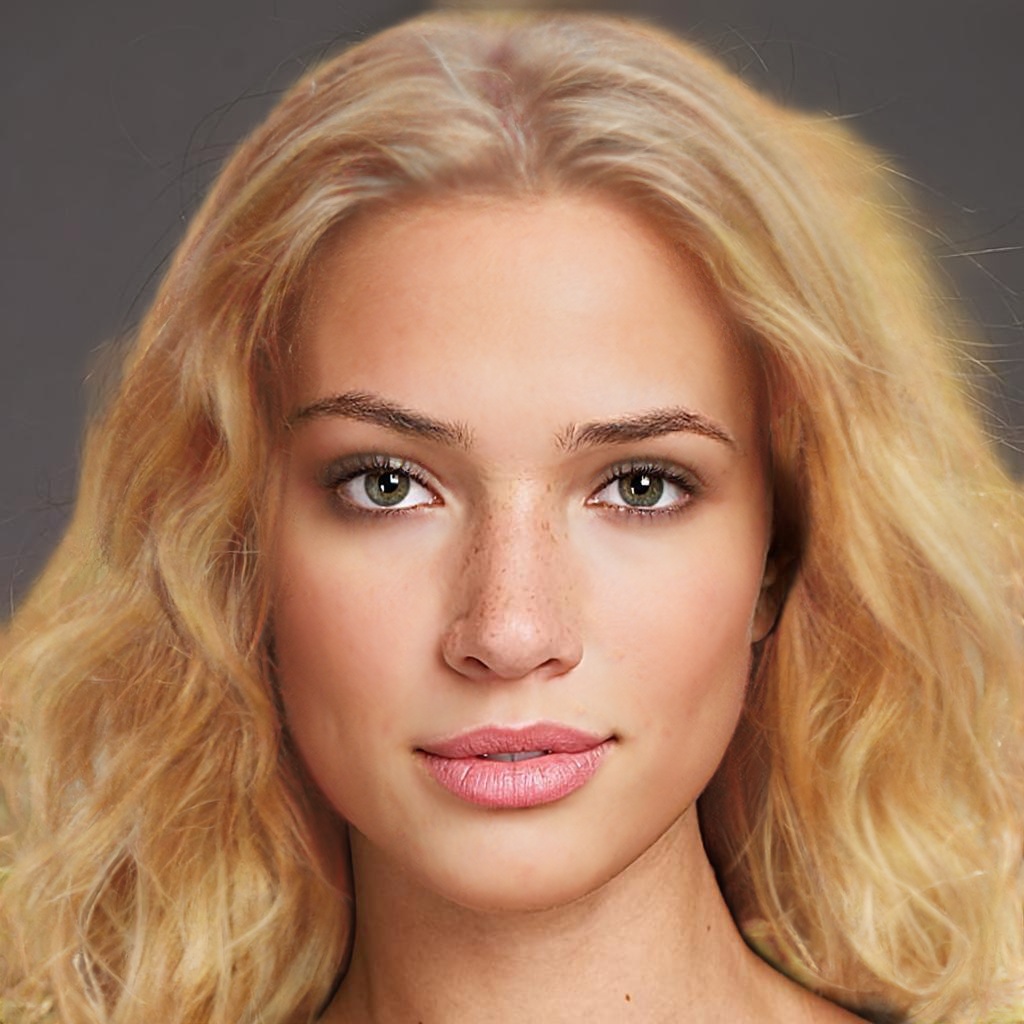} \\
Arched eyebrows & \im{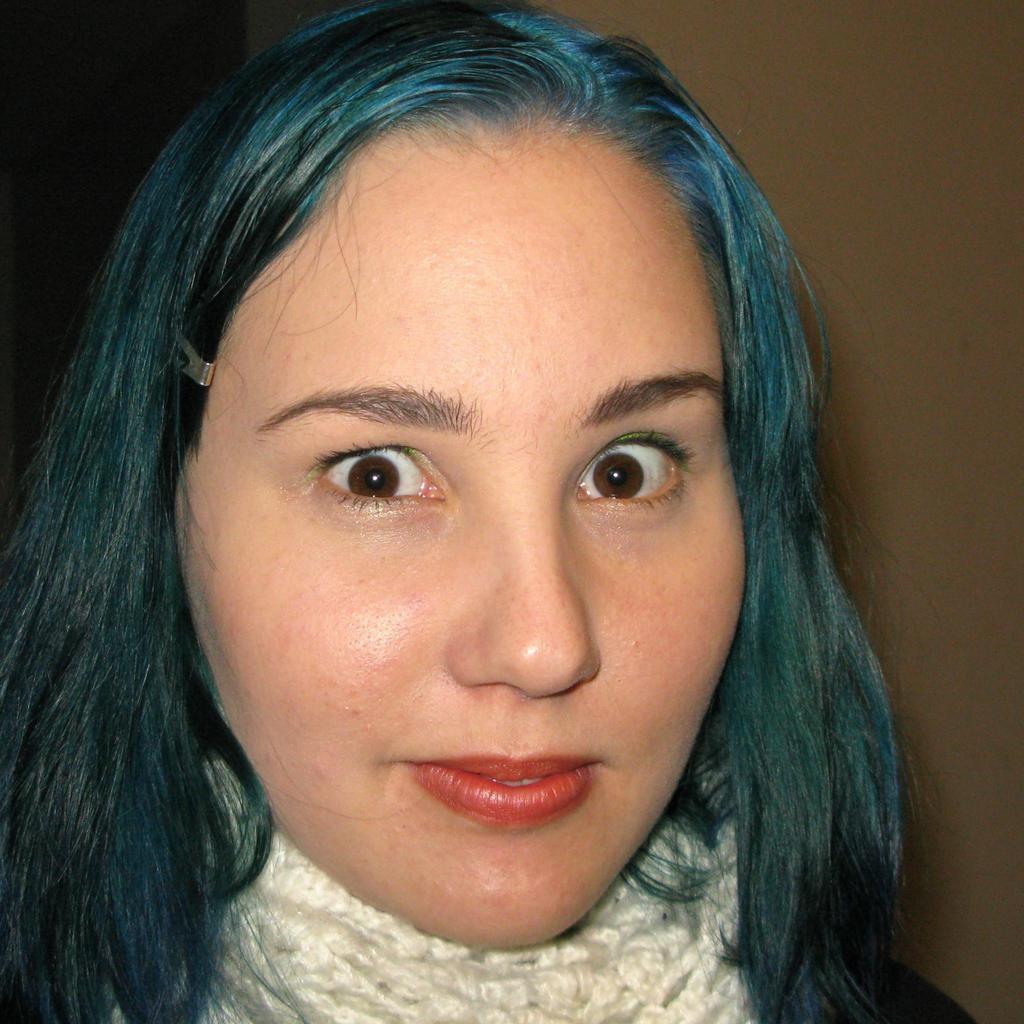} & \im{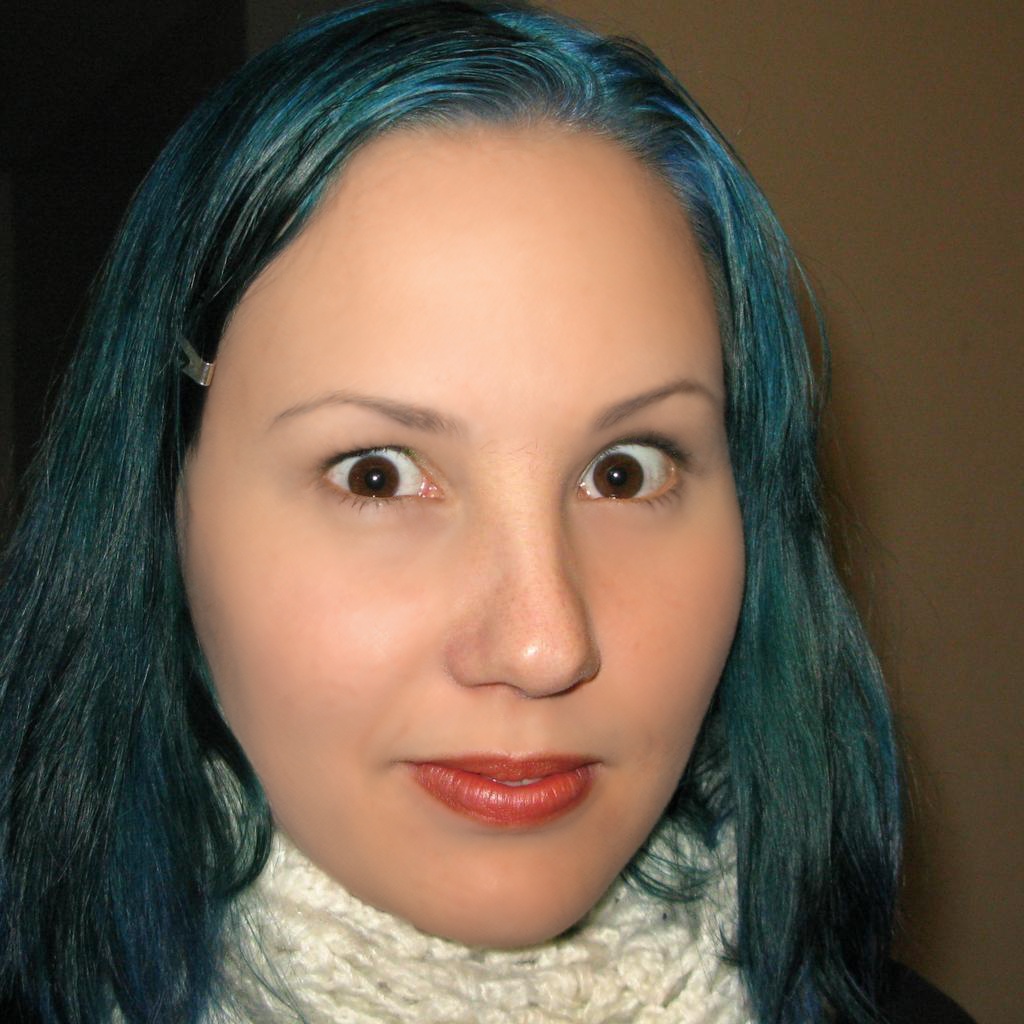} & \im{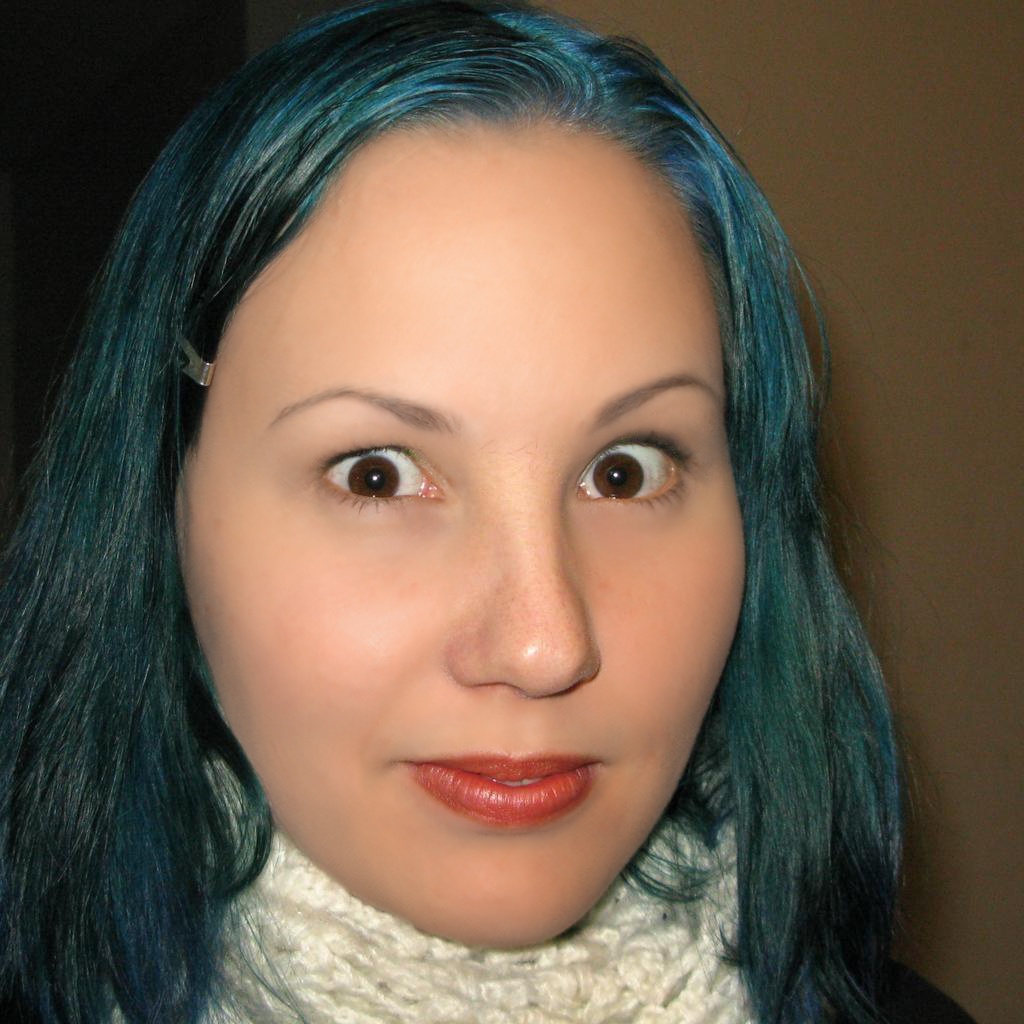} & \im{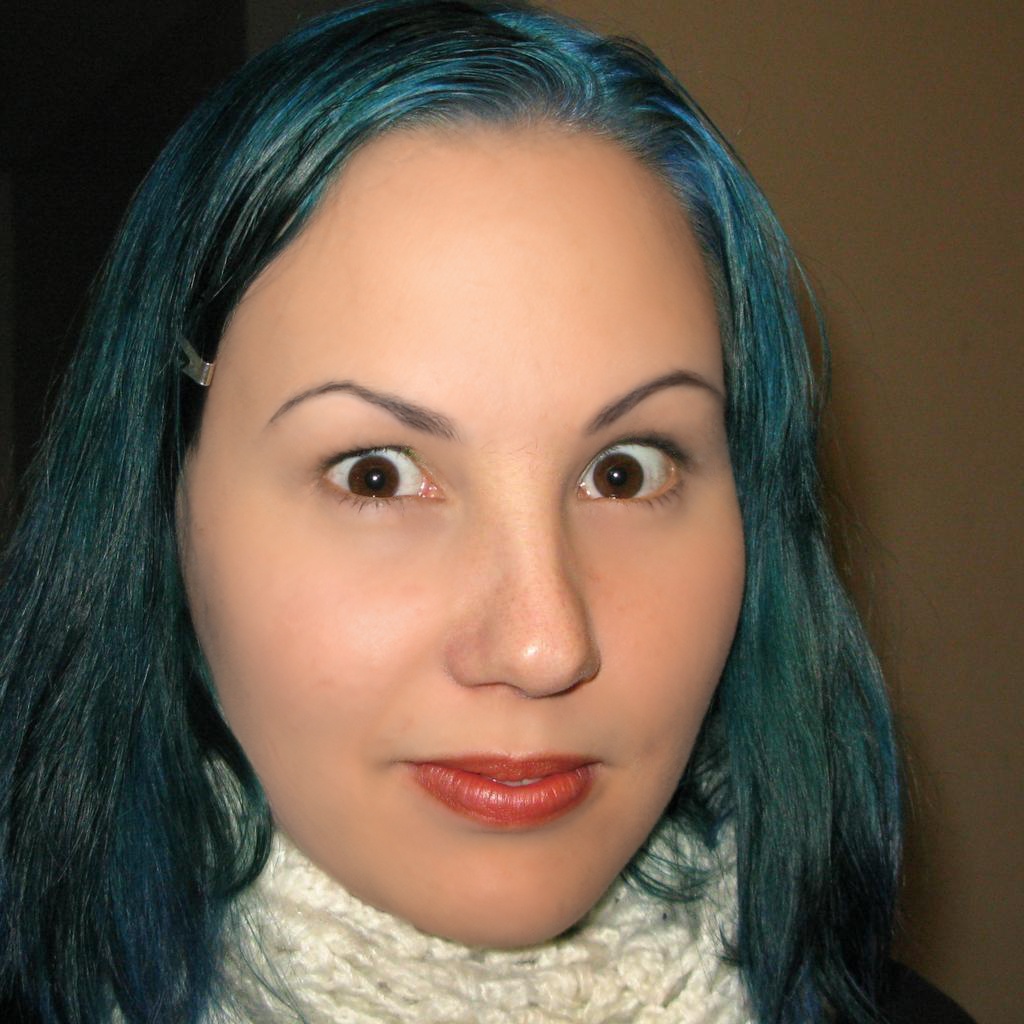} \\
Smiling & \im{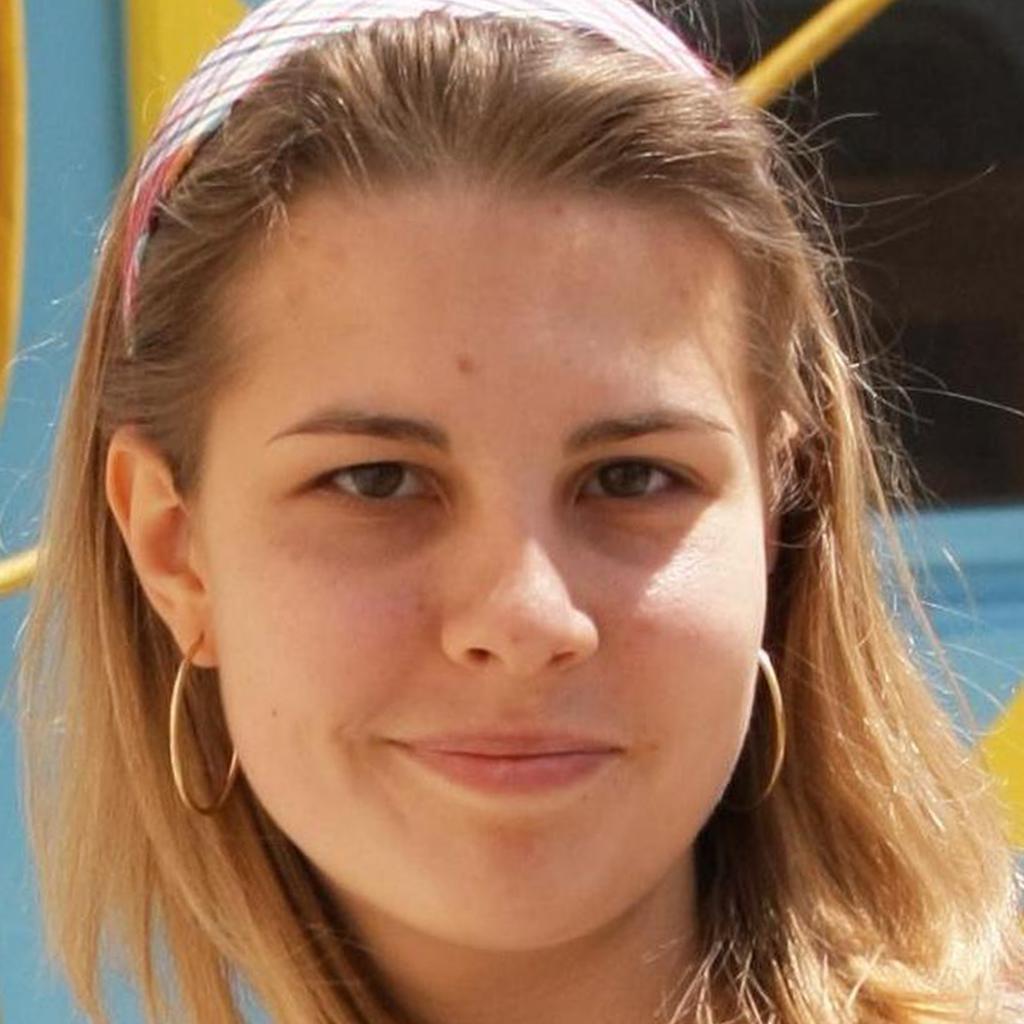} & \im{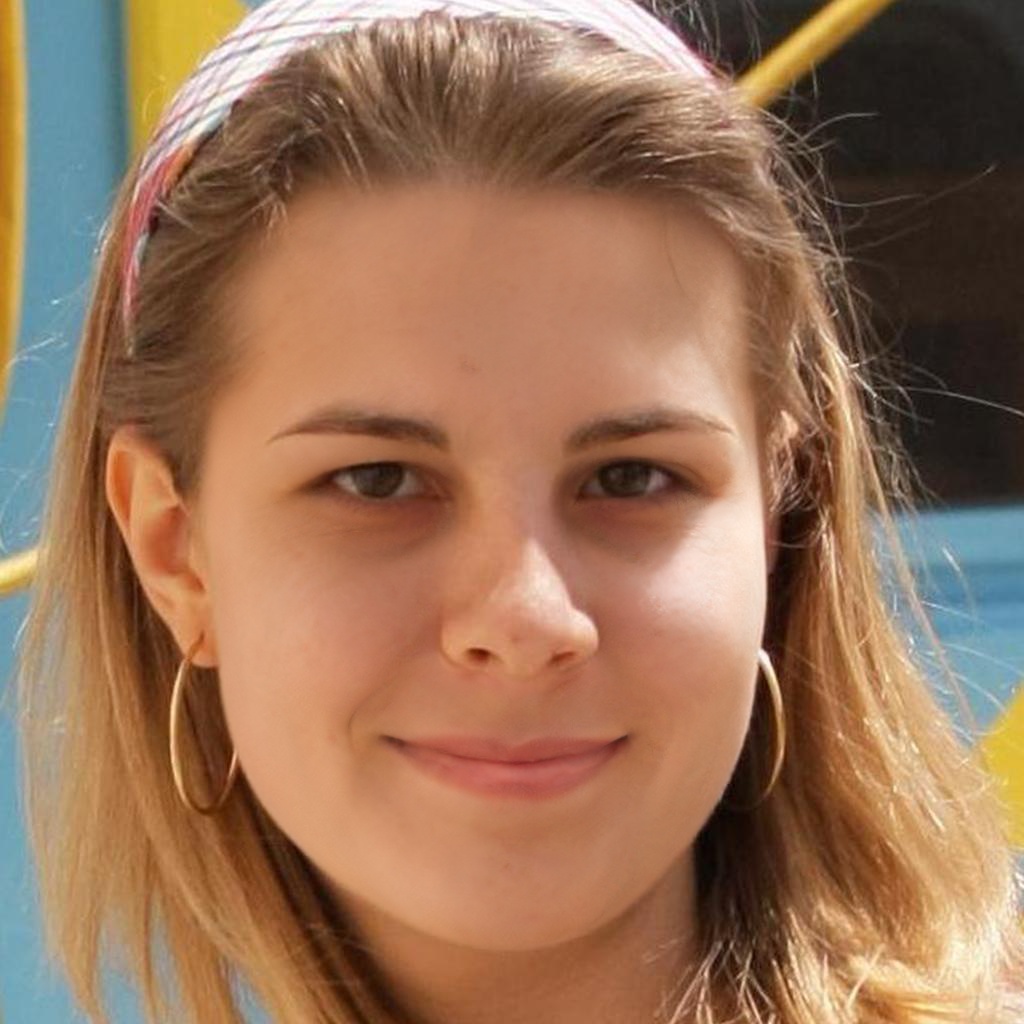} & \im{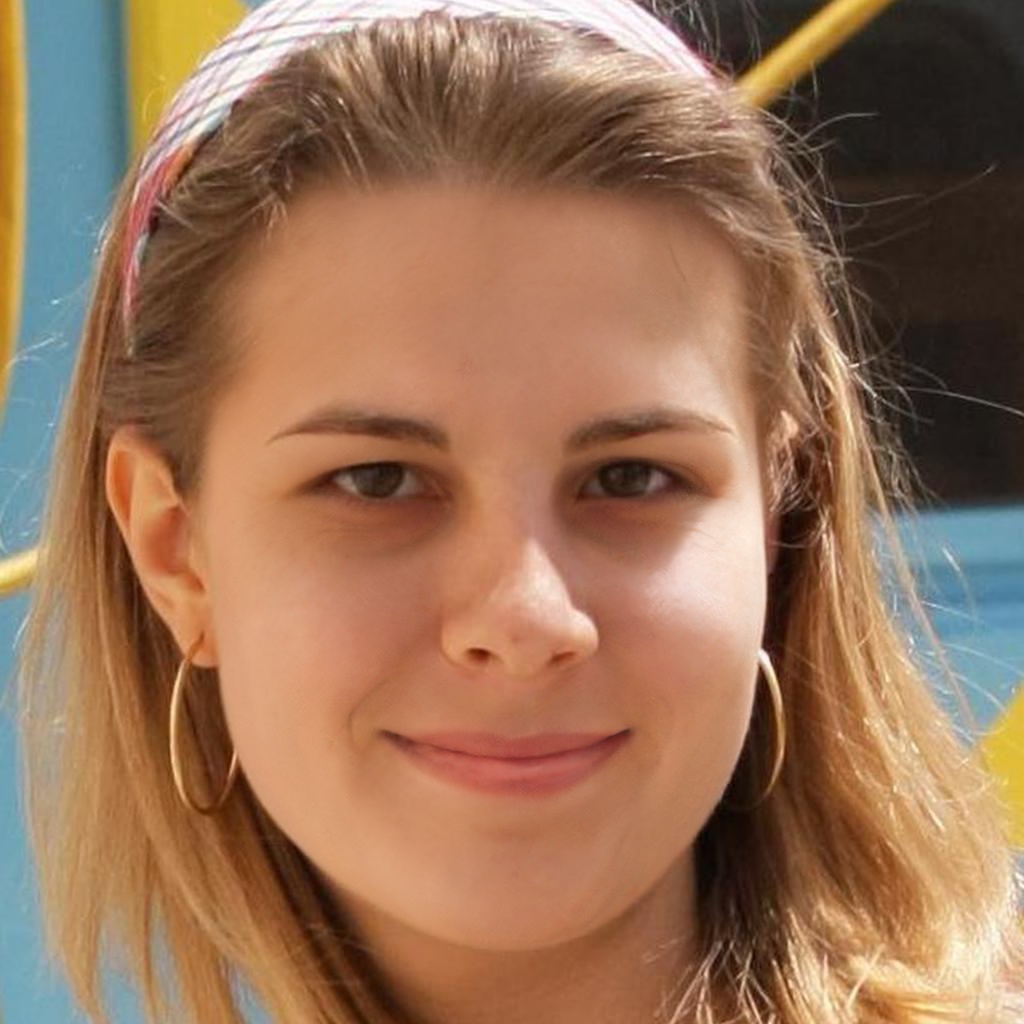} & \im{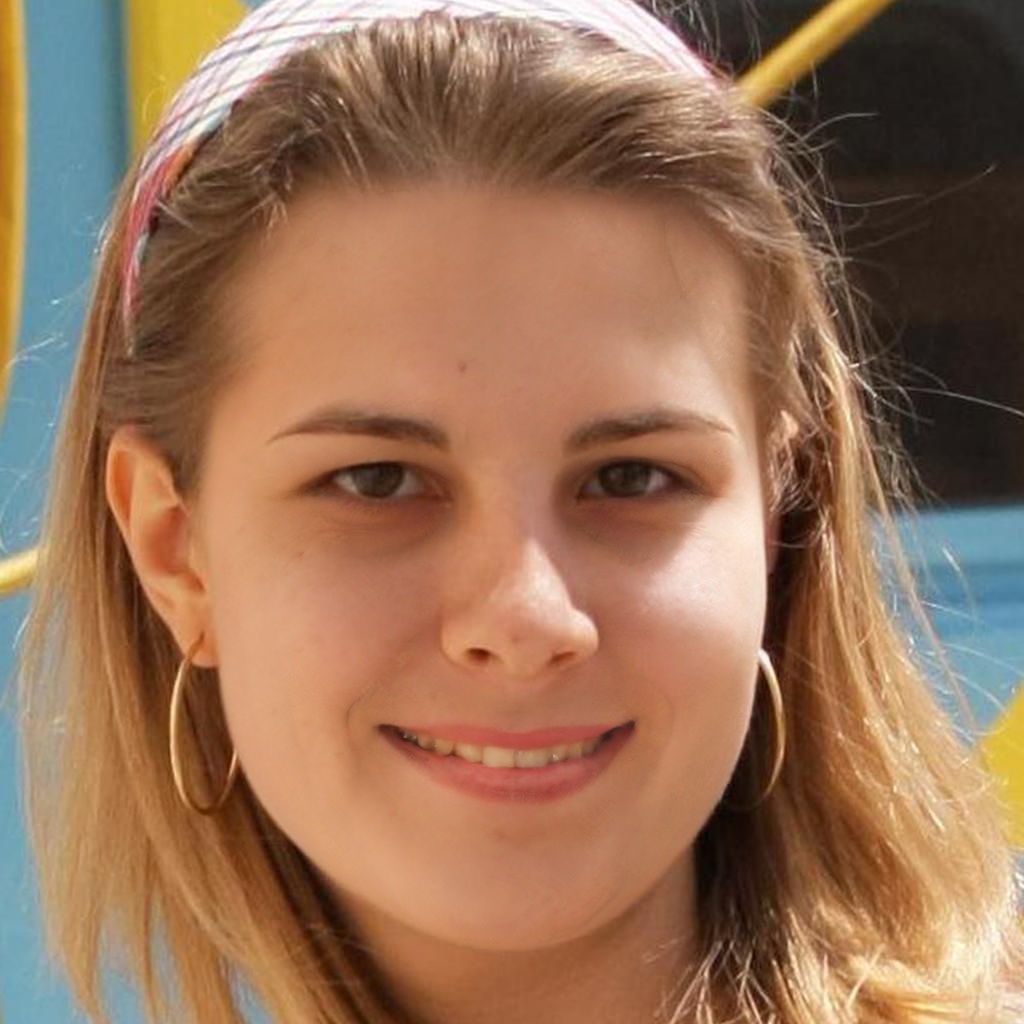} \\
Big nose & \im{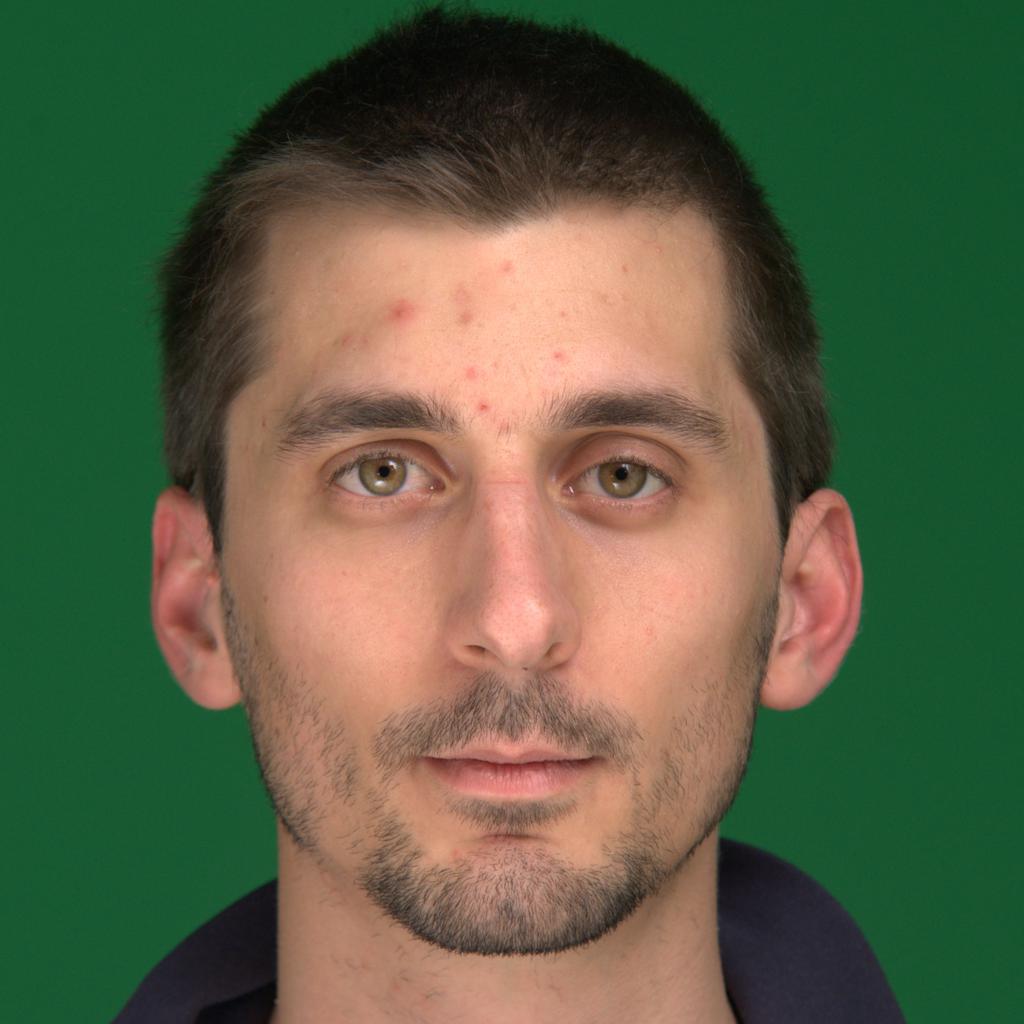} & \im{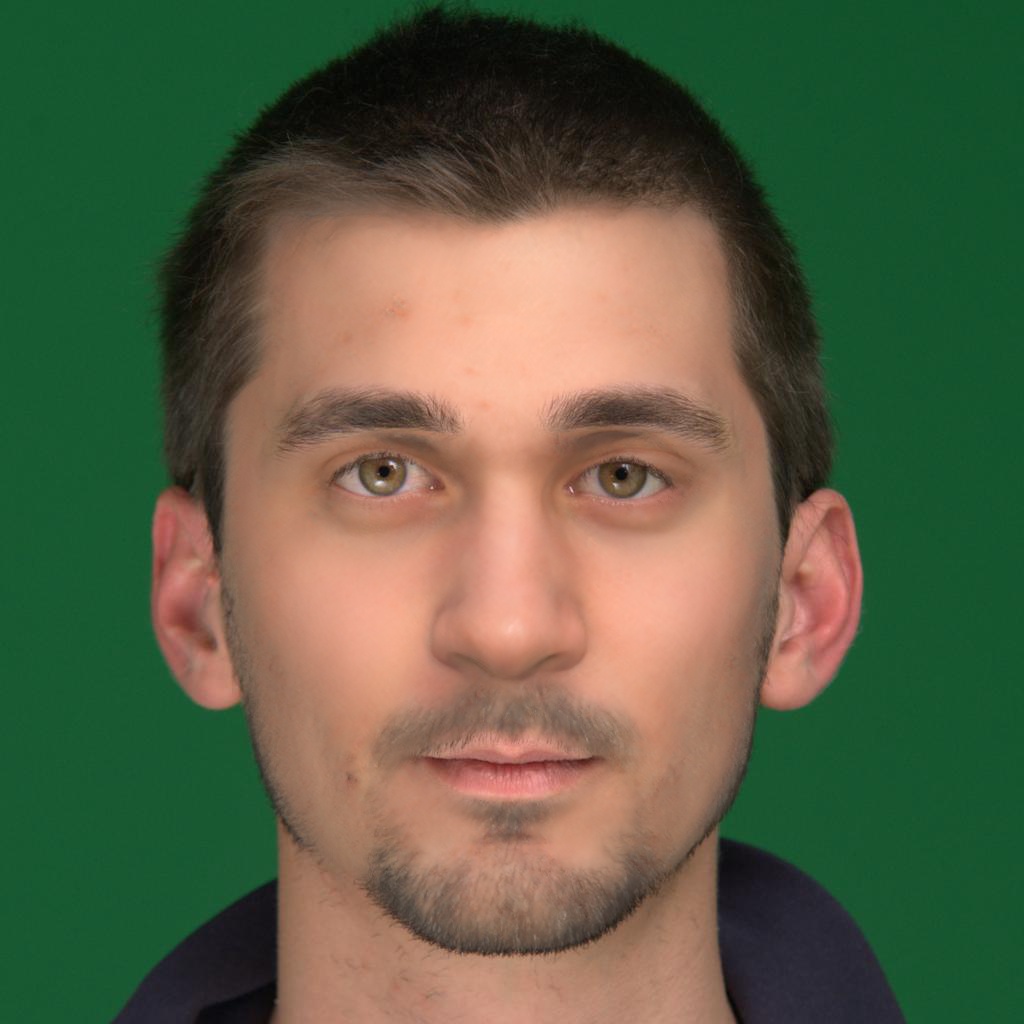} & \im{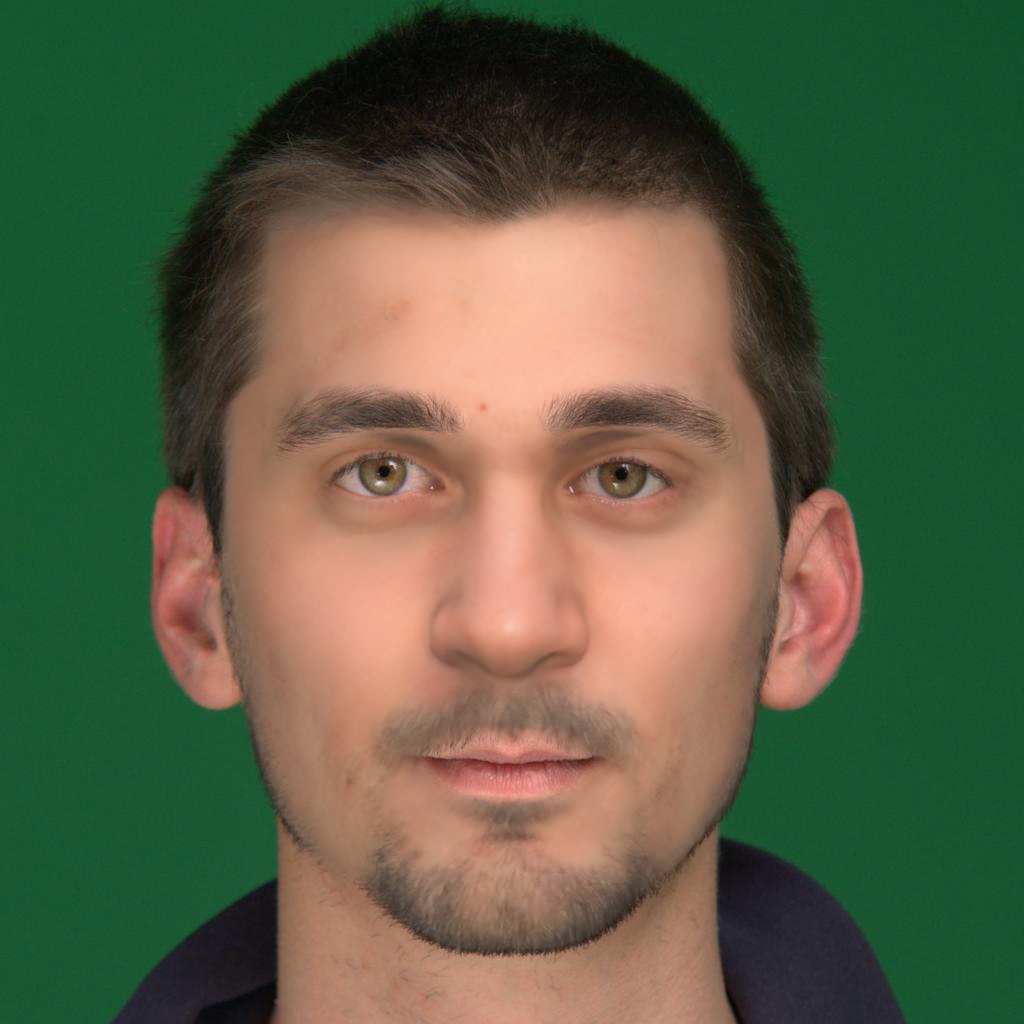} & \im{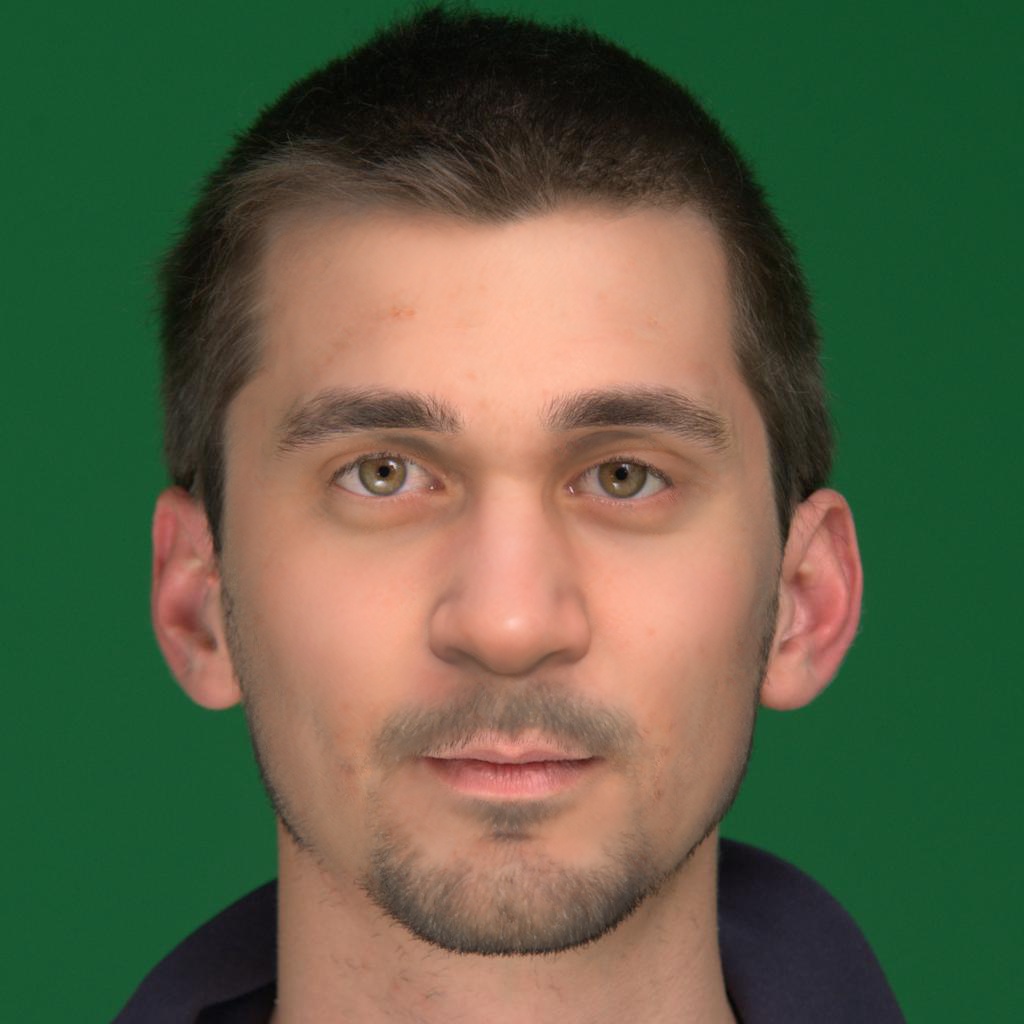} \\
Pointy nose & \im{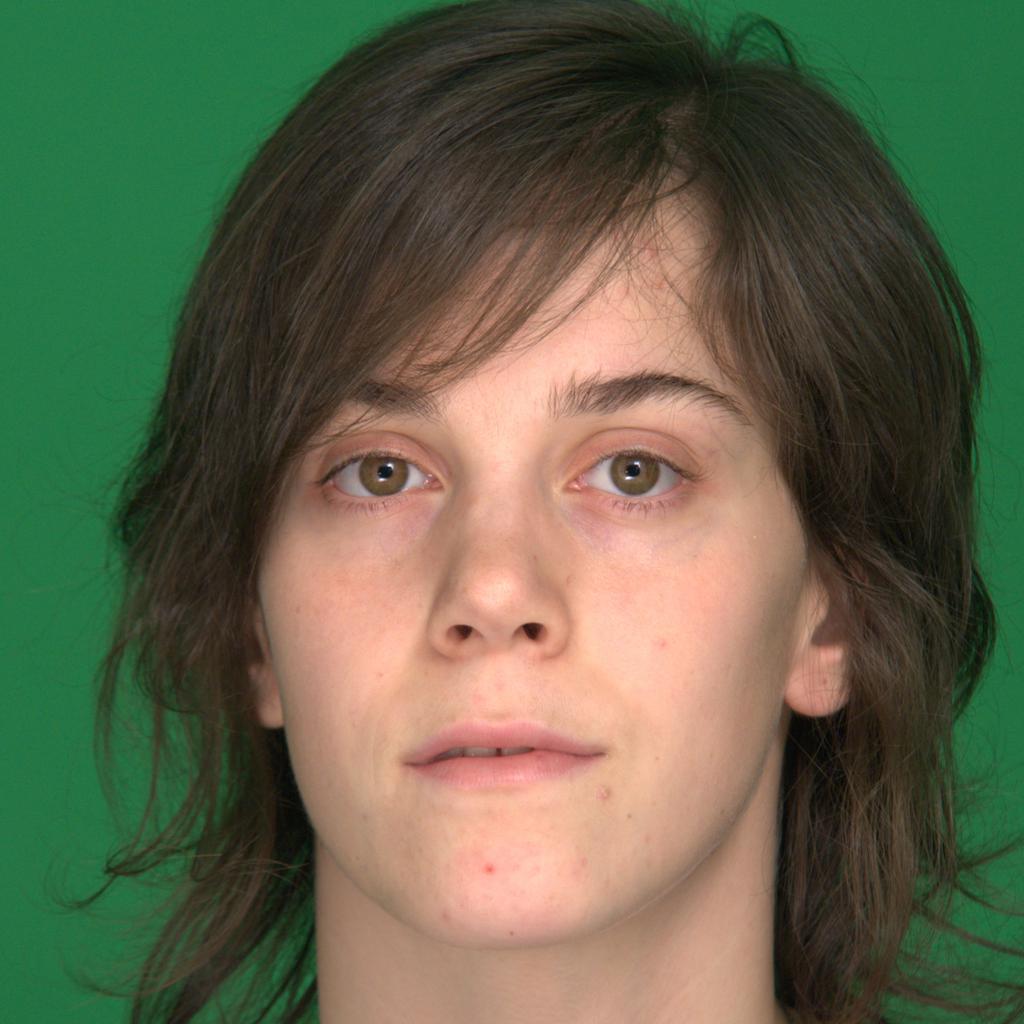} & \im{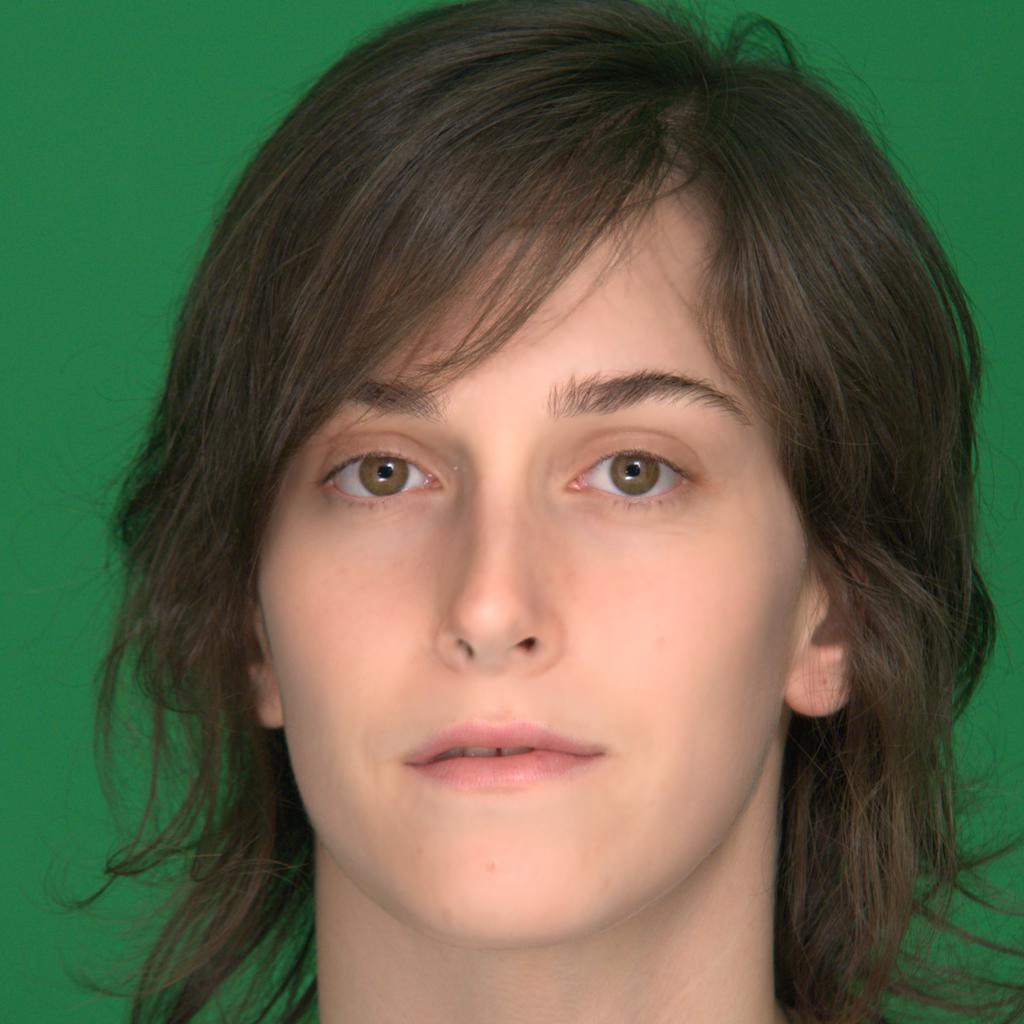} & \im{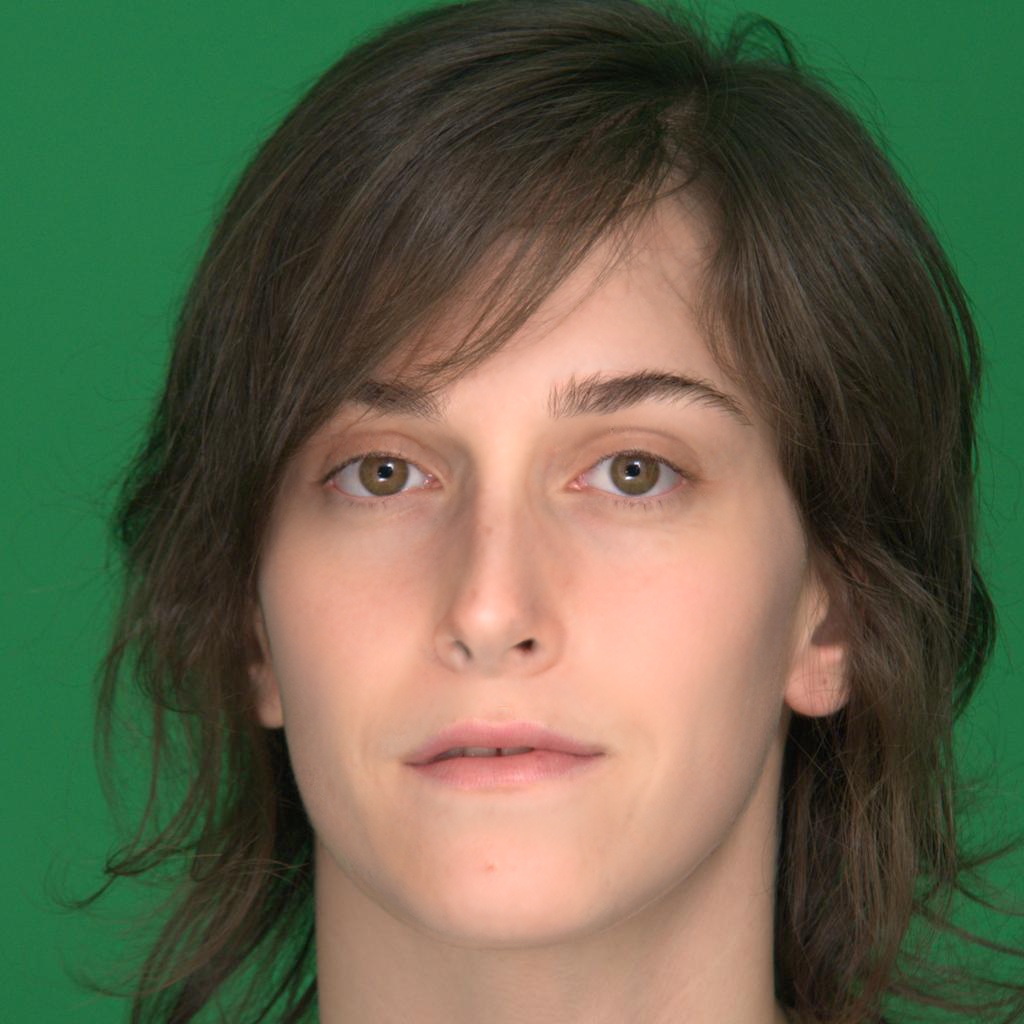} & \im{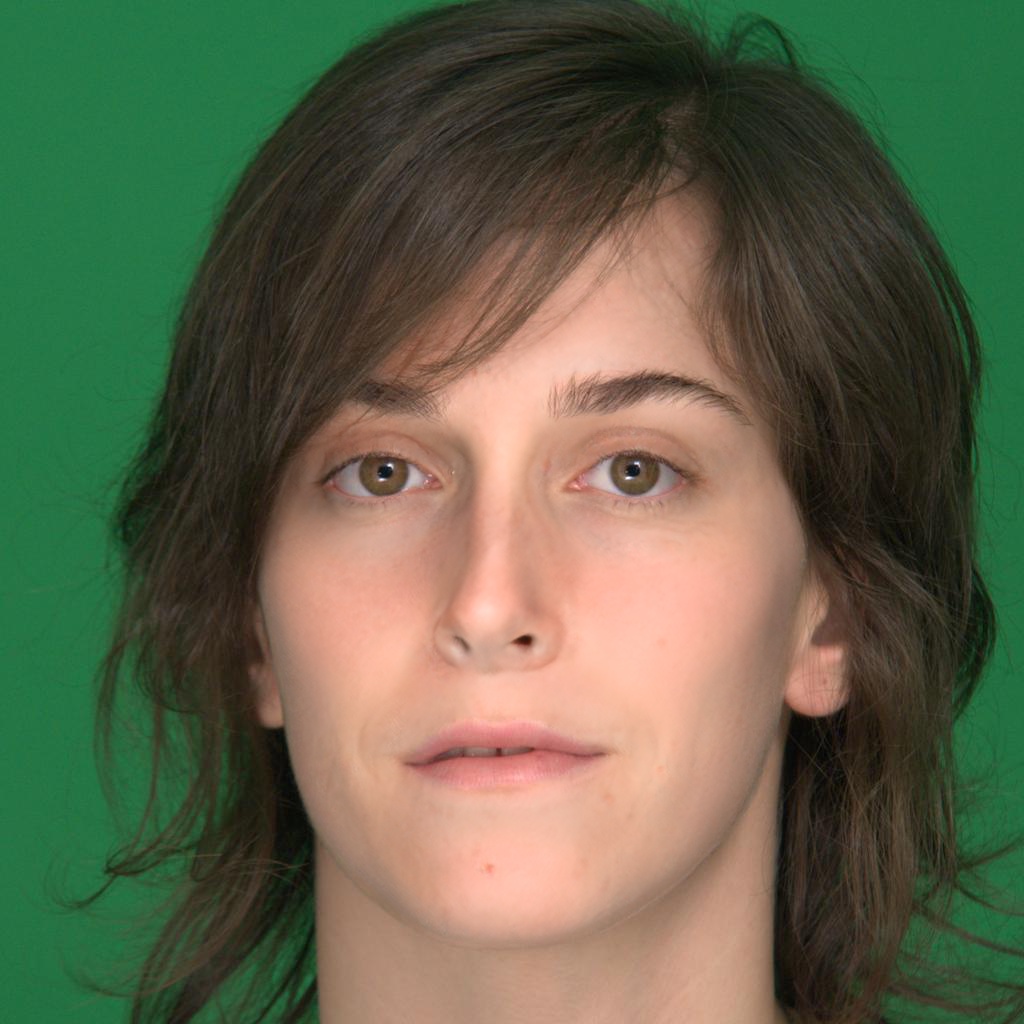} \\
& Original & $\epsilon = 0.10$ & $\epsilon = 0.05$ & $\epsilon = 0.00$ \\
\end{tabular}
\caption{Examples of intensity control for various values of the smoothing parameter $\epsilon$. MaskFaceGAN allows for fine-grained control over attribute appearance despite being trained with binary attribute labels only. Intensity control can be applied to a wide variety of facial attributes that affect color, shape or even more complex changes of facial appearance.}
\label{fig:intensity}
\end{figure*}

\renewcommand{\im}[1]{\includegraphics[width=\figureimagewidth]{images/attribute_size/appendix/#1}}
\renewcommand{\imrow}[2]{\im{#2--original.jpg} & \im{#2--#1--0,5.jpg} & \im{#2--#1--1,0.jpg} & \im{#2--#1--1,5.jpg}}

\subsection{Component Size Manipulation}
\begin{figure*}[!th!]
\begin{tabular}{DCCCC}
Straight hair & \imrow{straight_hair}{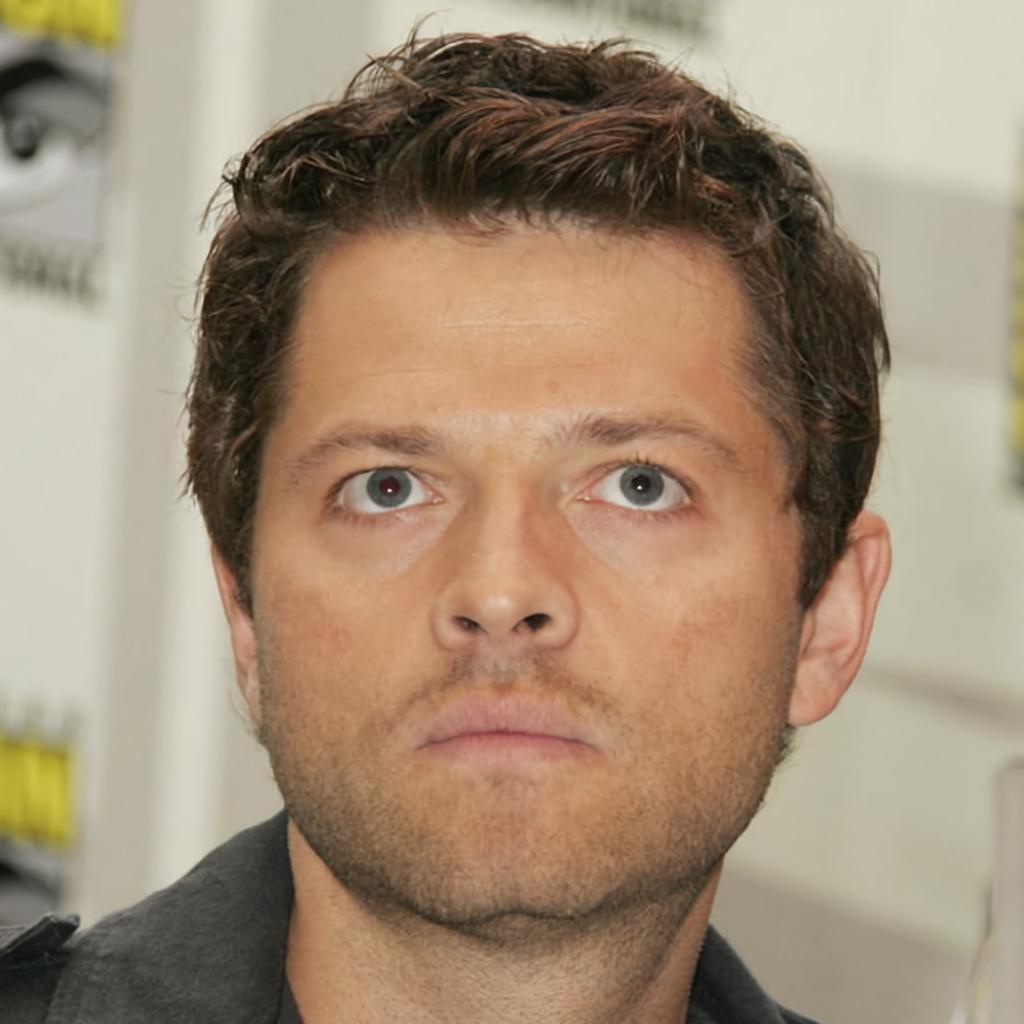} \\
Pointy nose & \imrow{pointy_nose}{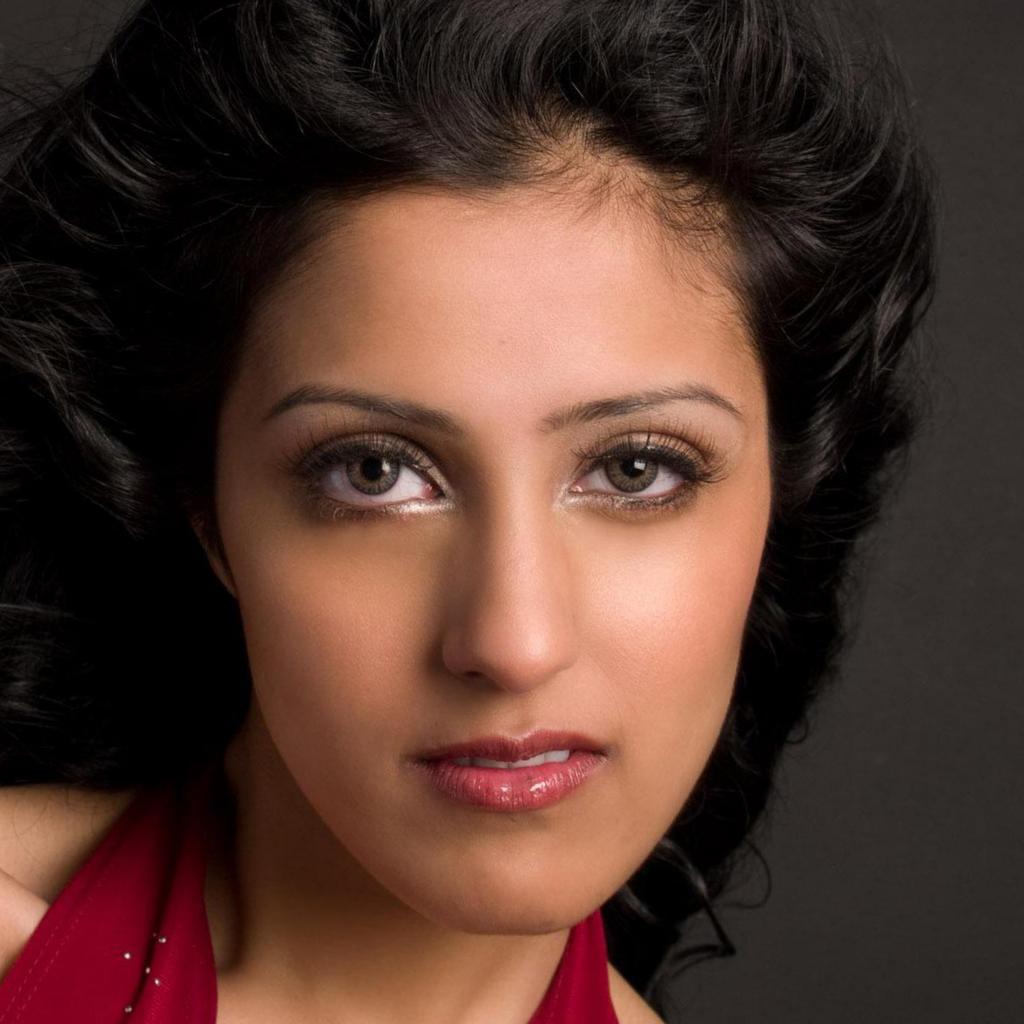} \\
Arched eyebrows & \imrow{arched_eyebrows}{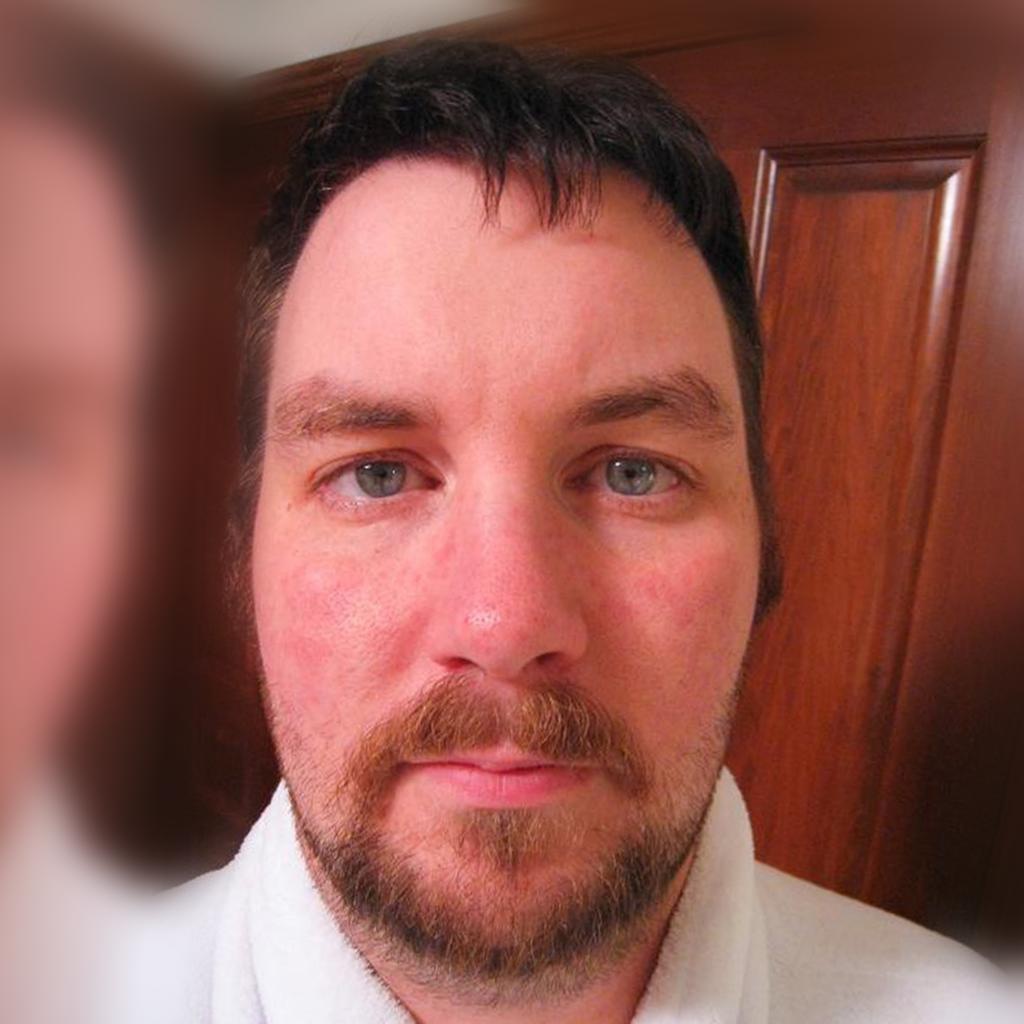} \\
Big nose & \imrow{big_nose}{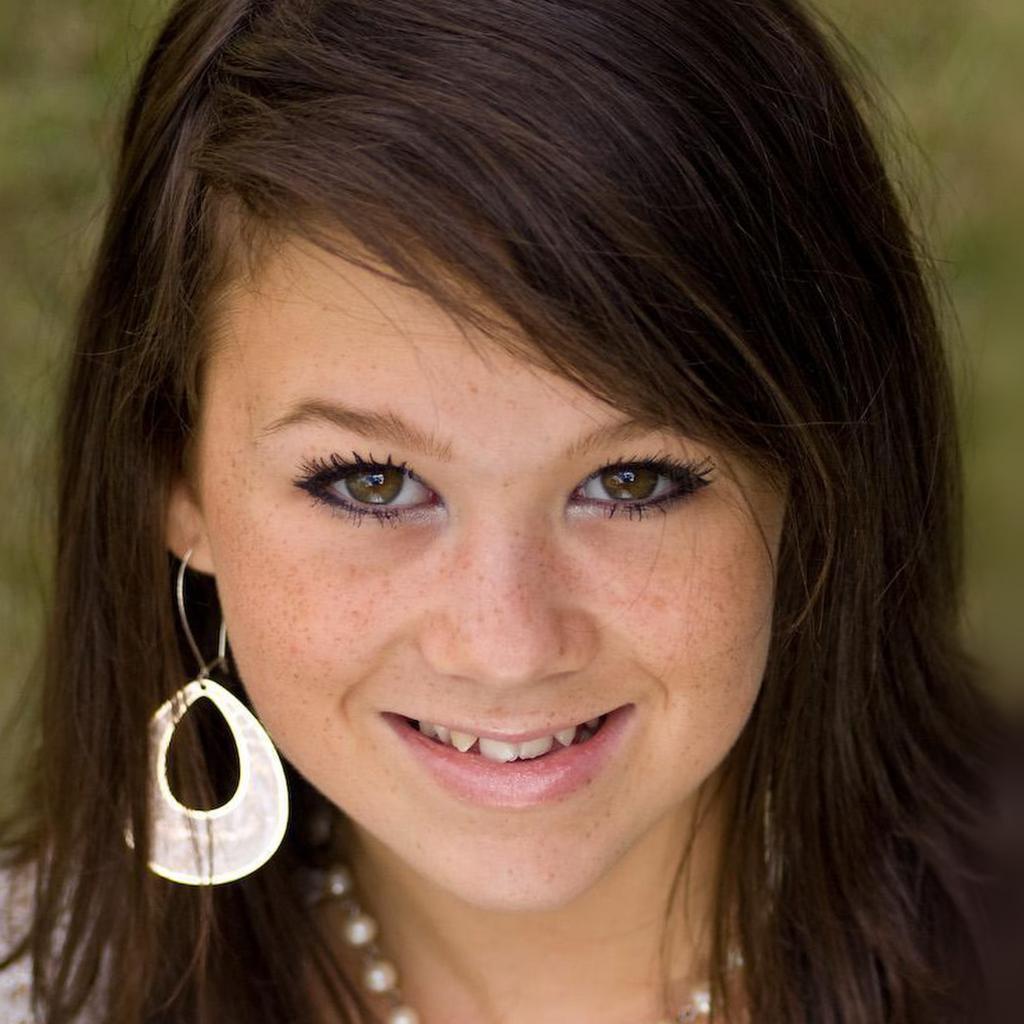} \\
Smiling & \imrow{smiling}{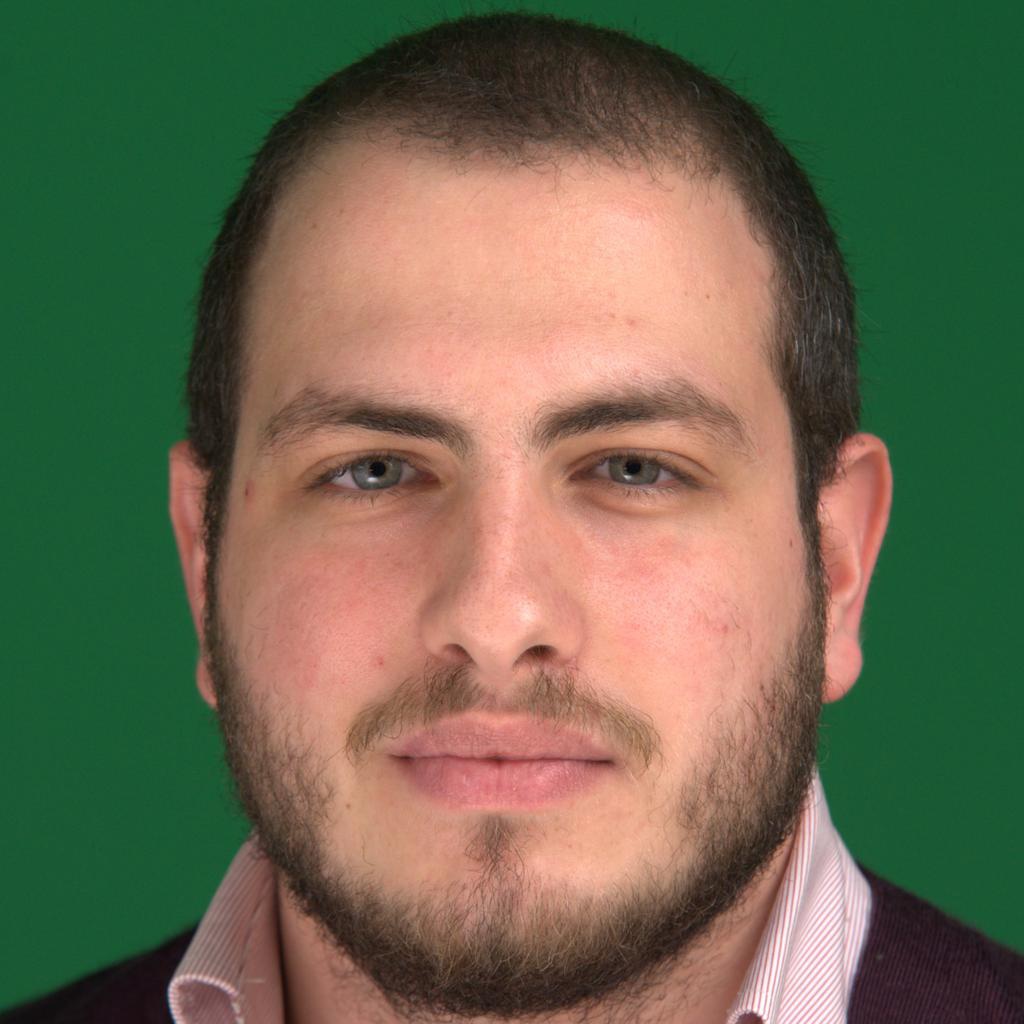} \\

& Original & $\alpha = 0.5$ & $\alpha=1.0$ & $\alpha= 1.5$ \\
\end{tabular}
\caption{Examples of component size manipulation for various scaling factors $\alpha$. MaskFaceGAN allows growing the spatial region within which a targeted attribute is edited. This results in flexible image editing that can generate images with different variations of a targeted attribute in terms of size.}
\label{fig:size}
\end{figure*}

Fig. \ref{fig:size} presents results that demonstrate MaskFaceGAN's ability to manipulate the size of the spatial region to be edited for a given target attribute. Specifically, the figure shows results for five targeted attributes and different scaling factors $\alpha$, which defines the target attribute size with respect to the initial  size of the region associated with a given attribute. Varying the scaling factor $\alpha$ allows MaskFaceGAN to grow or shrink certain face parts. Due to the shape term in the loss equation that assures consistent blending, we only allow growing of hair (not shrinking). Other face components, such as the nose, eyebrows or mouth, on the other hand, can be shrunk as well. As can be seen from the presented examples, the ability of MaskFaceGAN to control the size of the spatial region during editing enables diverse image manipulations around the same targeted attribute and represents a unique feature of the proposed editing procedure. 

\subsection{Multiple Attribute Editing}
Fig. \ref{fig:multiple_attributes} presents visual results when editing multiple attributes with a single optimization procedure. We show examples for jointly editing two attributes but our experiments suggest that considering more than two attributes leads to realistic results as well. As illustrated in Fig. \ref{fig:multiple_attributes}, the editing procedure can also \textit{target the same facial components/region} even if multiple attributes are edited at the same time -- see the (\textit{Wearing lipst.}, Mouth sl. open), (Bushy eyeb., Arched eyeb.), (\textit{Smiling}, \textit{Wearing lipst.}) and (Smiling, Mouth sl. open) results. Despite having to accomodate different target semantics in the same spatial region, MaskFaceGAN generates realistic facial appearances and produces virtually no visual artefacts.   

\renewcommand{\figureimagewidth}{.24\textwidth}

\renewcommand{\miniwidth}{.22\textwidth}
\renewcommand{\im}[1]{\includegraphics[width=\textwidth]{images/multiple_attributes/two_attributes/#1}}
\newcommand{\mysubfigure}[2]{\begin{subfigure}{\miniwidth}
    		\centering
    		\im{#1}\vspace{-1mm}
    		\caption{\small #2}
    	\end{subfigure}}

\begin{figure*}[!th!]
    \captionsetup[subfigure]{labelformat=empty}    

\begin{tabular}{CCCC}

    		\mysubfigure{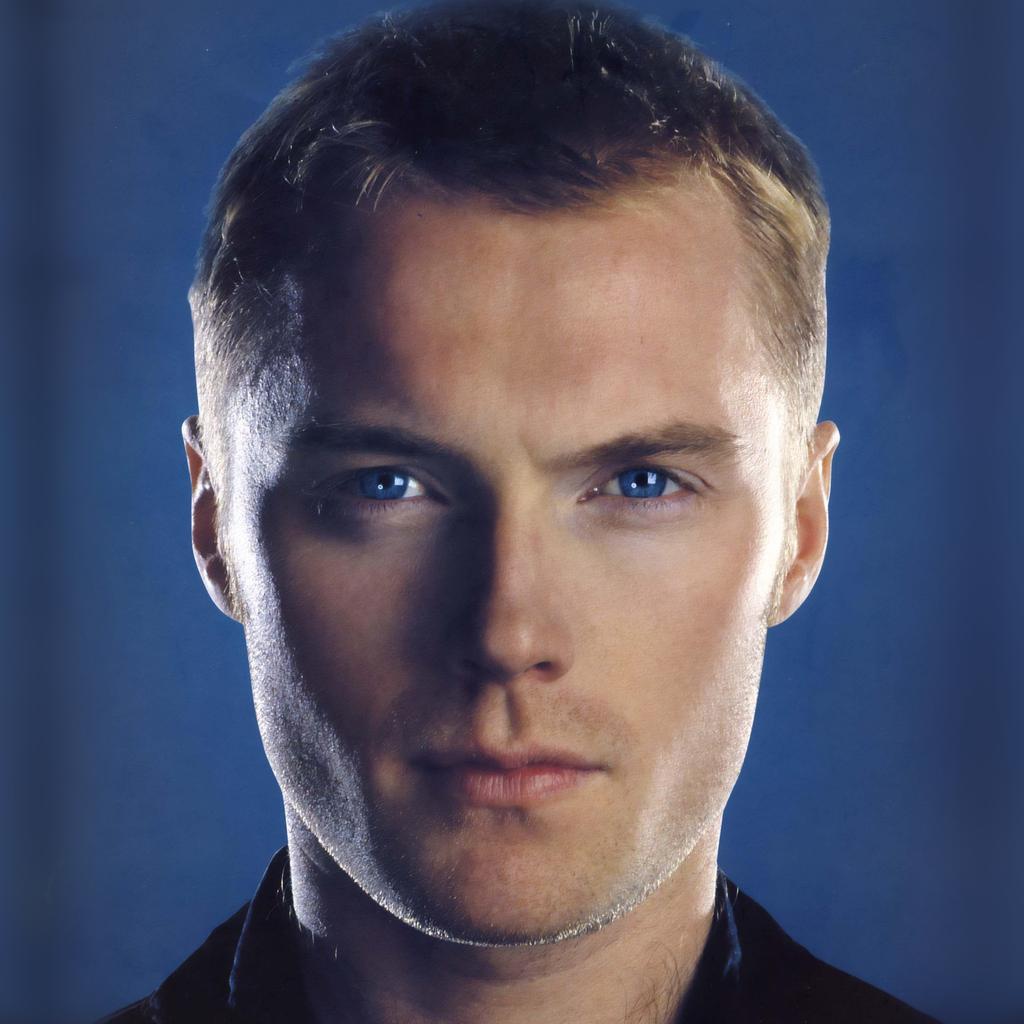}{Original} &
   		\mysubfigure{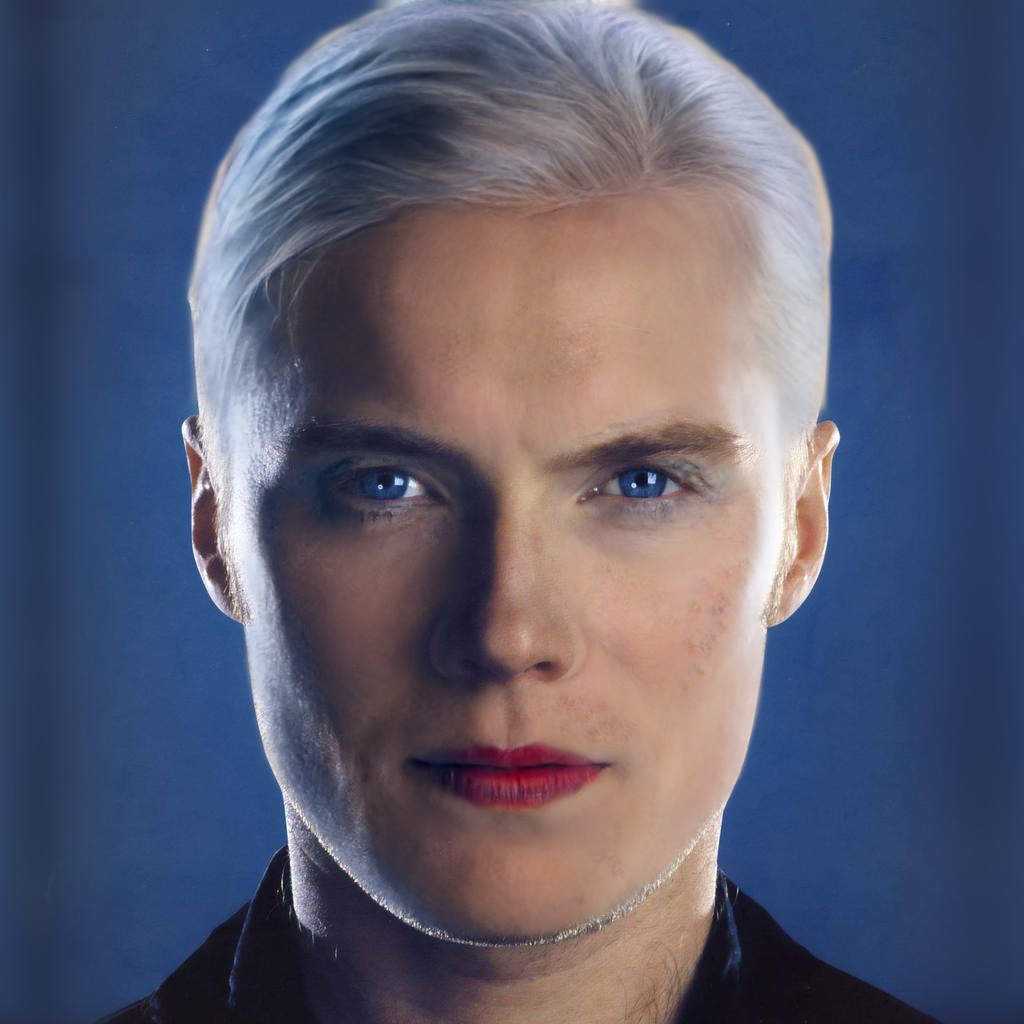}{Wearing lipst., Grey hair} &
    		\mysubfigure{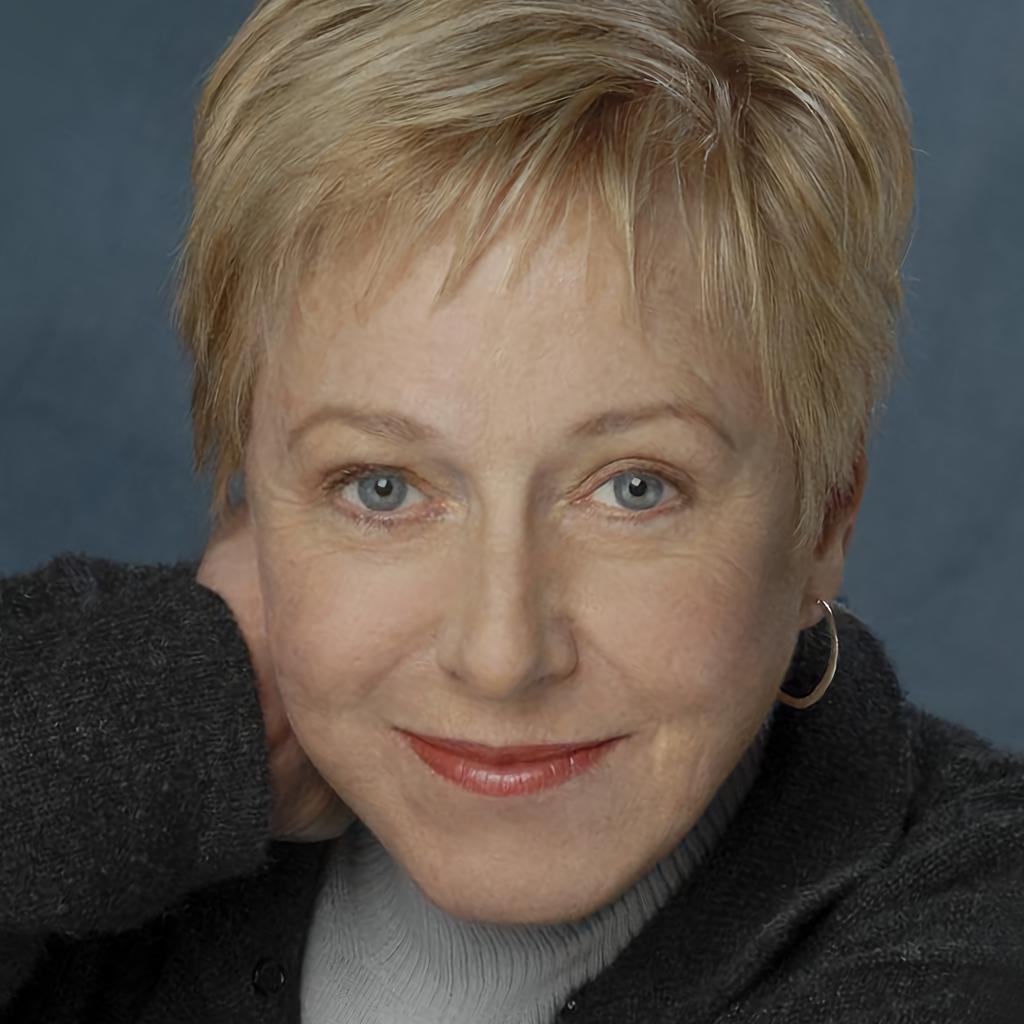}{Original} &
    		\mysubfigure{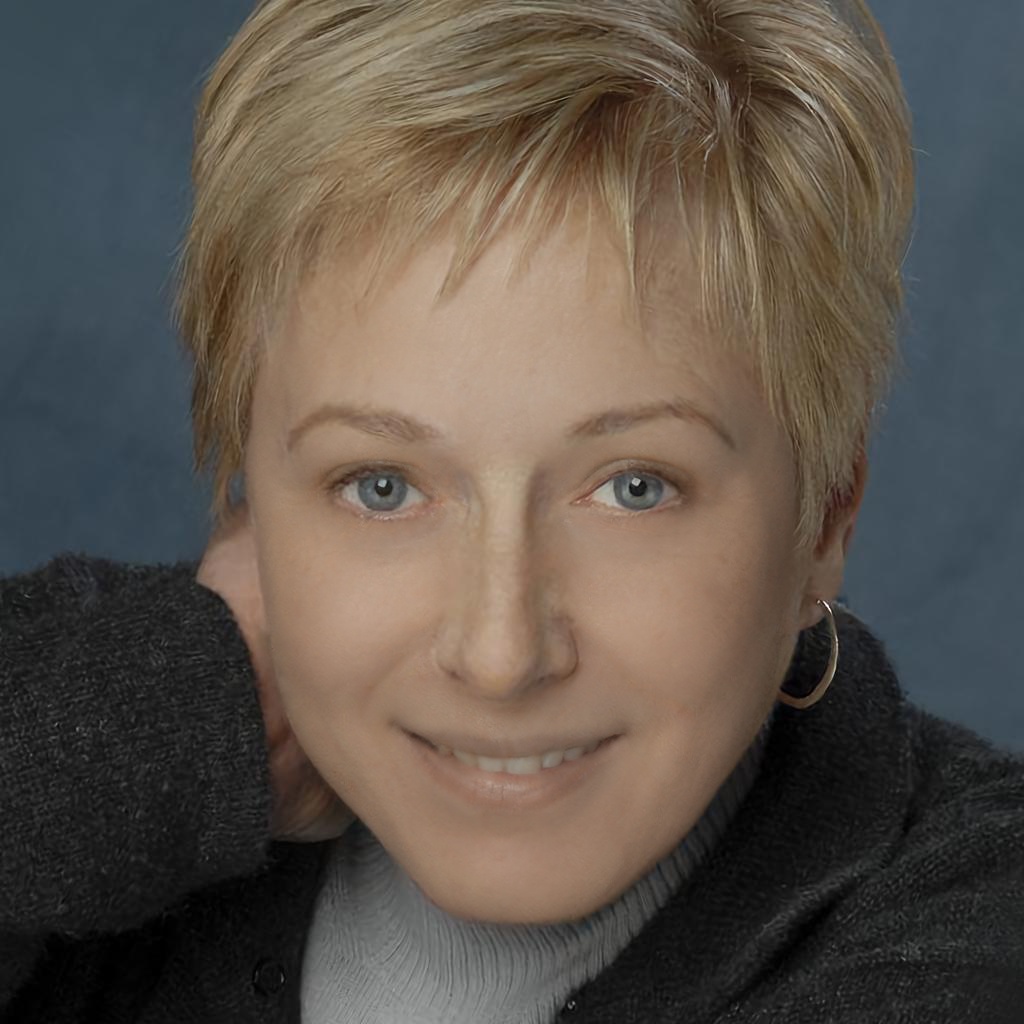}{\textit{Wearing lipst.}, Mouth sl. open} \\
    		&\\[-3mm]
    		\mysubfigure{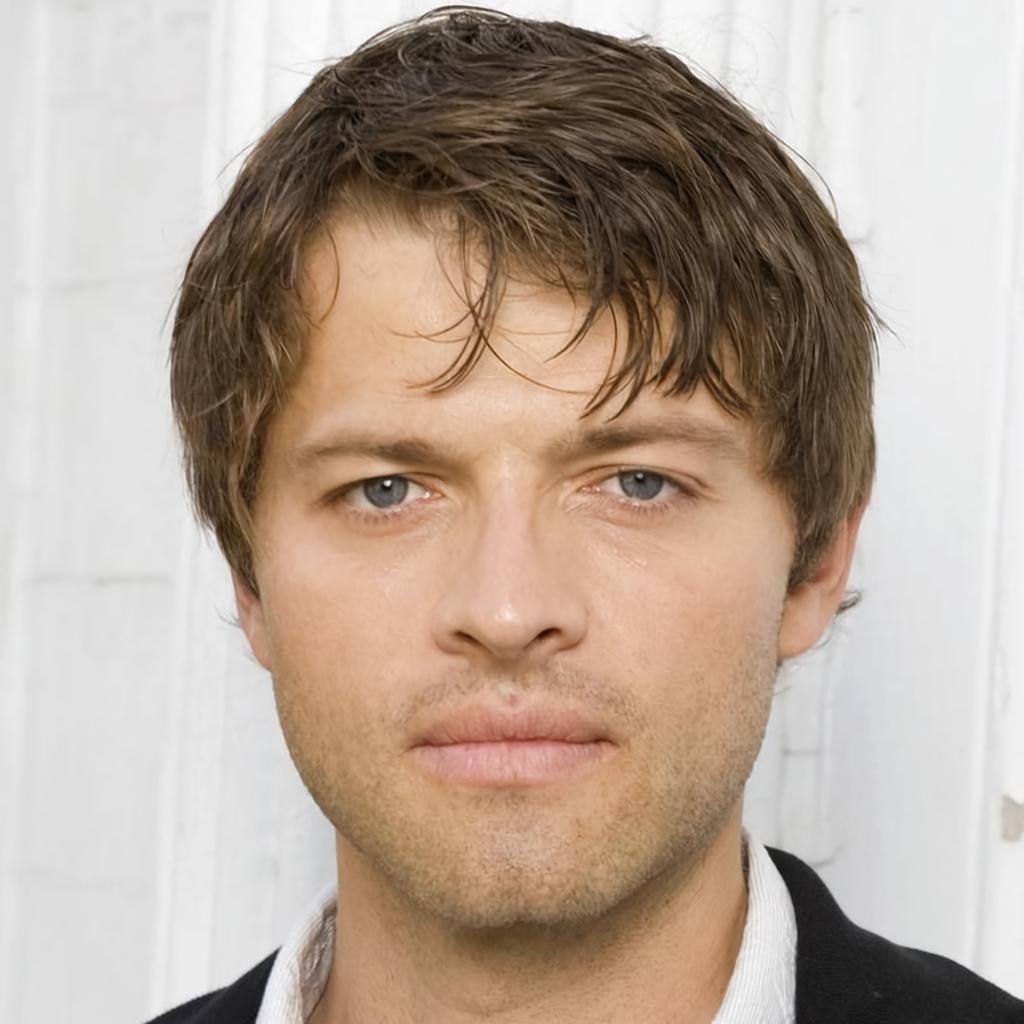}{Original} &
    		\mysubfigure{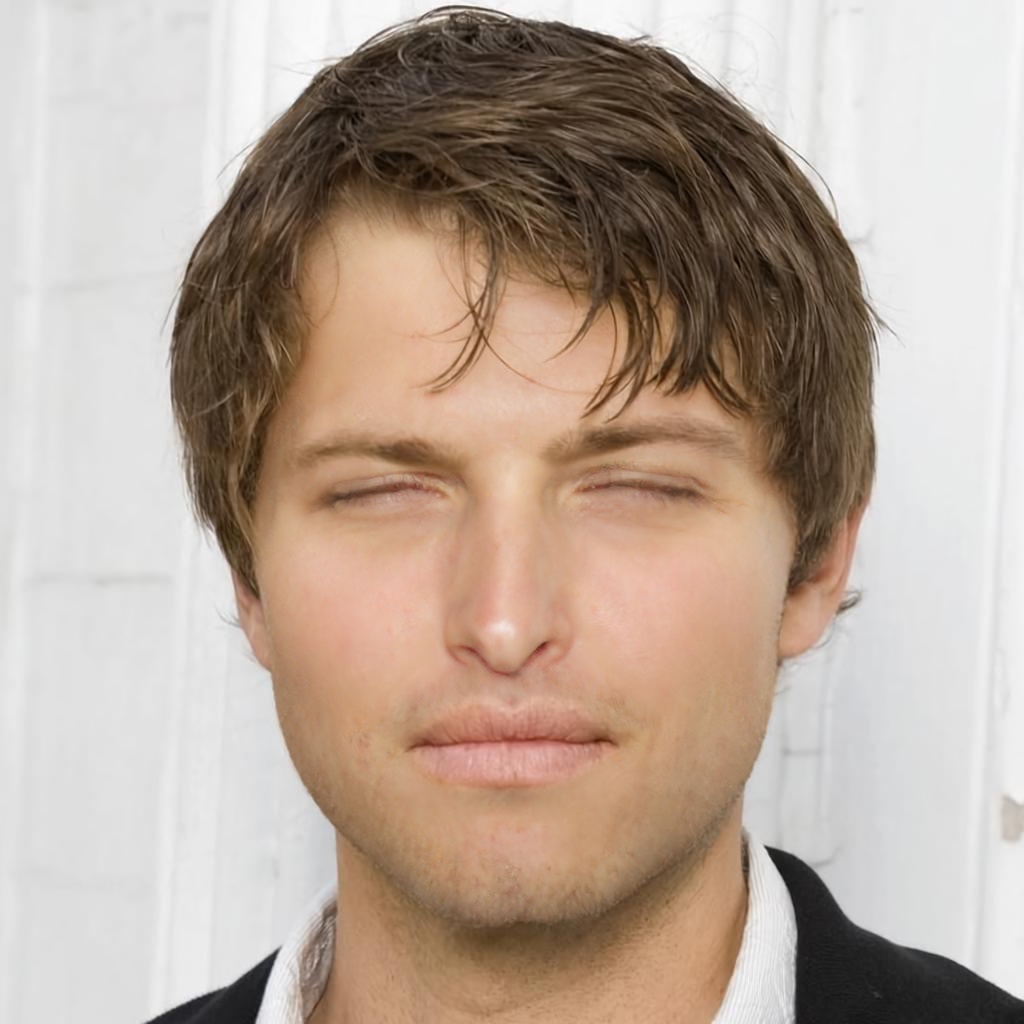}{Narrow eyes, Pointy nose}&
    		\mysubfigure{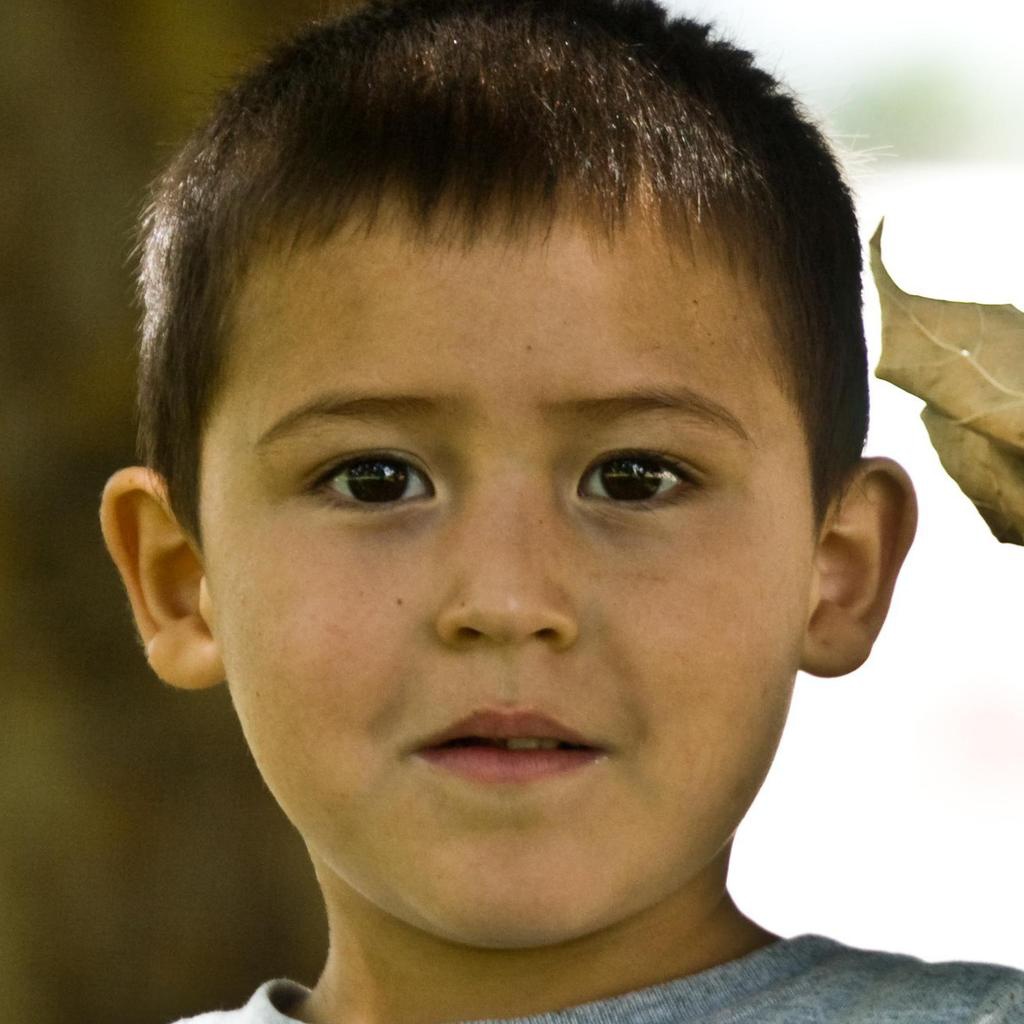}{Original} &
    		\mysubfigure{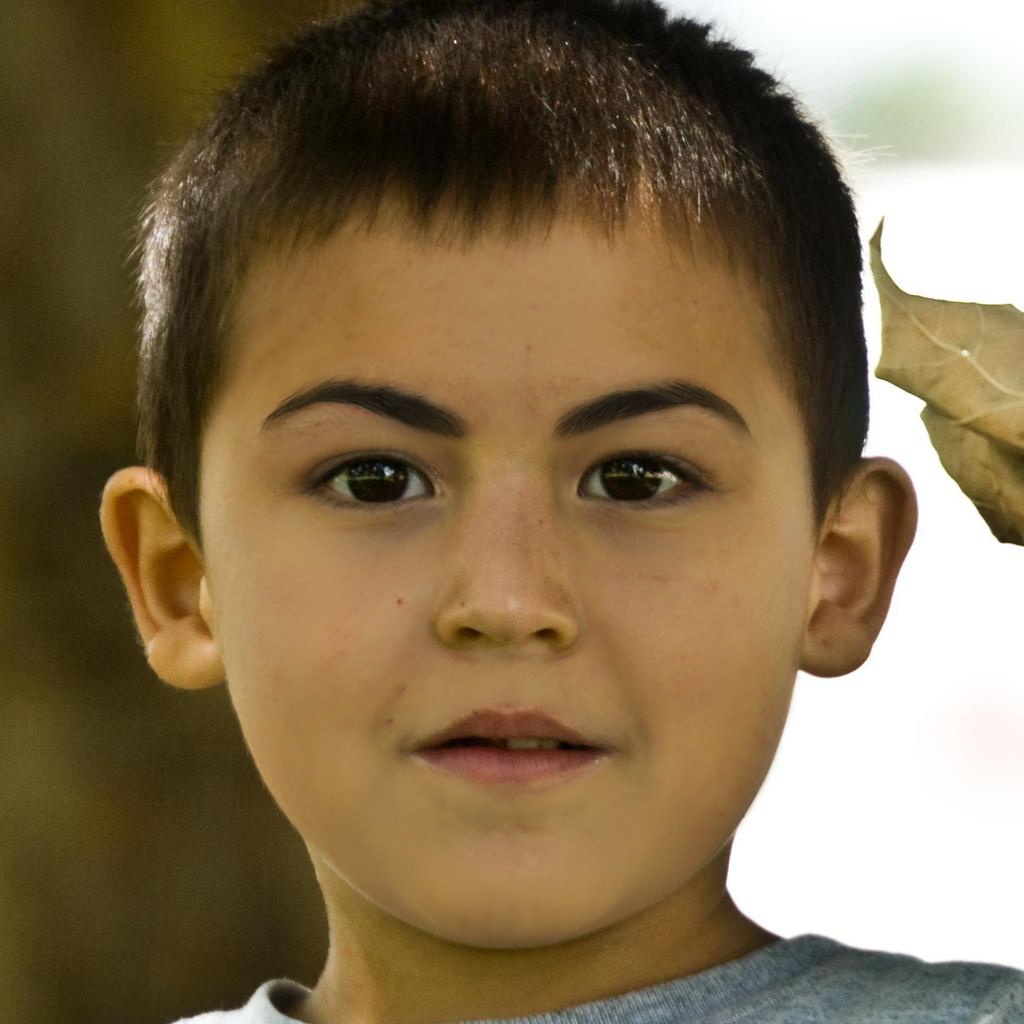}{Bushy eyeb., Arched eyeb.} \\
    		&\\[-3mm]
    		\mysubfigure{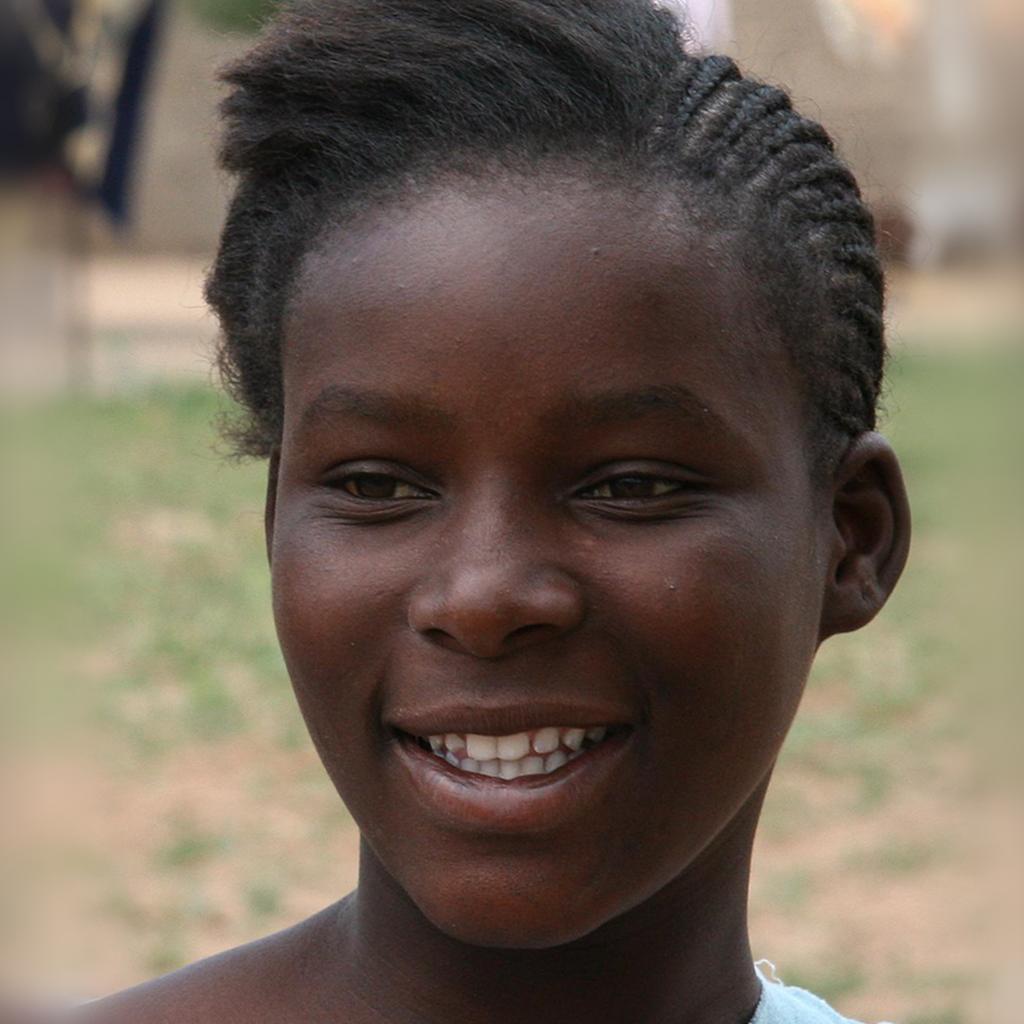}{Original} &
    		\mysubfigure{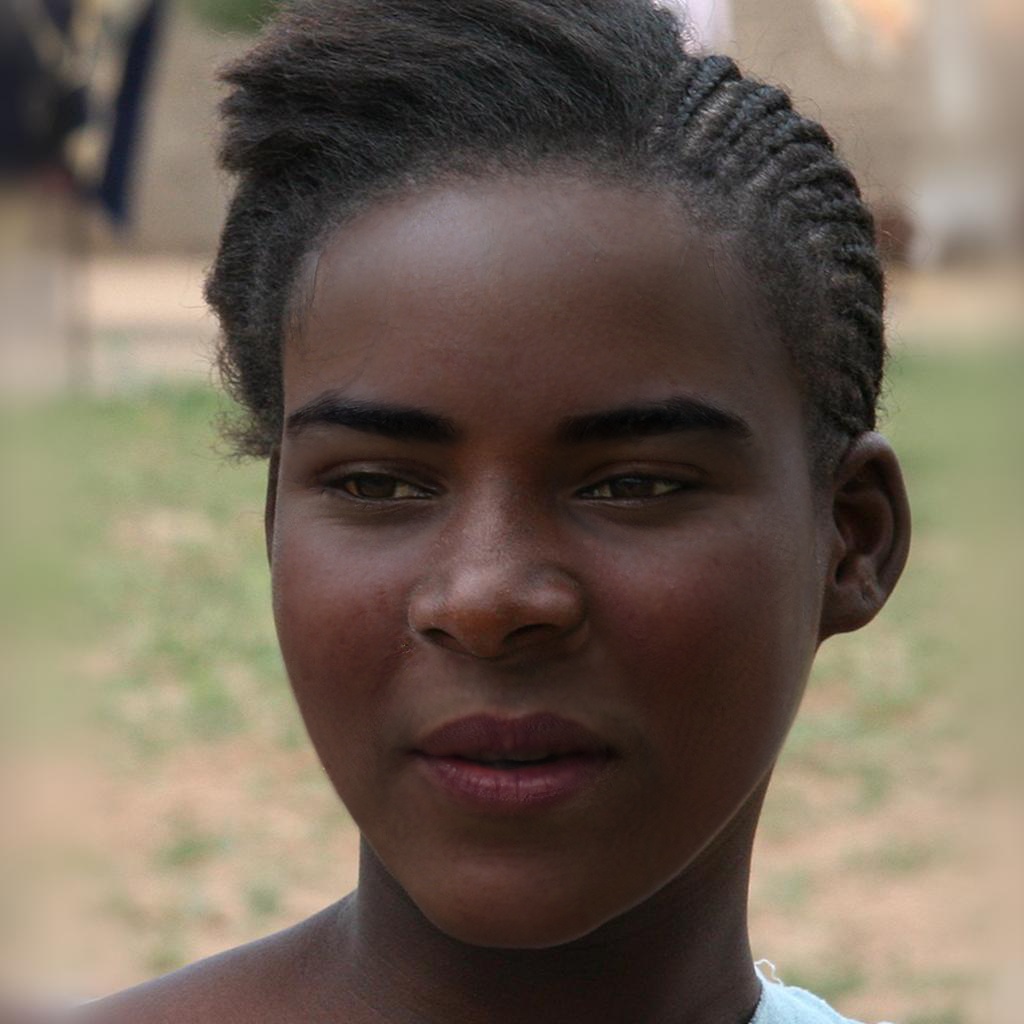}{Bushy eyeb., \textit{Mouth sl. open}} &
    		\mysubfigure{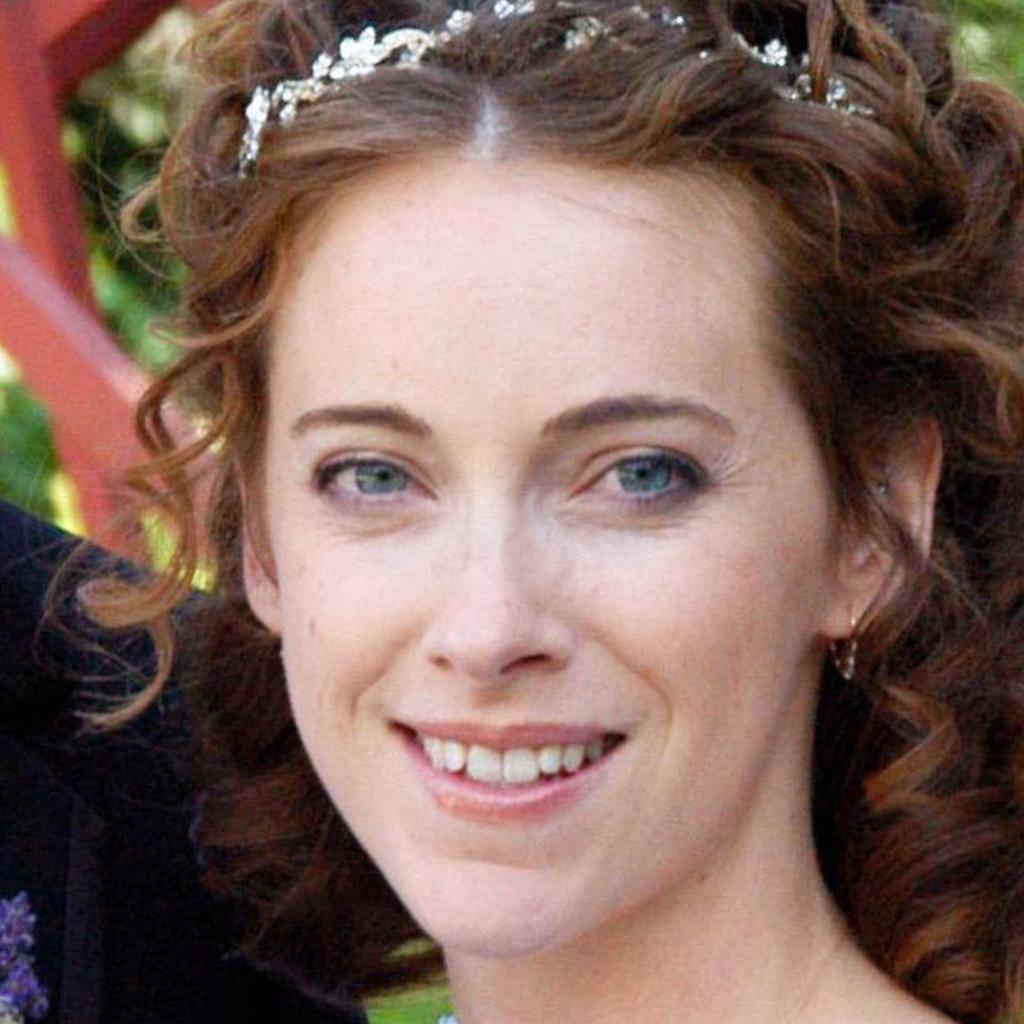}{Original} &
		\mysubfigure{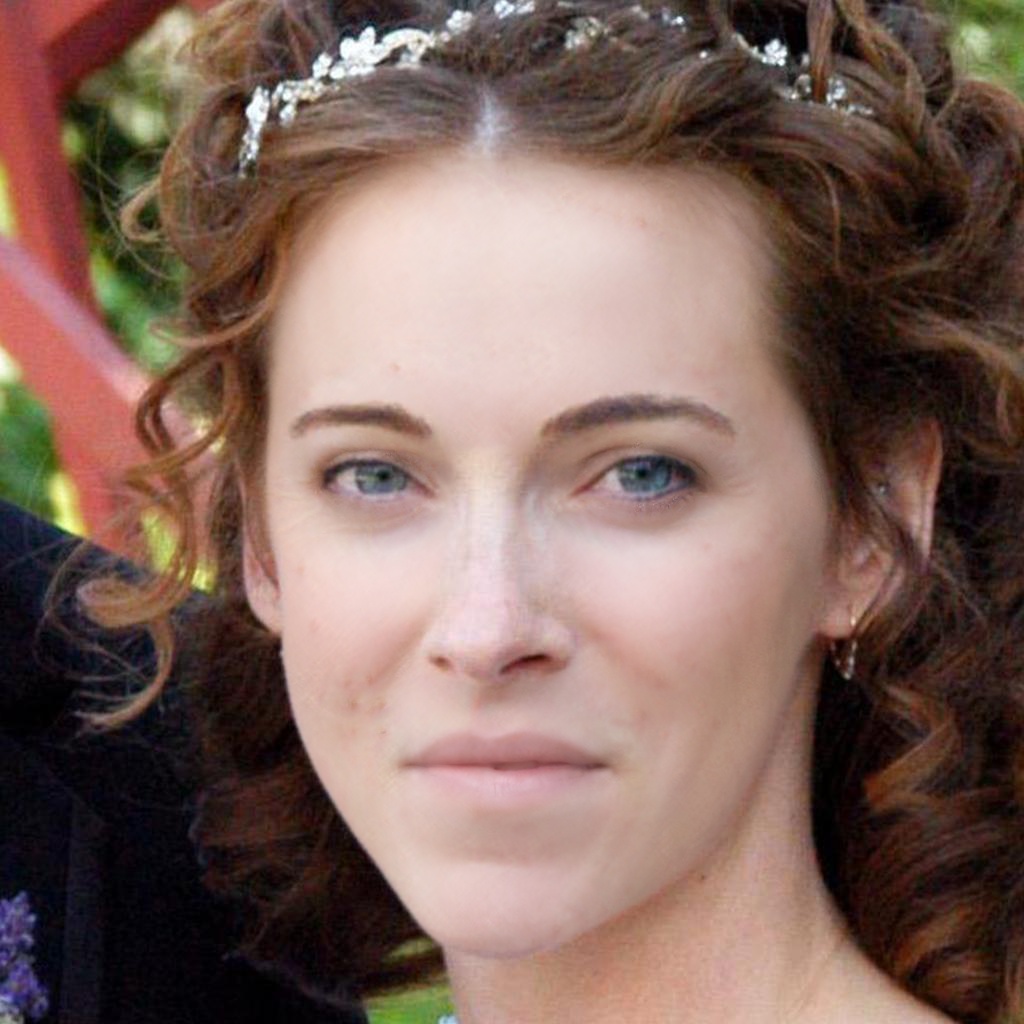}{\textit{Smiling}, \textit{Wearing lipst.}} \\
		&\\[-3mm]
		\mysubfigure{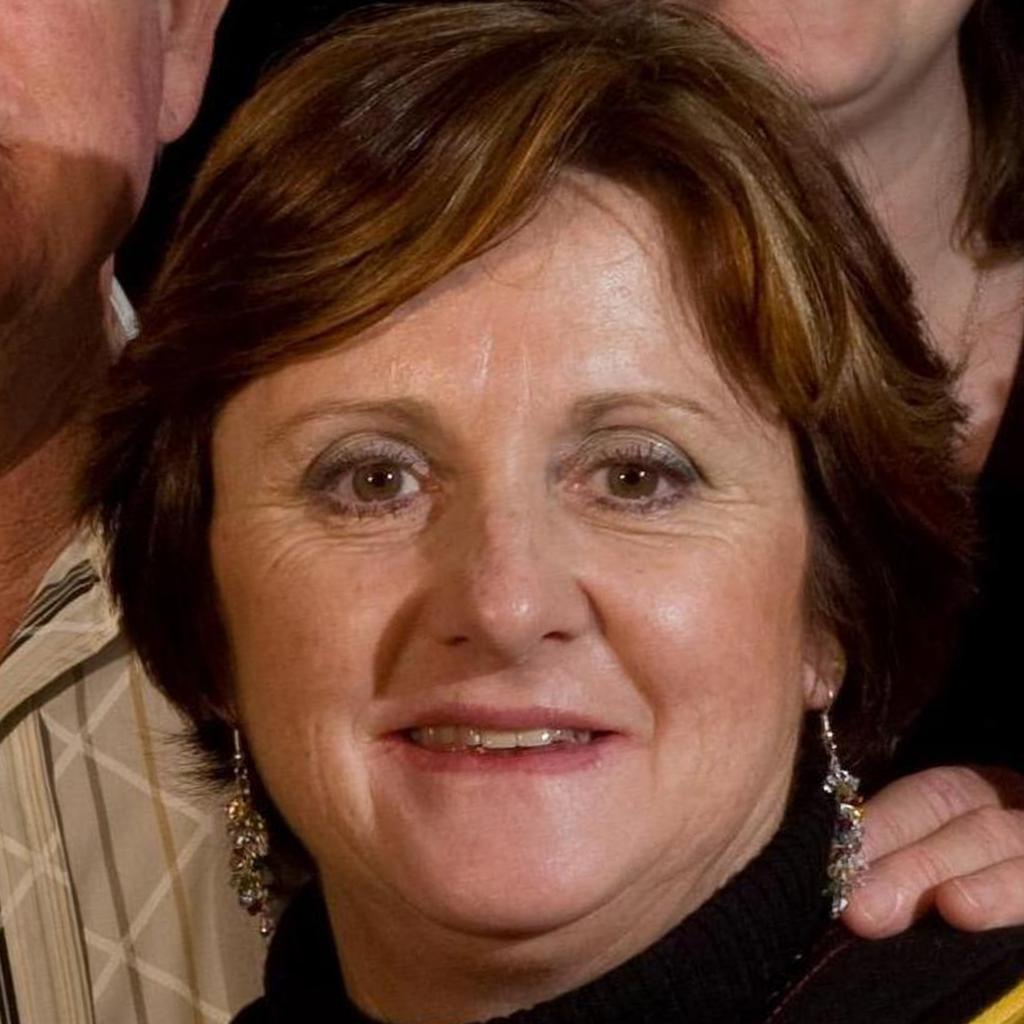}{Original} &
		\mysubfigure{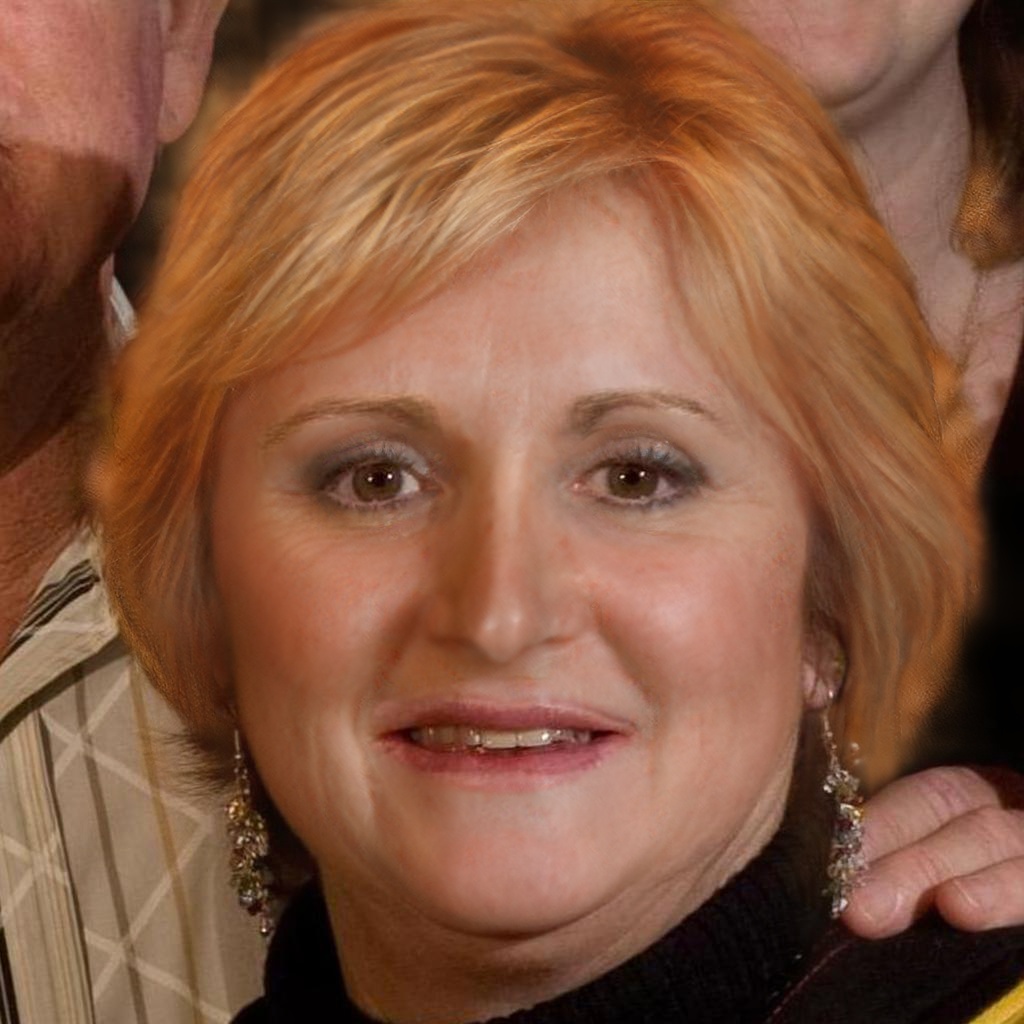}{Blond hair, Pointy nose} &
		\mysubfigure{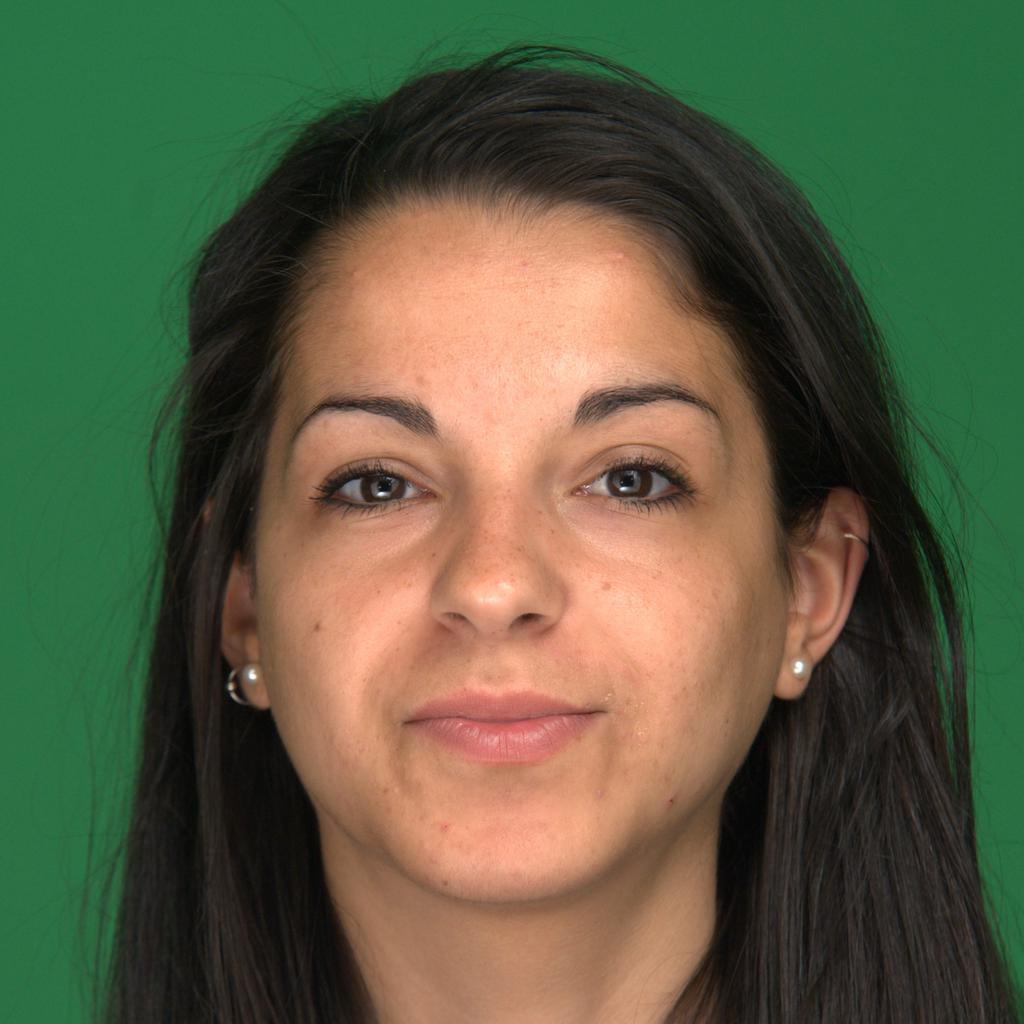}{Original} &
		\mysubfigure{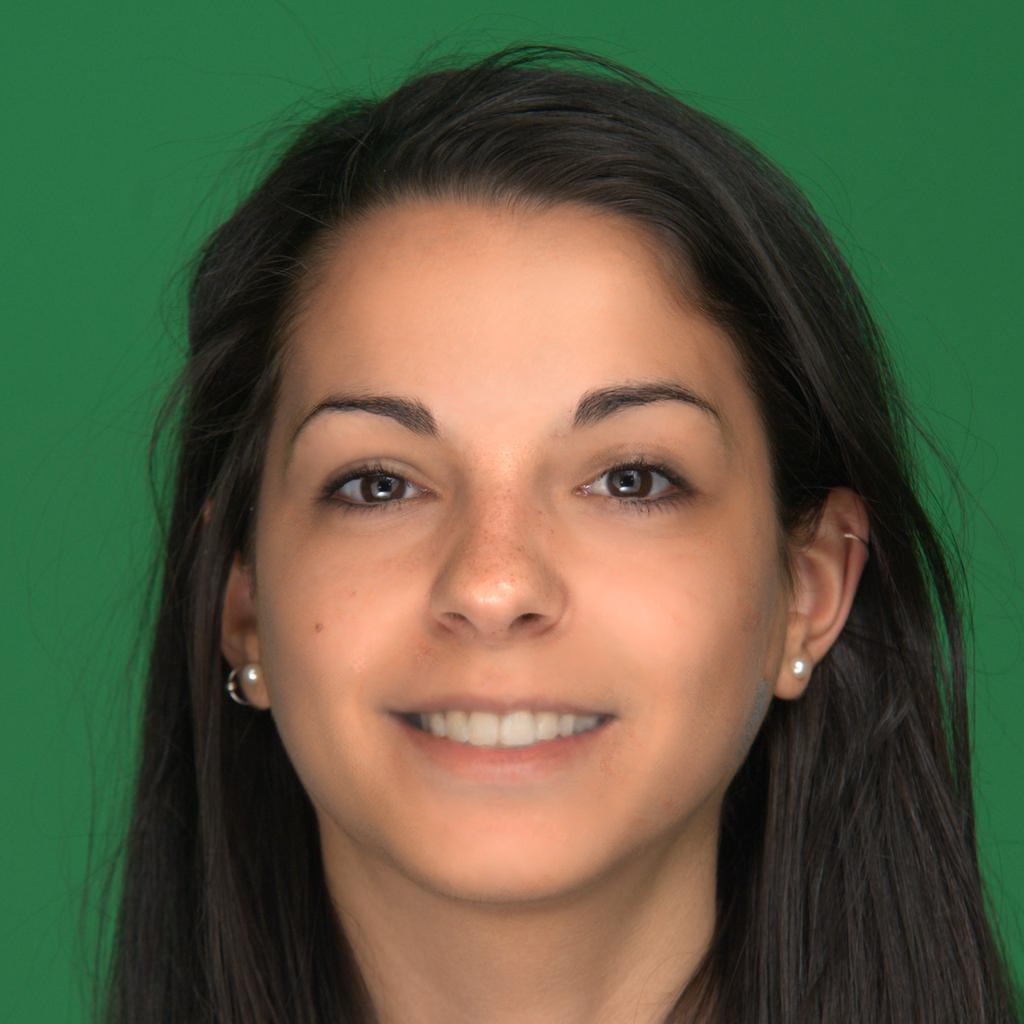}{Smiling, Mouth sl. open} \\
		\end{tabular}
\caption{Visual examples of image editing using MaskFaceGAN where multiple attributes are edited with a single optimization procedure. The presented examples show results when editing two attributes at the same time. Note that even for cases when the attributes correspond to the same spatial region, the generated results appear realistic and exhibit minimal distortions. Inverted (removed) attributes are displayed in italic.}
\label{fig:multiple_attributes}
\end{figure*}
\begin{table}[!th!]
\centering
\caption{Ablation study results on CelebA--HQ. Reported are FID scores computed for each targeted attribute separately ($\downarrow$ -- lower is better). Note that both the noise optimization procedure as well as the shape term contribute toward higher quality results. The shape term affects only editing procedures associated with the hair region.}
\label{tab:ablation}
  \begin{tabular}{l|ccc} 
  \hline\hline
  \multirow{2}{*}{\textbf{Target Attribute}} & \multicolumn{3}{c}{\textbf{FID scores} $\downarrow$}\\ \cline{2-4}
  & $n = 0, \lambda_S=0$ & $\lambda_S=0$ & MaskFaceGAN \\
  \hline
 Arched eyebrows & $49.69$ & $24.17$  & $24.17$ \\
 Big nose & $44.59$ & $19.84$ & $19.84$ \\
   Black hair & $78.77$ & $55.27$ & $43.39$ \\
   Brown hair & $72.14$ & $50.53$ & $37.12$ \\
   Bushy eyebrows & $52.67$ & $22.80$ & $22.80$ \\
   Grey hair & $81.35$ & $47.98$ & $37.47$ \\
   Mouth sl. open & $46.24$ & $23.27$ & $23.27$ \\
   Narrow eyes & $61.89$ & $28.69$ & $28.69$ \\
   Pointy nose & $47.47$ & $23.38$ & $23.38$ \\
   Smiling & $46.91$ & $23.87$ & $23.87$ \\
   Straight hair & $75.23$ & $54.58$ & $52.58$ \\
    Wavy hair & $73.00$ & $52.00$ & $59.83$ \\
    Wearing lipstick & $48.79$ & $26.93$ & $26.93$ \\
    \hline
    \hline
\end{tabular}
\end{table}
\setlength{\tabcolsep}{6pt} 
\renewcommand{\arraystretch}{1.3}  
\begin{table*}[!b!]
\renewcommand{\arraystretch}{1.3}
\caption{FID scores per attribute computed for the CelebA-HQ dataset -- lower is better. The table shows a  comparison with competing state--of--the--art models. The best score for each attribute is highlighted in bold.}
\label{tab:FID-CelebA}
\centering
    \begin{tabular}{l|rrrrrr} 
    \hline \hline 
     \textbf{Attribute} & \textbf{StarGAN} & \textbf{AttGAN} & \textbf{STGAN} & \textbf{InterFaceGAN} & \textbf{InterFaceGAN--D} & \textbf{MaskFaceGAN}  \\
     \hline 
Arched eyebrows & $133.49$ & $52.22$ & $40.34$ & $73.01$ & $77.77$ & $\boldsymbol{24.17}$ \\ Big nose & $137.24$ & $57.78$ & $46.94$ & $82.30$ & $76.20$ & $\boldsymbol{19.84}$ \\ Black hair & $143.31$ & $77.26$ & $57.05$ & $80.51$ & $77.57$ & $\boldsymbol{43.39}$ \\ Brown hair & $140.61$ & $58.88$ & $46.88$ & $72.97$ & $75.21$ & $\boldsymbol{37.12}$ \\ Bushy eyebrows & $125.90$ & $50.67$ & $38.22$ & $78.34$ & $74.31$ & $\boldsymbol{22.80}$ \\ Grey hair & $172.74$ & $139.82$ & $98.05$ & $98.45$ & $78.50$ & $\boldsymbol{37.47}$ \\ Mouth slightly open & $137.74$ & $61.47$ & $38.75$ & $80.88$ & $76.24$ & $\boldsymbol{23.27}$ \\ Narrow eyes & $137.76$ & $61.63$ & $39.42$ & $82.20$ & $77.63$ & $\boldsymbol{28.69}$ \\ Pointy nose & $131.32$ & $50.51$ & $39.15$ & $73.41$ & $73.87$ & $\boldsymbol{23.38}$ \\ Smiling & $141.69$ & $61.63$ & $42.83$ & $78.88$ & $75.41$ & $\boldsymbol{23.87}$ \\ Straight hair & $137.27$ & $60.57$ & $\boldsymbol{43.62}$ & $73.65$ & $76.94$ & $52.58$ \\ Wavy hair & $143.40$ & $67.06$ & $\boldsymbol{48.65}$ & $73.62$ & $75.24$ & $59.83$ \\ Wearing lipstick & $148.91$ & $78.70$ & $63.99$ & $89.00$ & $84.45$ & $\boldsymbol{26.93}$ \\ \hline \hline 
\end{tabular}
\end{table*}

\begin{table*}[!th!]
\renewcommand{\arraystretch}{1.3}
\caption{FID scores per attribute computed for the Helen dataset -- lower is better. The table shows a  comparison with competing state--of--the--art models. The best score for each attribute is highlighted in bold.}
\label{tab:FID-Helen}
\centering
    \begin{tabular}{l|rrrrrr} 
    \hline 
    \hline 
     \textbf{Attribute} & \textbf{StarGAN} & \textbf{AttGAN} & \textbf{STGAN} & \textbf{InterFaceGAN} & \textbf{InterFaceGAN--D} & \textbf{MaskFaceGAN}  \\
     \hline 
Arched eyebrows & $148.26$ & $68.22$ & $47.38$ & $87.90$ & $91.24$ & $\boldsymbol{21.65}$ \\ Big nose & $145.10$ & $70.57$ & $56.79$ & $94.05$ & $90.71$ & $\boldsymbol{19.19}$ \\ Black hair & $151.61$ & $85.47$ & $60.52$ & $92.19$ & $92.26$ & $\boldsymbol{42.74}$ \\ Brown hair & $149.86$ & $70.38$ & $52.14$ & $87.42$ & $89.12$ & $\boldsymbol{37.88}$ \\ Bushy eyebrows & $145.54$ & $66.99$ & $44.20$ & $89.00$ & $91.61$ & $\boldsymbol{21.29}$ \\ Grey hair & $180.83$ & $140.68$ & $102.65$ & $102.65$ & $91.60$ & $\boldsymbol{43.04}$ \\ Mouth slightly open & $150.45$ & $72.40$ & $46.06$ & $88.16$ & $89.00$ & $\boldsymbol{21.87}$ \\ Narrow eyes & $154.31$ & $81.84$ & $50.00$ & $90.56$ & $93.07$ & $\boldsymbol{23.67}$ \\ Pointy nose & $147.12$ & $66.93$ & $49.04$ & $84.23$ & $89.09$ & $\boldsymbol{20.98}$ \\ Smiling & $150.66$ & $74.05$ & $48.94$ & $90.17$ & $88.20$ & $\boldsymbol{22.47}$ \\ Straight hair & $148.26$ & $78.32$ & $\boldsymbol{55.92}$ & $88.24$ & $92.06$ & $66.85$ \\ Wavy hair & $161.97$ & $81.34$ & $\boldsymbol{65.33}$ & $89.95$ & $90.39$ & $76.04$ \\ Wearing lipstick & $150.33$ & $81.43$ & $67.74$ & $97.29$ & $93.76$ & $\boldsymbol{24.33}$ \\ 
\hline 
\hline 
\end{tabular}
\end{table*}
\begin{table*}[!th!]
\renewcommand{\arraystretch}{1.3}
\caption{FID scores per attribute computed for the SiblingsDB-HQf dataset -- lower is better. The table shows a  comparison with competing state--of--the--art models. The best score for each attribute is highlighted in bold. }
\label{tab:FID-Siblings}
\centering
    \begin{tabular}{l|rrrrrr} 
    \hline 
    \hline 
     \textbf{Attribute} & \textbf{StarGAN} & \textbf{AttGAN} & \textbf{STGAN} & \textbf{InterFaceGAN} & \textbf{InterFaceGAN--D} & \textbf{MaskFaceGAN}  \\
     \hline 
Arched eyebrows & $181.24$ & $66.69$ & $38.29$ & $75.42$ & $81.55$ & $\boldsymbol{29.95}$ \\ Big nose & $166.84$ & $68.64$ & $43.69$ & $85.97$ & $80.07$ & $\boldsymbol{16.93}$ \\ Black hair & $176.82$ & $87.83$ & $49.60$ & $85.91$ & $83.24$ & $\boldsymbol{37.45}$ \\ Brown hair & $175.56$ & $64.17$ & $37.08$ & $78.72$ & $81.90$ & $\boldsymbol{31.92}$ \\ Bushy eyebrows & $171.77$ & $62.65$ & $32.34$ & $81.63$ & $82.24$ & $\boldsymbol{28.43}$ \\ Grey hair & $199.54$ & $141.63$ & $106.99$ & $90.54$ & $83.61$ & $\boldsymbol{36.37}$ \\ Mouth slightly open & $173.15$ & $76.05$ & $36.66$ & $82.04$ & $86.61$ & $\boldsymbol{23.80}$ \\ Narrow eyes & $182.95$ & $94.33$ & $39.37$ & $84.46$ & $87.40$ & $\boldsymbol{35.49}$ \\ Pointy nose & $169.58$ & $67.07$ & $36.67$ & $77.21$ & $81.77$ & $\boldsymbol{24.55}$ \\ Smiling & $175.40$ & $82.80$ & $43.63$ & $75.63$ & $81.38$ & $\boldsymbol{22.81}$ \\ Straight hair & $172.94$ & $76.65$ & $\boldsymbol{49.29}$ & $74.36$ & $81.75$ & $59.24$ \\ Wavy hair & $184.82$ & $84.39$ & $\boldsymbol{62.69}$ & $77.47$ & $84.81$ & $70.06$ \\ Wearing lipstick & $168.28$ & $84.55$ & $59.19$ & $91.06$ & $87.84$ & $\boldsymbol{31.84} $\\
\hline \hline 
\end{tabular}
\end{table*}

\setlength{\tabcolsep}{6pt} 
\renewcommand{\arraystretch}{1.3}  

\section{Results by Attributes}
In this section we present additional quantitative results grouped by individual attributes. Specifically, we report per attribute results for the ablation study in terms of FID scores, analyze FID scores over the three experimental datasets and elaborate on per attribute user--study ratings.

\subsection{Ablation Study}

To ensure better insight into the contribution of individual components on the performance MaskFaceGAN, we again remove the noise optimization procedure and shape terms from our model and investigate the impact of these components on the FID scores generated per attribute on CelebA-HQ.

The  FID scores calculated for the ablation study are presented in Table \ref{tab:ablation}. We  observe that the noise optimization procedure improves the FID scores of every attribute. This can be attributed to the fact that noise optimization further optimizes the quality of the embedded images and generates high frequency image details, such as hair. The inclusion of the shape term $\lambda_S$ only affects attributes associated with the hair region. Most of these attributes achieve a significantly lower FID score when the shape term is included in the loss function. This can be attributed to the more natural looking results of the blending procedure, since the hair components from the generated and the original image are spatially aligned.

\subsection{FID results}
The FID scores are used as a measure of quality for the edited images. FID scores, grouped by attributes, are shown in Tables \ref{tab:FID-CelebA}, \ref{tab:FID-Helen} and \ref{tab:FID-Siblings} for CelebAHQ, Helen and Siblings dataset, respectively. 

Note that MaskFaceGAN outperforms all competing models with a significant margin at all attributes except when editing the ``Straight hair'' and ``Wavy hair'' attributes. For these attributes, STGAN achieves better results than MaskFaceGAN. The single--attribute editing results shown in Figs. \ref{fig:celebahq}, \ref{fig:helen} and \ref{fig:siblings} show that STGAN only slightly changes the hair shape, which preserves the quality of the initial image -- but at the cost of a convincing semantic presence of the targeted attribute. MaskFaceGAN, on the other hand, generates stronger semantic content, but also hallucinates some of the image details. 

\subsection{User Study Results}
Similarly as in the main part of the paper, we report user study scores for two experiments in this section: $(i)$ in the first experiment, users were asked to rate editing results on a $5$--point Likert scale, and $(ii)$ in the second experiment, users were asked to select the best result among the edited images generated by the evaluated models. 

The results for model ratings are shown in Tables \ref{tab:user-study-by-attribute-CelebAHQ}, \ref{tab:user-study-by-attribute-Helen} and  \ref{tab:user-study-by-attribute-Siblings} for the CelebA--HQ, Helen and Siblings datasets, respectively and the results for the best model selection ratings are shown in Tables \ref{tab:user-study-best_models-attribute-CelebAHQ}, \ref{tab:user-study-best_models-attribute-Helen} and \ref{tab:user-study-best_models-attribute-Siblings} with the same dataset order. The proposed MaskFaceGAN model is consistently rated higher than other models on most of the considered attributes with the notable exception of the ``Narrow eyes'' attribute, where AttGAN achieves better performance across all datasets. The difference most likely stems from  MaskFaceGAN's tendency to close the eyes instead of narrowing them. A similar effect can also be observed in Figs. \ref{fig:celebahq}, \ref{fig:helen} and \ref{fig:siblings} for the InterFaceGAN and InterFaceGAN--D results.

\begin{table*}[!th!]
\caption{Results of the user study on test images from the CelebA--HQ dataset. Users were asked to rate the models on a $5$--point Likert scale, where $5$ stands for a perfect result. The model with the best user rating for each attribute is presented in bold.}
\label{tab:user-study-by-attribute-CelebAHQ}
\centering
\begin{tabular}{l|rrrrrr}
\hline\hline
\textbf{Attribute} & \textbf{StarGAN} & \textbf{AttGAN} & \textbf{STGAN} & \textbf{InterFaceGAN} & \textbf{InterFaceGAN--D} & \textbf{MaskFaceGAN} \\
\hline
Arched eyebrows  & $1.44 \pm 0.94$ & $2.75 \pm 0.96$ & $3.00 \pm 1.02$ & $3.05 \pm 1.03$ & $3.38 \pm 1.19$ & $\boldsymbol{4.03 \pm 1.24}$\\
Big nose  & $1.37 \pm 0.84$ & $3.09 \pm 1.05$ & $2.87 \pm 0.96$ & $2.72 \pm 0.99$ & $2.60 \pm 0.99$ & $\boldsymbol{4.35 \pm 1.14}$\\
Black hair  & $1.68 \pm 1.02$ & $2.74 \pm 1.23$ & $3.02 \pm 1.12$ & $3.16 \pm 0.81$ & $3.50 \pm 1.06$ & $\boldsymbol{4.00 \pm 1.23}$\\
Blond hair  & $1.74 \pm 1.15$ & $2.81 \pm 1.15$ & $2.97 \pm 1.06$ & $2.98 \pm 1.02$ & $3.24 \pm 1.06$ & $\boldsymbol{4.12 \pm 1.12}$\\
Brown hair  & $1.39 \pm 0.82$ & $2.73 \pm 1.10$ & $3.00 \pm 0.98$ & $3.13 \pm 0.95$ & $3.37 \pm 0.97$ & $\boldsymbol{4.41 \pm 0.86}$\\
Bushy eyebrows  & $1.63 \pm 1.21$ & $2.61 \pm 1.12$ & $3.18 \pm 0.97$ & $3.17 \pm 0.95$ & $3.53 \pm 1.03$ & $\boldsymbol{4.42 \pm 1.17}$\\
Grey hair  & $1.41 \pm 0.84$ & $3.15 \pm 1.15$ & $2.58 \pm 1.11$ & $2.10 \pm 1.15$ & $2.07 \pm 1.18$ & $\boldsymbol{3.91 \pm 1.22}$\\
Mouth slightly open  & $1.27 \pm 0.57$ & $2.97 \pm 1.21$ & $2.88 \pm 0.93$ & $2.44 \pm 1.10$ & $2.68 \pm 1.30$ & $\boldsymbol{4.11 \pm 1.28}$\\
Narrow eyes  & $1.38 \pm 0.86$ & $\boldsymbol{3.60 \pm 1.27}$ & $3.00 \pm 1.08$ & $2.52 \pm 0.98$ & $2.62 \pm 1.11$ & $3.52 \pm 1.46$\\
Pointy nose  & $1.63 \pm 1.09$ & $3.00 \pm 1.09$ & $3.00 \pm 0.96$ & $3.14 \pm 1.05$ & $3.41 \pm 1.15$ & $\boldsymbol{4.34 \pm 1.01}$\\
Smiling  & $1.36 \pm 0.89$ & $2.84 \pm 1.20$ & $3.17 \pm 1.08$ & $2.66 \pm 1.12$ & $2.80 \pm 1.31$ & $\boldsymbol{4.11 \pm 1.18}$\\
Straight hair  & $1.31 \pm 0.81$ & $3.05 \pm 1.04$ & $3.36 \pm 1.02$ & $3.12 \pm 0.94$ & $3.64 \pm 1.03$ & $\boldsymbol{3.80 \pm 1.28}$\\
Wavy hair  & $1.52 \pm 0.97$ & $2.71 \pm 0.93$ & $3.13 \pm 1.01$ & $3.05 \pm 1.18$ & $3.31 \pm 1.27$ & $\boldsymbol{3.74 \pm 1.32}$\\
Wearing lipstick  & $1.27 \pm 0.72$ & $3.17 \pm 1.13$ & $2.89 \pm 1.16$ & $2.67 \pm 0.85$ & $2.80 \pm 1.08$ & $\boldsymbol{4.04 \pm 1.35}$\\\hline\hline
\end{tabular}
\end{table*}
\begin{table*}[!th!]
\caption{Results of the user study on test images from the Helen dataset. Users were asked to rate the models on a $5$--point Likert scale, where $5$ stands for a perfect result. The model with the best user rating for each attribute is presented in bold. }
\label{tab:user-study-by-attribute-Helen}
\centering
\begin{tabular}{l|rrrrrr}
\hline\hline
\textbf{Attribute} & \textbf{StarGAN} & \textbf{AttGAN} & \textbf{STGAN} & \textbf{InterFaceGAN} & \textbf{InterFaceGAN--D} & \textbf{MaskFaceGAN} \\
\hline
Arched eyebrows  & $1.46 \pm 0.88$ & $2.79 \pm 1.18$ & $2.95 \pm 1.06$ & $2.53 \pm 0.94$ & $2.72 \pm 1.09$ & $\boldsymbol{4.10 \pm 1.15}$\\
Big nose  & $1.25 \pm 0.58$ & $2.92 \pm 1.15$ & $3.07 \pm 1.07$ & $2.16 \pm 0.88$ & $2.42 \pm 1.00$ & $\boldsymbol{4.39 \pm 0.89}$\\
Black hair  & $1.44 \pm 0.76$ & $2.54 \pm 1.19$ & $3.24 \pm 1.19$ & $2.79 \pm 0.91$ & $2.89 \pm 1.12$ & $\boldsymbol{3.64 \pm 1.28}$\\
Blond hair  & $1.30 \pm 0.55$ & $2.73 \pm 1.18$ & $2.92 \pm 1.14$ & $2.88 \pm 1.11$ & $2.71 \pm 1.05$ & $\boldsymbol{3.66 \pm 1.14}$\\
Brown hair  & $1.36 \pm 0.65$ & $2.70 \pm 1.22$ & $2.59 \pm 1.08$ & $3.00 \pm 1.07$ & $2.79 \pm 1.11$ & $\boldsymbol{4.05 \pm 1.18}$\\
Bushy eyebrows  & $1.37 \pm 0.72$ & $2.35 \pm 1.23$ & $3.13 \pm 1.21$ & $2.40 \pm 1.02$ & $2.67 \pm 1.14$ & $\boldsymbol{3.93 \pm 1.21}$\\
Grey hair  & $1.34 \pm 0.65$ & $2.74 \pm 1.21$ & $2.77 \pm 1.20$ & $1.94 \pm 1.02$ & $2.07 \pm 1.06$ & $\boldsymbol{3.65 \pm 1.34}$\\
Mouth slightly open  & $1.24 \pm 0.58$ & $2.78 \pm 1.32$ & $3.14 \pm 1.22$ & $1.98 \pm 0.88$ & $2.12 \pm 1.07$ & $\boldsymbol{3.73 \pm 1.51}$\\
Narrow eyes  & $1.32 \pm 0.61$ & $\boldsymbol{3.18 \pm 1.43}$ & $2.98 \pm 1.34$ & $1.84 \pm 0.92$ & $1.65 \pm 0.77$ & $2.99 \pm 1.38$\\
Pointy nose  & $1.16 \pm 0.52$ & $3.02 \pm 1.11$ & $3.16 \pm 1.04$ & $2.56 \pm 0.98$ & $2.70 \pm 0.95$ & $\boldsymbol{4.30 \pm 0.96}$\\
Smiling  & $1.15 \pm 0.41$ & $2.82 \pm 1.49$ & $\boldsymbol{3.55 \pm 1.33}$ & $2.15 \pm 1.03$ & $2.14 \pm 1.11$ & $3.43 \pm 1.33$\\
Straight hair  & $1.42 \pm 0.71$ & $3.14 \pm 1.01$ & $\boldsymbol{3.36 \pm 1.22}$ & $2.39 \pm 1.01$ & $2.62 \pm 1.19$ & $3.23 \pm 1.32$\\
Wavy hair  & $1.27 \pm 0.63$ & $2.64 \pm 1.06$ & $2.92 \pm 1.13$ & $2.52 \pm 1.17$ & $2.28 \pm 1.16$ & $\boldsymbol{4.05 \pm 1.14}$\\
Wearing lipstick  & $1.19 \pm 0.48$ & $2.62 \pm 1.21$ & $2.65 \pm 1.21$ & $2.34 \pm 0.98$ & $2.37 \pm 1.08$ & $\boldsymbol{4.08 \pm 1.20}$\\\hline\hline
\end{tabular}
\end{table*}
\begin{table*}[!th!]
\caption{Results of the user study on test images from the SiblingsDB--HQf dataset. Users were asked to rate the models on a $5$--point Likert scale, where $5$ stands for a perfect result. The model with the best user rating for each attribute is presented in bold. }
\label{tab:user-study-by-attribute-Siblings}
\centering
\begin{tabular}{l|rrrrrr}
\hline\hline
\textbf{Attribute} & \textbf{StarGAN} & \textbf{AttGAN} & \textbf{STGAN} & \textbf{InterFaceGAN} & \textbf{InterFaceGAN--D} & \textbf{MaskFaceGAN} \\
\hline
Arched eyebrows  & $1.48 \pm 0.78$ & $3.00 \pm 1.15$ & $3.27 \pm 1.01$ & $2.43 \pm 0.90$ & $2.64 \pm 0.96$ & $\boldsymbol{4.14 \pm 1.12}$\\
Big nose  & $1.51 \pm 0.87$ & $3.15 \pm 1.13$ & $3.00 \pm 0.86$ & $2.29 \pm 0.81$ & $2.50 \pm 0.88$ & $\boldsymbol{4.27 \pm 1.03}$\\
Black hair  & $1.55 \pm 0.88$ & $2.99 \pm 1.10$ & $\boldsymbol{3.64 \pm 1.08}$ & $2.75 \pm 0.87$ & $3.15 \pm 1.03$ & $3.60 \pm 1.12$\\
Blond hair  & $1.62 \pm 0.88$ & $3.10 \pm 1.15$ & $3.20 \pm 0.97$ & $2.71 \pm 0.86$ & $2.97 \pm 0.87$ & $\boldsymbol{3.82 \pm 1.15}$\\
Brown hair  & $1.67 \pm 1.02$ & $3.13 \pm 1.12$ & $3.14 \pm 0.94$ & $2.94 \pm 0.94$ & $3.04 \pm 1.04$ & $\boldsymbol{3.62 \pm 1.10}$\\
Bushy eyebrows  & $1.42 \pm 0.73$ & $2.93 \pm 1.08$ & $3.27 \pm 1.02$ & $2.57 \pm 0.98$ & $2.97 \pm 1.06$ & $\boldsymbol{3.92 \pm 1.23}$\\
Grey hair  & $1.72 \pm 1.10$ & $3.05 \pm 1.11$ & $2.99 \pm 0.98$ & $2.30 \pm 0.96$ & $2.40 \pm 1.03$ & $\boldsymbol{3.55 \pm 1.30}$\\
Mouth slightly open  & $1.47 \pm 0.88$ & $3.11 \pm 1.15$ & $3.38 \pm 0.95$ & $2.08 \pm 0.86$ & $2.14 \pm 0.93$ & $\boldsymbol{4.15 \pm 1.18}$\\
Narrow eyes  & $1.50 \pm 0.89$ & $\boldsymbol{3.39 \pm 1.23}$ & $3.23 \pm 1.19$ & $2.20 \pm 0.97$ & $2.20 \pm 0.95$ & $3.20 \pm 1.29$\\
Pointy nose  & $1.43 \pm 0.79$ & $3.20 \pm 1.11$ & $3.32 \pm 0.98$ & $2.61 \pm 0.84$ & $2.91 \pm 1.04$ & $\boldsymbol{4.20 \pm 1.03}$\\
Smiling  & $1.38 \pm 0.85$ & $3.04 \pm 1.04$ & $3.84 \pm 0.97$ & $2.36 \pm 0.81$ & $2.46 \pm 0.84$ & $\boldsymbol{4.11 \pm 0.99}$\\
Straight hair  & $1.51 \pm 0.85$ & $3.27 \pm 0.98$ & $3.34 \pm 1.07$ & $2.67 \pm 0.90$ & $2.77 \pm 0.96$ & $\boldsymbol{3.47 \pm 1.22}$\\
Wavy hair  & $1.53 \pm 0.93$ & $3.19 \pm 1.10$ & $3.35 \pm 1.13$ & $2.50 \pm 0.93$ & $2.56 \pm 1.08$ & $\boldsymbol{3.70 \pm 1.21}$\\
Wearing lipstick  & $1.53 \pm 1.00$ & $3.15 \pm 0.96$ & $3.17 \pm 1.02$ & $2.33 \pm 0.89$ & $2.54 \pm 1.07$ & $\boldsymbol{4.16 \pm 1.12}$\\\hline\hline
\end{tabular}
\end{table*}

\begin{table*}[!th!]
\caption{Results of the user study on test images from the CelebA--HQ dataset in terms of percentage of selected images. Users were asked to select the best editing results when presented with image examples generated by all evaluated models. The model with the best user rating for each attribute is presented in bold.}
\label{tab:user-study-best_models-attribute-CelebAHQ}
\centering
\begin{tabular}{l|rrrrrr}
\hline\hline
\textbf{Attribute} & \textbf{StarGAN} & \textbf{AttGAN} & \textbf{STGAN} & \textbf{InterFaceGAN} & \textbf{InterFaceGAN--D} & \textbf{MaskFaceGAN} \\
\hline
Arched eyebrows & $1.43 \%$ & $4.29 \%$ & $7.14 \%$ & $7.14 \%$ & $24.29 \%$ & $\boldsymbol{55.71 \%}$\\
Big nose & $8.82 \%$ & $10.29 \%$ & $1.47 \%$ & $2.94 \%$ & $1.47 \%$ & $\boldsymbol{75.00 \%}$\\
Black hair & $6.15 \%$ & $3.08 \%$ & $7.69 \%$ & $3.08 \%$ & $26.15 \%$ & $\boldsymbol{53.85 \%}$\\
Blond hair & $5.71 \%$ & $4.29 \%$ & $8.57 \%$ & $7.14 \%$ & $18.57 \%$ & $\boldsymbol{55.71 \%}$\\
Brown hair & $2.94 \%$ & $1.47 \%$ & $2.94 \%$ & $4.41 \%$ & $16.18 \%$ & $\boldsymbol{72.06 \%}$\\
Bushy eyebrows & $3.12 \%$ & $3.12 \%$ & $4.69 \%$ & $1.56 \%$ & $9.38 \%$ & $\boldsymbol{78.12 \%}$\\
Grey hair & $3.08 \%$ & $12.31 \%$ & $10.77 \%$ & $6.15 \%$ & $10.77 \%$ & $\boldsymbol{56.92 \%}$\\
Mouth slightly open & $0.00 \%$ & $11.43 \%$ & $4.29 \%$ & $2.86 \%$ & $12.86 \%$ & $\boldsymbol{68.57 \%}$\\
Narrow eyes & $0.00 \%$ & $\boldsymbol{42.65 \%}$ & $10.29 \%$ & $1.47 \%$ & $7.35 \%$ & $38.24 \%$\\
Pointy nose & $1.45 \%$ & $11.59 \%$ & $2.90 \%$ & $1.45 \%$ & $13.04 \%$ & $\boldsymbol{69.57 \%}$\\
Smiling & $5.80 \%$ & $8.70 \%$ & $13.04 \%$ & $2.90 \%$ & $5.80 \%$ & $\boldsymbol{63.77 \%}$\\
Straight hair & $1.47 \%$ & $8.82 \%$ & $11.76 \%$ & $14.71 \%$ & $20.59 \%$ & $\boldsymbol{42.65 \%}$\\
Wavy hair & $1.52 \%$ & $6.06 \%$ & $6.06 \%$ & $9.09 \%$ & $19.70 \%$ & $\boldsymbol{57.58 \%}$\\
Wearing lipstick & $0.00 \%$ & $11.76 \%$ & $13.24 \%$ & $4.41 \%$ & $8.82 \%$ & $\boldsymbol{61.76 \%}$\\\hline\hline
\end{tabular}
\end{table*}
\begin{table*}[!th!]
\caption{Results of the user study on test images from the Helen dataset in terms of percentage of selected images. Users were asked to select the best editing results when presented with image examples generated by all evaluated models. The model with the best user rating for each attribute is presented in bold.}
\label{tab:user-study-best_models-attribute-Helen}
\centering
\begin{tabular}{l|rrrrrr}
\hline\hline
\textbf{Attribute} & \textbf{StarGAN} & \textbf{AttGAN} & \textbf{STGAN} & \textbf{InterFaceGAN} & \textbf{InterFaceGAN--D} & \textbf{MaskFaceGAN} \\
\hline
Arched eyebrows & $3.74 \%$ & $14.95 \%$ & $9.35 \%$ & $5.61 \%$ & $9.35 \%$ & $\boldsymbol{57.01 \%}$\\
Big nose & $7.62 \%$ & $13.33 \%$ & $11.43 \%$ & $5.71 \%$ & $5.71 \%$ & $\boldsymbol{56.19 \%}$\\
Black hair & $3.74 \%$ & $11.21 \%$ & $19.63 \%$ & $9.35 \%$ & $15.89 \%$ & $\boldsymbol{40.19 \%}$\\
Blond hair & $0.93 \%$ & $10.28 \%$ & $16.82 \%$ & $16.82 \%$ & $14.02 \%$ & $\boldsymbol{41.12 \%}$\\
Brown hair & $7.55 \%$ & $9.43 \%$ & $7.55 \%$ & $8.49 \%$ & $6.60 \%$ & $\boldsymbol{60.38 \%}$\\
Bushy eyebrows & $0.00 \%$ & $11.43 \%$ & $20.00 \%$ & $6.67 \%$ & $9.52 \%$ & $\boldsymbol{52.38 \%}$\\
Grey hair & $9.43 \%$ & $6.60 \%$ & $19.81 \%$ & $5.66 \%$ & $6.60 \%$ & $\boldsymbol{51.89 \%}$\\
Mouth slightly open & $0.96 \%$ & $17.31 \%$ & $22.12 \%$ & $2.88 \%$ & $6.73 \%$ & $\boldsymbol{50.00 \%}$\\
Narrow eyes & $0.00 \%$ & $\boldsymbol{47.17 \%}$ & $22.64 \%$ & $4.72 \%$ & $1.89 \%$ & $23.58 \%$\\
Pointy nose & $1.89 \%$ & $7.55 \%$ & $9.43 \%$ & $7.55 \%$ & $8.49 \%$ & $\boldsymbol{65.09 \%}$\\
Smiling & $0.00 \%$ & $25.47 \%$ & $\boldsymbol{45.28 \%}$ & $0.94 \%$ & $6.60 \%$ & $21.70 \%$\\
Straight hair & $0.93 \%$ & $16.82 \%$ & $\boldsymbol{32.71 \%}$ & $7.48 \%$ & $12.15 \%$ & $29.91 \%$\\
Wavy hair & $0.00 \%$ & $3.85 \%$ & $11.54 \%$ & $6.73 \%$ & $9.62 \%$ & $\boldsymbol{68.27 \%}$\\
Wearing lipstick & $1.90 \%$ & $10.48 \%$ & $12.38 \%$ & $2.86 \%$ & $8.57 \%$ & $\boldsymbol{63.81 \%}$\\\hline\hline
\end{tabular}
\end{table*}
\begin{table*}[!th!]
\caption{Results of the user study on test images from the SiblingsDB--HQf dataset in terms of percentage of selected images. Users were asked to select the best editing results when presented with image examples generated by all evaluated models. The model with the best user rating for each attribute is presented in bold. }
\label{tab:user-study-best_models-attribute-Siblings}
\centering
\begin{tabular}{l|rrrrrr}
\hline\hline
\textbf{Attribute} & \textbf{StarGAN} & \textbf{AttGAN} & \textbf{STGAN} & \textbf{InterFaceGAN} & \textbf{InterFaceGAN--D} & \textbf{MaskFaceGAN} \\
\hline
Arched eyebrows & $2.11 \%$ & $10.56 \%$ & $14.79 \%$ & $4.23 \%$ & $8.45 \%$ & $\boldsymbol{59.86 \%}$\\
Big nose & $7.04 \%$ & $4.23 \%$ & $6.34 \%$ & $6.34 \%$ & $5.63 \%$ & $\boldsymbol{70.42 \%}$\\
Black hair & $2.80 \%$ & $10.49 \%$ & $\boldsymbol{32.17 \%}$ & $4.90 \%$ & $31.47 \%$ & $18.18 \%$\\
Blond hair & $4.26 \%$ & $5.67 \%$ & $12.77 \%$ & $11.35 \%$ & $15.60 \%$ & $\boldsymbol{50.35 \%}$\\
Brown hair & $11.43 \%$ & $14.29 \%$ & $5.71 \%$ & $8.57 \%$ & $17.14 \%$ & $\boldsymbol{42.86 \%}$\\
Bushy eyebrows & $2.13 \%$ & $10.64 \%$ & $15.60 \%$ & $7.09 \%$ & $16.31 \%$ & $\boldsymbol{48.23 \%}$\\
Grey hair & $5.59 \%$ & $11.89 \%$ & $8.39 \%$ & $10.49 \%$ & $13.99 \%$ & $\boldsymbol{49.65 \%}$\\
Mouth slightly open & $1.41 \%$ & $7.04 \%$ & $15.49 \%$ & $4.93 \%$ & $4.93 \%$ & $\boldsymbol{66.20 \%}$\\
Narrow eyes & $4.32 \%$ & $30.22 \%$ & $\boldsymbol{33.81 \%}$ & $5.76 \%$ & $1.44 \%$ & $24.46 \%$\\
Pointy nose & $3.57 \%$ & $10.71 \%$ & $6.43 \%$ & $5.71 \%$ & $23.57 \%$ & $\boldsymbol{50.00 \%}$\\
Smiling & $2.84 \%$ & $3.55 \%$ & $\boldsymbol{63.12 \%}$ & $1.42 \%$ & $0.71 \%$ & $28.37 \%$\\
Straight hair & $2.86 \%$ & $20.00 \%$ & $25.00 \%$ & $7.14 \%$ & $9.29 \%$ & $\boldsymbol{35.71 \%}$\\
Wavy hair & $3.55 \%$ & $5.67 \%$ & $13.48 \%$ & $3.55 \%$ & $13.48 \%$ & $\boldsymbol{60.28 \%}$\\
Wearing lipstick & $5.76 \%$ & $6.47 \%$ & $10.07 \%$ & $7.91 \%$ & $14.39 \%$ & $\boldsymbol{55.40 \%}$\\
\hline\hline
\end{tabular}
\end{table*}
\begin{table}[!th!]
\renewcommand{\arraystretch}{1.3}
\setlength{\tabcolsep}{1.5mm}
\caption{Implementation details for the InterFaceGAN--D model used in the experiments. The table shows which attribute was disentangled for a given target attribute. The  target attributes presented in italic font denote the attributes, for which the disentangled attribute was selected based on the maximum absolute correlation, computed over the CelebA training dataset.}
\label{tab:interface-d}
\centering
    \begin{tabular}{l|l} 
    \hline\hline
     \textbf{Target attribute} & \textbf{Disentangled attribute}\\
     \hline 
     Arched eyebrows & Male \\
     \textit{Big nose} & Chubby \\
     Black hair & Pale skin \\
     Blond hair & Pale skin \\
     Brown hair & Pale skin \\
     \textit{Bushy eyebrows} & Mustache \\
     Grey hair & Young \\
     Mouth slightly open & Smiling \\
     \textit{Narrow eyes} & Bald \\
     \textit{Pointy nose} & Rosy cheeks \\
     Smiling & Mouth slightly open \\
     \textit{Straight hair} & Bald \\
     \textit{Wavy hair} & Heavy makeup \\
     Wearing lipstick & Male \\
     \hline\hline
    \end{tabular}
\end{table}

\section{User study details}
The user study was performed on the Amazon Mechanical Turk platform. Access was only granted to ``Master workers'', that is, workers that have had good performance on other tasks at the time of the study. All images shown were of the same resolution ($1024 \times 1024$) to discourage annotation based on resolution. Lower resolution images were resampled using bilinear interpolation. The question posed to worker was \textit{"Choose the image that changes the attribute more successfully, is of higher image quality and better preserves the identity and fine details of the source image"}. The workers then had to score the editing results of each of the considered models. The images were reshuffled for every worker task to avoid cheating. The same setup was also used for the Likert scale study.

\section{Implementation details}

\subsection{Encoder-decoder methods (StarGAN, AttGAN, STGAN)}
For the encoder-decoder methods we strictly follow the implementation details specified by the source code of each model. The only hyperparameter specified is the input and output resolution of each model. For StarGAN model, the highest suggested resolution advocated by the authors for this model ($256 \times 256$) resulted in considerable visual degradations of the editing results. We, therefore, trained the model with a smaller resolution ($128 \times 128$) that ensured better performance. For AttGAN and STGAN model, we chose the highest advocated resolution, that is, $384 \times 384$ pixels.

\subsection{InterFaceGAN}
The InterFaceGAN method is implemented in the latent space of the StyleGAN2 model, similarly to MaskFaceGAN. We follow the procedure as outlined in \cite{shen2020interpreting}, i.e.: first, we generate $500,000$ pairs of latent codes $w$ and corresponding images. An attribute classifier scores the facial attribute of each image and the top $10,000$ positive and negative samples are selected for the next step. A linear SVM is then applied over the latent codes of the selected images to obtain the normal vector of the hyperplane needed for editing. The SVM regularization term is set to $1$.

For InterFaceGAN, the magnitude of the movement in the latent space needs to be specified. A small magnitude barely changes the presence/absence of a facial attribute, but preserves the facial identity. A large magnitude tends to make larger changes that correspond to the desired facial attribute edit, however it usually comes at the expense of a change in identity. We experimented with  several different magnitudes and chose $1$ as the optimal value that provided the best trade-off between attribute editing performance and preservation of original image characteristics on our sample images. 

\subsection{InterFaceGAN--D}
When implementing the disentangled version of the InterFaceGAN model, denoted as InterFaceGAN--D, one can choose multiple attributes that can be disentangled from the targeted attribute to be edited. In our experiments, we found that disentanglement works best when  disentangling only a single attribute, i.e. with a simple vector projection method, as described in \cite{shen2020interpreting}. When choosing the attribute to disentangle for a given target attribute, we choose the attribute that displays the most entanglement issues in the generated results. For the attributes, where entanglement was not that problematic or the target attribute is hard to specify, the most correlated attribute (positive or negative correlation) from the CelebA--HQ training dataset was chosen. The disentangled attributes are presented in Table \ref{tab:interface-d}.

\section{Reproducibility}

To foster reproducibility, all experiments presented in the main part of the paper as well as in the supplementary material are implemented with publicly available source code. For the competing models considered in the comparative experiments the official code released by the authors was used as detailed below. 
\begin{itemize}
    \item The StarGAN model was implemented with the following code: \url{https://github.com/yunjey/stargan}
    \item The AttGAN model was implemented with the following code: \url{https://github.com/LynnHo/AttGAN-Tensorflow}
    \item The STGAN model was implemented with the following code: \url{https://github.com/csmliu/STGAN}
     \item The InterFaceGAN(-D) model was implemented with the following code: \url{https://github.com/genforce/interfacegan}
     \item The StyleGAN2 model used as the generator for MaskFaceGAN was implemented with the following code: \url{https://github.com/NVlabs/stylegan2}
\end{itemize}
The source code for the MaskFaceGAN is also made publicly available via the following URL: \url{https://github.com/MartinPernus/MaskFaceGAN}.

\end{document}